\documentclass
	[paper=17cm:24cm,
	DIV=14,
	fontsize=10pt, 
	BCOR=15mm,
	parskip=half,
	cleardoublepage=empty,
	headinclude, 
	pagesize]
	{scrbook} 

\usepackage[ngerman,english,american]{babel} % Englische Version
\usepackage{palatino}
\usepackage{amsmath}
\usepackage{amsthm}
\usepackage{amssymb}
\usepackage{makeidx}
\usepackage{url}
\usepackage{hyperref}
\usepackage{graphicx}
\usepackage{url}
\usepackage{listings}
\def\TReg{\textsuperscript{\textregistered}}

\def\TTra{\textsuperscript{\texttrademark}}
\newcommand{\sgn}{\operatorname{sgn}}
\newcommand{\etal}{\textit{et al.\,}}
\newcommand{\eg}{\textit{e.g.}}
\newcommand{\ie}{\textit{i.e.}}
\newcommand{\app}{App.,\,}

\usepackage[sort&compress]{natbib} % im text nach nummer sortierte literaturverweise
\bibpunct{[}{]}{,}{n}{,}{,} % trennzeichen für literaturverweise
\usepackage{tocbibind} % \usepackage[nottoc]{tocbibind}: ohne "Inhaltsverzeichnis" im Inhaltverzeichnis

\usepackage{listings} % Für Quellcodes

\usepackage{cleveref}
\crefname{appendix}{App.\negthinspace\,}{App.\negthinspace\,}
\crefname{chapter}{Chap.\negthinspace\,}{Chap.\negthinspace\,}
\crefname{equation}{Eq.\negthinspace\,}{Eq.\negthinspace\,}
\crefname{algorithm}{Alg.\negthinspace\,}{Alg.\negthinspace\,}
\crefname{section}{Sec.\negthinspace\,}{Sec.\negthinspace\,}
\crefname{subsection}{Sec.\negthinspace\,}{Sec.\negthinspace\,}
\crefname{subsubsection}{Sec.\negthinspace\,}{Sec.\negthinspace\,}
\crefname{figure}{Fig.\negthinspace\,}{Fig.\negthinspace\,}
\crefname{table}{Tab.\negthinspace\,}{Tab.\negthinspace\,}
\crefname{subfigure}{Fig.\negthinspace\,}{Fig.\negthinspace\,}
\crefname{subsubfigure}{Fig.\negthinspace\,}{Fig.\negthinspace\,}
\crefname{lstlisting}{Lst.\negthinspace\,}{Lst.\negthinspace\,}
\crefname{listing}{Lst.\negthinspace\,}{Lst.\negthinspace\,}
\crefname{lstlistingname}{Lst.\negthinspace\,}{Lst.\negthinspace\,}
\usepackage{amssymb}    % für weitere Symbole (durchgestrichene Pfeile)
\usepackage{amsmath}
\usepackage{bm}
\usepackage{algorithm2e}
\usepackage{url}
\usepackage{tabularx}
\usepackage{resizegather}

\usepackage{lscape}

\usepackage{rotating}
\usepackage{supertabular}
\usepackage{booktabs}
\usepackage{longtable}
\usepackage{ifthen}
\usepackage{theorem}
\usepackage{pbox}
\usepackage[table]{xcolor}
\PassOptionsToPackage{table}{xcolor}

\usepackage[justification=justified, %centerlast
            format=plain,
            labelfont={small,bf},
            textfont=small,
            width=\textwidth]{caption}

\newcommand{\argmax}{\operatornamewithlimits{argmax}}
\newcommand{\argmin}{\operatornamewithlimits{argmin}}

\lstdefinelanguage{TXT}
{
  basicstyle=\ttfamily,
  morestring=[s]{"}{"},
  morecomment=[s]{?}{?},
  morecomment=[s]{!--}{--},
  commentstyle=\color{black},
  moredelim=[s][\color{black}]{>}{<},
  moredelim=[s][\color{black}]{\ }{=},
  stringstyle=\color{black},
  identifierstyle=\color{black}
}

\lstdefinelanguage{XML}
{
  basicstyle=\ttfamily,
  morestring=[s]{"}{"},
  morecomment=[s]{?}{?},
  morecomment=[s]{!--}{--},
  commentstyle=\color{darkgreen},
  moredelim=[s][\color{black}]{>}{<},
  moredelim=[s][\color{red}]{\ }{=},
  stringstyle=\color{blue},
  identifierstyle=\color{maroon}
}

\definecolor{maroon}{rgb}{0.5,0,0}
\definecolor{darkgreen}{rgb}{0,0.5,0}
\definecolor{lightgray}{rgb}{.95,.95,.95}

\usepackage{pdfpages}

\makeindex

\begin{document}

%\frontmatter
%\input{preface}
%\maketitle
%\thispagestyle{empty}

\thispagestyle{empty}
\pagenumbering{Alph}
\begin{center}

\vskip 8cm

\huge \textbf{ \textrm{New Methods to Improve Large-Scale Microscopy Image Analysis with Prior Knowledge and Uncertainty} } \normalsize

\vskip 2cm

\large Zur Erlangung des akademischen Grades\\ 
\vskip 0.25cm
\large \textbf{Doktor der Ingenieurwissenschaften} \\ 
\vskip 0.25cm
\large der Fakult\"at f\"ur Maschinenbau \\ 
\vskip 0.25cm
\large Karlsruher Institut f\"ur Technologie (KIT)

\vskip 3cm %\large Diplomarbeit \normalsize \vskip 3cm

\large genehmigte \\
\vskip 0.25cm
\large \textbf{Dissertation} \\
\vskip 0.25cm
von \\
\vskip 0.5cm
\large \textbf{Johannes Stegmaier, M.Sc.} \\
\vskip 0.25cm
\large geboren am 16. November 1985 in Stuttgart \\
\vfill

\begin{tabular}{ll}
\textrm{Hauptreferent:} & \textrm{apl. Prof. Dr.-Ing. Ralf Mikut}\\
\textrm{Korreferenten:}	 & \textrm{Prof. Dr. Uwe Str\"ahle}\\
  & \textrm{Prof. Dr. Jan G. Korvink} \\
	\\
\textrm{Tag der m\"undlichen Pr\"ufung:} & \textrm{3. Juni 2016} \\	
\end{tabular}

\newpage
\thispagestyle{empty}
%\cleardoublepage
\clearpage

\end{center}

\pagenumbering{Arabic}

\includepdf[pages=-]{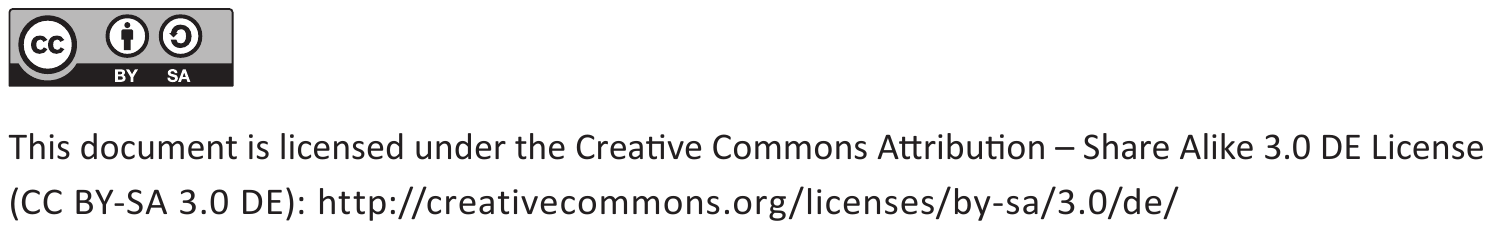}

% Abstracts
\setcounter{page}{1} \pagenumbering{roman}

\selectlanguage{ngerman}

\chapter*{Zusammenfassung}
\addcontentsline{toc}{chapter}{Zusammenfassung} 
J\"ungste Entwicklungen im Bereich der mehrdimensionalen Mikroskopie bieten ein gro{\ss}es Potential f\"ur die Beantwortung vielerlei Fragestellungen in wissenschaftlichen Bereichen. Beispielsweise bieten neue Verfahren wie die zeitaufgel\"oste 3D Konfokal- und Lichtscheibenmikroskopie oder die Transmissionselektronenmikroskopie weitreichende M\"oglichkeiten im Bereich der Biologie, die von der ganzheitlichen Analyse embryonaler Entwicklung \"uber die Betrachtung subzellul\"arer Prozesse bis hin zur Rekonstruktion von Verschaltungen im Nervensystem von Modellorganismen reichen. Die dabei routinem\"a{\ss}ig anfallenden Datenmengen im Terabyte-Bereich k\"onnen allerdings nur unzureichend manuell ausgewertet werden und eine wichtige Komponente f\"ur die erfolgreiche Auswertung solcher bildbasierter Experimente ist daher eine gr\"o{\ss}t-m\"ogliche Anzahl von Analyseschritten durch Bildanalyseverfahren zu automatisieren. Bestehende Verfahren f\"ur die automatische Bildauswertung sind hierbei jedoch meist nicht unmittelbar auf die gro{\ss}en Datenmengen anwendbar und die Analysen m\"ussen daher auf kleine Ausz\"uge der Daten beschr\"ankt werden, falls die enormen Anforderungen an Rechenleistung und Verarbeitungszeit nicht gew\"ahrleistet werden k\"onnen. Zudem wird vorhandenes \textit{a priori} Wissen oft nur unzureichend in automatische Verfahren eingebettet und somit ein bedeutender Teil an Zusatzinformationen vernachl\"assigt, der zu einer verbesserten Ergebnisqualit\"at beitragen kann.

Die Hauptbeitr\"age der vorliegenden Arbeit sind ein neues Konzept zur Absch\"a-tzung und Weiterleitung von Unsicherheiten in Bildverarbeitungsketten sowie die Entwicklung neuer Segmentierungsverfahren, die f\"ur eine effiziente Analyse von 3D Mikroskopbildern im Terabyte-Bereich eingesetzt werden k\"onnen. Basierend auf unscharfen Mengen (engl. fuzzy sets) wurde zur Verf\"ugung stehendes Vorwissen systematisch in eine mathematische Repr\"asentation \"uberf\"uh-rt, die anschlie{\ss}end f\"ur effiziente Datenselektion, eine Unsicherheitsabsch\"at-zung von automatisch extrahierten Daten sowie f\"ur eine gezielte Verbesserung von Bildverarbeitungsoperatoren eingesetzt werden konnte. Um den Bedarf an effizienten Bildverarbeitungsalgorithmen zu reduzieren wurden drei neue Segmentierungsalgorithmen entwickelt, die sich f\"ur eine Extraktion von sph\"ari-schen, linienf\"ormigen und lokal planaren Objekten eignen. Die neuen Segmentierungsmethoden wurden dabei gezielt f\"ur den Einsatz in der automatisierten Analyse von gro{\ss}en 3D Bilddatens\"atzen optimiert, insbesondere durch die systematische Ausnutzung von vorhandenem \textit{a priori} Wissen w\"ahrend der Algorithmenentwicklung und durch die Beschr\"ankung auf rechen- und speichereffiziente Teilkomponenten in der Implementierung.  Anhand einer exemplarischen Bildverarbeitungskette wurde veranschaulicht, wie Unsicherheiten in bestehende Operatoren integriert werden und zur Verbesserung der Ergebnisqualit\"at beitragen k\"onnen. Um die Funktionalit\"at der vorgestellten Verfahren zu validieren, wurden erweiterte oder zum Teil neu erstellte, simulierte Benchmarkdatens\"atze verwendet, die eine Vielzahl m\"oglicher Einsatzszenarien systematisch abdecken. Die effizienten Implementierungen werden zum einen innerhalb der vorliegenden Arbeit vorgestellt und zum anderen als plattformunabh\"angige Open-Source Software zur allgemeinen Verf\"ugung bereitgestellt. Eine Reihe von Problemen im Bereich der Entwicklungsbiologie wurde mittels der theoretisch eingef\"uhrten Verfahren erfolgreich ausgewertet. So wurden die Methoden beispielsweise f\"ur eine automatisierte, quantitative Analyse der Auswirkungen von bekannten und unbekannten Chemikalien auf die neuronale Entwicklung im R\"uckenmark von Zebrab\"arblingen, f\"ur die Detektion, Segmentierung und zeitliche Verfolgung von fluoreszenzmarkierten Zellkernen in Zebrab\"arblingsembryos sowie zur quantitativen Charakterisierung von Zellmorphologie\"anderungen in zeitaufgel\"osten 3D Mikroskopbildern von Fruchtfliegen-, Zebrab\"arblings- und Mausembryos eingesetzt.

\selectlanguage{american}

‌

\selectlanguage{american}

\chapter*{Abstract}
\addcontentsline{toc}{chapter}{Abstract}

Recent developments in the area of multidimensional imaging techniques provide powerful ways to examine various kinds of scientific questions. For instance, in biological applications, time-resolved 3D light-sheet microscopy and serial section electron microscopy provide unprecedented possibilities ranging from \textit{in toto} analyses of embryonic development down to investigations of subcellular processes or reconstructions of the nervous system. The routinely produced datasets in the terabyte-range, however, can hardly be analyzed manually. Thus, the extensive use of image analysis-based automation is an essential key to the success of the performed imaging experiments. Existing algorithms for such analysis tasks are mostly not directly applicable to these large-scale datasets and either have to be confined to small excerpts of the data or require an immense amount of computation capacities and execution time. Moreover, available prior knowledge that could be exploited for advanced analyses is often not sufficiently considered by automatic processing pipelines.

The major contributions of the present thesis are a new concept for the estimation and propagation of uncertainty involved in image analysis operators and the development of new segmentation algorithms that are suitable for terabyte-scale analyses of 3D+t microscopy images. Based on fuzzy set theory, available \textit{a priori} knowledge was transformed into a mathematical representation and extensively used to enhance the performance of processing operators by data filtering, uncertainty propagation and explicit exploitation of information uncertainty for result improvements. To target the need for efficient image analysis operators, three new segmentation algorithms were specifically developed to detect a generalized geometric class of objects, namely, spherical objects, line-like objects and locally plane-like objects. The developed pipelines were specifically tuned to be applicable to large-scale analyses, \ie, only fast and memory efficient processing operators were used in the implementation. Using an exemplary pipeline, it is demonstrated how a combination of both the fast algorithms and the proposed uncertainty framework could be used to further enhance the overall quality of the considered processing operators. All developed methods were thoroughly validated on existing and newly developed simulated benchmarks, to be able to quantitatively assess their applicability to different imaging conditions. In addition, the efficient implementations of all developed algorithms are presented and were made accessible to the community as platform independent open-source software tools. The new methods were successfully applied to multiple large-scale analyses of fluorescence microscopy images in the field of developmental biology. In particular, the proposed pipelines were used to quantify the impact of both known and unknown chemical substances on the neuronal development in the spinal cord of zebrafish in 2D images. Furthermore, the developed methods were applied to time-resolved 3D images to detect, segment and track fluorescently labeled cellular nuclei of entire zebrafish embryos and to quantitatively characterize cell morphology dynamics using fluorescently labeled cellular membranes in 3D+t microscopy images of fruit fly, zebrafish and mouse embryos.
 \cleardoublepage

% Acknowledgements

\chapter*{Acknowledgements}
\addcontentsline{toc}{chapter}{Acknowledgements}
In the first place I want to thank Prof.~Dr.-Ing.~habil.~Georg Bretthauer for the great opportunity to spend my time as a PhD student at the Institute for Applied Computer Science (IAI) at the Karlsruhe Institute of Technology (KIT) and for his supervision of the thesis. Special thanks to my direct supervisor apl.~Prof.~Dr.-Ing.~Ralf Mikut for the guidance, encouragement, constructive discussions and the continuous support throughout the entire time at the IAI. I deeply appreciate having had the chance to work freely and responsibly on a highly captivating project. I want to thank Prof.~Dr.~Uwe Str\"ahle and Prof.~Dr.~Jan G. Korvink for reviewing the thesis and Prof.~Dr.~Barbara Deml for heading the examination board.
Thanks to all colleagues, Bachelor's students and trainees at the IAI, especially, to R\"udiger Alshut, Thomas Antritter, Andreas Bartschat, Dr.~Christian Bauer, Wolfgang Doneit, Eduard H\"ubner, Arif ul Maula Khan, Jorge Angel Gonzalez Ordiano, Nico Peter, Willis Pinaud, Dr.~Markus Reischl, Benjamin Schott, Manuel Traub, Michele Rene Tuga and Simon Waczowicz for the pleasant, cooperative and exciting working atmosphere in the group for biosignal analysis.
Many thanks to all the collaboration partners from the Institute of Toxicology and Genetics (ITG) and the Institute for Applied Physics (APH), especially, for igniting my interest in developmental biology. In particular I want to thank Dr.~Thomas Dickmeis, Dr.~Andrei Kobitski, Prof.~Dr.~G.~Ulrich Nienhaus, Dr.~Jens~C.~Otte, Dr.~Sepand Rastegar, Dr.~Maryam Shahid, Prof.~Dr.~Uwe Str\"ahle, Dr.~Masanari Takamiya, Dr.~Benjamin Weger, Dr.~Meltem Weger and Dr.~Lixin Yang for all the exciting projects I was allowed to contribute to during the time being at KIT.
I really appreciate all the infrastructural support such as data storage management and cluster computing received by Serguei Bourov, Dr.~Ariel Garc\'ia, Volker Hartmann, Dr.~Rainer Stotzka, Jos van Wezel and all the persons behind the scenes that made the required large-scale analyses technically feasible.
For the support, guidance and exciting advanced training courses I want to thank my thesis advisory committee (apl.~Prof.~Dr.-Ing.~Ralf Mikut, Dr.~Markus Reischl, Dr.~Ute Schepers and Prof.~Dr.~Uwe Str\"ahle) and the BioInterfaces International Graduate School (BIF-IGS).
Moreover, I'm grateful for the unique opportunity to do an inspiring internship at Howard Hughes Medical Institute's Janelia Farm Research Campus (JFRC). For all their support and supervision during the internship I want to thank Dr.~Philipp J. Keller, Dr.~Fernando Amat and the whole Keller Lab. For financial support during my time as a PhD student at KIT and during the internship at JFRC, I want to thank the Helmholtz Association in the program BioInterfaces, the Howard Hughes Medical Institute (HHMI) and the Karlsruhe House of Young Scientists (KHYS).
For their contributions to the implementation and improvements of XPIWIT I want to thank Dr.~Fernando Amat (fast GPU implementations), Andreas Bartschat (XML functionality, filter implementations and pipeline workflow), Eduard H\"ubner (graphical user interface) and apl.~Prof.~Dr.-Ing.~Ralf Mikut (Gait-CAD interface). Additionally, I want to thank R\"udiger Alshut, Andreas Bartschat, apl.~Prof.~Dr.-Ing.~Ralf Mikut, Dr.~Sebastian Pfeiffer, Dr.~Markus Reischl and Karl W\"oll for their contributions to the IMVid and Tracking toolboxes.
Last but not least, I want to cordially thank my girlfriend, family and friends for their continuous support and understanding when I was buried in thoughts once in a while.
 \cleardoublepage

\tableofcontents \cleardoublepage

%% Content
\setcounter{page}{1} \pagenumbering{arabic}

%% Introduction

%%%%%%%%%%%%%%%%%%%% INTRODUCTION %%%%%%%%%%%%%%%%%%%%
%\part{Introduction (20)}
\chapter{Introduction}

Three-dimensional (3D) imaging techniques have lately revolutionized the possibilities of in-depth analyses in almost any area of natural sciences. Techniques like radar, to\-mo\-gra\-phy, microscopy and other image-based techniques allow to probe various unanswered questions in unprecedented detail, \eg, in the fields of material science, physics, biology or medicine \cite{Midgley03, Moosmann13, Amat14}. An inherent problem to volumetric imaging, however, is the tremendous amount of data that is produced by these techniques. Especially, experiments that involve a spatio-temporal acquisition of volumetric image data (3D+t) easily produce datasets reaching the terabyte-range. Besides the infrastructural challenges, datasets of this size render manual investigations almost impossible and require an extensive use of automatic image processing and analysis for the successful assessment of experimental results. A multitude of algorithms for automatic image analysis have been presented in the past. However, most of these methods were developed and tested solely on 2D images and their immediate application to 3D is not possible in many cases. In addition to implementation-based hurdles, many mathematically elegant solutions require an immense amount of computing power and computation time to generate results, and are thus often not practically applicable. Furthermore, available prior knowledge is mostly not systematically incorporated into the algorithms, \ie, a significant amount of extra information is wasted instead of being used to improve algorithmic performance.

The aim of this work was to derive a generic and efficient way to enhance image analysis pipelines by an extensive use of prior knowledge and to assess the uncertainty of produced results with respect to predefined, prior knowledge-based criteria. The generally introduced concept was successfully used to filter erroneous detections, to improve flawed results and to generally resolve ambiguities occurring in automatic processing operators, and had only a low impact on the computation time. Moreover, three new segmentation algorithms were developed in order to provide efficient new ways to extract various geometric shapes such as line-like, spherical or locally plane-like objects from large-scale 3D images. The performance improvements were achieved by maximizing the amount of prior knowledge used for the algorithmic development. Furthermore, the involved processing operators were carefully selected and optimized with respect to both time complexity and memory efficiency, to make large-scale analyses feasible. All methods were validated on extended and newly developed artificial benchmarks, in order to proof their robustness against disturbances and their applicability in a variety of different imaging scenarios. In all scenarios, the algorithms provided excellent results comparable to state-of-the-art algorithms, while lowering the processing times by at least an order of magnitude compared to existing methods. Finally, all methods were used for quantitative analyses of multidimensional fluorescence microscopy images from the field of developmental biology. The presented approaches were successfully used to (1) quantitatively analyze the impact of known and unknown small molecules to neuron populations in the spinal cord of zebrafish embryos, (2) to detect, segment and track fluorescently labeled cell nuclei in terabyte-scale 3D+t image data of zebrafish embryos and (3) to obtain a quantitative characterization of cell shape dynamics in time-resolved 3D microscopy images of fruit fly, zebrafish and mouse embryos. 

The next section introduces related work and the context of the thesis, such as fluorescence microscopy, the required image analysis components as well as previous approaches to quantify and use uncertain information in image-based processing pipelines. Furthermore, a summary of existing benchmarks and software solutions is provided. Although most of the developed methodology is generally applicable to problems observed in 3D imaging experiments, the presentation of recent methods and the considered applications mainly focuses on the automatic analysis of 3D fluorescence microscopy images as demanded by the biological application field considered throughout this thesis. An overview of open questions, aims of the thesis as well as a coarse outline of the major attempts how the problems were solved is provided at the end of this chapter.

%%%%%%%%%%%%%%%%%%%% RELATED WORK %%%%%%%%%%%%%%%%%%%%
\section{Theoretical Background and Related Work}

%%%%%%%%%%%%%%%%%%%%%%% IMAGE REPRESENTATION %%%%%%%%%%%%%%%%%%%%%%%%
\subsection{Image Acquisition}
\label{sec:chap1:ImageAcquisition}
In the beginning of image-based experiments the object of interest has to be recorded using the microscopy technique of choice, such as stereo, compound, phase contrast, confocal, light-sheet or electron microscopy to name but a few. Each of the methods has its own benefits and drawbacks and the appropriate technique has to be carefully selected based on the questions to be answered. A great overview of various microscopy approaches can be found in \cite{Murphy12}. In this thesis, the main emphasis is put on fluorescence microscopy with a special focus on imaging biological specimens using light-sheet fluorescence microscopy, as all application examples considered in this work stem from the field of fluorescence microscopy.

In principle, fluorescence microscopy represents a special area of light microscopy. Based on illumination at a specific excitation wavelength, emitted fluorescence can be imaged through the detection optics using a special filter that is matched to the emission spectrum of the fluorophore the investigated specimen has been labeled with \cite{Misteli97}. The emitted and filtered fluorescent signal can then be captured using detectors like charge-coupled devices (CCD) or complementary metal oxide semiconductor-based detectors (CMOS) and is subsequently digitized for archival, processing and analysis \cite{Murphy12}. Besides fluorescent staining methods like Hoechst or DAPI dyes to label nuclear DNA and immunofluorescence approaches using antibodies \cite{Otto90, Eissing14}, the use of fluorescent proteins has unveiled enormous possibilities for detailed image-based studies of gene expression and protein targeting in intact cells and whole organisms \cite{Tsien98}. Fluorescent proteins like the green or red fluorescent protein (GFP, RFP) can be fused to a gene of interest and are subsequently detectable upon expression of this particular gene in the transgenic target organism using fluorescence microscopy \cite{Misteli97}.

\begin{figure}[htb]
\centerline{\includegraphics[width=0.5\columnwidth]{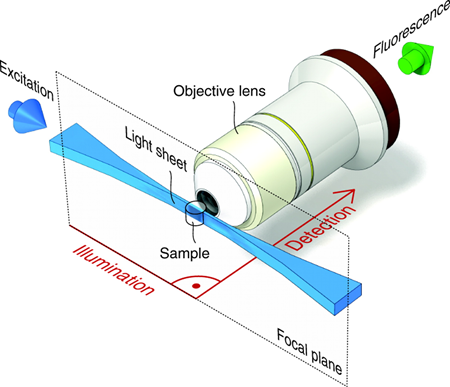}}
\caption[Fundamental principle of light-sheet fluorescence microscopy]{Fundamental principle of light-sheet fluorescence microscopy. The specimen is located in a tube-like holder in the center of the microscope and is illuminated using a movable thin sheet of laser light. Illumination is performed orthogonal to the detection objective and is positioned, such that only the focal plane is excited. Three-dimensional images can be acquired with this technique by moving the light-sheet or the sample (reproduced with permission from \cite{Huisken09}).}
\label{fig:chap1:SPIMConcept}
\end{figure}

Approaches such as confocal microscopy and the recently established light-sheet microscopy perform an optical sectioning of the investigated probe to obtain detailed 3D image stacks, possibly of the entire specimen. Moreover, the improved imaging speed of these techniques enables the acquisition of time-series of 3D images, \ie, to obtain 3D videos of dynamic objects \cite{Bao06, Huisken09, Weber11}. The basic principle of a light-sheet microscope is illustrated in \cref{fig:chap1:SPIMConcept}. Based on a focused laser light-sheet, sections of a few micrometers are illuminated at a time and captured by orthogonally arranged detection optics \cite{Huisken04}. The microscope is adjusted such that only the focused plane is illuminated, to minimize fluorescence in out-of-focus areas, \ie, to effectively reduce light scattering, photobleaching of fluorophores as well as phototoxicity \cite{Kobitski15}. A 3D image comprised of sequentially acquired planes can be generated by moving the specimen through the light-sheet (Selective Plane Illumination Microscopy, SPIM, \cite{Huisken04}) or vice versa by scanning the laser light-sheet through the specimen (Digital Scanned Laser Light Sheet Fluorescence Microscopy, DSLM, \cite{Keller08}).
Problems such as light scattering and absorption can be successfully compensated by multiview acquisition schemes that rotate the specimen \cite{Huisken04, Keller08SC} or use multiple detection paths \cite{Krzic12, Tomer12}. Furthermore, a more homogeneous illumination and a reduced light absorption of the imaged plane can be obtained by the use of double-sided illumination \cite{Huisken09, Keller08SC, Kobitski15}.

Successfully imaged model organisms using both confocal and light-sheet microscopy are embryos of nematodes (\textit{Caenorhabditis elegans}) \cite{Bao06}, fruit flies (\emph{Drosophila melanogaster}) \cite{Tomer12, Kausler12}, crustaceans (\emph{Parhyale hawaiensis}) \cite{Benton13}, zebrafish (\emph{Danio rerio}) \cite{Keller08SC, Lou11, Maerz11, Kausler12, Mikut13, Kobitski15} and mice (\emph{Mus musculus}) \cite{Amat14, Stegmaier16}. \cref{tab:chap1:3DScreens} provides an impression of the typical amount of images produced in 3D+t imaging experiments of different model organisms in the field of developmental biology using confocal or light-sheet microscopy.
The acquired 3D+t images routinely produced by these imaging systems easily reach the terabyte-range. Since the first commercial system has become available (Zeiss Z1) and detailed building plans for self-assembled microscopes have been published recently \cite{Pitrone13, Gualda13}, the amount of image data produced by light-sheet microscopes will further increase. Most commercially available platforms, however, only provide solutions for archival and viewing of data but offer only limited image analysis capabilities, \ie, there is an urgent need for an elaborate toolbox of automatic image analysis approaches.
\begin{table}[htb]
%\rowcolors{2}{white}{gray!25}
%\centering
\resizebox{\textwidth}{!}{
\begin{tabular}{lccccc}
\toprule
\textbf{Author} & \textbf{Modality} & \textbf{Organism} & \textbf{Duration} & \textbf{Size (Per Image/Embryo)} \\
\midrule
Bao \etal (2006) \cite{Bao06} & Confocal & Nematode & 6.6~h & 10.2~MB / 4.0~GB \\
Tomer \etal (2012) \cite{Tomer12} & SiMView & Fruit Fly & 20~h & 5.1~GB / 11.0~TB \\
Keller \etal (2008)  \cite{Keller08SC} & DSLM & Zebrafish & 24~h & 3.9~GB / 15.3~TB \\
Kobitski \etal (2015) \cite{Kobitski15} & DSLM & Zebrafish & 16~h & 5.1~GB / 9.0~TB \\
Amat \etal (2014) \cite{Amat14} & SiMView & Mouse & 2~h & 332~MB / 8.4~GB \\
\bottomrule
\end{tabular}}
\caption[Exemplary 3D+t datasets produced in embryomics]{Exemplary datasets that were used to study the early embryonic development of various model organisms using confocal or light-sheet microscopy. Acquisition durations range from a few hours up to days and produce 3D+t image datasets in the terabyte-range for a single embryo (values taken from \cite{Bao06, Tomer12, Keller08SC, Kobitski15, Amat14}).}
\label{tab:chap1:3DScreens}
\end{table}

%%%%%%%%%%%%%%%%%% IMAGE ANALYSIS %%%%%%%%%%%%%%
\subsection{Image Analysis}
\label{sec:chap1:ImageAnalysis}

%%%%%%%%%%%%%%%%%% DIGITAL IMAGE REPRESENTATION %%%%%%%%%%%%%%
\subsubsection{Digital Image Representation and the Processing Pipeline Concept}
\label{sec:chap1:ImageRepresentation}
Digital images are usually represented as a 2D, 3D or more generally an N-D matrix, if multiple channels, slices or time points of a scene are captured. Each of the matrix entries directly correlates with the amount of photons that were absorbed by the respective sensors and the analog signal is quantized using an analog-to-digital converter, \ie, a mapping of an analog signal to a finite set of positive integer or floating point values is performed. Common dynamic ranges used for digital images are 8 bit (256 intensity levels), 16 bit (65536 intensity levels) or 32 bit (floating point, usually scaled to the interval $[0,1]$). For a 16-bit image this corresponds to $0$ being the minimum intensity (black) and $65535$ being the maximum intensity (white) \cite{Burger09}. Consequently, RGB color images store a separate intensity value for each of the color components red, green and blue. In cases where the physical detector operates with a lower dynamic range (\eg, 12 bit), only a subset of the actually available intensity range provided by the image format is used. On the other hand, if the number of photons hitting a pixel during the exposure time exceeds its full-well capacity, a saturation of the pixel occurs and should be avoided by any means. Of course, the used bit depth directly affects the physical size of the datasets, \ie, converting an 8 bit image to 16 or 32 bit results in a doubling or even quadrupling of the input image and might require compression strategies to tackle the large file sizes. This becomes especially important when dealing with large-scale multidimensional image data on an infrastructure with limited data storage capacity, network bandwidth or main memory on the processing nodes. However, some data storage policies require the archival of uncompressed, raw original data, \ie, in these cases it is inevitable to directly preserve the acquired image data as it was produced by the respective microscope and to provide the required infrastructure.

With respect to image filtering, neighborhood relations based on pixel connectivity often have to be considered \cite{Jaehne12}. Common ways to define the pixel connectivity in 2D images are the 4-neighborhood (pixels that share an edge with the central pixel) and the 8-neighborhood (all pixels surrounding the central pixel). Analogously, the voxel connectivity for 3D images can be defined using a 6-neighborhood (voxels that share a face with the central voxel) or a 26-neighborhood (all voxels surrounding the central pixel). As these definitions fail in border regions of the image, different methods like padding with constant values, the intensity of the closest border pixel or a cyclic mirroring of the image content can be used to allow neighborhood operations in these areas \cite{Burger09}.

An important measure to quantify the quality of a digital imaging system is its signal-to-noise ratio. As the name already indicates, this measure connects the signal intensity observed for the investigated objects to the respective background noise. For the simulated benchmark data considered in this thesis, the following formulation was used to estimate the signal-to-noise ratio:
\begin{gather}
	\text{SNR} = \frac{\mu_{\text{fg}}}{\sigma_{\text{bg}}},
	\label{eq:chap1:SNR}
\end{gather}
with $\mu_{\text{fg}}$ being the foreground mean intensity and $\sigma_{\text{bg}}$ being the standard deviation of the background signal \cite{Smith97, Voigtman97, Welvaert13}. Using the available ground truth image as a mask, the bright foreground and low intensity background regions could be optimally separated. An alternative measure to assess the image quality is given by the contrast-to-noise ratio (CNR), which is defined as:
\begin{gather}
	\text{CNR} = \frac{|\mu_{\text{fg}}-\mu_{\text{bg}}|}{\sigma_{\text{bg}}},
	\label{eq:chap1:CNR}
\end{gather}
with $\mu_{\text{fg}}$ being the foreground mean intensity, $\mu_{\text{bg}}$ being the background mean intensity and $\sigma_{\text{bg}}$ being the standard deviation of the background signal \cite{Welvaert13}. Although, many variations of these criteria exist, the definition in \cref{eq:chap1:SNR} apparently is the most widely used formulation and was thus used assess the image quality throughout in this thesis. For background mean intensities close to zero, the two measures yield similar values.

A further relevant aspect to consider is the handling of metadata. For instance, information about the imaged objects, the acquisition procedure, the physical size of the voxels, image resolution and the like, can be used to adjust algorithmic parameters and to convert object properties such as size or volume from image space to physical units. Metadata is often packed into the file header or put into a separate metadata file \cite{Allan12}.

\begin{figure}[htb]
\centerline{\includegraphics[width=\columnwidth]{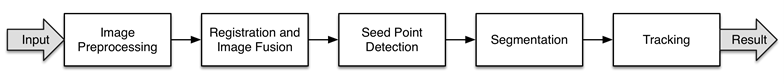}}
\caption[Exemplary image analysis pipeline]{Exemplary image analysis pipeline consisting of image preprocessing, registration and multiview fusion, seed point detection, segmentation and tracking.}
\label{fig:chap1:PipelineOverview}
\end{figure}

After the images and the associated metadata found their way to a digital representation, the processing of images as well as the information extraction is usually formulated as a set of linearly arranged processing operators that modify the image or retrieve information about its content. An overview of exemplary processing steps in a sequential order is shown in \cref{fig:chap1:PipelineOverview}. Information passed between the processing steps can be images, extracted features or both of them. Of course, some filters may also require the input of multiple preceding processing operators. However, for simplicity, these special cases were omitted in the figure. In the following sections each of the mentioned processing operators is described in more detail and an overview of available methods to accomplish the respective analysis tasks is provided therein.

%%%%%%%%%%%%%%%%%%%%%%% PREPROCESSING %%%%%%%%%%%%%%%%%%%%%%%%
\subsubsection{Image Preprocessing}
\label{sec:chap1:ImagePreprocessing}
\begin{figure}[htb]
\centerline{\includegraphics[width=\columnwidth]{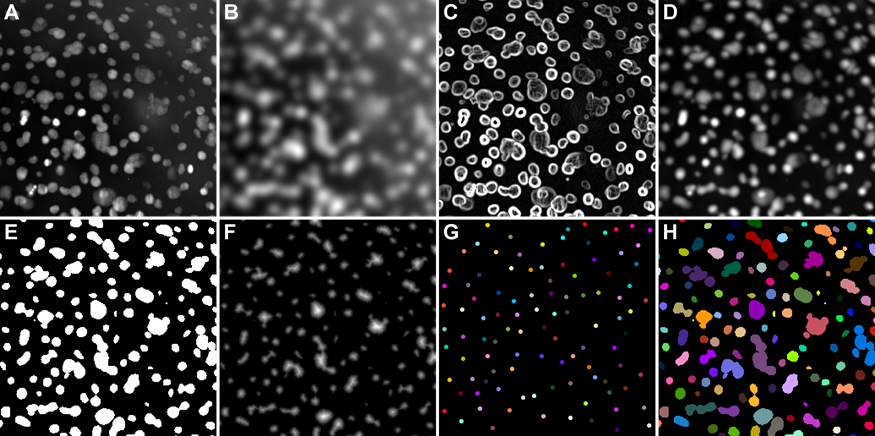}}
\caption[Exemplary results of several image analysis methods]{Exemplary results of various image analysis operators. Based on a raw input image of stained nuclei of a zebrafish embryo (A), the individual panels show the results of a Gaussian low-pass filter (B), a Sobel edge detector (C) and a Laplacian-of-Gaussian filter (D). In (E), a binarized version of the input image is shown and the image was used to calculate an Euclidean distance map (F), seed point locations (G) and the uniquely labeled connected components (H). All panels were generated using the open-source software Fiji \cite{Schindelin12}.}
\label{fig:chap1:AnalysisOperators}
\end{figure}

Image preprocessing represents an important step in the automated image analysis pipeline of microscopy images. Mostly, the acquired images are flawed by noise, illumination variations, dirt, limited resolution or simply contain unnecessary image regions that need to be discarded. To cope with such flawed image material, the general idea of preprocessing is to enhance and transform the images such that the further processing is facilitated. Using an appropriate preprocessing of the images can have a significant impact on the achieved processing quality of subsequent processing steps such as the seed detection, segmentation and tracking as described in the next sections. One of the most important enhancements is the suppression of noise in the images, \eg, using Gaussian low-pass filtering (\cref{fig:chap1:AnalysisOperators}B), mean filtering, median filtering, quantile-based noise removal or more advanced methods such as anisotropic diffusion filtering or objectness filters that selectively emphasize desired structures based on image derivatives \cite{Perona90, Frangi98, Antiga07}. Besides image noise, illumination inhomogeneities and contrast adaptions can be optimized using techniques like surfaces fitting, histogram equalization, high-pass filtering, morphological filtering or prior knowledge-based feedback techniques \cite{Vovk07, Sintorn10, Jaehne12, Khan13Krusebook}. However, if quantitative comparisons of intensity values need to be performed, it has to be ensured that the applied transformations preserve the comparability of different images. To emphasize properties such as edges, \ie, strong local intensity changes, methods like the Sobel filter, the Canny edge detector (\cref{fig:chap1:AnalysisOperators}C), the Laplacian-of-Gaussian (LoG, \cref{fig:chap1:AnalysisOperators}D) or the gradient magnitude can be used \cite{Canny1986, Gonzalez03, Marr80}.
Another frequently used family of efficient preprocessing methods are morphological operations such as erosion, dilation and their combinations like opening, closing and the top-hat filter, which are perfectly suited to improve imperfect raw images for a specific application \cite{Beucher93, Soille03, Gonzalez03}. Based on binary structuring elements that define the influence range of the operators, morphological operations can be used for image region separation, filling of holes, structure smoothing, efficient noise reduction, illumination correction and even for gradient approximations \cite{Soille03, Fernandez10, Luengo-Oroz12a, Khan14}. In some cases, only a specific part of the image is important for further consideration. Based on manual selection, intensity distributions, geometrical properties or template matching, the images can be trimmed to the requested region \cite{Peravali11, Chen11a}.

%%%%%%%%%%%%%%%%%%%%%%% REGISTRATION FUSION %%%%%%%%%%%%%%%%%%%%%%%%
\subsubsection{Registration and Image Fusion}
\label{sec:chap1:RegistrationAndFusion}
In many applications, images of different sources have to be spatially aligned to each other, such that the same pieces of information are located at the same places. Popular applications that require image registration are image stitching \cite{Brown07}, movement estimation \cite{Se01}, the combination of biomedical images from different modalities \cite{Hill01}, the registration of multiple views obtained from 3D microscopy \cite{Preibisch08a, Preibisch10, Tomer12} or the alignment of tissue sections in connectomics projects \cite{Chklovskii10}. In the scope of this thesis only rigid registrations are considered (translation, rotation and reflection). However, in different imaging scenarios it might also become necessary to scale or morph images in order to perfectly align the respective content, \ie, to perform an elastic registration \cite{Kybic03, Sorzano05, Gehrig09}.
Generally, a rigid registration of images can be performed either based on intensity value matching or on the registration of extracted features. The former method tries to identify a transformation that maximizes a predefined similarity measure of the intensity values in both images \cite{Preibisch08a}. As this method performs an iterative optimization of the transformation and needs to evaluate the similarity measure at each given step, it is inherently slow and involves the risk of being stuck in a local optimum. Additionally, intensity-based registration heavily relies on matching image content, which might not be satisfied for large developing specimen \cite{Preibisch09}. Nevertheless, the method can be sped up by restricting the degrees of freedom for the transformation and by choosing a good initialization that might be available from information about the acquisition apparatus. In addition, no landmarks are needed for registration and it is thus possible to perform a markerless multimodal image registration, which is a common task in medical image analysis \cite{Maes97}. Valid similarity measures are, for instance, the sum of squared differences, normalized cross correlation, gradient correlation, difference image entropy or mutual information \cite{Penney98}. 
A second registration approach is to match extracted landmarks onto each other. These landmarks can for example be scale-invariant features extracted from the actual content of the images \cite{Se01, Lowe04, Brown07} or manually placed markers such as fluorescent beads that have to be identifiable in each of the acquired images \cite{Preibisch10, Kobitski15}. Given the landmarks of two complementary views, the registration task comes down to finding an optimal transformation to align the obtained point clouds of landmarks. This approach is computationally much more efficient than intensity-based registration techniques, as the amount of data that has to be aligned is significantly reduced. After having identified corresponding beads in both views, \eg, by local nearest neighbor-based bead descriptors \cite{Preibisch10} or by approaches like SIFT descriptors \cite{Lowe04, Liu14}, the transformation can be calculated iteratively using a random sample consensus (RANSAC) \cite{Preibisch10} or by using direct solutions like least-squares estimations \cite{Umeyama91}.

After having identified an appropriate transformation of the different images, some applications require a fusion of the registered image material in order to obtain a single high-quality image. The simplest method to accomplish a fusion of the images is to perform a local averaging of intensity values. In many cases, however, multiple images are acquired in order to compensate acquisition deficiencies of other views, \ie, it is desirable to only use the high-quality content of the images for the fused result image. More advanced techniques to fuse different images are, \eg, linear blending of the intensity values \cite{Tomer12}, and entropy-based blending \cite{Preibisch10} or wavelet-based fusion approaches where a high-quality image is formed by combining the best wavelet coefficients obtained from multiple views \cite{Li95, Rubio-Guivernau12, Tomer12}.

%%%%%%%%%%%%%%%%%%%%%%% SEED DETECTION %%%%%%%%%%%%%%%%%%%%%%%%
\subsubsection{Seed Point Detection}
\label{sec:chap1:SeedDetection}
Seed point detection represents an important component in many automated image analysis pipelines. The main intention is to obtain approximate locations of objects of interest, such as pedestrians or vehicles in surveillance videos \cite{Hu07, Bauml10}, specific organs or tissues in medical imaging applications \cite{Das09}, particles \cite{Genovesio06} or labeled nuclei and cells as discussed in this work \cite{Mikula10, Al-Kofahi10}. Object detection methods are often used as an additional preprocessing step to obtain a precise localization of objects of interest or regions of interest (ROI) that can be further investigated in proceeding analysis steps like seeded watershed, level-sets or graph-cuts segmentations \cite{Pinidiyaarachchi05, Li07b, Mikula10, Al-Kofahi10}. The major benefit of this preprocessing step is that complex and usually time consuming methods can be intelligently initialized or even guided to the correct image location \cite{Al-Kofahi10}. In addition to using seed point detection as a preprocessing step, several questions can be reasonably answered using the detected seed points, without the necessity of a subsequent segmentation of the objects of interest. For instance, counting the number of objects, tracking the spatio-temporal dynamics of objects of interest, extraction of registration landmarks or extracting average intensity levels within fixed window are common tasks that can be performed solely based on object centroids and their surrounding \cite{Lempitsky10, Sbalzarini05, Preibisch10}.

A common method for seed point detection is based on the second-order derivatives of the images using filters such as the Laplacian-of-Gaussian (LoG) filter (\cref{fig:chap1:AnalysisOperators}D) \cite{Marr80} and its approximations Difference-of-Gaussian (DoG) \cite{Preibisch10} and Difference-of-Mean (DoM) \cite{Grabner06}. These filters efficiently emphasize intensity changes such as edges or bright objects in images of arbitrary dimension at an adjustable scale. The actual seed points are then detected by searching for local extrema in the 8-neighborhood and the 26-neighborhood of the filtered 2D and 3D images, respectively. In order to detect differently sized objects, scale-space approaches use a digitally generated set of successively filtered images to simulate a scaling of the images and thus a scaling of the contained objects \cite{Lindeberg98, Lowe04}. The local extrema detection is then performed within neighboring scales, to determine the scale with the maximum response \cite{Lowe04}. Similarly, the wavelet transform can be used to preprocess the images and to detect seeds by multiscale products \cite{Olivo-Marin02}. Another approach that is frequently used is the concept of Euclidean distance maps (EDM) \cite{MaurerJr03}. In an EDM image (\cref{fig:chap1:AnalysisOperators}F) of a binarized input image (\cref{fig:chap1:AnalysisOperators}E) each foreground pixel value encodes the Euclidean distance to the closest background pixel. Objects of a certain size can be isolated by extracting the $h$-maxima of the EDM image \cite{Soille03}. A further example for efficient seed detection is the hotspot operator, which performs an analysis of concentric shells surrounding each pixel in an image \cite{Wassenberg10}. Finally, some more complex seed detection approaches are based on partial differential equations (PDE), such as shrinking level-sets \cite{Mikula11, Mosaliganti13} or gradient vector diffusion to identify centroids of objects of interest \cite{Li07}. An exemplary result of a seed detection algorithm is depicted in \cref{fig:chap1:AnalysisOperators}G.

%%%%%%%%%%%%%%%%%%%%%%% SEGMENTATION %%%%%%%%%%%%%%%%%%%%%%%%
\subsubsection{Segmentation}
\label{sec:chap1:Segmentation}
In various application scenarios similar to those listed in the previous section, the localization of an object of interest alone is not sufficient and region boundaries obtained by a segmentation algorithm potentially provide much more information about an object's properties. A detailed analysis of the entire region of an object can be used to extract precise statistical values, regional mean intensities, principal components, covered area and volume, geometrical properties and possible interaction sites between neighboring objects to name but a few. Thus, the goal of the segmentation step is to compute a binary image with a clear and reasonable separation of the objects of interest in the foreground regions of the image (\cref{fig:chap1:AnalysisOperators}E). If multiple objects of interest reside within a single image, it us usually desirable to identify the connected components within the binary image and to assign a unique integer label to each of the individual objects (\cref{fig:chap1:AnalysisOperators}H). Popular examples for the application of segmentation algorithms in microscopy images are ranging from segmenting fluorescent beads, nano particles or single molecules over labeled cell nuclei and cell membranes through to entire organs or organisms \cite{Preibisch10, Fent10, Keller08SC, Mosaliganti12, Kokel13, Khan14b}.

The easiest way to retrieve a segmented image of bright objects on a dark background is the use of a binary threshold, where all intensity values below the threshold are labeled as background and the remaining pixels get the foreground label assigned. In the case of dark objects on a bright background, the image can be inverted to use the same notion of background and foreground separation. Besides manual adjustment of the thresholds, various methods for the automatic threshold identification have been presented \cite{Sezgin04, Korzynska13}. Most methods for the automatic threshold adjustment are based on an analysis of the histogram mode distribution \cite{Otsu79} and can be further improved by additionally considering an image inherent noise model \cite{Pal92}. Khan \etal presented a feedback approach to optimize the threshold selection based on annotated ground truth images \cite{Khan13smps}. However, the application of these straightforward thresholding methods is limited to images with sufficiently high contrast, high signal-to-noise ratio and low object densities and may require pre- or postprocessing steps to improve the result quality. A further frequently used segmentation method is the watershed transform \cite{Beucher93, Vachier05}, which considers the image intensities as a topographic surface and performs a flooding of the surface starting at local intensity minima. The unseeded version of the algorithm usually strongly over-segments the image and requires additional tricks to correct these flaws \cite{Navlakha13, Amat14, Stegmaier16}. Based on provided seed points of a preprocessing step, segmentation algorithms can be targeted to regions of interest and perform a local extraction of the information present in the images. Examples for seeded segmentation techniques are region growing approaches where seed points are extended based on regional intensity statistics \cite{Gonzalez07}, seeded watershed segmentations that start from seeds instead of local intensity minima \cite{Beare06, Straehle12}, graph partitioning methods that may use seed points as shape priors \cite{Al-Kofahi10, Sharma12, Lou11a, Ciesielski13} or PDE-based methods that can use the provided seed points or seed regions for the initialization of level-set functions or parametric curves \cite{Yu09, Bourgine10, Li10b, Mikula11, Mosaliganti13}. Furthermore, gradient vector flow tracking methods that trace gradient vectors to local minima and use the respective attraction basins as final segmentation \cite{Li07b, Liu08} or model-based segmentation approaches that try to identify characteristic shapes in the images, for instance, based on elastic registration \cite{Kelemen99, Collins95} proofed to provide reasonable segmentation results for particular applications.
Recently, classification-based approaches to segmentation have become popular. Based on manually annotated input data, features like image derivatives, intensity distributions or local texture histograms can be used to train random forest classifiers \cite{Breiman01, Lou11b, Kaynig15, Gul-Mohammed14} or the features can be even learned automatically from the provided image material using deep learning approaches \cite{Jain10a, Ciresan12, Krizhevsky12}. The retrieved probability maps of the image classification can then be used to obtain the actual segmentation via thresholding and similar techniques. Obtained segments and the underlying grayscale image regions can be further analyzed, categorized and compared, \eg, using classification approaches as described in \cite{Reischl10}. In this work, the focus is put on the segmentation of grayscale images instead of directly analyzing multicolor images in the color space. Nevertheless, it should be noted that many additional algorithms exist for the special task of color image segmentation \cite{SiangTan11, Wang11a, Zhuang12}.

%%%%%%%%%%%%%%%%%%%%%%%% TRACKING %%%%%%%%%%%%%%%%%%%%%%%
\subsubsection{Tracking}
\label{sec:chap1:Tracking}
The tracking of objects is a frequently observed task in spatio-temporal analysis of image data. Generally, the goal of tracking is to determine reasonable one-to-one correspondences in subsequent video frames. Tracking approaches can be found in many different areas, such as surveillance video analysis \cite{Song08, Bowden05}, target tracking \cite{Mazor98, Blackman04}, pedestrian tracking \cite{Philomin00}, tracking of virus particles in a host cell \cite{Godinez09, Godinez11} or animal tracking for behavioral analyses \cite{Branson09, Straw10}. In the focus of this work, tracking of thousands of moving cell nuclei was the main application and it was required to quantitatively investigate cellular dynamics in early stages of embryonic development \cite{Melani07, McMahon10, Tomer12, Amat14}. The following presentation of tracking algorithms thus primarily focuses on valid approaches for the tracking of nuclei and cells in microscopy images.

Depending on the acquisition method, tracking is mostly performed on the results of a preceding segmentation procedure and can either be based on identified segment centroids or on the actual segments. The simplest way to obtain the linkage of the objects in adjoint frames is the identification of the nearest neighbor for each of the contained objects \cite{Meijering09}. The performance of these algorithms is reasonably well if the temporal sampling and the segmentation quality is high enough to guarantee unambiguous associations. In addition, these methods can be easily equipped with various correction heuristics and feature matching approaches to improve the quality \cite{Bao06, Meijering09}. The approach can be further improved using global data association schemes as demonstrated in \cite{Bise11, Kausler12} or global consistency constraints to compensate under-segmentation errors \cite{Schiegg13}. In addition, some recently presented new tracking approaches identify globally optimal solutions in over-segmented data out of multiple hypotheses \cite{Schiegg14} or perform the object associations based on structured learning \cite{Lou11b}.

Another set of tracking algorithms is based on non-parametric and parametric contour evolution \cite{Amat13a}. Using an initial shape approximation of the objects, the segmentation masks obtained from approaches like level-sets \cite{Li08a, Bourgine10} or parametric contours like Gaussian mixture models \cite{Tomer12, Amat14} are successively propagated to initialize subsequent time points. A related set of methods are state-space models, such as Kalman filters \cite{Kalman60, Bishop01} and particle filters \cite{Carranza11}. These predictor-corrector schemes estimate future object locations based on a given movement model and the previous object state \cite{Amat13a}. A further extension of this approach is given by interacting multiple model (IMM) approaches that additionally allow probabilistic transitions between different movement models based on the given observations \cite{Genovesio06, Li08a}. Successfully identified object associations can finally be used to extract spatio-temporal object dynamics, object fate maps and lineage trees in order to visualize and analyze the temporal behavior, ancestry and neighborhood relations of dynamic objects \cite{Meijering09, McMahon10, Tomer12, Amat14}.

%%%%%%%%%%%%%%%%%%%%%%%% BENCHMARKING %%%%%%%%%%%%%%%%%%%%%%%%%%%
\subsection{Benchmarking}
\label{sec:Introduction:Benchmarking}
To enable a thorough validation of newly developed algorithms and to quantitatively compare different algorithms for a certain task, the extensive use of benchmarking is inevitable. Various benchmarks have already been presented in the past and offer a great toolbox of testing environments for different application scenarios. The benchmarks presented in this section are restricted to the analysis operators relevant to this thesis.

The most straightforward but also most tedious way to obtain a benchmark dataset is the manual annotation of representative image regions. Examples for such benchmarks are manually annotated images for seed point detection \cite{Kvilekval10}, for segmentation \cite{Gelasca09, Coelho10} and for tracking purposes \cite{Kausler12, Maska14}. In order to circumvent the tedious manual labeling step, many of the recently presented benchmarks rely on simulated data with a directly available ground truth. For instance, in the field of fluorescence microscopy images, various approaches for the simulation of cells and cellular nuclei in 2D and 3D images have been presented \cite{Lehmussola08, Svoboda09, Ghaye12, Svoboda12}. In addition, locally plane-like structures such as cellular membranes can be simulated, \eg, using a Voronoi tessellation of randomly initialized seed points \cite{Mosaliganti12}. To additionally add authentic movement behavior of interacting objects, physical phenomena such as repulsion and adhesion interaction can be embedded into the simulations \cite{Macklin12}. A great selection of both manually annotated and simulated tracking benchmarks was introduced in the scope of the challenges for particle and cell tracking \cite{Chenouard14, Maska14}. Similarly, a set of simulated particle tracking benchmarks can be generated within ICY \cite{Chaumont12}. A simulated benchmark to assess the quality of multiview deconvolution approaches using spherical objects, point spread function simulations and different noise models is given in \cite{Preibisch14}. In this work, existing simulated benchmarks were extended, such that entire processing pipelines could be validated on a single comprehensive benchmark.

%%%%%%%%%%%%%%%%%%%%%%%% UNCERTAINTY ESTIMATION AND PROPAGATION %%%%%%%%%%%%%%%%%%%
\subsection{Prior Knowledge and Uncertainty}
\label{sec:Introduction:Uncertainty}
The use of prior knowledge is an important component of cutting-edge algorithms. For instance, the approaches described in \cite{Al-Kofahi10, Bourgine10} incorporate information about the number of expected objects and their associated physical size in order to adjust and improve a seed point detection algorithm. Similarly, to perform an image segmentation into foreground and background regions, properties like size, shape, geometry, intensity distributions and the like can be used to improve the algorithmic performance \cite{Fernandez10, Lou11, Khan14a}. A frequently used technique to incorporate this prior knowledge is based on shape penalization terms, \eg, by adding additional terms to the energy functional of a graph-cuts \cite{Lou11, Vu08} or a level-sets segmentation \cite{Leventon00} or by generalized Hough transforms that can detect arbitrary shapes \cite{Ballard81}. Information about size, shape and movement properties of objects can additionally be used to specify efficient tracking correction heuristics \cite{Bao06, Schiegg13, Amat14}.

Besides the use of prior knowledge in image analysis, a great potential lies in the estimation of uncertainties of the automatically produced results. Most automatic image analysis operators do not produce flawless results and contain a measurable degree of uncertainty that should ideally be considered by subsequent processing steps \cite{Santo04}. A general introduction to the expression of uncertainty inherent to visual sensor signals can be found in \cite{GUM, Wilson84}. On the pixel level, this uncertainty can be used to assess the information quality of a single pixel due to sensor imperfections or temperature dependence \cite{Santo04, Maji11}. Furthermore, a lot of previous work was performed to assess the localization uncertainty of geometric features such as corners, centroids, edges and lines in images \cite{Pal01, Chen09, Chen10, Anchini07} or to evaluate the quality of image registration algorithms \cite{Kybic10}. Besides many applications from the field of quality quantification, a part of research focuses on uncertainty quantification in areas such as face recognition and other biometric technologies \cite{Callejas06, Betta11, Betta12}, the tracking of shapes in ultrasound images \cite{Zhou05} or to evaluate the impact of noisy measurements on the validity of diagnosis results \cite{Mencattini10}. An uncertainty formulation based on fuzzy set theory has been employed to perform pixel-based classification tasks \cite{Boskovitz02, Tizhoosh05} or to detect specific structures in the image \cite{Radojevic14, Radojevic15}. A further possibility to exploit the uncertainty information is to optimize parameter values of a respective operator in a feedback fashion such that the outcome minimizes a previously defined optimization criterion as demonstrated in \cite{Khan13smps, Khan13Krusebook}. A further example is the improvement of a graph-based watershed implementation, where uncertainties are used to assess the influence of individual edges on the final segmentation outcome \cite{Straehle12}. In the present thesis, uncertainty is considered as the imperfect knowledge about the validity of a piece of extracted information produced by the respective image analysis operators and was used for efficient improvement heuristics to enhance image analysis pipelines \cite{Bouchon00, Stegmaier12a}.

%%%%%%%%%%%%%%%%%%%%%%%% SOFTWARE TOOLS %%%%%%%%%%%%%%%%%%%%%%%
\subsection{Available Software Solutions}
\label{sec:Introduction:Software}
Of course, a tremendous amount of both commercial and open-source software tools has already been presented to facilitate automatic evaluations of microscopy images and a subset of the most useful ones for analysis tasks similar to those considered in the present thesis is presented here \cite{Eliceiri12}. Many existing tools such as CellProfiler and ZFIQ \cite{Carpenter06, Liu08} were specifically developed for 2D applications. Even though they are not directly applicable to 3D analysis problems, they feature various processing operators that can be arranged in pipelines and executed in batch processing mode, to evaluate large amounts of 2D images and to get an impression of potentially useful algorithms.  
For both 2D and 3D image analysis tasks, several great open-source software tools have been presented that condense frequently used methods to analyze biological experiments in user-friendly graphical user interfaces (GUI) and are developed and advanced by active communities. Examples for these tools are ImageJ/Fiji \cite{Schindelin12}, ICY \cite{Chaumont12}, BioImage XD \cite{Kankaanpaa12} and Vaa3D \cite{Peng14}. A somewhat different approach is pursued by the ilastik application, which essentially uses learning approaches to accomplish image segmentation tasks \cite{Sommer11}.

Moreover, several high-quality commercial tools like Imaris (Bitplane AG), Volocity 3D (PerkinElmer Inc.), Amira 3D (FEI) and arivis vision (arivis AG) have been developed to facilitate in-depth automated analyses in the life sciences with sophisticated GUIs, special-purpose visualizations and statistical evaluation toolboxes. For rapid prototyping, MATLAB (The MathWorks Inc.) and the associated image processing toolbox represents a valuable tool with an easy-to-learn scripting language, comprehensive image analysis filters and powerful mathematical features for quantitative analyses and visualizations.

To facilitate new implementations, various software development kits (SDK) such as the Insight Toolkit (ITK) \cite{Ibanez05}, the Visualization Toolkit (VTK) \cite{Schroeder01}, OpenCV \cite{Bradski00a} and the Point Cloud Library (PCL) \cite{Rusu11} have been developed and are freely available to the community. Furthermore, BioView 3D \cite{Kvilekval10}, VisBio \cite{Rueden04} and ParaView \cite{Cedilnik06} are valuable tools for visualization purposes. Some tools that mainly focus on data analysis are, \eg, RapidMiner \cite{Mierswa06}, KNIME \cite{Berthold08}, WEKA \cite{Frank05} or the MATLAB toolbox Gait-CAD \cite{Mikut08Biosig}. A comparison of the newly developed XPIWIT software tool to existing software solutions is provided in \cref{tab:chap5:SoftwareComparison}.

%%%%%%%%%%%%%%%%%%%% OPEN QUESTIONS %%%%%%%%%%%%%%%%%%%%
\section{Open Questions}
Although previous research in image analysis offers a tremendous amount of elaborate methodology, there are several open questions in the domain of multidimensional image analysis that still are largely unsolved:
\begin{itemize}
	\item A feasible approach to improve the result quality of automated image analysis operators is the use of prior knowledge. However, available prior knowledge is mostly not sufficiently incorporated to image analysis operators, \ie, a great amount potentially usable extra information is wasted. Furthermore, there is no uniform approach to transform, embed and use the available prior knowledge to improve both existing and new algorithms.

	\item Although the sequential arrangement of processing operators to image analysis pipe\-lines is a broadly used concept, results propagated through the pipelines are mostly not assessed by the individual pipeline components with respect to their uncertainty. Errors of early processing steps tend to accumulate and may negatively affect the final result quality.
	
	\item Existing image analysis methods such as seed point detection and segmentation were mostly developed and tested in 2D scenarios or only work on small image datasets. Thus, they only offer limited applicability to large-scale image analysis tasks and either require to use fiercely resized datasets or immense processing capabilities to obtain quantitative results reliably and fast.
	
	\item Many benchmarks for the quality assessment of individual image analysis pipeline components have already been presented. However, comprehensive benchmarks that are crucial to consistently evaluate the quality of all involved steps on the same 3D image data are largely missing. In addition, manually annotated ground truth images may be biased and differently rated by different investigators, \ie, realistic simulated benchmarks are preferable to use in the first place.

	\item Existing approaches to multiview image fusion are usually based on the assumption that the temporal sampling is sufficiently high or that investigated objects are fixed. If these preconditions are missed, the quality of the fusion images is compromised and may heavily affect the quality of subsequent analysis steps.
	
	\item Most available tracking approaches heavily rely on a high-quality segmentation data as their input. However, most methods are not capable of dealing with merged objects resulting in a high number of avoidable linkage errors.
		
	\item Even though many usable software tools for multidimensional image analysis have been presented in the past, many applications only offer a limited applicability to terabyte-scale image datasets. Frequently observed problems arise from specific hardware requirements, closed implementations, insufficient parallelization, infeasible me\-mo\-ry demands or the inability of being executed in distributed computing environments.
	
	\item State-of-the-art microscopy techniques produce a tremendous amount of data with unprecedented quality. However, the investigations and feasible questions have been often compromised by the capabilities of automated processing routines or unattainable infrastructural demands. Hence, many scientific areas demand for sophisticated new approaches to investigate, proof and unveil hypotheses in multidimensional image data reliably, quantitatively and fast.
\end{itemize}

\section{Objectives and Thesis Outline}
Based on the open problems mentioned in the previous section, the central objectives of the present thesis are:

\begin{enumerate}
	\item To derive a new concept for assessing and exploiting the uncertainty produced by individual image analysis operators using prior knowledge. The framework should be able to propagate identified uncertainty values through an entire image analysis pipeline and thereby improve the result quality awareness of each processing operator or trigger adapted processing strategies.
	
	\item The development of accurate and fast seed detection and segmentation algorithms that are applicable to detect different geometric shapes in large multidimensional images.
	
	\item New validation benchmarks that closely resemble the real imaging scenarios and allow validating entire processing pipelines with a single consistent dataset.
	
	\item The analysis of seed detection, segmentation, multiview fusion and tracking techniques in different acquisition scenarios and an extension of the algorithms based on the developed uncertainty framework.
	
	\item The implementation of all developed methods in fast, efficient and user-friendly open-source software tools that are applicable to large-scale image analysis problems.
	
	\item The proof of successful operation and performance of all developed methods on large-scale datasets from the field of developmental biology.
\end{enumerate}

To generally improve the incorporation of prior knowledge into image analysis operators and to enable a quantitative assessment of the produced result quality, an extension to the classical image analysis pipeline concept with respect to uncertainty estimation, propagation and usage is presented in \cref{sec:chap2:UncertaintyFramework}. To advance the possibilities of automatic segmentation in terabyte-scale 3D microscopy images, \cref{sec:chap3:EfficientSegmentation} presents three new segmentation algorithms that were specifically developed to extract predefined geometrical classes of objects and feature highly efficient subroutines, a high degree of parallelization and a low memory footprint to obtain high-quality segmentation results reliably and fast. \cref{sec:chap4:AlgorithmEnhancement} illustrates how the uncertainty-based pipeline concept could be used for the enhancement of image analysis components such as seed point detection, segmentation, multiview fusion and tracking and introduces a comprehensive simulated validation benchmark dataset that was suitable for the quantitative assessment of all involved steps. All methods presented in this thesis were implemented as platform-independent open-source software tools as introduced in \cref{sec:chap5:Implementations}. In addition to the systematic validation of the new algorithms on synthetic benchmarks, the presented approaches were applied to large-scale microscopy data originating from experiments in developmental biology, which emphasizes the practical applicability of the presented methods (\cref{sec:chap6:Applications}). Finally, a summary of the accomplished tasks and an outlook on potential follow-up projects is provided in \cref{sec:chap7:Conclusions}.

\cleardoublepage

%% Uncertainty Framework

%\part{Methods (80)}

%%%%%%%%%%%%%%%%%%% UNCERTAINTY ESTIMATION AND PROPAGATION %%%%%%%%%%%%%%%%%%%
\chapter[Uncertainty Estimation and Propagation in Image Analysis Pipelines]{A New Concept for Uncertainty Estimation and Propagation in Image Analysis Pipelines} % \footnote{This chapter is partly based on a publication by Stegmaier \etal \cite{Stegmaier12a}.}
\label{sec:chap2:UncertaintyFramework}
Many automatically analyzable scientific questions are well-posed and offer a variety of information about the expected outcome \textit{a priori}. Although often being neglected, this prior knowledge can be systematically exploited to make automated analysis operations sensitive to a desired phenomenon or to evaluate extracted content with respect to this prior knowledge. Hence, the performance of processing operators could be greatly enhanced, for instance, by a more focused detection strategy and the direct information about the ambiguity inherent in the extracted data. Using natural language expressions to describe and rate observed phenomena, a new approach to extend image analysis pipelines by the quantification and propagation of uncertainty is introduced in this chapter. Initially, the general image analysis pipeline concept is introduced and an exemplary overview of potentially usable prior knowledge in the domain of microscopy images is provided. To transfer the available background information into a mathematical representation, the concept of fuzzy set membership is presented and it is demonstrated how the obtained feature classification can be used for an improved performance and consciousness of utilized processing operations, \eg, by filtering, propagating and using available information appropriately. The major emphasis is put on the automated analysis of images by means of dedicated processing pipelines. However, the developed concepts can theoretically be applied to similar data processing pipelines observed in other scientific or technical fields with few required adaptions.

%%%%%%%%%%%%%%%%%%% THE IMAGE ANALYSIS PIPELINE CONCEPT %%%%%%%%%%%%%%%%%%%%%
\section{The Image Analysis Pipeline Concept}
\label{sec:chap2:UncertaintyPropagation:PipelineConcept}
\begin{figure}[htb]
\begin{centering}
\includegraphics[width=1.0\textwidth]{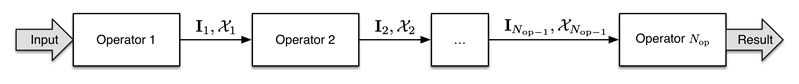}
\caption[General image analysis pipeline concept]{A common image analysis pipeline comprised of $N_{\text{op}}$ processing operators that are sequentially arranged. The performance of each operator is directly dependent on the quality of the input provided by its predecessor. Propagated information can be either images ($\mathbf{I_\ast}$), extracted features ($\mathcal{X}_\ast$) or both of them (adapted from \cite{Stegmaier16Arxiv}).}
\label{fig:chap2:OperatorPipelineReduced}
\end{centering}
\end{figure}

A general concept inherent to most image analysis pipelines is the use of multiple processing operators that perform a specialized task, \eg, to improve the image signal, to transform the image to a representation that can be more easily processed by downstream processing operators or to extract information from the images (\cref{sec:chap1:ImageAnalysis}). The basic layout of such an image analysis pipeline is depicted in \cref{fig:chap2:OperatorPipelineReduced}. Based on a raw input image, the $N_{\text{op}}$ sequentially connected operators receive either a processed version of the input image, extracted features or both from their preceding analysis operator. In \cref{fig:chap2:OperatorPipelineReduced}, extracted features that are passed between the pipeline components are denoted by a set $\mathcal{X}_i$ and processed images by $\mathbf{I}_i$, for the output of the $i$-th processing operator.
More formally, let $i \in \{1,...,N_{\text{op}}\}$ be the number of an operator generating a set $\mathcal{X}_i = \lbrace \mathbf{x}^{\top}_{i}[n] : n = 1,..., N_i \rbrace$ with $N_{i}$ data tuple vectors containing $N_{\text{f},i}$ features each. The generation of the output set $\mathcal{X}_i$ produced by operator $i$ is directly dependent on the input set $\mathcal{X}_{i-1}$ provided by its predecessor.

The output set $\mathcal{X}_i$ can also be written in matrix notation as an ${(N_i \times N_{\text{f},i})}$ matrix:
\begin{gather}
\mathbf{X}_{i} =
	\left( \begin{array}{c}
			\mathbf{x}^{\top}_{i}[1] \\
      \vdots \\
      \mathbf{x}^{\top}_{i}[N_i]
  \end{array} \right) 
	=      
  \left( \begin{array}{ccc}
			x_{i}[1,1]         & \cdots & x_{i}[1, N_{\text{f},i}] \\
			\vdots             &        & \vdots \\
			x_{i}[N_{i},1]     & \cdots & x_{i}[N_{i}, N_{\text{f},i}]
	\end{array} \right).
\label{eq:chap2:InitialOutputMatrix}
\end{gather}
The convention used throughout in this chapter is to store each data tuple in a separate row of the table and the individual features to the respective columns. For processing operators such as image preprocessing filters that do not produce any features, $\mathbf{X}_i$ is an empty matrix and only the processed image is passed to the next operator. Of course, image analysis pipelines can be much more complex than the linear arrangement introduced here. For instance, multiple input images may be given or an analysis operator may require extracted features from multiple sources to generate its results. For convenience, the uncertainty framework introduced in this chapter is illustrated on this basic pipeline structure. The developed concepts, however, are generally applicable and transferring them to more complex pipeline structures is straightforward. An extension of the general result matrix shown in \cref{eq:chap2:InitialOutputMatrix} by the quantified object uncertainties derived in \cref{sec:chap2:FuzzySetMembershipFunctions} is provided in \cref{sec:chap2:UncertaintyPropagation}.

\section{Identification of Suitable Prior Knowledge}
\label{sec:chap2:PriorKnowledgeIdentification}
Generally, prior knowledge can be obtained from literature, expert knowledge, experimental evidence or knowledge databases. In the field of microscopy, the instantly available visual representation of the observed data helps to facilitate the identification of recurring patterns, intensity properties or the appearance of objects contained in the images. An exemplary overview of potentially usable prior knowledge from microscopy images is given in the following listing, where italic terms in the brackets correspond to exemplary natural language expressions that could be used to characterize the respective features (adapted from \cite{Stegmaier16Arxiv}):
\begin{itemize}
	\item \textbf{Microscope}: Acquisition specific prior knowledge such as illumination conditions, detection path, image resolution, physical spacing of voxels, high-quality image regions, point spread function (PSF) or the detection path (\textit{Image quality decreases from XY to YZ}).
	
	\item \textbf{Intensity}: Signal dependent information like the intensity range, fluorescence properties, signal-to-noise ratio and global statistical properties of the image intensity values (\textit{Valid objects are brighter than XY}).
	
	\item \textbf{Localization}: Positional information of the objects or object properties in absolute image coordinates. Furthermore, localization of extracted properties or objects relative to each other can be used to define neighborhood relations (\textit{Object type XY only appears close to location YZ}).
	
	\item \textbf{Spatial Extent}: Object properties such as size, volume, principal components, convex hull extents or bounding volumes (\textit{Object type X is larger than Y but smaller than Z}).
	
	\item \textbf{Geometry}: Geometrical properties like dimensionality, symmetry, shape, proportions and relative localization of features within an object (\textit{Object type XY has a line-like shape with a central symmetry axis}).
	
	\item \textbf{Morphology}: Combination of intensity-based and geometrical properties, \eg, to link information about patterning, texture, structure and color to geometrical properties such as shape and symmetry (\textit{Object XY is spherical, bright and has a textured surface}).

	\item \textbf{Object Interaction}: Characterization of between-object properties like clustering, adhesion, repulsion, division or regional density changes (\textit{Object behavior XY rather appears in dense regions}).
	
	\item \textbf{Spatio-Temporal Coherence}: Dynamically changing quantities such as object growth, movement direction, speed, object appearance and disappearance (\textit{Object moves maximally XY pixels between two subsequent frames}).
\end{itemize}
The properties are listed in bottom-up order, going from the acquisition stage over the actual content of a single image through to the spatio-temporal comparison between time series of images. Of course, the listing can only provide a coarse overview of potentially usable prior knowledge and the suitable features have to be carefully selected to match the underlying image material and analysis problem. As introduced in the next sections, these linguistic expressions can be used to transform the prior knowledge of different sources to a consistent mathematical representation using the concept of fuzzy sets.

%%%%%%%%%%%%%%%%%%% PRIOR KNOWLEDGE IN IMAGE BASED BIOLOGICAL SCREENS %%%%%%%%%%%%%%%%%%%
\section{Prior Knowledge-based Uncertainty Quantification}
\label{sec:chap2:PriorKnowledge}
To fully make use of the prior knowledge presented in the previous section, it has to be transformed into a mathematical representation. A valuable tool for the formulation of prior knowledge in terms of mathematical equations is the use of fuzzy sets that have been introduced by Zadeh in 1965 \cite{Zadeh65}. Fuzzy sets extend the concept of classical sets by the incorporation of membership degree, \ie, allowing gradual association of elements to various sets. Linguistic terms are used to describe the fuzzy sets, which makes it a very intuitive approach that is based on human language \cite{Kiendl97, Bandemer93, Kroll16}. Considering the example of gene expression, the use of linguistic expressions enables to quantify vague biological properties like \textit{gene X is slightly over-expressed} and additionally allows soft transitions between expression states instead of using artificially generated sharp thresholds \cite{Windhager11}.

The concept of fuzzy sets has already been employed in several scientific applications. The technique has been used in the modeling of biological networks and biochemical pathways \cite{Le04, Aldridge09, Morris10}. Another example is the use of fuzzy clustering with prior knowledge-based distance measures \cite{Huang06, Tari09}. Moreover, a fuzzy formulation of prior knowledge has been used by Khan \etal to automatically adjust parameters of contrast enhancement and segmentation algorithms \cite{Khan13smps, Khan13Krusebook}. A similar formulation of background knowledge using fuzzy set theory is used in this work to assess the uncertainty of information that has been extracted by image analysis operators. An alternative to assess the result quality is, \eg, a Bayesian approach that tries to estimate posterior probabilities for all observed quantities \cite{Pohorille02}. However, in most cases it is not straightforward to calculate the respective probabilities based on the available data. Soft transitions between feature classifications as well as the possibility to have partial membership associations make fuzzy sets perfectly suited to model the imprecise knowledge observed in many scientific applications \cite{Bouchon00}. Furthermore, the simplicity of knowledge transformation and the intuitive concepts help, \eg, non-technical users to adjust algorithmic parameters on the foundation of acquired expert knowledge. Based on these benefits the main focus of this work is put on the use of fuzzy sets for the formulation of imprecise information, \ie, imprecisions of available prior knowledge are reflected by the width of the plateau regions with full membership of the respective fuzzy sets. Fuzzy set membership degrees (FSMD) are used to measure uncertainties, which means that uncertainties are considered proportional to the degree of believe a certain statement holds true or a linguistic term is fulfilled \cite{Bouchon00}. Consequently, high uncertainty corresponds to a low FSMD value to the considered fuzzy set and vice versa, when referring to low uncertainty high FSMD values are assumed. 

The fuzzy set framework used in this thesis is mainly employed for a coarse classification of information extracted from images based on vague prior knowledge, in order to decide if produced results match the expectation or may require further processing to correct flaws. In addition, the assessed uncertainty in form of FSMD values of extracted information to predefined fuzzy sets is shared between multiple pipeline components to allow more elaborate decisions in each step. It should be noted, though, that the use of the term uncertainty propagation does not refer to the statistical error propagation theory but only indicates that estimated FSMD values are shared between operators of a processing pipeline.

\subsection{Quantifying Prior Knowledge using Fuzzy Set Membership Functions}
\label{sec:chap2:FuzzySetMembershipFunctions}
\begin{figure}[htb]
	\begin{centering}
	\includegraphics[width=1.0\columnwidth]{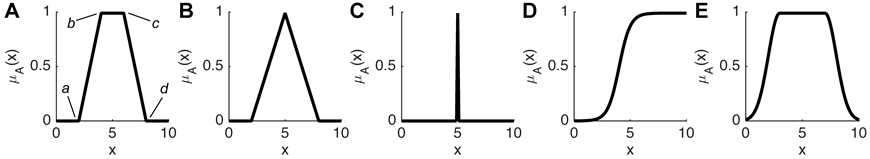}
	\caption[Different types of fuzzy set membership functions]{Exemplary membership functions that can be used to define fuzzy set membership degrees. The panels show a trapezoidal (A), a triangular (B), a singleton (C), a sigmoidal (D) and a Gaussian membership function (E). The triangular and singleton membership functions are special cases of the trapezoidal membership function and all of them can be parameterized using the parameters $a, b, c, d$ that define the locations of the respective transition regions (A).}
	\label{fig:chap2:MembershipFunctions}
	\end{centering}
\end{figure}

\begin{figure}[htb]
	\begin{centering}
	\includegraphics[width=1.0\textwidth]{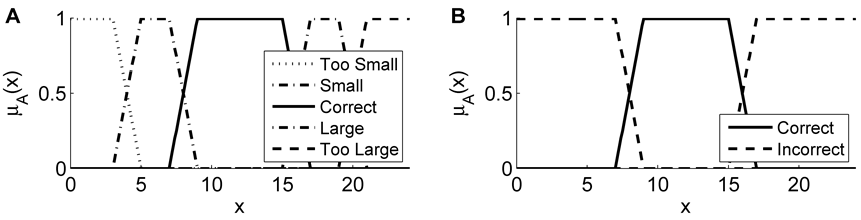}
	\caption[Feature space partitioning using trapezoidal membership functions]{Different possibilities to partition the input space of a feature $x$ using trapezoidal membership functions. In (A), each of the linguistic terms has a separate fuzzy set and (B) shows a reduced version with only three fuzzy sets that correspond to the desired class and its complement. In (B), different possibilities to summarize the correct objects arise. Besides solely restricting the class to the correct set as done in (B), the correct fuzzy set could be extended by the potentially useful classes (\textit{Small} and \textit{Large}). However, the appropriate formulation has to be chosen application dependent (adapted from \cite{Stegmaier16Arxiv}).}
	\label{fig:chap2:TrapezoidalMembershipFunctionComparison}
	\end{centering}
\end{figure}

In the classical sense, a set $\mathcal{C}$ is characterized by clearly defined boundaries that can be described with its associated characteristic function \cite{Bouchon00}:
\begin{equation}
\chi_\mathcal{C}(x) = \begin{cases} 1, x\in \mathcal{C} \\ 0, x\notin \mathcal{C},\end{cases}
\end{equation}
which means that an element can either be part of the set or not. An extension to this definition of a set are so called fuzzy sets, which allow a gradual membership of elements. Analogous to the characteristic function of a classical set, a fuzzy set $\mathcal{A}$ can be defined by its associated membership function (MBF) $\mu_{\mathcal{A}}:\mathcal{X}\mapsto [0,1]$ that maps each element of a universe of discourse $\mathcal{X}$ to a value in the range $[0,1]$ \cite{Zadeh65}. This assigned value in turn directly reflects the degree of membership (FSMD) of the respective element to the fuzzy set $\mathcal{A}$. In the special case $\mu_{\mathcal{A}}(x)=1$, the element $x$ is fully included in $\mathcal{A}$, whereas $\mu_{\mathcal{A}}(x)=0$ indicates that $x$ is not part of the fuzzy set $\mathcal{A}$ at all \cite{Bede13}. Examples for possible membership functions are singletons, triangular, rectangular, trapezoidal, sigmoidal and Gaussian MBFs (\cref{fig:chap2:MembershipFunctions}). The most common membership functions used in practice are trapezoidal membership functions, which can additionally be parameterized to model singletons, triangular and rectangular MBFs \cite{Kiendl97}. A trapezoidal membership function can be formulated as:
\begin{gather}
	\mu_\mathcal{A}(x,\pmb{\theta}) = \max\left( \min\left( \frac{x-a}{b-a}, 1, \frac{d-x}{d-c} \right), 0\right),
	\label{eq:chap2:TrapezoidalMBF}
\end{gather}
with the parameter vector $\pmb{\theta}=\left(a,b,c,d\right)^\top$ that is used to control the start and end points of the respective transition regions. An appropriate MBF type has to be selected on the basis of the underlying problem and mostly the use of a standard partition represents a reasonable choice \cite{Dubois00, Kroll16}. In a standard partition, maximally two neighboring fuzzy sets overlap and have non-zero membership values for a certain value of $x$. At the same time, the membership degrees for any input value of $x$ to the corresponding fuzzy sets have to sum up exactly to $1$ with plateau regions indicating that the respective element is fully included in the fuzzy set $\mathcal{A}$ (\cref{fig:chap2:TrapezoidalMembershipFunctionComparison}).

The soft transitions of an object's membership between different fuzzy sets are perfectly applicable to model imprecise experimental observations, \eg, obtained from biological experiments \cite{Bouchon00}. In particular, a fuzzy set $\mathcal{A}$ can be used to express the gradual validity of a linguistic term describing a certain phenomenon by one or more feature values. Considering, for instance, the detection of a specific kind of object it might make sense to use the size feature of the object as an indicator of its appropriateness. Linguistic terms could in this case for example be restricted to five possible outcomes, where the extracted feature ...
\begin{enumerate}
	\item ... perfectly matches the expected value (\textit{Correct}).
	\item ... is smaller than expected but might be useful (\textit{Small}).
	\item ... is larger than expected but might contain useful information (\textit{Large}).
	\item ... is too small and not useful (\textit{Too Small}, \eg, noise or artifacts).
	\item ... is too large and not useful (\textit{Too Large}, \eg, segments in background regions).
\end{enumerate}
The parameterization of the associated fuzzy sets can be performed on the basis of available prior knowledge and an exemplary standard partition for these five linguistic terms using trapezoidal membership functions is shown in \cref{fig:chap2:TrapezoidalMembershipFunctionComparison}A.

Considering an entire processing pipeline that may be comprised of $N_\text{op}$ processing operators, it is theoretically possible to specify multiple linguistic terms for each extracted feature of each processor $i\in \lbrace 1,...,N_{\text{op}}\rbrace$ if the respective prior knowledge is available and reasonably usable. Put more generally, a feature $f \in \lbrace 1,...,N_{\text{f},i} \rbrace$ produced by the image analysis operator $i$ is fuzzified by $N_\text{l}$ linguistic terms, which may be specified based on \textit{a priori} knowledge. The fuzzy set membership functions for image analysis operator $i$, feature $f$ and linguistic term $l \in \lbrace 1,...,N_\text{l} \rbrace$ are then denoted by $\mu_{\mathcal{A}_{ifl}}: \mathbb{R} \rightarrow [0,1]$ (\cref{fig:chap2:TrapezoidalMembershipFunctionComparison}A). A data tuple $\mathbf{x}_{i}[n]$ obtains the degree of membership $\mu_{\mathcal{A}_{ifl}}(x_i[n,f])$ to the fuzzy set $\mathcal{A}_{ifl}$ for each feature $f$ and each linguistic term $l$.

\begin{figure}[htb]
	\begin{centering}
	\includegraphics[width=1.0\columnwidth]{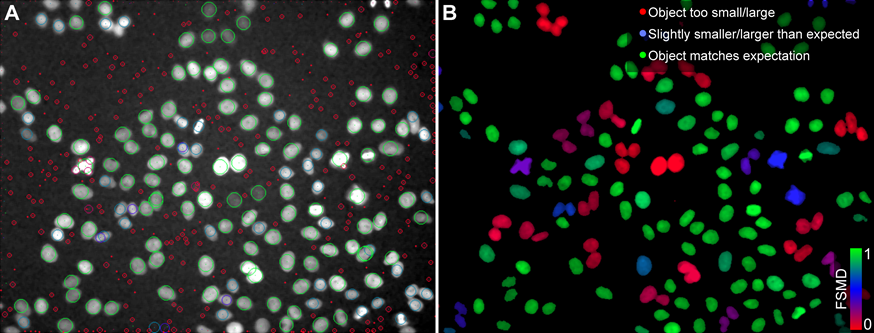}
	\caption[Visualization of fuzzy set membership degree for seed detection and segmentation]{Color-based visualization of the fuzzy set membership degree for a seed detection algorithm (A) and a segmentation algorithm (B). Red objects have low membership values due to being noise or merged regions, blue objects are potentially useful but may contain special object properties and green objects perfectly match the expectation  (adapted from \cite{Stegmaier12a}).}
	\label{fig:chap2:UncertaintyMaps}
	\end{centering}
\end{figure}

If only one outcome of the operators is of importance (\eg, case 1 in the above-mentioned example), it is possible to use only one linguistic term and aggregate all other cases by its complement, which is given by $\mu_{A_{if2}}(x) = 1-\mu_{A_{if1}}(x)$ (\cref{fig:chap2:TrapezoidalMembershipFunctionComparison}B). Of course, this formulation only allows to determine if an object belongs to case 1 or not. However, depending on the field of application it might also make sense to explicitly use the remaining cases, too. Considering for instance an image segmentation operator that erroneously extracts fused image regions due to under-segmentation. Although the information might not directly be useful, the information that some of the objects are classified to case 3, \ie, having a large region that potentially contains desired objects, might become important at later processing stages (\cref{sec:chap2:UncertaintyPropagation}). To facilitate the implementation this can be achieved by using only a single fuzzy set to determine the objects that need to be corrected (low FSMD values to case 1) and subsequently determine if they are too large or too small by a simple comparison to the boundaries of the \textit{Correct} fuzzy set. \cref{fig:chap2:UncertaintyMaps} shows exemplary results of two image analysis operators that were applied on a microscopy images showing stained nuclei of HeLa cells\footnote{Raw data taken from \url{http://www.cbi-tmhs.org/Dcelliq/files/051606_HeLaMCF10A_DMSO_1.rar}}. The degree of membership to the desired class of an appropriately detected cell is indicated by the color code following the fuzzy sets depicted in \cref{fig:chap2:TrapezoidalMembershipFunctionComparison}B. Objects with size feature values lying in the expected range are colored in green, potentially useful objects that are slightly smaller or larger than expected are colored in blue and finally noise segments as well as merged segments are colored in red. The five different fuzzy sets used to characterize the objects by their size (\cref{fig:chap2:TrapezoidalMembershipFunctionComparison}A) can be exploited, \eg, to split merged objects that are considered too large (case 3) or to filter out false positive detections in background regions (case 4).
The next section describes how the membership values of different features can be combined to form more complex object descriptors.

\subsection{Combination of Fuzzy Set Membership Functions}
\label{sec:chap2:MembershipFunctionCombination}
In many cases a specific phenomenon and the associated linguistic term are not only characterized by a single feature but by a combination of concurrently valid assumptions. Therefore, it is desirable to be able to combine the individually specified fuzzy sets to a single expression that yields a combined membership function for a particular linguistic term. For instance, if an extracted object has to be structured and rectangular to be a desired object, the corresponding FSMD values obtained for both features should be high, \ie, they should be combined using a fuzzy pendant to a logical conjunction \cite{Bouchon00}. A conjunction of $N_{\text{f},i}$ fuzzy set membership functions for linguistic term $l$ based on the multiplication operator can be formulated as:
\begin{gather}
	\mu_{\text{lc},\mathcal{A}_{il}}(\mathbf{x}_{i}[n]) = \prod_{f=1}^{N_{\text{f},i}} \mu_{A_{ifl}}(x_{i}[n,f]),
	\label{eq:chap2:MembershipFunctionCombination}
\end{gather}
where the index of $\mu_{\text{lc}}$ simply indicates that the membership function is a combination of multiple linguistic terms. Similarly, a conjunction can be defined using the minimum operator:
\begin{gather}
	\mu_{\text{lc},\mathcal{A}_{il}}(\mathbf{x}_{i}[n]) = \min_{f=1,...N_{\text{f},i}} \left( \mu_{A_{ifl}}(x_{i}[n,f]) \right).
%	\mu_{\text{lc},\mathcal{A}_{il}}(\mathbf{x}_{i}[n]) = \max_{f=1,...N_{\text{f},i}} \left( \mu_{A_{ifl}}(x_{i}[n,f]) \right).
	\label{eq:chap2:MembershipFunctionCombinationMin}
\end{gather}
Features that should be excluded from the multiplication (\cref{eq:chap2:MembershipFunctionCombination}) or the minimum calculation (\cref{eq:chap2:MembershipFunctionCombinationMin}) can be disabled by setting the corresponding MBFs to the constant value $1$ (identity element of the conjunction). The complement of the combined linguistic term is simply given by $1-\mu_{\text{lc},\mathcal{A}_{il}}(\mathbf{x}_{i}[n])$ as illustrated in \cref{fig:chap2:TrapezoidalMembershipFunctionComparison}B. In both formulations, high values of the combined functions are obtained only if all individual feature fuzzy sets yielded high values, \ie, a low value of the combined MBFs indicates that the investigated object strongly deviates from the expected properties of the respective linguistic term. The minimum-based formulation in \cref{eq:chap2:MembershipFunctionCombinationMin} is somewhat more informative than the multiplication, as a non-zero value directly represents the lowest FSMD value of the considered fuzzy sets. Of course, other logical combinations like disjunctions or implications are in principle possible but the conjunction expressions mentioned above were sufficient for the applications of this thesis.

%%%%%%%%%%%%%%%%%%% UNCERTAINTY ESTIMATION AND PROPAGATION %%%%%%%%%%%%%%%%%%%
\section{Uncertainty Propagation in Image Analysis Pipelines}
\label{sec:chap2:UncertaintyPropagation}
In the previous section it has been shown, how prior knowledge can be quantified using fuzzy set theory to characterize and assess the extracted information of a single processing operator according to defined linguistic terms. However, image analysis pipelines are mostly comprised of multiple processing operators that are serially linked and directly dependent on the results provided by previous calculations. Thus, an important extension is to pass the quantified uncertainty associated with each data tuple through the pipeline, \ie, performing a feed-forward propagation of the reliability of data shared between operators. For each data vector that is produced by operator $i$, the degree of membership to the respective fuzzy sets is calculated and the initial feature output matrix defined in \cref{eq:chap2:InitialOutputMatrix} is replaced by an augmented matrix of size ${(N_{i} \times (N_{\text{f},i}+N_{\text{l}}))}$ that is passed to the subsequent operators for further processing steps:
\begin{gather}
\mathbf{X}_{i} = 
	\left( \begin{array}{cccccc}
		x_{i}[1,1] 		 &	\cdots 	& x_{i}[1, N_{\text{f},i}] 		& \mu_{\text{lc},\mathcal{A}_{i1}}(\mathbf{x}_i[1]) & \cdots & \mu_{\text{lc},\mathcal{A}_{iN_{\text{l}}}}(\mathbf{x}_i[1]) \\
		\vdots 				 &	       	& \vdots 								& \vdots  								     & 				& \vdots  \\
		x_{i}[N_{i},1] &  \cdots  & x_{i}[N_{i}, N_{\text{f},i}] & \mu_{\text{lc},\mathcal{A}_{i1}}(\mathbf{x}_i[N_{i}]) & \cdots & \mu_{\text{lc},\mathcal{A}_{iN_{\text{l}}}}(\mathbf{x}_i[N_i]) \end{array} \right).
\label{eq:chap2:AugmentedOutputMatrix}
\end{gather}
In \cref{eq:chap2:AugmentedOutputMatrix}, it is theoretically allowed to add an arbitrary number of combined linguistic terms $N_{\text{l}}$ of additional criteria $\mu_{\text{lc},\mathcal{A}_{il}}(\mathbf{x}_i[n])$. If only a classification into correct/incorrect objects needs to be performed, one linguistic term is sufficient ($N_{\text{l}} = 1$). Note that the combined linguistic terms $\mu_\text{lc}$ obtain the entire feature vector $\mathbf{x}_i[n]$ of data tuple $n$ as an argument. This notation is chosen for brevity and it does not necessarily mean that each entry of $\mathbf{x}_i[n]$ contributes to the combined fuzzy set membership function. If the conjunction operator is used for the combined linguistic terms, the fuzzy sets of features that should not contribute to the combined linguistic term are simply set to a constant value of $1$.

%%%%%%%%%%%%%%%%%%%%%%% COPIED FROM CIWORKSHOP %%%%%%%%%%%%%%%%%%%%
\subsection{Uncertainty-based Object Rejection}
\label{sec:chap2:ObjectRejection}
The first application of the uncertainty framework is to filter the extracted output information $\mathcal{X} _i$ produced by an operator $i$ using thresholds $\alpha_{il} \in [0,1]$. Each processing operator $i$ that is part of the pipeline obtains an individual threshold parameter $\alpha_{il}$ for each linguistic term $l$, \eg, parameter $\alpha_{31}$ would be used to threshold the FSMD value of the first linguistic term of the third processing operator. According to the FSMD values $\mu_{\text{lc},\mathcal{A}_{il}}(\mathbf{x}_i[n])$ calculated for each data tuple $\mathbf{x}_i[n] \in \mathcal{X} _i$, $\mathbf{x}_i[n]$ is only passed to the next pipeline component if $\mu_{\text{lc},\mathcal{A}_{il}}(\mathbf{x}_i[n]) \geq \alpha_{il}$ for the membership to a desired set. The reduced set which serves as input for operator $i+1$ can be written as:
\begin{gather}
\tilde{\mathcal{X}}_{i} := \{\mathbf{x}_{i}[n] \in \mathcal{X}_{i}: \mu_{\text{lc},\mathcal{A}_{il}}(\mathbf{x}_{i}[n]) \geq \alpha_{il}, \forall l \in \lbrace 1,...,N_\text{l} \rbrace \}.
\end{gather}
This object suppression based on the respective FSMD values allows to easily filter out false positive detections based on predefined criteria of the expected object properties. For instance, to improve the sensitivity of an object detection algorithm it might be necessary to use parameters that produce many false positive detections in background regions but offer a high detection rate. Prior knowledge can be used to obtain a coarse classification of the objects and to reject inappropriate objects from further processing. Examples for the application of this filtering approach to improve the precision of a seed point detection and a segmentation algorithm are provided in \cref{sec:chap4:SeedDetection} and \cref{sec:chap4:Segmentation}, respectively.

\subsection{Extended Information Propagation to Compensate Operator Flaws}
\label{sec:chap2:InformationPropagation}
In a usual image analysis pipeline, each processing operator mostly only uses the input of its direct predecessor. However, in some cases it might be advantageous to the processing operators to be able to access results obtained by earlier processing steps. For instance, if an operator $i$ fails to sufficiently extract information from its provided input data (\eg, missing, merged or misshapen objects), it can inform downstream operators about these flawed results. The characterization of the produced results is again performed on the basis of FSMD values to the prior knowledge-based fuzzy sets obtained by the respective extracted objects. For extracted objects that exhibit a low degree of membership to the desired set, it might be beneficial to keep information $\mathbf{x}_{i-1}[n] \in \tilde{\mathcal{X}}_{i-1}$ from one of the previous steps that produced the uncertain information $\mathbf{x}_{i}[n] \in \mathcal{X} _i$. Using a second threshold $\beta_{il} \in \left[\alpha_{il}, 1\right]$ for each operator, the FSMD level below which the information of the previous steps should be additionally propagated can be controlled. Analogously to the forward threshold, each processing operator $i$ that is part of the pipeline obtains an individual threshold parameter $\beta_{il}$ for each linguistic term $l$, \eg, parameter $\beta_{42}$ would be used to propagate information based on the FSMD value of the second linguistic term of the fourth processing operator. More formally this means that instead of only forwarding the $\alpha_{il}$-filtered set $\tilde{\mathcal{X}}_{i} \subseteq \mathcal{X}_{i}$ produced by operator $i$ to operator $i+1$ a set
\begin{gather}
	\Omega_i = \tilde{\mathcal{X}}_{i} \cup \tilde{\Omega}_{i-1},
\label{eq:chap2:PropagationSet}
\end{gather}
with $i \geq 2$ and $\Omega_1 = \tilde{\mathcal{X}}_1$ is passed through the pipeline. In \cref{eq:chap2:PropagationSet}, $\tilde{\Omega}_{i-1}$ is defined as:
\begin{gather}
	\tilde{\Omega}_{i-1} = \{ \mathbf{x}_{i-1}[n] \in {\Omega}_{i-1}: \alpha_{il} \leq \mu_{\text{lc},\mathcal{A}_{il}}(\mathbf{x}_{i}[n]) < \beta_{il}, \mathbf{x}_{i}[n] \in \tilde{\mathcal{X}} _{i}, \forall l \in \lbrace 1,...,N_\text{l} \rbrace \},
\label{eq:chap2:ReducedPropagationSet}
\end{gather}
and thus represents the subset of elements in $\Omega_{i-1}$ which were not successfully transferred into useful information by operator $i$, \ie, elements $\mathbf{x}_{i-1}[n] \in {\Omega}_{i-1}$ that generated output $\mathbf{x}_{i}[n] \in \tilde{\mathcal{X}} _{i}$ with $\alpha_{il} \leq \mu_{\text{lc},\mathcal{A}_{il}}(\mathbf{x}_{i}[n]) < \beta_{il}$. Such elements characterize information of operator $i-1$ that might be useful in later steps to correct flawed results of operator $i$. Although there is in principle no limitation on how many predecessors provide their information in a feed-forward fashion, it mostly makes sense to only use a part of the available data depending on the information that is actually needed by downstream operators to avoid unnecessary communication overhead. In the current version of the framework it was assumed to be the responsibility of the respective processing operators to calculate $\Omega_i$ and to appropriately calculate the respective FSMD values. This approach was successfully applied to improve the outcome of a tracking algorithm that was allowed to use both extracted segmentation masks and seed points that produced the segments. In cases where the provided segmentation data was insufficient, the algorithm could fall back on seed point information of the penultimate processing operator to resolve tracking conflicts that originated from under-segmentation errors as described in \cref{sec:chap4:Tracking}. 

%%%%%%% 
\subsection{Resolve Ambiguities using Propagated Uncertainty}
\label{sec:chap2:ResolvingAmbiguities}
Next to filtering and propagating the information of the operators within the pipeline, the operators can explicitly make use of the provided uncertainty information to improve their results. Depending on the degree of uncertainty of provided information, parameters or even whole processing methods might be adapted if needed. Of course, the adaptions required by a particular algorithm cannot be generalized and only exemplary applications are introduced in this section. An important potential application is the correction of under-segmentation errors. The evaluated FSMD values can be used to identify segments that deviate from the expectation (larger than expected, corresponding to case 3 in \cref{sec:chap2:FuzzySetMembershipFunctions}) and the identified segments can then be separated using additional information such as seed points or distance maps. An exemplary segmentation algorithm based on adaptive thresholding was successfully equipped with the proposed extension as demonstrated in \cref{sec:chap4:Segmentation}.

Another promising application is the fusion of redundant information that might originate from different sensors, multichannel or multiview experiments or to combine the results of different algorithms applied on the same dataset. After having identified a likely grouping of the data using common classification or clustering approaches, the combination of redundant information within a class or cluster can be performed using the propagated FSMD values. Possible ways to combine the redundant information based on the respective FSMD values are a convex combination of the feature vectors with weights proportional to the FSMD values (\eg, for the fusion of redundant seed points) or to simply select the object that yielded the highest FSMD value (\eg, for multiview segmentation fusion). Using this approach, it was possible to fuse redundant data extracted from complementary multiview images as shown in \cref{sec:chap4:MultiviewFusion} and \cref{sec:chap6:TWANG}.

%%%%%% SUMMARY %%%%%%
\subsection{Improved Processing Pipeline by Uncertainty Propagation}
\begin{figure}[htb]
	\begin{centering}
  \includegraphics[width=1.0\textwidth]{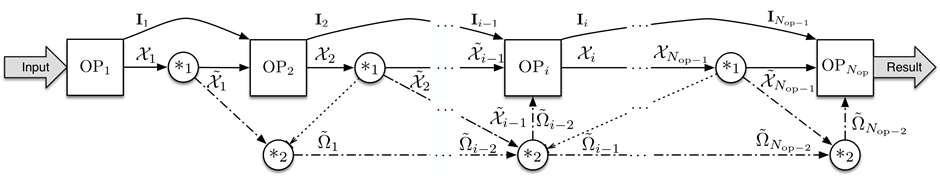}
	\caption[Extended image analysis pipeline concept with explicit uncertainty handling]{In most analysis pipelines each of the $N_{\text{op}}$ processing objects directly relies on the output $\mathcal{X} _i$ of its predecessor. Using thresholds the extracted output information of each operator can be filtered according to its uncertainty leading to a reduced set $\tilde{\mathcal{X}}_{i} \subseteq \mathcal{X} _i$ that is passed to the successor ($\ast_1$, threshold $\alpha_{il}$). If an operator $i$ produces uncertain outcome $\mathbf{x}_i[n] \in \mathcal{X} _i$ its successor may access preserved information from one of the previous steps. Therefore, each element $\mathbf{x}_{i-1}[n] \in \Omega_{i-1}$ that led to an uncertain result in processing object $i$ is kept in the set $\tilde{\Omega}_{i-1} \subseteq \Omega_{i-1}$ and additionally passed to the successor, \ie, a subset of $\Omega_i = \tilde{\mathcal{X}}_{i} \cup \tilde{\Omega}_{i-1}$ with $i \geq 2$ and $\Omega_1 = \tilde{\mathcal{X}}_1$ is used as input for operator $i+2$ ($\ast_2$, threshold $\beta_{il}$). Solid lines indicate the main information flow, dash-dotted lines the propagation of previously calculated results and dotted lines emphasize influence on the selection of propagated information. In addition to the flow of extracted features ($\mathcal{X}_\ast$, $\Omega_\ast$, $\mathcal{\tilde{X}}_\ast$, $\tilde{\Omega}_\ast$), the operators may pass processed images, image parts or simply forward the input image ($\mathbf{I}_\ast$) to the subsequent processing operator (adapted from \cite{Stegmaier16Arxiv}).}
	\label{fig:chap2:OperatorPipeline}
	\end{centering}
\end{figure}

The general scheme for the proposed uncertainty propagation framework is summarized in \cref{fig:chap2:OperatorPipeline}. All in all, the uncertainty framework described in this chapter encompasses filtering, propagation and exploitation of information produced by and between individual pipeline components based on fuzzy set membership functions.

The key ideas of the proposed framework are summarized in the following enumeration:
\begin{enumerate}
	\item Filter information based on their uncertainty, \ie, define uncertainty thresholds $\alpha_{il}$ for each pipeline component. To keep all extracted information the threshold is set to $\alpha_{il}=0$. Contrary, for $\alpha_{il}=1$, no uncertain information is passed to operator $i+1$ (\cref{sec:chap2:ObjectRejection}).
	\item Allow operators to fall back on information of penultimate processing steps if predecessors do not deliver good results. The parameter $\beta_{il}$ specifies the level up to which operator $i$ should preserve previous information. If $\beta_{il}=1$ all information of $i-1$ that produced an uncertain outcome is propagated to the successor $i+1$. If $\beta_{il} = \alpha_{il}$ only the information $\tilde{\mathcal{X}_i}$ produced by operator $i$ is propagated (\cref{sec:chap2:InformationPropagation}).
	\item Specifically handle information provided by previous operators based on the respective uncertainty, \eg, use the $\beta_{il}$-threshold to inform an algorithm which of its produced results might need to be revised or to resolve ambiguities when combining redundant information (\cref{sec:chap2:ResolvingAmbiguities}).
\end{enumerate}
The proposed uncertainty framework is exemplarily applied to some of the newly developed segmentation approaches of \cref{sec:chap3:EfficientSegmentation}. As described in \cref{sec:chap4:AlgorithmEnhancement}, an existing image analysis pipeline comprised of seed point detection, segmentation, multiview fusion and object tracking greatly benefited from the uncertainty-based improvements and the extensions only had a low impact on the processing times due to the lightweight adaptions of the propagated feature matrices.

 \cleardoublepage

%% Efficient Segmentation

%%%%%%%%%%%%%%%%%%%%%%% SEGMENTATION %%%%%%%%%%%%%%%%%%%%%%%%
\chapter[Efficient Segmentation in Multidimensional Image Data]{New Approaches for Efficient Segmentation in Multidimensional Image Data}
\label{sec:chap3:EfficientSegmentation}
In the previous chapter, the estimation and propagation of prior knowledge-based validity of extracted information was presented as a general technique to enhance arbitrary image analysis operators. Here, three newly developed segmentation algorithms are introduced, which on the one hand serve as exemplary processing operators to flesh out the general concepts presented in the last chapter and on the other hand represent efficient new solutions to three important problems observed in large-scale image analysis.

Arguably, image segmentation is one of the most important processing steps in numerous practical applications that basically tries to determine an image partition into a finite set of regions. In many cases the segmentation task consists of extracting objects that belong to a specific class of geometry and one way to achieve this segmentation goal is the explicit incorporation of prior knowledge into the developed algorithms. Examples for algorithms that can be guided to detect specific predefined shapes are the Hough-transform \cite{Ballard81}, shape priors for graph-cut and level-set segmentation approaches \cite{Freedman05, Mosaliganti08} or eigenvalue-based approaches that emphasize a specific type of geometry \cite{Frangi98, Antiga07}. Existing segmentation algorithms, however, were mostly developed for 2D applications and many mathematically elegant concepts and implementations were optimized with a focus on segmentation quality rather than computational efficiency. On the other hand, computationally efficient methods such as adaptive thresholding or blob detection methods mostly disregard prior knowledge and require sophisticated pre- and postprocessing procedures to correct erroneous results. To enable analyses of the tremendous amount of image data produced by state-of-the-art multidimensional imaging techniques, however, particularly the computational efficiency is a mandatory criterion that has to be met while at the same time providing high segmentation quality. Another common issue is the lack of thorough validation, \ie, the performance of new methods is often only tested on a particular type of data without consideration of possible image flaws like noise or out-of-focus effects.

\begin{figure}[htb]
\centerline{\includegraphics[width=\columnwidth]{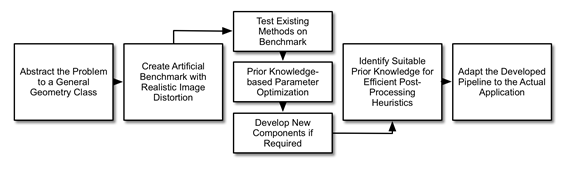}}
\caption[General procedure for segmentation algorithm development]{Schematic overview of a general procedure for the development of a new segmentation algorithm. Initially, the problem is analyzed and abstracted to create a realistic simulated benchmark. Flaws and benefits of existing methods are systematically identified based on the benchmark and prior knowledge-based parameter optimization. A new algorithm or pipeline is then developed using existing components where possible and new components where computational efficiency or quality improvements are required. Available prior knowledge is exploited to generate sophisticated postprocessing heuristics and finally the developed pipeline is adapted to the actual application scenario.} 
\label{fig:chap3:AlgorithmDevelopment}
\end{figure}

The approaches presented in this chapter were specifically designed with these limitations of existing algorithms in mind. Each algorithm was tailored to detect a specific geometric class of objects and thereby maximally exploits available prior knowledge to obtain accurate segmentation results. First, the problem was analyzed thoroughly and it was generalized to an abstract geometrical class. To be able to assess the performance of both existing and new algorithms under various image quality conditions, simulated benchmarks were used in combination with different image distortion techniques. This approach of extensive benchmarking throws light on the particular strengths and weaknesses of the algorithms and helps to narrow down reusable existing components and unsuitable components that require improvements with respect to time performance or segmentation quality. Parameterization of existing methods should of course be performed with the maximal amount of available prior knowledge to guarantee their optimal operation. Performance improvements were achieved by incorporating efficient processing operators and by heavily parallelizing the calculations where possible, to make large-scale analyses feasible. Developed pipelines were then further optimized by the explicit consideration of available prior knowledge in postprocessing heuristics to filter and improve extracted information (\cref{sec:chap4:AlgorithmEnhancement}). Finally, the pipelines were adapted and optimized to the particular application they were developed for (\cref{sec:chap6:Applications}). The general concept that was used for the algorithmic development of the new segmentation methods is sketched in \cref{fig:chap3:AlgorithmDevelopment}.

The first project considered in this thesis required a detection and extraction of the spinal cord in laterally oriented zebrafish embryos, \ie, more generally the extraction of elongated, possibly curved structures in 2D images including a characterization of the local environment (\cref{sec:chap3:SpinalCord}, \cite{Stegmaier14a}). Besides compensating the usual deficiencies of fluorescence microscopy images, the algorithm needed to be able to accurately extract comparable regions of interest from possibly thousands of 2D images on a single desktop computer. 

Furthermore, an efficient pipeline for the detection, segmentation and tracking of fluorescently labeled cellular nuclei and membranes in large-scale 3D+t microscopy images of various model organisms had to be developed. To be able to even analyze multiple terabytes of 3D+t image data, the algorithms had to be both highly efficient and reasonably accurate to validate biological hypotheses. Generally, these developed methods are suitable for a fast and accurate segmentation of spherical objects (\cref{sec:chap3:TWANG}, \cite{Stegmaier14}) and locally plane-like objects (\cref{sec:chap3:RACE}, \cite{Stegmaier16}) in large-scale 3D images. Although the algorithms were developed with these particular applications in mind, the thorough validation on more general synthetic benchmark datasets confirms their usability and robustness, and the general concepts can be easily adapted to similar problems in other scientific fields. As all presented segmentation approaches make use of a seed detection stage for an approximated object localization, two seed detection algorithms that are well suited for this purpose are presented in \cref{sec:chap3:SeedDetection}.

%%%%%%%%%%%%%%%%%% SEED DETECTION %%%%%%%%%%%%%%%%%%%
\section{Seed Point Detection}
\label{sec:chap3:SeedDetection}
In \cref{sec:chap1:SeedDetection}, the conceptual idea of a seed detection stage in the image analysis pipeline was presented. Although the projects considered in this thesis specifically required to extract fluorescently labeled nuclei and cells in 3D+t microscopy images, the described methods can in principal be used to detect (hyper-)spherical objects in arbitrary dimensions. Here, a blob detection using Laplacian-of-Gaussian (LoG) filtering \cite{Stegmaier14, Bartschat13} and a second approach based on the Euclidean distance transform \cite{Stegmaier16} are described. Although both approaches are established methods for seed point detection, some modifications were necessary to improve their applicability to large-scale datasets and to enhance their detection capabilities.

\subsection{Validation Benchmark}
\label{sec:chap3:SeedDetection:Benchmark}
\begin{figure}[!htb]
\centerline{\includegraphics[width=\columnwidth]{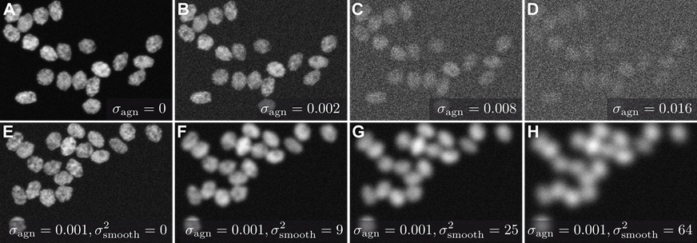}}
\caption[Benchmark Overview]{Maximum intensity projections of differently flawed benchmark images with different levels of additive Gaussian noise (A-D) and Gaussian smoothed versions with fixed noise level (E-H). While the low noise levels in (B) can usually be handled by seed detection algorithms, higher noise levels (C, D) cause a lower signal-to-noise ratio and thus the structures of interest become eventually almost indistinguishable from the background noise. Increasing the variance of the Gaussian smoothing causes initially separated objects to merge and sharp edges of the objects are successively blurred.}
\label{fig:chap3:SeedDetection:BenchmarkOverview}
\end{figure}

The validation of the presented seed detection algorithms was performed on a set of 3D benchmark images that contained simulated spherical objects as described by Svoboda \etal and were generated using the CytoPacq simulation toolbox (HL60 cell line, low SNR, 75\% clustering probability) \cite{Svoboda09}\footnote{\url{http://datasets.gryf.fi.muni.cz/cytometry2009/HL60_HighNoise_C75_3D_TIFF.zip}}. To additionally assess the performance of the seed detection algorithms in relation to the signal-to-noise ratio, a series of images with increasing zero-mean additive Gaussian noise levels with a standard deviation $\sigma_{\text{agn}}$ in the range of $[0.0, 0.128]$ have been used. Furthermore, out-of-focus effects were simulated using Gaussian blurred versions of the benchmark images with different variances $\sigma^2_{\text{smooth}}$ in the range of $[0.0,100.0]$ and a fixed zero-mean additive Gaussian noise with $\sigma_\text{agn}=0.001$.
The ranges of both image distortion approaches were empirically determined, where the largest values produced images with signal-to-noise ratios close to the visual detection limit \cite{Murphy12}. The specifiers \texttt{SBDS1}, \texttt{SBDS2} and \texttt{SBDS3} are used to refer to the raw, the noisy and the smoothed datasets of the benchmark, respectively. Exemplary benchmark images are shown in \cref{fig:chap3:SeedDetection:BenchmarkOverview} and the settings of all datasets are summarized in \app\cref{tab:Appendix:BenchmarkDatasets}.

%%%%%%%%%%%%%%%%%% SEED DETECTION LOG %%%%%%%%%%%%%%%%%%%
\subsection{Seed Detection using a Laplacian-of-Gaussian Maximum Projection} %\footnote{This section is partly based on a publication by Stegmaier \etal \cite{Stegmaier14}.}
\begin{figure}[tb]
\centerline{\includegraphics[width=\columnwidth]{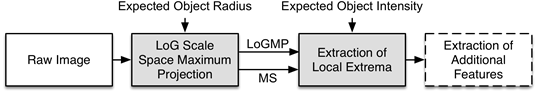}}
\caption[Processing steps of a LoG-based seed detection]{Pipeline overview of the LoG-based seed detection. Local extrema are extracted from a LoG scale-space maximum projection. Additional features such as the mean intensity in the vicinity of each local extremum can be used to further filter extracted seed points, \eg, to suppress false positive detections. Processing steps shaded in grey incorporate prior knowledge such as expected radius and intensity properties of the desired objects. The result filter is indicated by a dashed edge line.}
\label{fig:chap3:SeedDetection:PipelineOverview}
\end{figure}

\begin{figure}[tb]
\centerline{\includegraphics[width=\columnwidth]{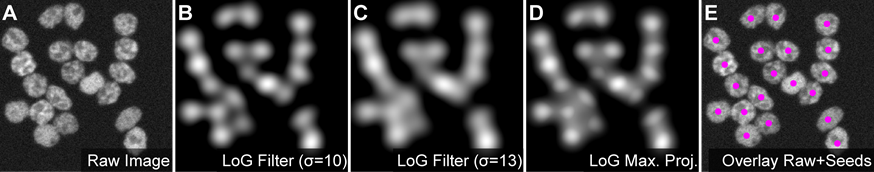}}
\caption[Processing steps of a LoG scale-space maximum projection for seed detection]{Visualization of the processing steps that are required to obtain a Laplacian-of-Gaussian (LoG) scale-space maximum projection for seed detection. (A) Original image showing spherical objects of the \texttt{SBDS1} benchmark images, (B, C) LoG filtered image ($\sigma=10$, $\sigma=13$), (D) LoG scale-space maximum projection ($\sigma_{\text{min}}=10$, $\sigma_{\text{max}}=13$, $\sigma_{\text{step}}=1$) and (E) a superposition of detected seeds on the raw image. As all contained objects had a similar radius, a single scale would have been sufficient ($\sigma=10$) and the additional scale as well as the maximum projection are only visualized for illustration purposes (adapted from \cite{Stegmaier14}).}
\label{fig:chap3:SeedDetection:LoGSSMP}
\end{figure}

\label{sec:chap3:SeedDetection:LoGSSMP}
The Laplacian-of-Gaussian filter and its approximations are well suited for the detection of spherical objects in images of arbitrary dimension (\cref{sec:chap1:SeedDetection}). To be able to detect differently sized objects, LoG-filtered images at different scales were used \cite{Lindeberg98, Al-Kofahi10} in combination with efficient recursive Gaussian filtering approaches that enabled seed detection even on large volumetric images \cite{Deriche90}. The LoG-filters were scale-normalized, \ie, the 3D filtering considered the physical spacing of the voxels to compensate the reduced axial resolution which is common in 3D microscopy and tomography approaches:
\begin{gather}
	\mathbf{I}^{\text{LoG}}(\mathbf{x}, \sigma) = \mathbf{I}^{\text{raw}}(\mathbf{x}) \ast \left(\sigma^3 \sum^{3}_{i=1}\frac{\partial^2 \mathbf{G}(\mathbf{x}, \sigma)}{\partial x_{i}^2}\right),
\end{gather}
with $\mathbf{I}^{\text{raw}}$ being the input image, $\mathbf{x} = (x_1,x_2,x_3)^{\top}$ being the voxel coordinates, $\mathbf{G}(\mathbf{x},\sigma)$ representing a Gaussian kernel with standard deviation $\sigma$ and $\ast$ being the convolution operator. For performance reasons, however, the automatic scale selection performed in \cite{Al-Kofahi10} was skipped and the calculations were restricted to a set of scales that was determined based on the radius of the objects of interest. Using the relationship that the radius of detected objects is $r=\sqrt{2}\cdot \sigma$, the appropriate set of scales could be determined by \textit{a priori} knowledge about the investigated biological content of the images \cite{Al-Kofahi10}. Using the radius of the smallest object of interest $r_{\text{min}}$ yields $\sigma_\text{min}=r_\text{min}/\sqrt{2}$. Analogously, the radius of the largest object of interest $r_\text{max}$ can be used to obtain $\sigma_\text{max}=r_\text{max} / \sqrt{2}$. If $\sigma_\text{min}$ is too small ($\sigma_\text{min} \ll r_\text{min} / \sqrt{2}$), an increased amount of false positives is detected. Analogously, if the parameter $\sigma_\text{max}$ is set too large ($\sigma_\text{max} \gg r_\text{max} / \sqrt{2}$), nearby objects may become indiscernible, which potentially causes missed detections. In most common scale-space approaches, points of interest are detected in the local surrounding of the investigated point within neighboring scales and possibly even at different octaves \cite{Lowe04}. Due to the enormous image size of several gigabytes confronted in the discussed application example, it was not feasible to keep multiple image stacks of the scale-space simultaneously in memory. Therefore, an iteratively calculated LoG scale-space maximum projection with a predefined discrete step size was used instead:
\begin{gather}
	\mathbf{I}^{\text{LoGMP}}(\mathbf{x}, \sigma_{\text{min}}, \sigma_{\text{max}}) = \max\limits_{\sigma_{\text{min}} \leq \sigma \leq \sigma_{\text{max}}} \mathbf{I}^{\text{LoG}}(\mathbf{x},\sigma).
\end{gather}
For each pixel the scale that yielded the maximum value was additionally stored:
\begin{gather}
	\mathbf{I}^{\text{MS}}(\mathbf{x}, \sigma_{\text{min}}, \sigma_{\text{max}}) = \argmax\limits_{\sigma_{\text{min}} \leq \sigma \leq \sigma_{\text{max}}} \mathbf{I}^{\text{LoG}}(\mathbf{x},\sigma).
	\label{eq:chap3:SeedDetection:MaxScale}
\end{gather}
This information could be exploited for down-steam processing operators as it provided an initial size approximation of the object about to be extracted. Finally, the actual seed extraction from the LoG scale-space maximum projection image came down to a simple local extrema detection in the direct neighborhood of each pixel (8-neighborhood and 26-neighborhood for 2D and 3D images, respectively). To discard detections in background regions, an intensity threshold was used to only consider seed points whose intensities were above the specified level. The intensity threshold can either be determined automatically, \eg, using Otsu's method \cite{Otsu79} or via manual adjustment as described in \cref{sec:chap5:FuzzyParameterAdjustment}. Instead of only using the single pixel value at the seed location, the robustness of the intensity-based filtering approach could be increased by extracting the average intensity in a small window centered at the seed location and by performing the threshold operation on this average intensity value. An efficient way to estimate the window mean intensity at a seed point location was to extract the intensity values from an integral image-based mean filtered raw image \cite{Viola01}. A schematic overview of the introduced pipeline is depicted in \cref{fig:chap3:SeedDetection:PipelineOverview}.
The processing steps needed to obtain the LoG scale-space maximum projection for an exemplary 2D image and an overlay of the detected seed points on the raw input image are shown in \cref{fig:chap3:SeedDetection:LoGSSMP}.

In \cref{sec:chap4:SeedDetection:SeedFusion} and \cref{sec:chap4:SeedDetection:UncertaintyIntegration}, respectively, an extension of the described LoG-based seed detection algorithm with respect to object fusion and uncertainty-based object rejection is introduced, which significantly improved the detection quality and additionally provided information about the validity of extracted seed points.

%%%%%%%%%%%%%%%%%% SEED DETECTION EUCLIDEAN DISTANCE MAP %%%%%%%%%%%%%%%%%%%
\subsection{Seed Detection using Thresholding and Euclidean Distance Maps} % \footnote{This section is partly based on a publication by Stegmaier \etal \cite{Stegmaier15}.}
\begin{figure}[!htb]
\centerline{\includegraphics[width=\columnwidth]{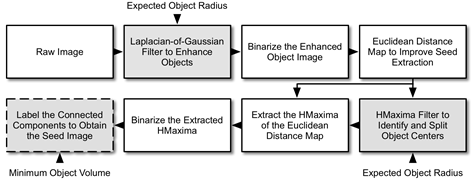}}
\caption[Processing steps of an EDM-based seed detection]{Schematic pipeline overview of the EDM-based seed detection. Seeds are extracted from the distance map of a preprocessed, binarized object image. Operators shaded in gray incorporate prior knowledge about minimum and maximum expected object radii and volume to improve the detection of desired objects and to discard noise detections. The result filter is indicated by a dashed edge line.}
\label{fig:chap3:SeedDetection:PipelineSchematicEDM}
\end{figure}

\begin{figure}[!htb]
\centerline{\includegraphics[width=\columnwidth]{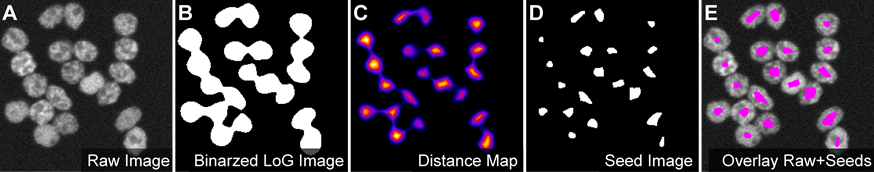}}
\caption[Euclidean distance map-based seed point detection]{The most important steps to detect spherical objects in the \texttt{SBDS1} dataset using an Euclidean distance map-based approach. Based on the raw input image (A), the binarized Laplacian-of-Gaussian of the spherical objects (B) was used to obtain initial object markers. Closely located objects that might be connected due to the binarization could be efficiently split using an Euclidean distance map on the binary images (C). The actual seeds were then extracted from the distance map images by a connected components labeling of the thresholded distance map or the extracted $h$-maxima of the distance map (D). An overlay of the detected seed segments and the raw image is depicted in (E). This approach is able to separate touching objects as long as the binary region connecting two objects is thinner than the actual object. If this condition is not met, overlapping objects may be detected as a single object.}
\label{fig:chap3:SeedDetection:EDMSeeds}
\end{figure}

\label{sec:chap3:SeedDetection:EDM}
A second approach to detect seed points for (hyper-)spherical objects like fluorescently labeled nuclei in multidimensional images is the use of Euclidean distance maps (EDM) of a binarized input image and is schematically illustrated in \cref{fig:chap3:SeedDetection:PipelineSchematicEDM}. The binarization can be achieved using a manually determined threshold, simple adaptive thresholding methods \cite{Otsu79} or more sophisticated techniques such as graph-cuts \cite{Al-Kofahi10}. However, this approach only yielded reasonable results if the individual objects could still be unambiguously distinguished in the binarized image. To improve the quality of the binarization, to reduce image noise and to possibly emphasize (hyper-) spherical objects of a certain size, a Laplacian-of-Gaussian-based preprocessing was used (\cref{sec:chap3:SeedDetection:LoGSSMP}). The possibly preprocessed and binarized image (\cref{fig:chap3:SeedDetection:EDMSeeds}B) was then used to calculate an Euclidean distance map (\cref{fig:chap3:SeedDetection:EDMSeeds}C), where each pixel value of the foreground corresponded to the distance in pixel to the closest background pixel of the binary input image. To guarantee a fast calculation of the EDM, a linear-time algorithm as described by Maurer \etal was used \cite{MaurerJr03}. In the ideal case, the distance was maximal in the center regions of the desired objects and the objects were extracted by subtracting an EDM with suppressed $h$-maxima from the initial EDM image \cite{Soille03}. The $h$ parameter suppressed local maxima of a height less than $h$ in the input image, \ie, subtracting the EDM with suppressed $h$-maxima from the initial EDM image yielded only the suppressed $h$-maxima. The labeled connected components of the binarized $h$-maxima delivered the actual seed segments (\cref{fig:chap3:SeedDetection:EDMSeeds}D, E). In cases where the objects are more or less equal in size, the $h$-maxima suppression could be substituted with a binary threshold, which reduced the processing times of the detection. Moreover, if the objects in the original image were well separated, a performance improvement could also be achieved by performing the entire seed detection step on a downscaled version of the image.

\subsection{Validation}
\subsubsection{Datasets and Compared Algorithms}
The performance assessment of the presented seed detection methods was performed on the simulated benchmark datasets \texttt{SBDS1}, \texttt{SBDS2} and \texttt{SBDS3} as described in \cref{sec:chap3:SeedDetection:Benchmark}.
In addition to the LoG-based (LoGSM) and the EDM-based (EDM) seed detection methods, another common seed detection approach based on the connected components of an eroded binary image that has been binarized using Otsu's method (EOCC, \cite{Otsu79}) was used for the comparison. Based on the available ground truth, the true positive (TP), false positive (FP) and false negative (FN) detections were counted to calculate recall, precision and the F-Score for each of the methods. True negatives (TN) are in principle all background pixels that were correctly not detect as seed points. However, this information is irrelevant for the seed detection step and was thus omitted. Furthermore, the average Euclidean distance to the correct object centroids as well as the processing time and voxels per second were measured. Additional details on the calculation of the utilized performance measures and the employed testing platform can be found in \cref{sec:Appendix:SeedDetectionPerformance} and \cref{sec:Appendix:EvaluationPlatform}. Implementation details are provided in \cref{sec:chap5:Implementations}.

\subsubsection{Performance Evaluation}
The seed detection performance of the investigated algorithms obtained on the \texttt{SBDS1} dataset is listed in \cref{tab:chap3:SeedDetection:DetectionPerformance}. All methods provided reasonable results on the benchmark images, where the best values were almost equally shared by the LoGSM and the EDM algorithm. The simpler EOCC approach missed a few objects and additionally produced a small number of false positive detections ($7.0\%$ lower F-Score compared to LoGSM and EDM). Regarding the processing times, EDM was the fastest approach, closely followed by the LoGSM method. As the EOCC method was both slower and more inaccurate compared to LoGSM and EDM, the latter methods should be favored. The slower execution time of EOCC was caused by the non-linear morphological erosion operation that cannot be implemented as efficient as a linear LoG filter used in LoGSM or the linear time distance transform performed by EDM.
\begin{table}[htb]
\begin{center}
%\rowcolors{2}{white}{gray!25}
\resizebox{\textwidth}{!}{
\begin{tabular}{lccccccccc} %{p{3.5cm}p{6cm}p{5cm}}
\toprule
\textbf{Method}& \textbf{TP} & \textbf{FP} & \textbf{FN} & \textbf{Recall} & \textbf{Precision} & \textbf{F-Score} & \textbf{Dist.} & \textbf{Time (s)} & \textbf{KVox./s} \\
\midrule
EOCC & 18.87 & 1.70 & 1.13 & 0.94 & 0.92 & 0.93 & 2.66 & 1.70 & 2031.76 \\
LoGSM & 19.93 & \textbf{0.00} & 0.07 & \textbf{1.00} & \textbf{1.00} & \textbf{1.00} & 1.37 & 1.34 & 2577.61 \\
EDM & \textbf{20.00} & 0.10 & \textbf{0.00} & \textbf{1.00} & \textbf{1.00} & \textbf{1.00} & \textbf{1.26} & \textbf{1.19} & \textbf{2902.52} \\
\bottomrule
\end{tabular}}
\caption[Quantitative assessment of the seed detection performance]{Quantitative performance assessment of the seed points detected in the \texttt{SBDS1} dataset by an eroded connected components of an Otsu thresholded image (EOCC), a LoG-based multi-scale approach (LoGSM) and an Euclidean distance map-based seed detection method (EDM). Each test image contained exactly $20$ objects and the criteria are true positives (TP), false positives (FP), false negatives (FN), recall, precision, F-Score and the distance to the reference (Dist., smaller values are better). Moreover, the achieved average time performance to process a single image was measured in seconds (smaller values are better) and voxels per second (larger values are better). EDM and LoGSM both provided excellent seed detection quality ($\text{F-Score} = 1$) with EDM being faster. As EOCC provided the worst results and was at the same time the slowest method due to non-linear filter operations, it should not be used in practice if the other methods are available.}
\label{tab:chap3:SeedDetection:DetectionPerformance}
\end{center}
\end{table}

\begin{figure}[htb]
\centerline{\includegraphics[width=\columnwidth]{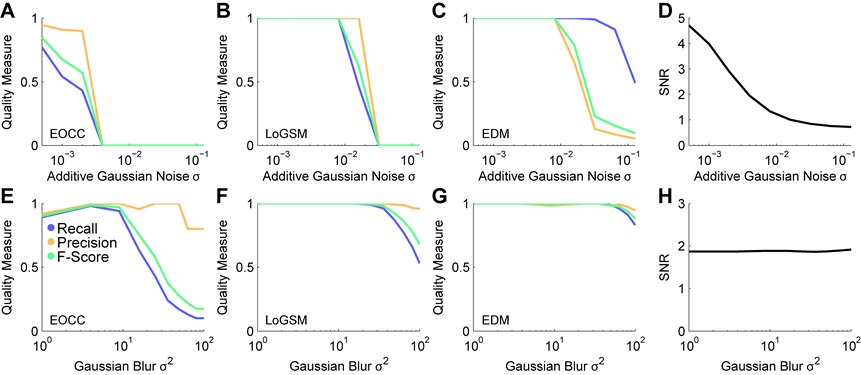}}
\caption[Assessment of the seed detection performance for different noise levels]{Assessment of the seed detection performance of EOCC (A, E), LoGSM (B, F) and EDM (C, G) for different noise levels (A-C) and different Gaussian blur levels (E-G) using the simulated 3D benchmark datasets \texttt{SBDS2} and \texttt{SBDS3}, respectively. The performance measures recall, precision and F-Score are plotted against the additive Gaussian noise level parameter $\sigma_{\text{agn}}$ (A-C)  and the Gaussian blur variance $\sigma^2_{\text{smooth}}$ (E-G). The influence of the noise and blur level on the signal-to-noise ratio (SNR) of the images is shown in (D, H).}
\label{fig:chap3:SeedDetection:NoiseLevelInfluence}
\end{figure}

The influence of different levels of both noise and blur of the images is depicted in \cref{fig:chap3:SeedDetection:NoiseLevelInfluence}. EOCC was very sensitive to image noise, as a global binary threshold is always a trade-off between false positive detections and merged objects for noisy images. Hence, the segmentation quality was already affected at the lowest noise levels where a higher amount of merged objects appeared and noise detections increased. At a noise level of $\sigma_\text{agn}=0.004$ no valid segments were found anymore by EOCC. This behavior also explains the improved performance of EOCC for small blur levels, as the Gaussian smoothing serves as a low-pass filter that efficiently suppresses image noise in this case. However, as soon as the variance of the smoothing filter surpassed a value of $10$, the detection quality again dropped instantly, as most of the real objects were missed. The variation in the precision values shown in \cref{fig:chap3:SeedDetection:NoiseLevelInfluence}E are caused by the adaptive threshold that automatically adjusts its level based on the image content. In contrast to this, LoGSM and EDM both produced reasonable results up to a noise level of $\sigma_\text{agn}=0.01$. Moreover, LoGSM and EDM reliably detected the centroids of the objects up to smoothing variances of $30$ and $50$, respectively. This is not surprising, however, as the LoGSM method essentially exploits image smoothing to detect the actual object, \ie, an increased blur level up to $\sigma^2_\text{smooth}=30$ is tolerated by the algorithm. Similarly, the EDM approach worked properly as long as the rough contours of the binarized objects were still distinguishable ($\sigma^2_\text{smooth} \leq 50$), which still resulted in unambiguous intensity maxima of the distance map in the center of the desired objects.

Further validation experiments of the seed detection methods were performed in combination with the uncertainty-based improvements and are covered in \cref{sec:chap4:SeedDetection:SeedFusion}. Additionally, LoGSM and EDM were used to provide seed points for the segmentation approaches discussed in the remainder of this chapter. Thus, the validation of the respective segmentation algorithms also represents an indirect evaluation of the seed detection performance.

\subsection{Discussion}
In this section, two seed detection methods for the automatic localization of seed points or seed segments in 3D images were presented. Both approaches are well suited to reliably extract the spatial coordinates of desired objects in both noisy and dense environments and the concepts presented for 3D images here can be easily extended to higher or lower dimensions, if necessary. Both approaches were highly robust with respect to increased image noise levels and blurred object contours. On the considered benchmark images, EDM provided the best results. However, due to the single-threaded implementation of the $h$-maxima filter, the performance of this operator decreased for large images on multicore machines, where the LoGSM operator still provides fast processing speed due to its parallel implementation \cite{Stegmaier14}.
The appropriate method has to be chosen according to the field of application. The LoG-based seed detection solely provides the object centroids, which is perfectly suited for object tracking or region of interest determination (\cref{sec:chap3:TWANG}). Contrary, the EDM-based method provides small seed segments that potentially range over several slices. For instance, for the fusion of 2D segments to 3D objects as performed by the proposed segmentation algorithm for locally plane-like structures (\cref{sec:chap3:RACE}). There it was desirable to have seeds spanning over several slices, as this instantly allowed to identify objects that corresponded to a certain segment. All segments that were intersecting with the seed segment could then be merged. Furthermore, having seeds that ranged over multiple slices lowered the chance of missing an object due to a single misdetection of a segment where the seed was located. However, both approaches had difficulties with elongated objects. The LoG-based seed detection tended to extract multiple seeds per desired object at the tips of elongated objects. Furthermore, the elongation occasionally caused intensity plateaus that did not have a single intensity maximum (see \cref{sec:chap4:SeedDetection}). Similarly, the EDM-based method tended to detect multiple $h$-maxima for a single elongated object. The detection behavior could be adjusted by the $h$ parameter of the $h$-maximum detection that directly depends on the radius of objects of interest by suppressing local maxima of a height smaller or equal to $h$ \cite{Soille03}. Values of $h$ that were smaller than the minimum object radius potentially yielded too many local minima, \ie, split objects. On the other hand, high values of $h$ larger than the minimal object radius tended to merge objects and rejected small objects. If shape and size of the desired objects were comparable, it was usually easy to determine suitable parameter values by manually measuring the radius of the smallest objects that still needed to be detected. However, if object shapes and sizes heavily varied, adjusting this parameter always had to be a trade-off that minimized missing, merged and split objects.
In \cref{sec:chap4:SeedDetection:SeedFusion}, an exemplary extension of the LoG-based seed detection method using the uncertainty and fusion framework (\cref{sec:chap2:UncertaintyFramework}) is presented. This extension helped to significantly improve the segmentation results in the case of redundant or missing detections.

%%%%%%%%%%%%%%%%%% SPINAL CORD SEGMENTATION %%%%%%%%%%%%%%%%%%%
%\newpage
\section[Accurate Extraction and Comparison of Elongated Shapes in 2D Images]{A New Algorithm for the Accurate Extraction and Comparison of Elongated Shapes in 2D Images} % \footnote{This section is partly based on a publication by Stegmaier \etal \cite{Stegmaier14a}.}
\label{sec:chap3:SpinalCord}
Elongated objects are an ubiquitous pattern found in many different image-based experiments. In technical applications, for instance, rods, ropes, cables or pipelines exhibit elongated structures that may need to be compared for automatic quality assurance. In material science, it might be required to quantify cracks in rocks, ice or any other material to draw conclusions about the stability of the investigated substance. Especially, numerous biological tissues such as bones, blood vessels, hair, axons, microtubuli to name but a few, exhibit elongated shapes. Besides extracting the actual elongated objects it is often required to obtain additional information about the local environment, \eg, to quantify dirt particles, neighboring molecules or cells, and basically any other entity that might be present in the close vicinity of the elongated object. In the particular application discussed in this thesis, the analysis was motivated by a toxicological screen of fluorescently labeled neurons in the spinal cord of laterally oriented zebrafish embryos that exhibited both a line-like arrangement of interneurons and blob-like sensory- and motoneurons located above and below the interneurons, respectively (\cref{sec:chap6:SpinalCord}, \cite{Stegmaier14a}). The specific requirements were not met by any existing software solution and the algorithm was therefore newly developed from scratch. In addition to robustly detecting and extracting the respective objects from the images, a standardized way to compare the extracted information of different images was derived.

\subsection{Validation Benchmark}
\label{sec:chap3:SpinalCord:Benchmark}
\begin{figure}[htb]
\centerline{\includegraphics[width=\columnwidth]{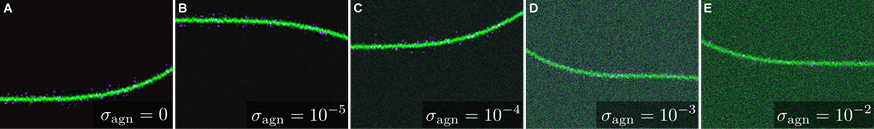}}
\caption[Exemplary images of the \texttt{SBDL} benchmark dataset]{Exemplary images of the \texttt{SBDL} benchmark dataset with different noise levels $\sigma_{\text{agn}}$. The elongated object is located in the green channel and the spherical objects in the vicinity of the elongated object occupy the other two channels. The standard deviation of the additive Gaussian noise increases from $\sigma_\text{agn}=0$ (A) up to $\sigma_\text{agn}=10^{-2}$ (E). Increased noise levels cause a successively decreasing signal-to-noise ratio and may affect the quality of automatic analysis algorithms.}
\label{fig:chap3:SpinalCord:BenchmarkOverview}
\end{figure}

To validate the proposed object extraction algorithm, a simulated benchmark that models both an elongated object and randomly distributed small objects in the vicinity of the elongated object was implemented (\texttt{SBDL}). The object types were rendered into two different channels, where channel 1 contained the small objects and channel 2 contained a single elongated object. A random parameterization of a third order polynomial that was sampled over the entire width of the desired output image was used to generate the rough shape of the elongated object. Each ordinate value of the regression curve was further disrupted with a normally distributed random offset. The ground truth image was then formed by iterating over all abscissa values and by adding a vertical bar centered at the ordinate value to form the ground truth image. The ground truth image was subsequently convolved with a Gaussian kernel ($\sigma=3$) and Poisson noise as well as zero-mean additive Gaussian noise were added to simulate acquisition deficiencies. Finally, an additional channel was added, which contained randomly sized roundish objects that were randomly distributed along the regression curve. Similar to the elongated object, the small object image was convolved with a Gaussian kernel ($\sigma=1$) and disrupted using the same noise settings as described before. To generate images with different noise levels, the standard deviation of the zero-mean additive Gaussian noise $\sigma_{\text{agn}}$ was varied in a range $[10^{-5}, 10^{-1}]$. Exemplary images with different levels of additive Gaussian noise are shown in \cref{fig:chap3:SpinalCord:BenchmarkOverview} and the benchmark settings are summarized in \app\cref{tab:Appendix:BenchmarkDatasets}. Details on the testing platform used for the time performance estimation are provided in \cref{sec:Appendix:EvaluationPlatform}.
For each noise level, $100$ different test images were generated and their quality was assessed with respect to the regression curve quality, the correlation between ground truth images and the extracted image regions and the detection quality of the small objects that were randomly distributed along the elongated object.

\subsection{Algorithmic Design}
\label{sec:chap3:SpinalCord:AlgorithmicDesign}
\begin{figure}[htb]
\centerline{\includegraphics[width=0.8\columnwidth]{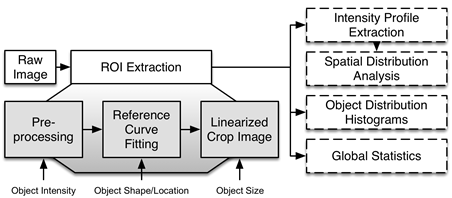}}
\caption[Schematic overview of the elongated shapes segmentation]{Schematic overview of the processing pipeline. The pipeline essentially consists of a regression-based region of interest (ROI) extraction and a straightening of the images to obtain comparable image regions. The actual comparison of extracted crop regions was performed using intensity profiles, spatial distribution analysis, object distribution histograms and general statistical properties of the images. Operators shaded in gray involve prior knowledge about the intensity distribution, object location, the expected size as well as the shape of the object of interest. Result filters are indicated by the dashed edge lines (adapted from \cite{Stegmaier14a}).}
\label{fig:chap3:SpinalCord:PipelineOverview}
\end{figure}

In order to quantify and identify alterations in the spatial appearance of the different object types with respect to an elongated reference object and to detect variations of the elongated objects themselves, a new image analysis pipeline was developed. The basic steps of the algorithm are summarized in \cref{fig:chap3:SpinalCord:PipelineOverview} and the individual steps are introduced in the following sections.

\subsubsection{Region of Interest Extraction}
\label{sec:chap3:SpinalCord:ROIExtraction}
\begin{figure}[!htb]
\centerline{\includegraphics[width=\columnwidth]{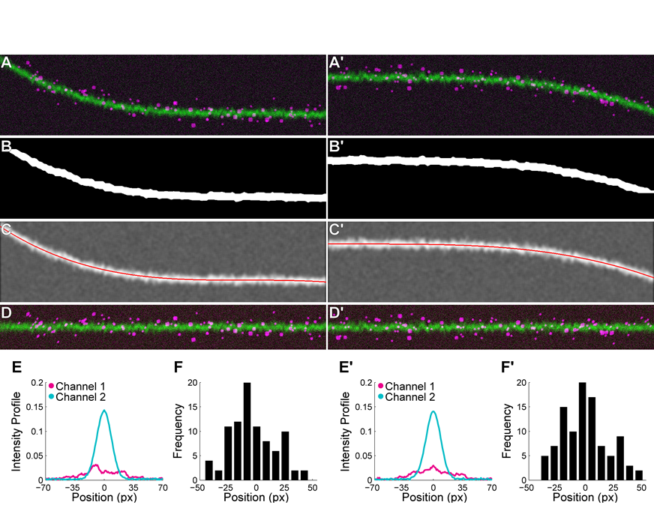}}
\caption[Processing steps and outcome of the elongated shape extraction]{Processing steps and exemplary outcome of the elongated shape extraction algorithm applied on two simulated benchmark images. The raw input image containing the elongated object (A, A') was binarized (B, B') and the remaining foreground pixels were used to perform a polynomial regression in a least squares sense (C, C'). A straightened region of interest was extracted along the regression curve using a radius $r=70$ to obtain a comparable image representation (D, D'). Based on the straightened crop region, intensity profiles (E, E', \cref{sec:chap3:SpinalCord:IntensityProfiles}) and object distribution histograms (F, F', \cref{sec:chap3:SpinalCord:ObjectDistributionHistograms}) were extracted with respect to the regression curve for further quantitative analysis. The abscissa of both the intensity profiles and the object distribution histograms are depicted relative to the regression curve in pixel (px) with the regression curve being located at the zero mark. Note that the spherical objects are randomly positioned around the zero mark and have a random size, \ie, the intensity profiles and histograms may differ substantially.}
\label{fig:chap3:SpinalCord:BenchmarkExample}
\end{figure}

The first step was to extract a comparable image region from the original image $\mathbf{I}^{\text{raw}}$ with a resolution of $N_x \times N_y$ (\cref{fig:chap3:SpinalCord:BenchmarkExample}). If multiple channels were available, this step was performed solely on the channel that contained the elongated object. An initial saturation of the lower $q_{\text{l}}$-quantile and the upper $q_{\text{u}}$-quantile of the intensity values was applied to the image for outlier rejection. Furthermore, an additional Gaussian smoothing with standard deviation of $\sigma_{\text{smooth}}$ was used to preserve only the low frequency content. Subsequently, Otsu's method was used to obtain a binarized image from the denoised, smoothed input image (\cref{fig:chap3:SpinalCord:BenchmarkExample}B) \cite{Otsu79}. A morphological closing was then used to further improve the segmented regions and to obtain a dense connected binary object \cite{Soille03}. If the thresholded image contained multiple connected components, the desired region was selected using prior knowledge about the expected size, location or intensity distribution \cite{Stegmaier14a}. If multiple features need to be combined for this regional characterization, the fuzzy set-based uncertainty framework presented in the previous chapter represents a perfectly suited approach for this task. To model the elongated binary object present in the image a polynomial regression curve was fit to the dichotomized image by using the image space coordinates of the foreground pixels of the object of interest and by minimizing the residual sum of squares. This was essentially accomplished by solving:
\begin{equation}
	\hat{\pmb\theta} = \argmin\limits_{\mathbf{\pmb\theta} \in \mathbb{R}^{N_p}} \sum_{(i_x,i_y) \in \mathcal{P}_{\text{fg}}} \left( i_y - f(i_x, \pmb\theta)\right)^2,
\label{eq:chap3:SpinalCord:RegressionSSD}
\end{equation}
where $f$ is an arbitrary regression function with unknown $N_p$-dimensional parameter vector $\pmb{\theta}$ and $\hat{\pmb\theta}$ is the optimal parameter vector that minimizes the residual sum of squares. Furthermore, $\mathcal{P}_{\text{fg}}$ is the set of foreground pixel coordinates that were used to perform the regression. For the analyses performed in this thesis, a third order polynomial $f(x, \pmb{\theta}) = \theta_1 \cdot x^3 + \theta_2 \cdot x^2 + \theta_3 \cdot x + \theta_4$, with parameter vector $\pmb{\theta} = (\theta_1,\theta_2,\theta_3,\theta_4)^\top$, was used to sufficiently describe the possible deformations of the imaged specimens. The selection of an appropriate regression model is important to obtain comparable results and has to be matched to the image content. Panels A-C in \cref{fig:chap3:SpinalCord:BenchmarkExample} summarize the individual processing steps performed for the region of interest extraction. The final linearized crop image $\mathbf{I}^{\text{cr}}_c$ was formed by extracting a band of predefined radius $r$ below and above the regression curve along the elongated object from the original image. If multiple channels were available, the extraction was performed separately for each of the channels using exactly the same regression curve and extraction radius for all image channels to preserve their colocalization. \cref{fig:chap3:SpinalCord:BenchmarkExample}D shows the final result of the ROI extraction on two exemplary benchmark images and has the dimensions $N_x \times (2 \cdot r + 1)$. Global statistical properties of the images were directly extracted from these linearized cropped images. Here, the min, max, mean, median and variance values from the intensity values of each of the channels $c = 1,...,N_c$ were extracted for the cropped image to characterize the environment of the elongated object within a distance of $r$ ($\texttt{gi}^{\text{min}}_c$, $\texttt{gi}^{\text{max}}_c$, $\texttt{gi}^{\text{mean}}_c$, $\texttt{gi}^{\text{\text{med}}}_c$, $\texttt{gi}^{\text{var}}_c$).

%\subsubsection{Rotation Correction}
%In cases where the elongated objects exhibited a strong curvature, a %rotation correction could be used to further optimize the region of %interest along the regression curve. For the correction the deviation %angle between the tangent to the regression curve and the horizontal image %axis (range: $\left[-\frac{\pi}{2},\frac{\pi}{2}\right]$) at each location %$i_x = 1,...,N_x-1$ was used according to
%\begin{equation}
%\alpha_{i_x} = \frac{180}{\pi} \cdot \operatorname{atan}\left( %\frac{\mathbf{r}_y\left[i_x+1\right]-\mathbf{r}_y\left[i_x\right]}{\mathbf%{r}_x\left[i_x+1\right] -\mathbf{r}_x\left[i_x\right]} \right),
%\label{eq:RotationCorrectionAngle}
%\end{equation}
%where $\mathbf{r}_x\left[i_x\right]$ and $\mathbf{r}_y\left[i_x\right]$ %represent abscissa and ordinate of the regression curve at each pixel %location $i_x$, respectively. Using a counter-clockwise rotation of the %original image at all pixel locations $i_x$ with the rotation angles %$\alpha_{i_x}$, the extraction of the respective bands was performed in %the individual rotated images. The tangent angle was estimated using %forward differences as noted in \cref{eq:RotationCorrectionAngle}. At the %rightmost border pixels where the forward difference scheme becomes %infeasible, an analogous calculation scheme using backward differences was %used.

\subsubsection{Intensity Profile Extraction}
\label{sec:chap3:SpinalCord:IntensityProfiles}
To quantitatively assess the distribution of different object types located near or on the elongated object and also to quantify the elongated object itself, the horizontal mean values of each row of pixels in each channel of the cropped image were used to form an average profile vector:
\begin{equation}
	\mathbf{p}_c\left[i_y\right] = \frac{1}{N_x} \sum_{i_x=1}^{N_x} \mathbf{I}^{\text{cr}}_c\left(i_x,i_y\right),
\label{eq:chap3:SpinalCord:IntensityProfile}
\end{equation}
for all rows $i_y=1,...,2r + 1$ and all channels $c = 1, ..., N_c$.
This vector $\mathbf{p}_c$, which is referred to as intensity profile in the remainder of this thesis, can be plotted in a cartesian coordinate system as depicted in \cref{fig:chap3:SpinalCord:BenchmarkExample}E. Another visualization possibility is to plot multiple intensity profiles of different observations vertically aligned to a single heat map image, where each row contains one profile and the intensity levels indicate the height of the respective intensity profiles. This representation was used to easily identify alterations in the intensity distributions of the investigated images \cite{Stegmaier14a}. Similar to the global statistical features, single value features like min, max, mean, median and variance of the intensity profiles were extracted ( $\texttt{ip}^{\text{min}}_c$, $\texttt{ip}^{\text{max}}_c$, $\texttt{ip}^{\text{med}}_c$, $\texttt{ip}^{\text{mean}}_c$, $\texttt{ip}^{\sigma}_c$).
Moreover, calculating the area under the curve yielded a single feature value that quantified the overall signal intensity in the respective channel according to:
\begin{eqnarray}
	\texttt{ip}^{\text{auc}}_c = \sum_{i_y=1}^{(2r+1)} \mathbf{p}_c[i_y].
\end{eqnarray}
This feature could be refined by using prior knowledge of the spatial location of different objects to constrain the summation only to a desired range. For instance, if a certain class of objects in a specific channel is known to reside above the regression curve, the summation can be adjusted accordingly, in order to use only the relevant information to calculate the respective feature value. In addition to the purely intensity-based features, the characteristics of the intensity profiles, such as peak locations, peak maxima values and the peak width were extracted ($\texttt{plp}_c$, $\texttt{pvp}_c$, $\texttt{pwp}_c$).

\subsubsection{Spatial Distribution Analysis}
\label{sec:chap3:SpinalCord:SpatialDistributionAnalysis}
The spatial distribution analysis was used to compare the intensity profiles of different experimental groups. In many applications the results obtained for a new experiment need to be compared to a control group $\mathbf{\bar{p}}^{\text{ctrl}}_c$ with known properties. For instance, the control group could be based on manual investigation. After having defined the control group, all respective intensity profiles of the control group images were combined to a reference intensity profile $\mathbf{\bar{p}}^{\text{ctrl}}_c$. The reference profile was specified by the average of all control profiles with the associated standard deviation series $\pmb{\sigma}^{\text{ctrl}}_c$ as described in \cite{Wolf06, Stegmaier14a}. This ensemble averaged reference profile could subsequently be used to identify differences to new observations, \eg, to identify quality deficiencies in automated quality assessment or treatment effects in toxicological experiments.
Technically, the comparison was performed by calculating the arithmetic mean of the normalized distance $\mathbf{nd}_c$, which reflects the variance normalized mean deviation of an arbitrary intensity profile $\mathbf{p}_{c}$ from the averaged reference $\mathbf{\bar{p}}^{\text{ctrl}}_c$:
\begin{equation}
	\mu^{\text{nd}}_c = \frac{1}{N_y} \sum_{i_y=1}^{N_y} \mathbf{nd}_c[i_y] = \frac{1}{N_y} \sum_{i_y=1}^{N_y} \frac{\left| \mathbf{p}_c\left[i_y\right] - \mathbf{\bar{p}}^{\text{ctrl}}_c\left[i_y\right] \right|}{\pmb{\sigma}^{\text{ctrl}}_c\left[i_y\right]}.
	\label{eq:chap3:SpinalCord:NormDeviation}
\end{equation}
The feature $\mu^{\text{nd}}_c$ measures the deviation of a new observation from the averaged reference profile of channel $c$ as n-fold variance. An average value below one indicates that on average all sample points of the investigated intensity profile deviate at most one standard deviation from the averaged reference curve. Hence, a value of zero corresponds to perfect agreement of the compared observation and the reference profile.

\subsubsection{Object Distribution Histograms}
\label{sec:chap3:SpinalCord:ObjectDistributionHistograms}
Another option that yielded more precise information about the spatial distribution of individual objects was the direct extraction of object locations relative to the calculated regression model. This was obtained using blob detection algorithms as described in \cref{sec:chap3:SeedDetection}. Extracted object locations were plotted using an object distribution histogram that summarized their frequency with respect to the relative spatial offset from the elongated object. Two exemplary histograms obtained on two validation benchmark images are depicted in \cref{fig:chap3:SpinalCord:BenchmarkExample}F. Additional features to quantify the object occurrence were given by the respective mode and maxima values of the histogram. Of course, this approach can be performed independently on multiple colocalized channels if available.

\subsection{Validation}
\label{sec:chap3:SpinalCord:Validation}

\subsubsection{Datasets and Compared Algorithms}
\label{sec:chap3:SpinalCord:Datasets}
The validation of the proposed segmentation approach was performed on the simulated benchmark dataset \texttt{SBDL} as described in \cref{sec:chap3:SpinalCord:Benchmark}. The regression-based approach (REG) was parameterized with a standard deviation of $\sigma=10$, a lower quantile saturation of $q_{\text{l}}=0.01$, an upper quantile saturation of $q_{\text{u}}=0.99$ and a region extraction radius $r=70$. The parameters were empirically determined, such that image noise was suppressed and the object of interest could be unambiguously segmented. The radius was chosen such that the entire region of interest is tightly contained in the cropped image. REG was compared to an alternative approach consisting of a median filter-based noise suppression ($5 \times 5$ neighborhood), a binary threshold using Otsu's method and an extraction of the elongated object based on the gradient information of the binary image (GRAD).

\subsubsection{Performance Evaluation}
\label{sec:chap3:SpinalCord:PerformanceEvaluation}
\begin{figure}[bt]
\centerline{\includegraphics[width=\columnwidth]{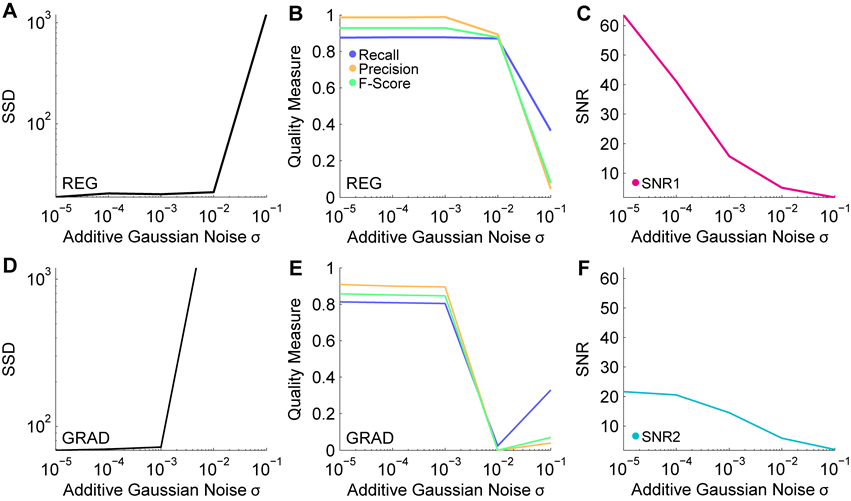}}
\caption[Extraction quality of elongated objects on different noise levels]{Quantitative assessment of the quality of extracted elongated objects as well as the objects in the vicinity of the elongated object using the methods REG and GRAD. The regression curve quality of REG and GRAD was assessed by the sum of squared distances (SSD) of the automatic regression curve and the ground truth regression curve (A, D). The detection quality of objects that were distributed around the regression curve is expressed by recall, precision and F-Score (B, E). The signal-to-noise ratio is plotted in (C, F), where SNR1 and SNR2 were obtained from the small object channel and the elongated object channel, respectively.}
\label{fig:chap3:SpinalCord:NoiseSensitivity}
\end{figure}

The quantitative comparison of REG and GRAD is listed in \cref{tab:chap3:SpinalCord:Validation} and depicted in \cref{fig:chap3:SpinalCord:NoiseSensitivity}. 
\begin{sidewaystable}[p]
%\rowcolors{2}{white}{gray!25}
\centering
\resizebox{\textwidth}{!}{
\begin{tabular}{lcccccccccccccc}
\toprule
\textbf{Method} & $\mathbf{SNR}_1$ & $\mathbf{SNR}_2$ & \textbf{SSD} & $\mathbf{CC}_1$ & $\mathbf{CC}_2$ & \textbf{TP} & \textbf{FP} & \textbf{FN} & \textbf{Rec.} & \textbf{Prec.} & \textbf{F-Sc.} & \textbf{Dist.} & $\mathbf{t_\text{roi} (s)}$ & $\mathbf{t_\text{seed} (s)}$ \\
\midrule
REG & 64 & 22 & 19 & 0.98 & 0.99 & 89 & 1.9 & 13 & 0.88 & 0.98 & 0.93 & 1.2 & 0.39 & 0.28 \\
GRAD & 64 & 22 & 69 & 0.86 & 0.97 & 83 & 8.4 & 19 & 0.81 & 0.91 & 0.86 & 1.8 & 0.31 & 0.30 \\

REG & 41 & 21 & 21 & 0.94 & 0.98 & 87 & 2 & 12 & 0.88 & 0.98 & 0.93 & 1.20 & 0.44 & 0.28 \\
GRAD& 41 & 21 & 70 & 0.79 & 0.93 & 80 & 9 & 19 & 0.81 & 0.90 & 0.85 & 1.80 & 0.31 & 0.31 \\

REG & 16 & 14 & 20 & 0.79 & 0.87 & 88 & 1.7 & 12 & 0.88 & 0.98 & 0.93 & 1.20 & 0.51 & 0.30 \\
GRAD & 16 & 14 & 72 & 0.48 & 0.69 & 80 & 9.40 & 20 & 0.80 & 0.89 & 0.85 & 1.90 & 0.31 & 0.32 \\

REG & 5.10 & 5.90 & 21 & 0.61 & 0.65 & 88 & 13 & 13 & 0.87 & 0.88 & 0.87 & 1.30 & 0.54 & 0.30 \\
GRAD & 5.10 & 5.90 & 4817 & 0.00 & 0.00 & 2.40 & 32 & 99 & 0.02 & 0.00 & 0.00 & - & 0.32 & 0.34 \\

REG & 1.80 & 2.10 & 1235 & 0.01 & 0.02 & 36 & 812 & 65 & 0.36 & 0.04 & 0.08 & 3.10 & 0.56 & 0.31 \\
GRAD & 1.80 & 2.10 & 4396 & 0.00 & 0.00 & 33 & 809 & 68 & 0.33 & 0.04 & 0.07 & 3.20 & 0.32 & 0.35 \\
\bottomrule
\end{tabular}}
\captionsetup{width=0.96\textheight}
\caption[Quantitative assessment of the detection quality for the regression curve estimation and the object detection quality]{Quantitative assessment of the ROI extraction performance of a regression-based approach (REG) and a gradient-based approach (GRAD). Additionally, the object detection quality obtained on the respective linearized crop images is evaluated. SNR1 and SNR2 are the signal-to-noise ratios of channel 1 and channel 2, respectively. SSD reflects the sum of squared distances of the true regression curve and the estimated regression curve. CC1 and CC2 are the correlation coefficients obtained for comparing the extracted crop regions with the ground truth regression curve and the estimated regression curve on the two image channels. The object detection quality was assessed by counting the true positives (TP), the false positives (FP), the false negatives (FN) and by the respective recall, precision and F-Score values. Furthermore, the average distance to the ground truth object centroids was estimated (Dist.). Average processing times for a single image were measured separately for the ROI extraction and the seed detection and are given in seconds.}
\label{tab:chap3:SpinalCord:Validation}
\end{sidewaystable}
Both the elongated structure as well as the surrounding objects were robustly detected by the regression-based approach (REG) up to an additive Gaussian noise level of $\sigma_{\text{agn}}=10^{-2}$ (SNR: $5-6$). The gradient-based approach (GRAD) provided reasonable results up to an additive Gaussian noise level of $\sigma_{\text{agn}}=10^{-3}$ (SNR: $14-16$). Generally, REG yielded more robust results with lower SSD values of $19-21$ and higher F-Score values of $0.87-0.93$ compared to SSD values of $69-72$ and F-Score values of $0.85-0.86$ for GRAD. For higher SNR values of $14-64$, most quality measures remained almost constant. The detection quality of both algorithms, however, abruptly dropped for signal-to-noise ratios lower than $5.9$. Starting at an SNR2 value of $5.9$, GRAD failed to correctly extract the elongated structures of interest as the kernel size of the involved median low-pass filter could not sufficiently reduce the image noise, whereas REG was still able to correctly identify the region. However, at an SNR2 level of $2.1$, REG failed to extract the correct region as well. As both approaches occasionally extracted entirely wrong regions of interest for these low SNR values, the SSD abruptly increased by up to two orders of magnitude and the correlation coefficients dropped to zero. The imprecision of the region of interest extraction step directly influenced the spherical object channel as well and caused an abrupt drop of TP values and a heavily increased amount of false positive detections. In some cases, however, the erroneous regions of interest coincidentally overlapped with the true regions of interest. Thus, the quality measures for the seed detection in \cref{fig:chap3:SpinalCord:NoiseSensitivity}E are not necessarily monotonically decreasing for noise levels $\sigma_\text{agn} > 10^{-3}$, if parts of the correct region of interest are extracted. 

The processing time of REG mainly depends on the number of sample points used for the registration curve fitting, \ie, larger binary regions caused by increased noise levels directly affected the processing time. Contrary, the performance of GRAD is independent of the image content and thus remained constant, even for high noise levels. In both cases, the processing time for the seed detection increased with increased noise levels, as the number of false positive detections in background regions raised. However, both algorithms required a only a few milliseconds to process a single 2D image, \ie, both of them are well suited even for the analysis of multiple thousands of images, which can be processed in a matter of minutes to maximally a few hours. As REG yielded better results and comparable processing times for the lower noise levels, this method should be favored.

\subsection{Discussion}
\label{sec:chap3:SpinalCord:Discussion}
In this section, a new approach for the robust extraction of elongated structures in 2D images was presented. Even for images with a low signal-to-noise ratio, a reliable extraction of both the elongated object and the surrounding objects was guaranteed. In comparison to an alternative approach, the regression-based approach was less sensitive to noise and generally extracted more exact regions of interest than the gradient-based approach. Several techniques that were used to compare the extracted image regions between each other have been introduced, namely, intensity profiles, spatial distribution analysis, object distribution histograms and global statistical values of the crop region. Depending on the actual scientific question, the list of extracted features can of course be arbitrarily extended. Although it appears that the object distribution histograms might provide the most quantitative features, the result quality is heavily reduced if the objects of interest are too small or if they form clusters that cannot be reliably resolved with the proposed seed detection methods. However, in these cases, the intensity profiles described in \cref{sec:chap3:SpinalCord:IntensityProfiles} essentially provided the same information directly on the intensity level rather than on the object level.

The current version of the algorithm assumed bright objects on a dark background. Depending on the raw image material, additional preprocessing steps such as intensity inversion, contrast adjustment or even more complex methods such as vessel enhancement filters might be necessary to produce reasonable results \cite{Frangi98}. The current implementation only searched for a single elongated object of interest in the images. The approach can be adapted to detect and extract multiple elongated structures by simply applying the regression-based ROI extraction on all connected components that contain elongated structures separately. In cases where the elongated structures overlap, a template matching approach that searches for line-like structures could be employed to perform the initial localization of the objects.

In \cref{sec:chap6:SpinalCord}, it is shown how to apply the presented methodology for the extraction of elongated shapes to a toxicological screen to investigate the impact of a large library of small molecules to the neuronal development in the spinal cord of zebrafish embryos.
%\clearpage

%%%%%%%%%%%%%%%%%% NUCLEUS SEGMENTATION %%%%%%%%%%%%%%%%%%%
\section[Efficient Segmentation of Roundish Objects]{A New Algorithm for the Efficient Segmentation of Roundish Objects} % \footnote{This section is partly based on a publication by Stegmaier \etal \cite{Stegmaier14}.}
\label{sec:chap3:TWANG}
A frequently observed segmentation task in many scientific and industrial applications is the detection and segmentation of circular, spherical or generally hyper-spherical objects. Possible objects are, for instance, cell nuclei, stars, nano particles, gas bubbles, coins and basically all other entities that exhibit a (hyper-) spherical shape and are well distinguishable from the background signal. Although the algorithm represents a general methodology to extract sphere-like structures in images of literally arbitrary dimensions, the focus discussed here is put on the fast extraction of fluorescently labeled cell nuclei in 3D microscopy images. Existing methods for this task were not applicable to large-scale time-resolved image data produced by state-of-the-art light-sheet microscopes and it was therefore necessary to find a fast and reliable algorithm to analyze these datasets in a reasonable amount of time \cite{Stegmaier14}. In the remainder of this section the new algorithm is referred to as TWANG (\textbf{T}hreshold of \textbf{W}eighted intensity \textbf{A}nd seed-\textbf{N}ormal \textbf{G}radient dot product image).

\subsection{Validation Benchmark}
The 3D segmentation quality evaluation was performed on the same dataset as described in \cref{sec:chap3:SeedDetection:Benchmark}, including raw benchmark images (\texttt{SBDS1}) and the disrupted image series with different noise levels (\texttt{SBDS2}) and different blur levels (\texttt{SBDS3}). The specific settings for the respective datasets are summarized in \app\cref{tab:Appendix:BenchmarkDatasets}. Additional validation examples performed on a 2D segmentation benchmark are provided in \cite{Stegmaier14}.

\subsection{Algorithmic Design}
\begin{figure}[htb]
\centerline{\includegraphics[width=\columnwidth]{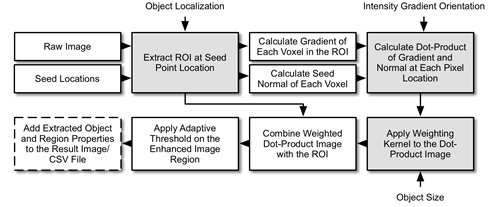}}
\caption[Schematic pipeline overview of the TWANG segmentation algorithm]{Schematic pipeline overview of the TWANG segmentation algorithm. Based on a raw input image and seed point locations of the desired objects, a region of interest of each seed point is transformed to a representation that can be segmented using Otsu's adaptive thresholding technique. All processing steps are performed in parallel for each of the seed points in a data parallel manner. Operators shaded in gray involve prior knowledge such as object localization, intensity gradient orientation and expected object size and the result filter of the pipeline is indicated by the dashed edge.}
\label{fig:chap3:TWANG:PipelineOverview}
\end{figure}

TWANG is a seed-based segmentation algorithm, which means that the initial locations of the objects about to be extracted should already be roughly known in advance and the respective coordinates have to be provided to the algorithm. In this work, the Laplacian-of-Gaussian scale space maximum projection approach was used for seed detection as described in \cref{sec:chap3:SeedDetection:LoGSSMP}. The central idea of the proposed segmentation algorithm is a transformation of the raw input image containing bright spherical objects on a dark background to a representation that can be handled by straightforward adaptive threshold techniques like Otsu's method \cite{Otsu79}. A schematic overview of the involved processing steps is shown in \cref{fig:chap3:TWANG:PipelineOverview}.

\begin{figure}[!htb]
\centerline{\includegraphics[width=\textwidth]{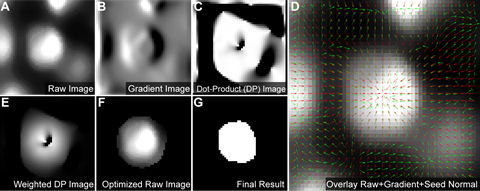}}
\caption[Processing steps of the spherical object segmentation]{Processing steps for the spherical object segmentation on a raw image similar to the \texttt{SBDS1} dataset. (A) Cropped raw image, (B) Gaussian smoothed left-right derivative image, (C) dot product of the seed normal and the normalized image gradient, (D) overlay of raw image and the smoothed gradient (green) and normal vector (red), (E) weighted dot product image, (F) transformed input image and (G) the final segmentation result (adapted from \cite{Stegmaier14}).}
\label{fig:chap3:TWANG:AlgorithmDetails}
\end{figure}

In order to increase the speed and to be able to detect objects under different illumination conditions, the input image transformation and the segmentation was performed only in the local vicinity of the provided seed points. The first step was to homogeneously distribute the seed points among the available threads and to perform further calculations independently and in parallel. In order to process as few voxels as possible, only small regions surrounding each seed point were cropped and analyzed. The initial size estimate provided by the seed detection stage was used to estimate the size of this region. For the 3D segmentation, a cuboid with side lengths of $(\sqrt{2} \cdot 3 \cdot \sigma_{\text{seed}}) \cdot (\frac{1}{s_1}, \frac{1}{s_2}, \frac{1}{s_3})^\top$ was used, where $\sqrt{2} \cdot \sigma_{\text{seed}}$ is the radius of the respective seed point at scale $\sigma_{\text{seed}}$ and $\mathbf{s} = (s_1, s_2, s_3)^\top$ corresponds to the physical spacing of the voxels. At each voxel location $\mathbf{x} = (x_1, x_2, x_3)^\top$ of the cropped image $\mathbf{I}^{\text{cr}}_s$ of the seed $s = 1,...,N_s$, the Gaussian smoothed gradient $\mathbf{g}_s$ was calculated: 
\begin{gather}
	\mathbf{g}_s(\mathbf{x}, \sigma_{\text{grad}}) = \nabla \left(\mathbf{I}^{\text{cr}}_s(\mathbf{x}) \ast \mathbf{G}(\mathbf{x},\sigma_{\text{grad}}) \right).
	\label{eq:chap3:TWANG:Gradient}
\end{gather}
In \cref{eq:chap3:TWANG:Gradient}, $\nabla$ is the nabla operator that is used to calculate the partial derivatives of the Gaussian smoothed cropped image. A regularization scale of $\sigma_{\text{grad}}=3.0$ of the Gaussian kernel $\mathbf{G}$ yielded good results for all investigated scenarios presented in this thesis (see \cref{fig:chap3:TWANG:NoiseLevelInfluence} and \cref{tab:chap3:TWANG:SegmentationEvaluation3D}). However, this parameter might need to be adjusted for higher noise levels, in order to guarantee a smooth gradient vector field (see \eg, SNR $\leq 2$ in \cref{fig:chap3:TWANG:NoiseLevelInfluence}). By iterating separately over each voxel $\mathbf{x}$ in the sub-region to the respective seed position $\mathbf{x}_{s}$, the optimized raw image is sequentially filled with the transformed voxel values. All remaining steps covered in this section are thus defined on the single voxel level. Initially, the difference vector pointing from the seed point location to current voxel was calculated as:
\begin{gather}
	\mathbf{d}_{s}\left(\mathbf{x}\right) = \mathbf{s} \circ \left( \mathbf{x}_s - \mathbf{x}\right),
\end{gather}
with $\circ$ being the Hadamard product \cite{Voigt07}, and the seed point normal $\mathbf{n}_{s}(\mathbf{x})$ at each voxel location is defined as:
\begin{gather}
	\mathbf{n}_{s}\left(\mathbf{x}\right) = \frac{\mathbf{d}_{s}\left(\mathbf{x}\right)}{\Vert \mathbf{d}_{s}\left(\mathbf{x}\right)\Vert_2}.
\end{gather}
The seed point normal is a vector pointing from the seed point location to the currently considered voxel $\mathbf{x}$. In the following step, the dot product ($\langle .,. \rangle$-operator) of each normal in the cropped region with the corresponding normalized gradient vector was calculated as:
\begin{gather}
	\phi_s\left(\mathbf{x}\right) = \frac{1}{2} \cdot \left( 1 + \langle \frac{\mathbf{g}_s\left(\mathbf{x}\right)}{\Vert \mathbf{g}_s\left(\mathbf{x}\right) \Vert_2}, \mathbf{n}_{s}\left(\mathbf{x}\right) \rangle \right).
	\label{eq:chap3:TWANG:DotProduct}
\end{gather}
This contrast invariant measure describes the angular dependency of the gradient and the seed point normal and is similar to the one described by Soubies \etal~\cite{Soubies12}, where it was used in the energy term of an ellipsoid fit segmentation approach. The transformed dot product of normalized vectors in \cref{eq:chap3:TWANG:DotProduct} has a value range of $[0,1]$, with $1$ being identical, $0.5$ being orthogonal and $0$ being antiparallel vectors. The illustration in \cref{fig:chap3:TWANG:AlgorithmDetails}C, D shows that this property can be exploited to discard voxels in transition regions of touching objects.

%The image transformation starts with a Gaussian smoothed gradient image, which we calculate the dot product of the normalized gradient at each location to the corresponding normal pointing from the seed point location to the direction of the investigated pixel.

% Pixels that potentially belong to the desired segment mostly have small angles for this  This dot product allows to identify the pixels potentially belonging to the region occupied by the nucleus. Additionally, pixels of neighboring cells cancel out each other due to opposing gradient and normal direction.

\begin{figure}[!htb]
\centerline{\includegraphics[width=\columnwidth]{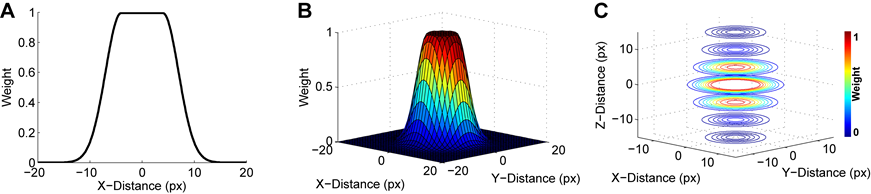}}
\caption[TWANG segmentation weighting kernel]{TWANG segmentation weighting kernel used to discard distant voxels and border artifacts of the cropped region. Exemplary weighting kernels plotted as a line for 1D (A), a surface for 2D (B) and a contour plot for 3D (C), respectively, using the parameters $r_s=4$, $\sigma_{\text{kernel}}=3.0$ and $\omega_{\text{kpm}}=1.0$. The weights are shown on the ordinate axis (A, B) or indicated by the color-code (C). The selected kernel size should encompass the region of interest, such that high weights are obtained in the close vicinity of the object (adapted from \cite{Stegmaier14}).}
\label{fig:chap3:TWANG:SmoothingKernels}
\end{figure}

To discard border artifacts and distant neighboring objects, the intensity value of each voxel in the sub-region was decreased if it was far away from the detected seed location. Based on the initial radius approximation $r_s = \sqrt{2} \cdot \sigma_{\text{seed}}$ and a Gaussian kernel standard deviation $\sigma_{\text{kernel}}$, the following weighting function was employed:
\begin{gather}
	w_s\left(\mathbf{x}\right) = \max \begin{cases}
		\sgn \left( (\omega_{\text{kpm}} \cdot r_s) - \Vert \mathbf{d}_s\left(\mathbf{x}\right)\Vert_2 \right) \\ 
		\exp \left( -\frac{((\omega_{\text{kpm}} \cdot r_s) - \Vert \mathbf{d}_s\left(\mathbf{x}\right)\Vert_2)^2}{2 \cdot (\sigma_{\text{kernel}})^2}\right)
	\end{cases}.
	\label{eq:chap3:TWANG:WeightingKernel}
\end{gather}
An exemplary Gaussian-based smoothing kernel that was calculated using \newline \cref{eq:chap3:TWANG:WeightingKernel} is depicted in \cref{fig:chap3:TWANG:SmoothingKernels}. The weighting kernel is sampled at the discrete voxel locations of the cropped image, with the weighting kernel's origin being located in the center of the cropped image. The signum function ($\sgn$) was used to specify the extent of the plateau region in the center of the weighting kernel. The initial size approximation $r_s$ of the seed detection stage (\cref{eq:chap3:SeedDetection:MaxScale}) was used to preserve regions in the close vicinity of the centroid that were smaller than $r_s$. The weighting kernel plateau radius can optionally be increased or decreased using the multiplier $\omega_{\text{kpm}}$ with values larger or smaller than the default value of $1.0$, respectively. Similarly, the kernel standard deviation $\sigma_{\text{kernel}}$ can be adjusted to flatten the kernel at the borders of the plateau region to increase the influence of voxels at larger distances to the seed center.

The weighted normalized dot product image (\cref{fig:chap3:TWANG:AlgorithmDetails}E) could then be obtained by the voxel-wise multiplication of $w_s\left(\mathbf{x}\right) \cdot \phi_s\left(\mathbf{x}\right)$, \ie, a multiplication of the weighting kernel image (\cref{fig:chap3:TWANG:SmoothingKernels}C) and the dot product image (\cref{fig:chap3:TWANG:AlgorithmDetails}C). The cropped raw image (\cref{fig:chap3:TWANG:AlgorithmDetails}A) and the weighted dot product image \newline (\cref{fig:chap3:TWANG:AlgorithmDetails}E) were combined by copying the original intensity values within the seed radius and by multiplying all remaining raw intensity values with the weighted dot product image. The result of this operation is shown in \cref{fig:chap3:TWANG:AlgorithmDetails}F. Finally, a simple adaptive threshold such as Otsu's method was applied on the weighted and cropped original image depicted in \cref{fig:chap3:TWANG:AlgorithmDetails}F to obtain the final segmentation shown in \cref{fig:chap3:TWANG:AlgorithmDetails}G.
Segment properties such as volume, mean intensity, bounding volumes and the like were immediately extracted from the cropped image regions in parallel. Compared to other segmentation algorithms, a final labeling step on the whole image level can thus be skipped to accelerate the pipeline execution. Furthermore, the algorithm directly extracts one object at a time per thread based on the predefined size criteria, \ie, literally no merged segments are produced and a watershed-based post-processing to separate merged objects can also be omitted.

\subsection{Validation}
To assess the quality of the new segmentation method (TWANG), a quantitative comparison to Otsu's method (OTSU) \cite{Otsu79}, Otsu's method with a watershed-based splitting of merged blobs (OTSUWW) \cite{Otsu79, Beare06}, a geodesic active contours method (GAC) \cite{Caselles97}, a gradient flow tracking segmentation (GFT) \cite{Liu08, Li07b} and a graph cut-based segmentation (GC) \cite{Al-Kofahi10} was performed. Representative 3D benchmark images of fluorescently labeled nuclei were used (\texttt{SBDS1}, \texttt{SBDS2} and \texttt{SBDS2}), which were similar to the primary application field of the newly developed algorithm. Nevertheless, all illustrations in the following section should be regarded as exemplary use-cases, without the restriction to only work on this particular set of  images. In order to work with other imaging modalities or objects, however, it might be necessary to initially transform the input image into a similar representation, \eg, using image inversion, binary thresholds or color space conversions and then applying the proposed algorithm on the transformed image.

\subsubsection{Datasets and Compared Algorithms}
Otsu's method was selected to demonstrate that the analysis of the respective benchmark images could not sufficiently be analyzed using this adaptive thresholding technique. All other algorithms represent reasonable approaches for the segmentation of fluorescently labeled nuclei, as described in the respective publications \cite{Beare06,Liu08,Li07b,Al-Kofahi10}. For OTSU and OTSUWW, an additional Gaussian low-pass filter was used to reduce high frequency noise. The level set function of the GAC pipeline was initialized using the LoG-based seed detection method described in \cref{sec:chap3:SeedDetection:LoGSSMP}. All algorithms were implemented in C++ (see \cite{Stegmaier14} for implementation details). Details on the performance evaluation measures and the employed testing platform can be found in \cref{sec:Appendix:SegmentationEvaluation} and \cref{sec:Appendix:EvaluationPlatform}, respectively. Processing times were measured in seconds.

\subsubsection{Performance Evaluation}
\begin{figure}[htb]
\begin{center}
\includegraphics[width=\columnwidth]{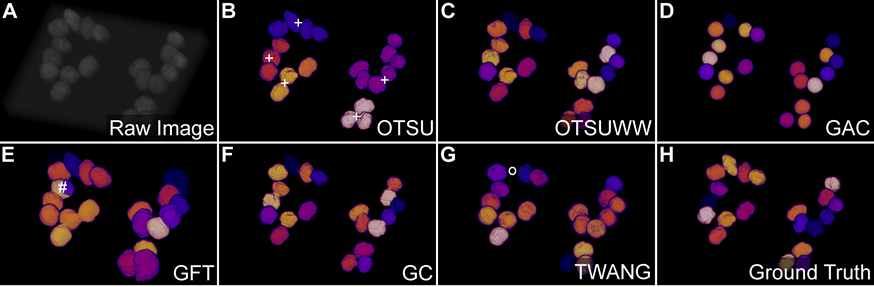}
\end{center}
\caption[Quality assessment of the TWANG segmentation on a simulated 3D benchmark]{Result quality achieved by the compared segmentation algorithms on a set of simulated 3D benchmark images (\texttt{SBDS1}). (A) Simulated original image, (B) adaptive thresholding using Otsu's method \cite{Otsu79}, (C) Otsu's method combined with watershed-based blob splitting \cite{Otsu79, Beare06}, (D) geodesic active contours \cite{Caselles97}, (E) gradient vector flow tracking \cite{Li07b}, (F) graph-cuts segmentation \cite{Al-Kofahi10}, (G) TWANG segmentation and (H) the ground truth. Segmentation errors are separated into split (\#), merged (+) or missing (o). Each detected object is uniquely colored (adapted from \cite{Stegmaier14}).}
\label{fig:chap3:TWANG:SegmentationComparison3D}
\end{figure}

The quality evaluation for the simulated 3D benchmark data is listed in \cref{tab:chap3:TWANG:SegmentationEvaluation3D}. 3D volume renderings of the false-colored segmentation results are depicted in \cref{fig:chap3:TWANG:SegmentationComparison3D}. Merged regions (\cref{fig:chap3:TWANG:SegmentationComparison3D}B), split regions (\cref{fig:chap3:TWANG:SegmentationComparison3D}E) and missing objects (\cref{fig:chap3:TWANG:SegmentationComparison3D}G) are highlighted by the respective symbols. Except of the plain OTSU method (\cref{fig:chap3:TWANG:SegmentationComparison3D}B) and the gradient flow tracking method (\cref{fig:chap3:TWANG:SegmentationComparison3D}E), most algorithms produced close to perfect topological segmentation results with heterogeneously distributed optimal values (all F-Score values $\geq 0.95$). The watershed-based post-processing (OTSUWW) could resolve the erroneous mergers produced by OTSU (\cref{fig:chap3:TWANG:SegmentationComparison3D}C) yielding a $15.2\%$ better recall value of $0.99$ compared to $0.84$ for the plain OTSU. Due to poor edge information, GAC and GFT produced rather too large segments resulting in $1.57$ and $1.13$ merged objects, respectively. Furthermore, GFT had an increased amount of added objects compared to the other algorithms, which caused the low precision value of $0.71$. In both cases, using a different parameterization resulted in heavy under- or over-segmentation. Although regions may not be captured as accurately as with the watershed-corrected adaptive thresholding, the graph-cut implementation provided very good segmentation results and properly split connected nuclei in most cases (F-Score value of $0.97$). TWANG offered a high precision (no merged, split or added nuclei) but missed a few cells, yielding a recall of $0.93$ and a combined F-Score of $0.96$. With respect to the shape based measurements (RI, JI, NSD and HM), OTSU, OTSUWW, GAC and GC produced the best results. The reduced values observed for TWANG and GFT were caused by the increased amount of missing and added nuclei, respectively. However, it can also be observed that OTSU obtained high values in these shape-based categories but is practically not usable without the watershed-based merge correction OTSUWW.
%% Figure 5
%% Table 3
\begin{sidewaystable}[p]
%\rowcolors{2}{white}{gray!25}
\resizebox{\textwidth}{!}{
\begin{tabular}{lccccccccccccc}
\toprule
\textbf{Method} & \textbf{RI} & \textbf{JI} & \textbf{NSD $(\times 10)$} & \textbf{HM} & \textbf{Split} & \textbf{Merg.} & \textbf{Add.} & \textbf{Miss.} & \textbf{Recall} & \textbf{Precision} & \textbf{F-Score} & \textbf{t (s)} & \textbf{t (s)}$^\ast$ \\
\midrule
OTSU & 97.35 & 6.00 & 22.82 & 5.72 & \textbf{0.00} & 3.27 & 0.87 & \textbf{0.00} & 0.84 & 0.95 & 0.89 & \textbf{0.49} & \textbf{0.44} \\ 
OTSUWW & 97.57 & 6.03 & \textbf{3.80} & \textbf{1.12} & 0.13 & \textbf{0.00} & \textbf{0.00} & \textbf{0.00} & 0.99 & 0.95 & \textbf{0.97} & 2.57 & 2.48 \\
GFT & 88.06 & 3.57 & 6.81 & 6.25 & 0.10 & 1.57 & 6.53 & 1.87 & 0.83 & 0.71 & 0.77 & 15.51 & - \\ 
GAC & 95.06 & \textbf{6.40} & 7.41 & 2.52 & \textbf{0.00} & 1.13 & \textbf{0.00} & 0.77 & 0.91 & \textbf{1.00} & 0.95 & 5.92 & - \\
GC & \textbf{97.78} & 6.37 & 5.66 & 1.69 & 1.34 & 0.07 & \textbf{0.00} & \textbf{0.00} & \textbf{1.00} & 0.94 & \textbf{0.97} & 5.92 & - \\ 
TWANG & 93.82 & 4.94 & 6.62 & 2.41 & \textbf{0.00} & \textbf{0.00} & \textbf{0.00} & 1.37 & 0.93 & \textbf{1.00} & 0.96 & 3.72 & 1.08 \\ 
\bottomrule
\end{tabular}}
\captionsetup{width=\textheight}
\caption[Comparison of the 3D Segmentation Quality for TWANG]{Segmentation quality of the compared algorithms on the \texttt{SBDS1} dataset. For quality assessment the Rand index (RI) in percent, the Jaccard index (JI), the Hausdorff metric (HM) and the normalized sum of distances (NSD, values multiplied by 10) as defined in \cite{Coelho09} and \cref{sec:Appendix:SegmentationEvaluation} were used. In addition, the average number of split, merged, erroneously added or missing objects is given. The table summarizes the arithmetic mean values of $30$ processed 3D benchmark images. The average performance of the algorithms to process a single image was measured without using threads and with 8 threads where possible (indicated by $^{\ast}$) \cite{Stegmaier14}. For GFT, GAC and GC no parallel version could be tested, \ie, the respective performance evaluations are missing.}
\label{tab:chap3:TWANG:SegmentationEvaluation3D}
\end{sidewaystable}

Required processing times significantly varied between the algorithms. The gradient flow tracking (GFT) was up to one order of magnitude slower than OTSU, OTSUWW and TWANG and did not produce better results that would justify the slower execution. When considering the single core performance, the best compromise of speed vs.\ quality was achieved by the OTSUWW segmentation method. However, this method does not easily scale as no parallel watershed implementation is available yet. Thus, if a multicore processor is available, the TWANG segmentation provides the best speed vs.\ quality trade-off. 

\begin{figure}[htb]
\centerline{\includegraphics[width=\columnwidth]{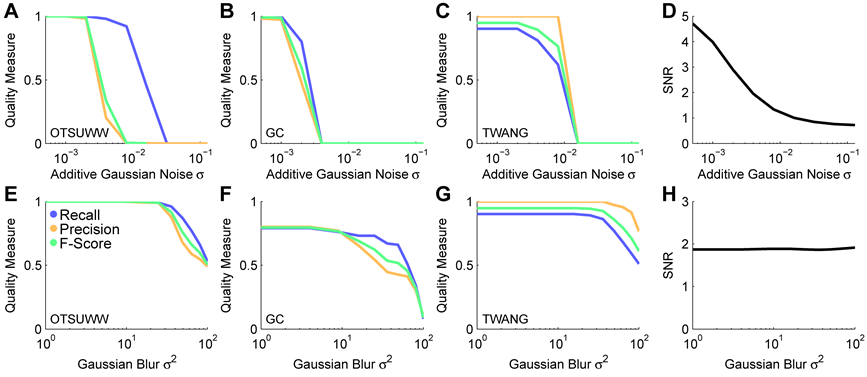}}
\caption[Assessment of the 3D segmentation performance for different noise levels]{Assessment of the 3D segmentation performance of OTSUWW (A, E), GC (B, F) and TWANG (C, G) for different noise and Gaussian blur levels using the simulated 3D benchmark datasets \texttt{SBDS2} and \texttt{SBDS3}, respectively. The performance measures recall, precision and F-Score are plotted against the additive Gaussian noise level parameter $\sigma_{\text{agn}}$ (A-C) and the Gaussian blur variance $\sigma^2_{\text{smooth}}$ (E-G). The influence of the noise and blur levels on the signal-to-noise ratio (SNR) of the images is shown in (D, H).}
\label{fig:chap3:TWANG:NoiseLevelInfluence}
\end{figure}

The three algorithms that provided the best segmentation results on the 3D benchmark (OTSUWW, GC and TWANG) were applied on the disrupted image series to assess their performance on differently flawed images. The recall, precision and F-Score values obtained by the three investigated algorithms on the disrupted image series are shown in \cref{fig:chap3:TWANG:NoiseLevelInfluence}. All investigated algorithms were at least able to tolerate a small amount of noise ($\sigma_\text{agn} \leq 0.001$). The segmentation quality of the GC algorithm, however, already significantly dropped for $\sigma_\text{agn}$ levels larger than $0.001$. OTSUWW was able to tolerate slightly higher noise levels up to $\sigma_\text{agn} \geq 0.002$. Above this level, the global Otsu threshold still recovered the correct segments (recall close to 1 up to $\sigma_\text{agn} = 0.01$), but at the same time detected a lot of false positive segments that abruptly lowered the precision of the algorithm. The most robust approach was the TWANG segmentation, which produced acceptable results almost up to noise levels of $\sigma_\text{agn}=0.01$ due to the robust seed detection method used to localize the actual objects of interest. Regarding the increasingly blurred images, OTSUWW and TWANG were able to reliably extract the objects up to a variance $\sigma^2_{\text{smooth}}=30$, whereas the quality of GC already slightly dropped at a variance of $\sigma^2_{\text{smooth}}=4$. As the blurred images were additionally disrupted with a constant Gaussian noise, the results of GC in \cref{fig:chap3:TWANG:NoiseLevelInfluence}F already started at a lower level compared to \cref{fig:chap3:TWANG:NoiseLevelInfluence}B. The results confirm the robustness of the proposed TWANG segmentation and render it a suitable approach even for noisy and possibly blurred images.

\subsection{Discussion}
The presented algorithm for the efficient segmentation of spherical objects in multidimensional images proved to be up to an order of magnitude faster than the compared methods when applied on multicore CPUs. Moreover, the quantitative and qualitative assessment of the segmentation quality showed that the TWANG algorithm produced label images that were comparable to other established methods. Even in the case of severe noise or blurring of the images, the proposed method robustly extracted the desired objects. For the assessment of the segmentation quality on disrupted images, the respective algorithmic parameters were kept constant throughout the experiment. A further extension might be an adaption of the parameter values individually for each noise or blur level. However, this is almost impossible using manual parameter tweaking and it is necessary to identify (semi-)automated parameter adjustment schemes to accomplish this task.

A remaining problem is the correct extraction of elongated 3D structures that do not match the sphere assumption of the weighting kernel (\eg, if the major axis of the object is twice as large as the two other axes). In these cases, the algorithm tended to clip the tips of ellipsoidal objects due to the spherical shape of the weighting kernel. This behavior could be compensated by adjusting the kernel size multiplier and the kernel standard deviation properly. Further work needs to be put into identifying the elongation properties of the objects already at the seed detection stage, to adapt the weighting kernels accordingly \cite{Stegmaier14}.

The quality and reliability of extracted seeds directly influenced the outcome of the segmentation and it was therefore crucial to provide high-quality seeds to the segmentation algorithm. As demonstrated in \cref{sec:chap4:SeedDetection:SeedFusion}, the extended LoG-based seed detection produced excellent results when combined with the uncertainty handling and the fusion of redundant information. The initial size estimate of the multiscale seed detection successfully guided the TWANG algorithm to extract differently sized objects properly \cite{Stegmaier14}. Furthermore, the extracted properties of segmented regions can be used to refine the segmentation results using the uncertainty framework as discussed in \cref{sec:chap4:AlgorithmEnhancement}. Possible features for the fuzzy set generation are the integrated region intensity, seed intensity, volume, bounding box extents, foreground vs. background ratio and many more.

To sum up, the presented segmentation algorithm is a reasonable choice for a fast analysis of terabyte-scale data that cannot be handled by methods such as \cite{Al-Kofahi10} due to given memory or time constraints. The presented segmentation algorithm was successfully applied to automatically analyze multiple terabytes of image data of developing zebrafish embryos (\cref{sec:chap6:TWANG}) in a distributed computing environment based on the Apache\TTra Hadoop\TReg framework (\cref{sec:Appendix:HadoopCluster}).
%\clearpage

%%%%%%%%%%%%%%%%%% MEMBRANE SEGMENTATION %%%%%%%%%%%%%%%%%%%
\section[Fast and Accurate Segmentation of Locally Plane-like Structures in 3D Images]{A New Pipeline for Fast and Accurate Segmentation of Locally Plane-like Structures in 3D Image Data} % \footnote{This section is partly based on a publication by Stegmaier \etal \cite{Stegmaier15}.}
\label{sec:chap3:RACE}

\begin{figure}[!htb]
\centerline{\includegraphics[width=\columnwidth]{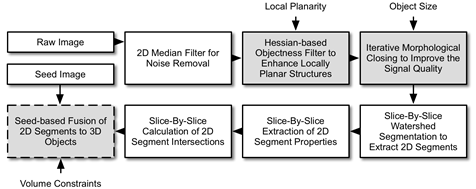}}
\caption[Schematic pipeline overview of the RACE segmentation algorithm]{Schematic pipeline overview of the RACE algorithm used to efficiently segment locally plane-like structures in 3D images. High-quality 2D segmentations are extracted individually for each of the slices on the basis of an enhanced input image. An additionally provided seed image is used to fuse the 2D segments in each of the slices to form complete 3D objects. Operators shaded in gray involve prior knowledge such as local planarity of desired structures, expected object sizes as well as volume constraints. The output operator is indicated by a dashed edge line.}
\label{fig:chap3:RACE:PipelineOverview}
\end{figure}

After having introduced algorithms for the seed detection, detection and extraction of elongated objects as well as segmentation of spherical objects, a further frequently observed problem is the segmentation of locally plane-like structures that encapsulate a hollow interior in 3D images. This segmentation task can, for instance, be required to quantitatively assess fluid foams \cite{Wang11}, formation of adipose tissue \cite{Weisberg03} or as in our particular case the segmentation of fluorescently labeled cellular membranes \cite{Fernandez10, Mosaliganti12, Khan14a, Liu14a}. In \cref{sec:chap1:Segmentation}, numerous methods that are in principle suitable for this class of segmentation problems were presented. However, existing methods are either only applicable for 2D images \cite{Krzic12} or involve complex processing operators, which in turn significantly limit their application to large-scale microscopy images as produced, \eg, by state-of-the-art light-sheet microscopy \cite{Mosaliganti12, Khan14a}. Furthermore, most 3D algorithms require isotropic image data to work properly. As many 3D acquisition techniques produce anisotropic image data, existing approaches perform an upscaling of the images to isotropic resolution, which further increases the processing time required by the algorithms. To overcome these drawbacks of existing algorithms, a hybrid approach of efficient image preprocessing, a slice-based segmentation and a subsequent fusion of extracted 2D segments is proposed, in order to apply the algorithm directly on anisotropic data and to be able to heavily parallelize the performed computations. A schematic overview of the involved processing operators is given in \cref{fig:chap3:RACE:PipelineOverview} and the new algorithm is referred to as
RACE (\textbf{R}eal-time \textbf{A}utomated \textbf{C}ell shape \textbf{E}xtractor) in the remainder of this chapter.

\subsection{Validation Benchmark}
\begin{figure}[!htb]
\centerline{\includegraphics[width=\columnwidth]{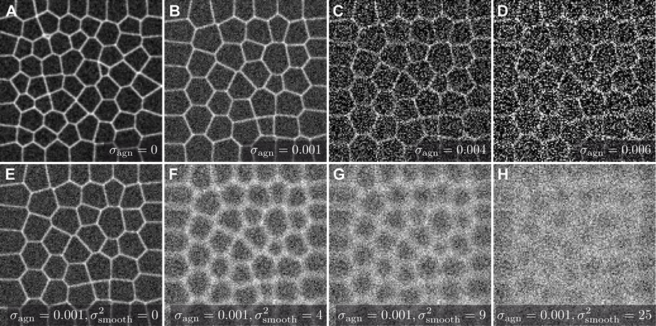}}
\caption[Benchmark Overview]{Exemplary slices of differently flawed 3D benchmark images with different levels of additive Gaussian noise (A-D) and Gaussian smoothed versions with fixed noise level (E-H).}
\label{fig:chap3:RACE:BenchmarkOverview}
\end{figure}

A set of simulated images by Mosaliganti \etal has been used to evaluate the performance of the developed algorithm on synthetic data of locally plane-like structures in 3D (\texttt{SBDP1}), as well as to be able to assess the segmentation quality under various noise levels and different blur levels \cite{Mosaliganti12}\footnote{\url{https://wiki.med.harvard.edu/pub/SysBio/Megason/ACME/SupplementaryInformation.pdf}}. Similar to the previous sections, the two sets of disrupted images were generated from the simulated benchmark data using zero-mean additive Gaussian noise with $\sigma_{\text{agn}} \in [0.0, 0.128]$ for the first series (\texttt{SBDP2}) and Gaussian blur with variance $\sigma^2_{\text{smooth}} \in [0, 100]$ and fixed additive Gaussian noise with $\sigma_{\text{agn}}=0.001$ for the second series (\texttt{SBDP3}). The disruption parameter range was determined manually, to obtain signal-to-noise ratio levels close to the visual detection limit. A set of exemplary benchmark images is shown in \cref{fig:chap3:RACE:BenchmarkOverview}. The signal-to-noise ratio was determined by the ratio of foreground mean intensity and background intensity standard deviation (\cref{eq:chap1:SNR}).
The quantitative measurements were averaged over the quality obtained independently on all annotated images. The segmentation performance was assessed based on the measures described in \cref{sec:Appendix:SegmentationEvaluation} and the benchmark settings are summarized in \app\cref{tab:Appendix:BenchmarkDatasets}. Further validation experiments on real microscopy images are provided in \cref{sec:chap6:RACE} and \cite{Stegmaier16}.

\subsection{Algorithmic Design}
\subsubsection{Noise Removal and Hessian-based Ridge Enhancement}
\begin{figure}[!htb]
\centerline{\includegraphics[width=\columnwidth]{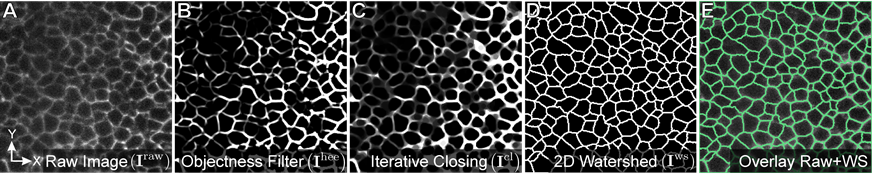}}
\caption[Processing steps for a high-quality 2D segmentation]{Performed processing steps to obtain a high-quality 2D segmentation for each slice of a 3D image stack. Starting with the raw image (A), the locally plane-like signal was enhanced using an eigenvalue-based objectness filter (B). An iterative morphological closing was applied in order to close small gaps in the signal (C). Finally, an accurate 2D segmentation of each slice was obtained using the watershed transform (D). An overlay of the segmentation result and the raw image is depicted in (E) (adapted from \cite{Stegmaier16}).}
\label{fig:chap3:RACE:PreprocessingSteps}
\end{figure}

As an initial step of the algorithm a shot noise reduction of the input image $\mathbf{I}^{\text{raw}}$ was performed using a 2D median filter approach (\cref{fig:chap3:RACE:PreprocessingSteps}A). For all images considered throughout this thesis, a filter radius of two pixels was sufficient. However, if the images are heavily corrupted by noise, it might be required to use larger filter radii. The median filtered image $\mathbf{I}^{\text{med}}$ was further processed using an objectness filter that emphasized locally plane-like structures in 3D images. Here, a generalization of Frangi's vessel enhancement filter was used \cite{Frangi98} as described by Antiga \etal \cite{Antiga07}. The Hessian matrix was approximated using finite differences of the image intensities of each voxel's direct neighbors \cite{Rudzki09}. In order to smooth the response of the filter, the Hessian matrix was calculated on a Gaussian-smoothed version of the input image $\mathbf{I}^{\text{med}}$ using a standard deviation $\sigma_{\text{hee}}$. Based on the eigenvalues of the Hessian matrix of the image at each voxel location, the filter response emphasized $M$-dimensional objects in $N$-dimensional images. For the particular case of enhancing locally plane-like structures, \ie, 2D plane-like structures in 3D images, the dimension parameters were set to $M=2$ and $N=3$. The generalized objectness function for this parameterization and the sorted eigenvalues $| \lambda_1 | \leq | \lambda_2 | \leq |\lambda_3 |$ of the Hessian matrix at each voxel location is given by \cite{Antiga07}:
\begin{gather}
	O \left(\pmb{\lambda}(\mathbf{x}, \sigma_{\text{hee}}) \right) = \begin{cases}
							e^{-\frac{R^2_B}{2\beta_{\text{hee}}^2}} \cdot \left(1-e^{-\frac{S^2}{2\gamma_{\text{hee}}^2}}\right) &,  \lambda_3 < 0 \\	
							0 &, \lambda_3 \geq 0. \\
						\end{cases}
\label{eq:chap3:RACE:ObjectnessFunction}
\end{gather}
In \cref{eq:chap3:RACE:ObjectnessFunction}, $\pmb{\lambda}(\mathbf{x}, \sigma_{\text{hee}})$ represents the vector of eigenvalues of the Hessian calculated at location $\mathbf{x}$ of the $\sigma_{\text{hee}}$-regularized input image $\mathbf{I}^{\text{med}}$ with the associated Frobenius norm  $S=\sqrt{\lambda_1^2+\lambda_2^2+\lambda_3^2}$ \cite{Voigt07}. As a large negative value of the largest eigenvalue $\lambda_3$ indicates a sheet-like structure, only this case was considered in \cref{eq:chap3:RACE:ObjectnessFunction} \cite{Rudzki09}. Furthermore, $R_B$ is defined as $|\lambda_2|/|\lambda_3|$ and $\beta_{\text{hee}}, \gamma_{\text{hee}}$ are user-defined weights to control the response of the objectness function (default values proposed in \cite{Antiga07}: $\sigma_{\text{hee}} = 10, \beta_{\text{hee}} = 0.5, \gamma_{\text{hee}} = 5.0$). The factor containing $R_A$, which is found in the original formulation of the generalized objectness filter was omitted here, as it evaluated to one for the considered parameterization \cite{Antiga07}. As the default parameters were not applicable for the image material considered in this thesis due to differently sized objects of interest, the parameters were manually optimized to emphasized the structures of interest and set to $\sigma_{\text{hee}} = 2, \beta_{\text{hee}} = 1, \gamma_{\text{hee}} = 0.1$. Sequentially calculating the objectness filter response for all voxels of the median filtered input image yielded an edge-enhanced image $\mathbf{I}^{\text{hee}}$ (\cref{fig:chap3:RACE:PreprocessingSteps}B).

\subsubsection{Iterative Morphological Closing and Slice-based Watershed Segmentation}
In order to close small gaps in the locally plane-like structures of the 3D images, an iterative closing of the edge-enhanced image $\mathbf{I}^{\text{hee}}$ was performed using grayscale morphology \cite{Soille03}. The morphological closing operation is defined as the erosion of the dilation of the respective input image. In the special case of grayscale morphology, erosion and dilation correspond to the minimum and maximum of the non-zero entries of the structuring element, respectively. Here, an Euclidean sphere was used as structuring element $\mathbf{S}_r$, where the respective radii $r$ were scaled according to the physical spacing of the input image.  
\begin{gather}
	\mathbf{I}^{\text{cl}}_r = \begin{cases} \mathbf{I}^{\text{hee}} &, r=0 \\ \mathbf{I}^{\text{cl}}_{r-1} \bullet \mathbf{S}_r &,r>0 \\ \end{cases}.
	\label{eq:chap3:RACE:IterativeClosing}
\end{gather}
The iteratively closed image up to radius $r$ is denoted by $\mathbf{I}^{\text{cl}}_r$ and calculated according to the recursion shown in \cref{eq:chap3:RACE:IterativeClosing}, where $\bullet$ is the morphological closing operation (\cref{fig:chap3:RACE:PreprocessingSteps}C). The maximum radius of the iterative closing operation was determined by the radius of the smallest objects that should still be detected by the algorithm. Note that the iterative closing can become a bottleneck in the processing pipeline if the used maximum radius is set too large. If only large objects are present in the images it makes sense to down-scale the image instead of using too large radii.
Applying a watershed segmentation onto an iteratively closed input image like $\mathbf{I}^{\text{cl}}_r$ is also known as the viscous watershed transform as described by Vachier \etal, which simulates the flooding of a topographic surface with a viscous fluid and helps to minimize leakage through porous object boundaries \cite{Vachier05}. Instead of a physically correct modeling of the viscous fluid, the iterative closing was restricted to a discrete set of predefined radii that was determined by prior knowledge about the desired objects in the images. For all considered images a set of $r=\{1,2,3,4\}$ sufficiently closed gaps in the locally plane-like object signal.

The subsequent watershed transform of the closed image was performed separately on the individual slices, \ie, in 2D instead of directly in 3D as performed by many existing approaches. This yielded a new 3D image $\mathbf{I}^{\text{ws}}$, where each of the slices contained a high-quality 2D segmentation of the desired objects (\cref{fig:chap3:RACE:PreprocessingSteps}D, E and \cref{fig:chap3:RACE:FusionSteps}B). The benefit of the slice-based watershed segmentation was the reduced leakage of background signal into the desired objects, which is a frequently observed issue in 3D watershed segmentations. In addition, the slice-based approach could be perfectly parallelized by performing the calculations of individual slices separately on different threads. The watershed segmentation used in the presented pipeline has a single parameter, which is the intensity dependent starting level of the flooding process, \ie, all local minima below this threshold value are suppressed \cite{Beare06}. Although the 2D segmentation approach helped to speed up the calculations and to mostly get rid of the leakage problem, a subsequent fusion of 2D segments was necessary to obtain the final 3D segmentation as described in the next paragraph.

\subsubsection{Seed Point Detection}
\label{sec:chap3:RACE:SeedDetection}
\begin{figure}[htb]
\centerline{\includegraphics[width=\columnwidth]{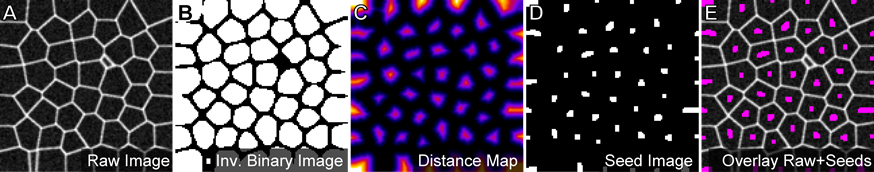}}
\caption[Euclidean distance map-based seed point detection]{The most important steps to detect centroids of locally plane-like objects in the \texttt{SBDP1} dataset using an Euclidean distance map-based approach. The inverted, binarized version of $\mathbf{I}^{\text{hee}}$ (B) was used to obtain initial object markers. Closely located objects that might be connected due to gaps in the plane-like objects could be efficiently split using an Euclidean distance map on the binary images (C). The actual seeds were then extracted from the distance map images by a connected components labeling of the thresholded distance map or the extracted $h$-maxima of the distance map (D). An overlay of the detected seed segments and the raw image is depicted in (E).}
\label{fig:chap3:RACE:EDMSeeds}
\end{figure}

As the synthetic dataset \texttt{SBDP1} only contained simulated membrane structures without central objects, the seeds were directly extracted from this channel. With minor adaptions, this can be achieved by applying the seed detection approach described in \cref{sec:chap3:SeedDetection:EDM} on an inverted version of the edge-enhanced image $\mathbf{I}^{\text{hee}}$ to calculate the EDM. In \cref{fig:chap3:RACE:EDMSeeds}, the individual processing steps to extract seeds from the \texttt{SBDP1} dataset are illustrated.

\subsubsection{Intersection Calculation and Segment Fusion}
\label{sec:chap3:RACE:IntersectionCalculation}

\begin{figure}[htb]
\centerline{\includegraphics[width=\columnwidth]{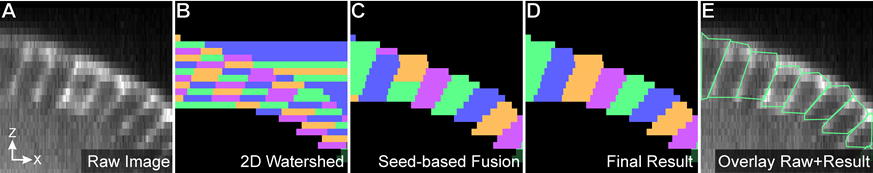}}
\caption[Processing steps for 3D segment fusion]{The processing steps to fuse high-quality 2D segments to a complete 3D object. An XZ-slice of an exemplary raw image containing the locally plane-like objects is depicted in (A). Based on the slice-based 2D segmentation (B), a seed-based combinatorial approach was used to merge intersecting 2D segments to single 3D object (C). Two fusion heuristics that incorporated both prior knowledge about intensity and expected object volume were employed to further improve the segmentation results and to correct over-segmentation errors (D). In (E), the final object silhouette is superimposed on the raw input image (adapted from \cite{Stegmaier16}).}
\label{fig:chap3:RACE:FusionSteps}
\end{figure}

Based on the slice-based 2D segmentations stored in the $\mathbf{I}^{\text{ws}}$ image (\cref{fig:chap3:RACE:FusionSteps}B), the correspondences between slices were determined on the basis of spatial overlap of neighboring segments. Here, the Jaccard index (JI) was used, which is defined as the ratio of the intersection of the two segments vs. the union of the two segments, \ie, it measures the percentage of overlap of two segments with $0$ for no overlap and $1$ for identical segments. More formally, considering two segments $A, B$ this can be formulated as \cite{Williams49}:
\begin{gather}
JI = \frac{A \cap B}{A \cup B}.
\label{eq:JaccardIndex}
\end{gather}

Assuming that segments with high spatial overlap are likely to belong to the same 3D structure, all intersection pairs were sorted in descending order based on the respective JI similarity value. To avoid fusing very unlikely intersections, a manually determined threshold was used, \ie, only the intersections with values larger than this threshold were considered for further segment fusion. A threshold value of $0.4$ produced reasonable results for all segmentation scenarios considered in this thesis. The fusion of the individual segments was obtained by calculating a seeded minimum spanning tree \cite{Kruskal56}. The seeds used for the segment fusion were derived either directly from the image containing the locally plane-like structures or optionally from an additional image that contained spherical objects that reside inside of the desired locally plane-like objects as described in \cref{sec:chap3:SeedDetection:EDM}. Initially, all 2D segments that intersected with a detected seed were directly labeled with the respective seed label. Intersections within the intersection list where already both segments had a seed label assigned were removed from the list as no further processing was necessary for these segments. All intersections where one of the two segments already had a label assigned were added to an intersection queue, which was again ordered in descending order based on the JI values. The queue was iteratively emptied by popping the top element and by propagating the label of the labeled segment to the unlabeled segment of the intersection. Subsequently, the new segment label was updated for all existing intersections with this newly labeled segment and the respective intersections were inserted into the intersection queue with respect to their similarity value. This process was repeated until the intersection queue was empty. Initialized by the provided seed points, all 2D segments were enlarged by this technique to the level where no intersections with a similarity above the specified threshold were present anymore (\cref{fig:chap3:RACE:FusionSteps}C).
If the provided seeds perfectly correspond to the number of objects present in the image, no further processing is required at this stage. However, due to potential imperfections of the seed detection, some objects may contain multiple seed points, which results in an over-segmentation of the respective object. To further improve the obtained 3D segmentation results, two fusion heuristics that reduce this kind of over-segmentation errors have been developed. The first heuristic is based on prior knowledge about the expected object size. By specifying a minimum and a maximum expected volume, sub-segments were further fused if they were smaller than the minimum volume and if the fusion additionally did not violate the specified maximum volume value. Similar techniques were also used in existing implementations, however, they were mostly directly performed on the image level rather than on the object level, which can have a significant impact on the time performance of the algorithms.

The second heuristic is based on calculating the minimum spanning tree directly on the similarity measure according to Kruskal's algorithm \cite{Kruskal56} with the additional constraint that no fusion of two segments should introduce a local intensity maximum, which would indicate an object boundary within a segment. The JI-based fusion produced a largely over-segmented result but literally no under-segmentation. Thus, fused segments in the JI-based fusion were a strong indicator of a required merger of two segments. In cases where a single 3D segment in the JI-based fusion intersected with two segments of the seeded fusion, the two segments of the seeded fusion were combined to a single segment. The final outcome of the fusion algorithm including the proposed fusion heuristics is depicted in \cref{fig:chap3:RACE:FusionSteps}D.

\subsection{Validation}
\label{sec:chap3:RACE:Validation}
\subsubsection{Datasets and Compared Algorithms}

The proposed RACE segmentation algorithm was compared to two state-of-the-art segmentation algorithms from the field of automatic segmentation of cellular membranes in fluorescence microscopy images. The ACME method applies a tensor voting operation on a ridge-enhanced input image and extracts segments with a 3D watershed algorithm \cite{Mosaliganti12}. The MARS method employs alternate sequential filtering combined with a size-dependent iterative watershed transform \cite{Fernandez10}. The respective version of the RACE algorithm is indicated in round brackets. The specifiers stand for seeds extracted from a spherical objects channel (NS) or seeds obtained directly from the locally plane-like object channel (MS). Moreover, different combinations of enabled and disabled small segment fusion heuristic (SSH) and the Jaccard index-based fusion heuristic (JI) are used. All measurements were performed on the testing platform described in \cref{sec:Appendix:EvaluationPlatform}. Processing times were measured in seconds and represent the average time needed to process a single image of the dataset.
%%Due to the limited automation possibilities of the graphical user interface of EDGE4D, it was not considered in the noise performance evaluation.

\subsubsection{Performance Evaluation}
\begin{figure}[htb]
\centerline{\includegraphics[width=\columnwidth]{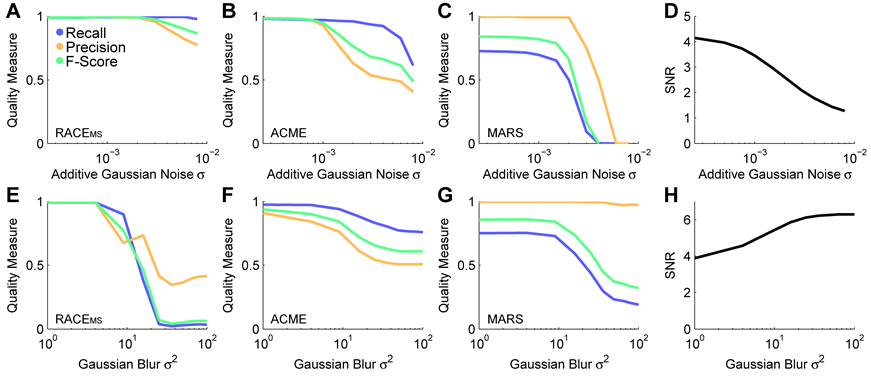}}
\caption[Assessment of the RACE 3D segmentation performance for different noise levels]{Assessment of the 3D segmentation performance of $\text{RACE}_\text{MS}$ (A, E), ACME (B, F) and MARS (C, G) for different noise (\texttt{SBDP2}) and Gaussian blur levels (\texttt{SBDP3}) using simulated 3D benchmark images by Mosaliganti \etal \cite{Mosaliganti12}. The performance measures recall, precision and F-Score are plotted against the additive Gaussian noise level parameter $\sigma_{\text{agn}}$ (A-C) and the Gaussian blur variance $\sigma^2_{\text{smooth}}$ (E-G). The influence of the noise and blur levels on the signal-to-noise ratio (SNR) of the images is shown in (D, H).}
\label{fig:chap3:RACE:NoiseLevelInfluence}
\end{figure}

In \cref{tab:chap3:RACE:SegmentationPerformanceBenchmark}, the segmentation performance achieved on the \texttt{SBDP1} benchmark dataset is listed. 
\begin{sidewaystable}[p]
\begin{center}
%\rowcolors{2}{white}{gray!25}
\resizebox{\textwidth}{!}{
\begin{tabular}{lccccccccccccc} %{p{3.5cm}p{6cm}p{5cm}}
\toprule
\textbf{Method} & \textbf{RI} & \textbf{JI} & \textbf{NSD $(\times 10)$} & \textbf{HM} & \textbf{Split} & \textbf{Merg.} & \textbf{Add.} & \textbf{Miss.} & \textbf{Rec.} & \textbf{Prec.} & \textbf{F-Sc.} & \textbf{t (s)} & \textbf{s (KVox./s)} \\
\midrule
$\text{RACE}_\text{MS}$ & 99.42 & \textbf{1.00} & 5.27 & 1.77 & 5.25 & 5.25 & \textbf{1.00} & 3.00 & \textbf{0.99} & \textbf{0.99} & \textbf{0.99} & 1.94 & 430.71 \\
$\text{RACE}_\text{MS, SSH}$ & 99.42 & \textbf{1.00} & 5.27 & 1.73 & 2.50 & 6.00 & \textbf{1.00 }& 3.00 & 0.98 & \textbf{0.99} & \textbf{0.99} & 1.69 & 494.43 \\
$\text{RACE}_\text{MS, JI}$ & 99.42 & \textbf{1.00} & 5.27 & 1.77 & 5.25 & \textbf{5.00} & \textbf{1.00} & 3.00 & \textbf{0.99} & \textbf{0.99} & \textbf{0.99} & \textbf{1.66} & \textbf{503.36} \\
$\text{RACE}_\text{MS, JI, SSH}$ & 99.42 & \textbf{1.00} & 5.27 & 1.73 & 2.50 & 6.00 & \textbf{1.00} & 3.00 & 0.98 & \textbf{0.99} & \textbf{0.99} & 1.71 & 488.65 \\

ACME & \textbf{99.48} & \textbf{1.00} & \textbf{2.63} & \textbf{1.39} & 4.75 & 10.00 & \textbf{1.00} & \textbf{2.50} & 0.98 & \textbf{0.99} & 0.98 & 28.15 & 74.21 \\
MARS & 98.31 & 0.98 & 3.73 & 1.67 & \textbf{1.25} & 90.25 & \textbf{1.00} & 61.25 & 0.73 & \textbf{0.99} & 0.84 & 12.41 & 168.33 \\
\bottomrule
\end{tabular}}
\captionsetup{width=0.96\textheight}
\caption[Quantitative assessment of the segmentation performance]{Quantitative performance assessment of the segmentation methods RACE \cite{Stegmaier16}, ACME \cite{Mosaliganti12} and MARS \cite{Fernandez10} on simulated benchmark images \cite{Mosaliganti12}. The criteria used to compare the algorithms are the Rand index (RI), the Jaccard index (JI), the normalized sum of distances (NSD) and the Hausdorff metric (HM) as described in \cref{sec:Appendix:SegmentationEvaluation} and \cite{Coelho09}. Additionally, the topological errors were assessed by counting the total number of split, merged, added and missing objects. Precision, recall and F-Score are based on the topological errors by considering split and added nuclei as false positives and merged and missing objects as false negatives, respectively. The average performance to process a single image was measured in seconds (smaller values are better) and voxels per second (larger values are better).}
\label{tab:chap3:RACE:SegmentationPerformanceBenchmark}
\end{center}
\end{sidewaystable}
Moreover, Both ACME and RACE produced almost perfect results with F-Score values of $0.98$ and $0.99$, respectively. MARS provided a high precision value of $0.99$ as it produced almost no split and added cells. On the other hand, due to insufficient border enhancement, MARS merged and missed way more cells compared to ACME and RACE, which resulted in a decreased recall value of only $0.73$ and a combined F-Score value of $0.84$. The average processing time of RACE was up to an order of magnitude faster compared to the other methods (less than $2$ seconds compared to $28$ seconds for ACME and $12$ seconds for MARS). The bottleneck of the ACME pipeline is the computationally expensive tensor voting followed by a single-threaded 3D watershed segmentation \cite{Mosaliganti12}. The main bottleneck of MARS is as well the single-threaded 3D watershed algorithm for the final segmentation \cite{Fernandez10}. These costly steps were intentionally omitted in RACE and replaced by a parallel slice-based watershed segmentation. However, the quality benefit of the 3D watershed segmentation and the tensor voting of ACME is reflected in the best scores for the shape-based measurements (RI, JI, NSD and HM). The analysis of the disrupted benchmark image datasets \texttt{SBDP2} and \texttt{SBDP3} unveiled that RACE was more robust to image noise, providing excellent results up to $\sigma_{\text{agn}}=0.002$ and smoothing variances of up to $\sigma^2_{\text{smooth}}=4$ (\cref{fig:chap3:RACE:NoiseLevelInfluence}A, E). However, for increased blur levels, the quality also significantly dropped. ACME's segmentation quality decreased already at $\sigma_{\text{agn}}=0.0007$ and was also more sensitive to smoothed object boundaries, which resulted in successively decreased precision already at low smoothing variance values of $\sigma^2_{\text{smooth}}\geq 1$ (\cref{fig:chap3:RACE:NoiseLevelInfluence}B, F). Although the results of MARS were worse compared to ACME and RACE with respect to overall recall value, it was at least able to robustly keep the detection level up to $\sigma_{\text{agn}}=0.001$ and $\sigma^2_{\text{smooth}}=4$, respectively (\cref{fig:chap3:RACE:NoiseLevelInfluence}C, G).

\begin{figure}[htb]
\centerline{\includegraphics[width=\columnwidth]{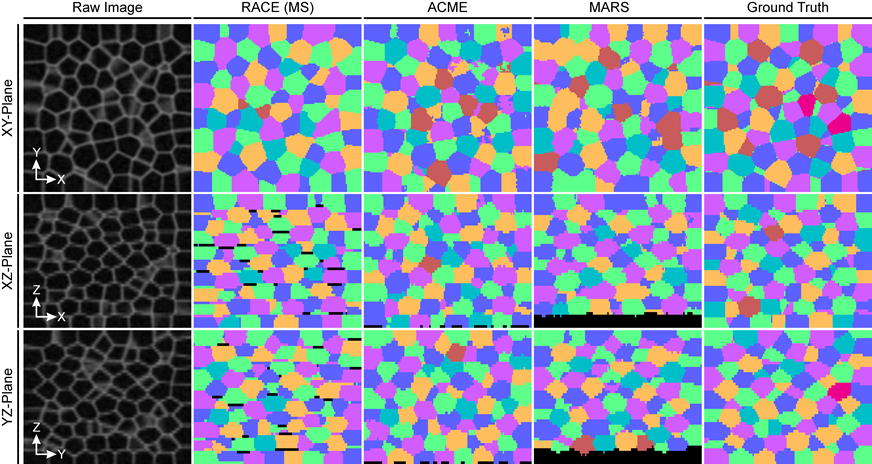}}
\caption[Qualitative comparison of the segmentation results of RACE, ACME and MARS]{Qualitative comparison of the segmentation quality obtained for the different algorithms on one of the \texttt{SBDP1} benchmark images. Besides the raw input image, the segmentation results for $\text{RACE}_{\text{MS}}$, ACME, MARS and the ground truth are shown. The exemplary XY, XZ and YZ slices of the segmentation label images were colored using a graph labeling scheme. Both RACE and ACME provided the best segmentation quality (F-Scores of $0.99$ and $0.98$, respectively), whereas MARS had issues with many merged objects as well as leakage between the individual objects that lead to many missed cells (F-Score of only $0.84$). Due to the 3D watershed segmentation, the axial transitions between cells are more smooth for ACME and MARS compared to RACE (black cavities between cells in XY-Plane and YZ-Plane). However, this influenced the overall quality only marginally (see \cref{tab:chap3:RACE:SegmentationPerformanceBenchmark}).}
\label{fig:chap3:RACE:SegmentationComparison}
\end{figure}

In \cref{fig:chap3:RACE:SegmentationComparison}, a graphical comparison of the obtained segmentation results is depicted for an exemplary benchmark image. RACE and ACME subjectively produced the best results, which was also confirmed by the quantitative analysis in \cref{tab:chap3:RACE:SegmentationPerformanceBenchmark}. Both algorithms reliably detected the objects present in the images. Due to the 3D watershed segmentation, the segmentation of ACME qualitatively looks better at the axial transition regions between cells, where RACE missed some slices. However, the general quality was only marginally affected (\cref{tab:chap3:RACE:SegmentationPerformanceBenchmark}). In contrast to this, MARS was not able to sufficiently close gaps in the membrane signal and produced many merged and erroneous regions. Note that increased blur levels cause an increasing SNR level in \cref{fig:chap3:RACE:NoiseLevelInfluence}H, due to a successively decreasing background signal standard deviation compared to an almost constant foreground mean intensity (see \cref{eq:chap1:SNR}). Although there is not yet a clear explanation for the precision variations observed in \cref{fig:chap3:RACE:NoiseLevelInfluence}E, this effect was most likely caused by local random noise variations that may have affected the 2D watershed segmentation and in turn the subsequent 3D segment fusion.

\subsection{Discussion}
The newly developed RACE algorithm presented in this section proved to provide similar or even better results compared to state-of-the-art algorithms. Even for noisy images or if object contours were slightly smoothed, the algorithm was able to robustly extract the desired objects. At the same time the processing times were at least one order of magnitude lower as for the comparative algorithms. Similar to the TWANG segmentation presented in the previous section, the proposed segmentation algorithm for the reliable extraction of locally plane-like structures was directly dependent on the quality of the provided seeds. It was therefore crucial to tune the seed detection stage to produce as few false positive detections as possible, in order to minimize the amount of split objects in the final segmentation. Possible ways to improve the overall quality of detected seeds in the case of redundant detections was to reject small seeds based on a minimum size criterion or to reject and fuse detected seeds based on prior knowledge-based fuzzy set membership values, as described in \cref{sec:chap4:SeedDetection}. The presented segmentation fusion heuristics represent a second feasible approach to partially get rid of over-segmentation errors. Especially, the SSH heuristic was able to successfully correct segmentation errors to a large extent, if object sizes did not vary too much (\eg, developmental stages of an embryo where all cells have comparable volumes that allow one to set tight volume constraints). If object sizes varied heavily (imprecise volume constraints), the improvements achieved by this heuristic were rather limited and it might make sense to add additional heuristics based on object intensity profiles or additional topological criteria. Although, the JI heuristic also slightly improved the segmentation results, its impact was rather small compared to the SSH heuristic \cite{Stegmaier16}. The segmentation quality of all investigated methods was strongly dependent on the sharpness of object contours as shown in \cref{fig:chap3:RACE:NoiseLevelInfluence}E-G, \ie, only relatively small smoothing of the filigree locally plane-like signal was tolerated. It is therefore crucial to ensure that acquired images are perfectly focused and that light scattering is minimized if possible. 

Memory limitations that were present, \eg, in the implementation of ACME, were solved by using single precision floating point operations as well as an explicit calculation of the eigenvalues at each pixel location, without calculating a Hessian image for the entire image first. These simple improvements combined with a dedicated GPGPU implementation instantly reduced the required amount of memory by a factor of 18 and enabled processing times that were two orders of magnitude faster compared to existing methods, and eventually enabled real-time data processing \cite{Stegmaier16}. Considering the significant improvement of required processing times as well as the achieved segmentation quality, the presented algorithm is particularly well suited for the automated analysis of large volumetric image stacks, \eg,  produced by live imaging of developing model organisms as demonstrated in \cref{sec:chap6:RACE}.

 \cleardoublepage

%% Algorithm Enhancement

%%%%%%%%%%% Enhancing Simple Algorithms with Uncertainty Treatment %%%%%%%%%%
\chapter{Enhancing Algorithms with Uncertainty Treatment}
\label{sec:chap4:AlgorithmEnhancement}
\begin{figure}[htb]
\centerline{\includegraphics[width=\columnwidth]{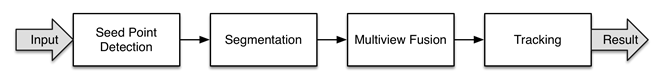}}
\caption[Exemplary image analysis pipeline]{Exemplary image analysis pipeline consisting of seed point detection, segmentation, multiview fusion and tracking.}
\label{fig:chap4:PipelineOverview}
\end{figure}

In \cref{sec:chap2:UncertaintyFramework}, the translation of available prior knowledge to a mathematical representation using fuzzy sets was introduced. It was shown, how the fuzzy sets can be used in order to determine the validity of extracted information by means of membership degree to a desired class of objects and potential applications such as object filtering, correction and fusion were outlined. Furthermore, three new segmentation algorithms were presented that are usable for efficient information extraction from large-scale multidimensional images (\cref{sec:chap3:EfficientSegmentation}). The present chapter partly combines the results of these two chapters and illustrates how the uncertainty framework can be incorporated into an existing image analysis pipeline for result quality enhancement. The investigated pipeline was inspired by the requirements of the automated analysis of embryonic development and essentially consisted of detection and segmentation of spherical objects (\cref{sec:chap4:SeedDetection}, \cref{sec:chap4:Segmentation}) that needed to be combined from different views (\cref{sec:chap4:MultiviewFusion}) and tracked (\cref{sec:chap4:Tracking}) in time series of 3D microscopy images. An overview of the considered pipeline is provided in \cref{fig:chap4:PipelineOverview}. 

Initially, a comprehensive, generic and artificial benchmark is introduced that served for the validation of all presented processing operators individually and entirely. Moreover, each of the processing operators used for the automatic extraction of spherical objects from three dimensional videos is presented. It is systematically indicated, at which steps of the pipeline the uncertainty framework was incorporated and how the operators could benefit from the additional information. At first the usage of uncertainties is discussed internally of processing operators, \eg, to exclude unlikely objects from further calculations. In the case of redundant information produced either by applying the same operator on different overlapping views of a scene or by the use of different detection methods on the same image, it is shown how to fuse the respective information appropriately. Furthermore, it is demonstrated how the uncertain information can be shared and propagated between multiple sequentially arranged processing operators to enhance the overall quality of the automated analysis. The processing operators presented in this chapter are entirely motivated by the automated analysis of biological experiments, which are detailed in \cref{sec:chap6:Applications} but the general concepts can be analogously applied to similar problems from other scientific fields.

%%%%%%%%%%%%%%%%%%%%%%%%% BENCHMARK %%%%%%%%%%%%%%%%%%%%%%%%%%
\section{A New Comprehensive Validation Benchmark}
\label{sec:chap4:Benchmark}
\begin{figure}[htb]
\centerline{\includegraphics[width=\textwidth]{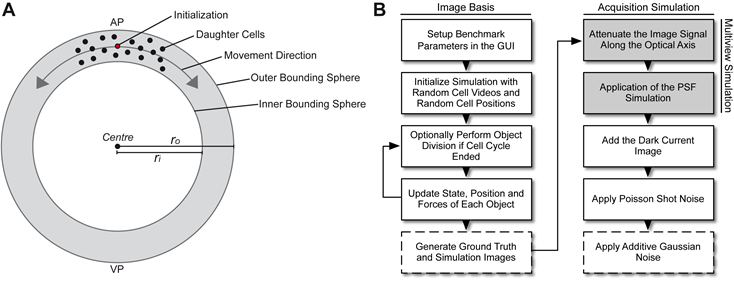}}
\caption[Pipeline schematic of the benchmark simulation]{(A) Illustration of the embryo simulation. Starting with an initial object, multiple division cycles were simulated including object interaction, object divisions and morphological constraints. Due to inner and outer bounding spheres and the between-object interactions, cells migrated from the animal pole (AP) to the artificial vegetal pole (VP), which produced a behavior similar to the epiboly movement of zebrafish embryogenesis. (B) The performed steps for a realistic simulated 3D+t benchmark. The left column reflects the object simulation and returned raw images that contained dynamic objects and the associated ground truth data. The right column contains the acquisition simulation, which distorted the simulated images by an artificial signal attenuation, a point spread function simulation (PSF), a dark current image simulation, Poisson distributed photon shot noise and additive Gaussian noise. Steps shaded in gray could optionally consider image rotation, if a multiview experiment was simulated and the output operators are indicated by dashed edge lines (adapted from \cite{Stegmaier16Arxiv}).}
\label{fig:chap4:Benchmark:Illustration}
\end{figure}

To provide a thorough validation of the proposed methods it was inevitable to make extensive use of benchmarking. Various benchmarks that could be employed for the individual components of image analysis pipelines such as seed detection \cite{Gelasca09, Ruusuvuori08}, segmentation \cite{Coelho09, Ruusuvuori08} or tracking \cite{Sbalzarini05,Rapoport11} have already been presented. An inherent problem of manually generated benchmark datasets, however, is that they often suffer from biased expert knowledge or contain ambiguous image content that is differently rated by different investigators. Hence, simulated data that incorporated all of the mentioned aspects in one comprehensive testing environment was desirable. The charm of using simulated data was the immediate availability of a reliable ground truth and the literally unrestricted possibilities to adjust parameters like noise levels, sampling rates or light attenuation. To overcome the problem of tedious and flawed manual labeling of microscopy images, it has been shown that biological phenomena can be realistically simulated if enough knowledge of the investigated probes was available \cite{Lehmussola06, Lehmussola07, Svoboda07, Rapoport11, Maska14}. However, none of the described approaches could directly be used for the analyses required to evaluate the performance of an entire pipeline comprised of seed detection, segmentation, multiview fusion and tracking with a single benchmark. The new benchmark required for a comprehensive analysis had to fulfill the following criteria:
\begin{itemize}
	\item Adjustable image size for empirical processing time estimations
	\item Realistic simulation of fluorescence properties of labeled nuclei
	\item Customizable number of cells, nucleus size, cell cycle duration and experimental duration
	\item Realistic cell behavior, such as cell migration and mitosis events
	\item Parameterized nucleus dynamics, \eg, different movement models, neighborhood related movement dynamics, spatial restrictions and randomized speed and direction
	\item Acquisition deficiencies such as an approximated point spread function, slice dependent illumination variations or uneven background illumination
	\item Simulated multiview generation including light attenuation along the virtual axial direction
	\item Simulation of detector and discretization related deficiencies like dark current, photon shot noise and signal amplification noise.
\end{itemize}
To achieve these requirements, existing components and methods were used where possible and condensed with new approaches to a comprehensive framework that provided the desired flexibility. Although, there may be more complex and more realistic approaches to model biologically correct behavior of the cells, the parameters for the generated images were manually tuned to get close to the target datasets of fluorescently labeled nuclei in light-sheet microscopy images while still providing the ease of use, flexibility and speed. This finally enabled testing a multitude of different imaging scenarios in detail and for an entire processing pipeline with a comprehensive benchmark instead of solely assessing the performance of the individual pipeline components on multiple specialized benchmarks. A schematic illustration of the simulated specimen and an overview of the involved simulation steps of the benchmark is shown in \cref{fig:chap4:Benchmark:Illustration}.

\subsection{Simulation of Fluorescently Labeled Objects}
\begin{figure}[htb]
\centerline{\includegraphics[width=\textwidth]{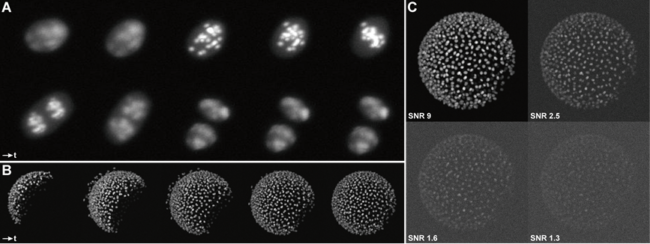}}
\caption[Maximum intensity projections of simulated benchmark images]{(A) Maximum projections of an extracted division cycle of one simulated nucleus \cite{Svoboda12}. Time increases from left to right and top to bottom. Single objects were randomly initialized and simulated for a predefined experimental duration. This approach yielded a simulated embryo including object movement, object interaction and object divisions with available ground truth (B). Generated raw image sequences were manipulated to simulate various acquisition conditions, such as different levels of additive Gaussian noise (C). The ground truth enabled a quantitative analysis of the algorithmic performance on realistic image data (adapted from \cite{Stegmaier16Arxiv}).}
\label{fig:chap4:Benchmark:TimeSeries}
\end{figure}

An object video template database was created from a small simulated time-lapse dataset comprised of eight dividing cells over two division cycles that was kindly provided by D.~Svoboda \cite{Svoboda12}. Sub-videos containing all frames of the division cycles of each cell were extracted and cropped to only contain a single cell (\cref{fig:chap4:Benchmark:TimeSeries}A). According to personal communication with D.~Svoboda, a higher cell density was not desired for their simulation of suspension cells. Hence, the object video template library described above was used to generate a more challenging benchmark, more precisely, to simulate an artificial developing organism with possibly thousands of dynamic objects. The size of the simulated objects was adjusted to match the size properties observed in representative images of developing zebrafish embryos (\cref{sec:chap6:TWANG}).

\subsection{Generating the Benchmark Images}
\begin{figure}[tb]
\centerline{\includegraphics[width=\textwidth]{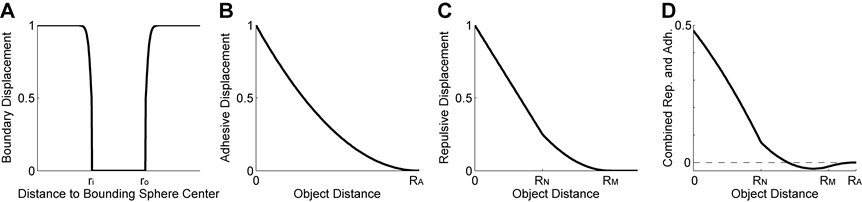}}
\caption[Potential functions of the object displacement]{Qualitative magnitude of the boundary displacement (A), the adhesive displacement (B), the repulsive displacement (C) and a combination of the repulsive and the adhesive displacement (D) using the weights $w_{\text{adh}}=0.52$ and $w_{\text{rep}}=1.0$ as proposed in \cite{Macklin12}. The boundary displacement vector depends on the distance of the considered object to the center of the bounding sphere. The other two displacement components act between objects, \ie, both depend on the distance between two objects. Note that the repulsive and adhesive displacement vectors point in the opposite direction and the respective influence can be controlled using weight parameters. The intersection of the combined repulsive and adhesive curves with the x-axis shown in (D) represents the equilibrium point of both forces, \ie, objects closer than this distance activate a repulsion and contrary, higher distances trigger an adhesion of two interacting objects.}
\label{fig:chap4:Benchmark:PotentialForces}
\end{figure}

The first step of the benchmark generation was the setup of the parameters that specified the desired content and the generated ground truth data (\cref{tab:chap4:Benchmark:Parameters}).
\bgroup
\def\arraystretch{1.5}
\begin{table}[htb]
\begin{center}
%\rowcolors{2}{white}{gray!25}
\begin{tabular}{p{0.15\textwidth}p{0.4\textwidth}p{0.35\textwidth}}
\toprule
\textbf{Simulation Step}& \textbf{Required Parameters} 		& \textbf{Ground Truth} \\
\midrule
Image Basis & Initial/Maximum Number of Objects, Maximum Speed of Objects, Movement Model, Time Series Length, Number of Frames, Image Size and Physical Spacing, Object Video Database	& Simulated Raw Images, Labeled Mask Images, Object Associations, Extracted Object Properties, Image Statistics \\
\midrule
Multiview Simulation &	View Transformations, Optical Axis, Simultaneous/Sequential Acquisition	& View Transformations, Optical Axis, Simultaneous/Sequential Acquisition \\
\midrule
Acquisition Simulation & Light Attenuation Model, Point Spread Function, Dark Frame Intensity, Noise Parameters & Attenuation Model, Point Spread Function, Dark Frame Intensity, Noise Parameters and SNR	\\
\bottomrule
\end{tabular}
\caption[Benchmark parameters and generated ground truth data]{Steps performed for the generation of the simulated benchmark set and the associated ground truth data.}
\label{tab:chap4:Benchmark:Parameters}
\end{center}
\end{table}
\egroup
Based on the specified image extents a virtual experimental volume was simulated. The selected amount of initial objects was randomly distributed within this experimental volume, each of the created objects got a randomly selected object video assigned and started at a randomly selected cell cycle state. After the initialization, the actual dynamic simulation was started, which was comprised of updating each object's cell cycle state and object video frame. If an object's cell cycle ended during the performed update an object division was performed. Each of the new objects was again randomly initialized and positioned relative to its ancestor.
To obtain a dynamically changing scene, the position of each object was updated at every simulation step by considering a set of simulated influences that were acting on it. The interactions were comprised of displacement vectors that originated from adhesive and repulsive behavior between nearby objects, such that objects were not freely floating around but formed clusters. However, objects had to maintain a minimum distance to avoid overlapping of simulated objects. This was accomplished using a repulsive component acting between two objects if their distance became smaller than the sum of their radii. Following the formulation by Macklin \etal \cite{Macklin12}, the displacement vector of the adhesive interaction between two objects was defined as:
\begin{gather}
	\Delta\mathbf{x}^{\text{adh}}(\mathbf{d}, R_A) = 
	\begin{cases} 
		\left(1-
			\frac{\Vert\mathbf{d}\Vert}{R_A} \right)^{2} \cdot \frac{\mathbf{d}}{\Vert\mathbf{d}\Vert} & 0 \leq \Vert\mathbf{d}\Vert < R_A \\
		\mathbf{0}, & \text{else}.
	 \end{cases} 
\label{eq:chap4:Benchmark:AdhesiveForce}
\end{gather}
In \cref{eq:chap4:Benchmark:AdhesiveForce}, $\mathbf{d}=\mathbf{x}_j-\mathbf{x}_i$ is the centroid difference vector of two interacting objects $i,j$ and $R_A$ is the maximum adhesive interaction distance \cite{Macklin12}. The displacement vector caused by the repulsive interaction was defined as \cite{Macklin12}:
\begin{gather}
	\Delta\mathbf{x}^{\text{rep}}(\mathbf{d}, R_N, R_M) = 
	\begin{cases} 
		-\left( \left( \left( 1- \frac{R_N}{R_M} \right)^2-1 \right) \cdot \frac{\Vert \mathbf{d} \Vert}{R_N}+1 \right) \cdot \frac{\mathbf{d}}{\Vert \mathbf{d} \Vert}, & 0 \leq \Vert \mathbf{d} \Vert \leq R_N \\
		- \left( 1 - \frac{\Vert \mathbf{d} \Vert}{R_M} \right)^2 \cdot \frac{\mathbf{d}}{\Vert \mathbf{d} \Vert}, & R_N < \Vert \mathbf{d} \Vert < R_M \\
				\mathbf{0}, & \text{else}.
	 \end{cases} 
\label{eq:chap4:Benchmark:RepulsiveForce}
\end{gather}
Next to the centroid difference vector $\mathbf{d}$, $R_N$ and $R_M$ are the radii of cellular nucleus and the membrane \cite{Macklin12}. The influence of the adhesive and the repulsive interaction can be controlled using the weights $w_{\text{adh}}$ and $w_{\text{rep}}$, respectively.
In addition to the cell-cell interactions, a boundary potential function $\Delta\mathbf{x}^{\text{bdr}}$ was defined to force the objects to move within a narrow volume between two bounding spheres (\cref{fig:chap4:Benchmark:Illustration}A):
\begin{gather}
	\Delta\mathbf{x}^{\text{bdr}}(\mathbf{x}, \mathbf{c}, r_i, r_o, a) = 
	\begin{cases} 
		\frac{\mathbf{x}-\mathbf{c}}{\Vert\mathbf{x}-\mathbf{c}\Vert} \cdot \left( 1-\frac{1}{e^{-a(\Vert\mathbf{x}-\mathbf{c}\Vert-r_i)}} \right), & \Vert\mathbf{x}-\mathbf{c}\Vert < r_i \\ 
		-\frac{\mathbf{x}-\mathbf{c}}{\Vert\mathbf{x}-\mathbf{c}\Vert} \cdot \left(1-\frac{1}{e^{a(\Vert\mathbf{x}-\mathbf{c}\Vert-r_o)}} \right), & \Vert\mathbf{x}-\mathbf{c}\Vert > r_o \\ 
		\mathbf{0}, & \text{else}.
	 \end{cases}
\label{eq:chap4:Benchmark:BoundaryForce}
\end{gather}
In \cref{eq:chap4:Benchmark:BoundaryForce}, $\mathbf{x}$ is the centroid of the considered object, $\mathbf{c}$ is the center of the bounding volumes, $r_i$ and $r_o$ are the radii of the inner and the outer sphere, respectively, and finally $a$ controls the shape of the sigmoidal boundary potential function. $\Delta\mathbf{x}^{\text{bdr}}$ only contributed to the displacement of a simulated object if the object was already out of the boundary. This additional displacement component prevented objects from entering the inner bounding sphere and from escaping the outer bounding sphere of the simulated embryo (\cref{fig:chap4:Benchmark:Illustration}A). Similar to the other two interaction types, the influence of the boundary term was controlled using a weight $w_{\text{bdr}}$. In \cref{fig:chap4:Benchmark:PotentialForces}, the magnitude of the three displacement terms is qualitatively illustrated.
The total displacement vector of a single object at a given time point can be summarized to:
\begin{gather}
	\Delta\mathbf{x}_i^{\text{tot}} = w_{\text{bdr}} \cdot \Delta\mathbf{x}^{\text{bdr}}(\mathbf{x}_i) + \sum_{j\in\{1,...,N\}}^{i\neq j}{\left[w_{\text{rep}} \cdot \Delta\mathbf{x}^{\text{rep}}(\Vert\mathbf{x}_i-\mathbf{x}_j\Vert) + w_{\text{adh}} \cdot \Delta\mathbf{x}^{\text{adh}}(\Vert\mathbf{x}_i-\mathbf{x}_j\Vert)\right]}.
	\label{eq:chap4:Benchmark:TotalForce}
\end{gather}
Depending on the desired movement model, additional terms can be added to \cref{eq:chap4:Benchmark:TotalForce}, for instance, to add directed movement components or to simulate Brownian motion. However, for the benchmark only passive movements that originated from object density variations, the between-object interaction and the boundary term were considered. As the simulation considered in this thesis was only performed on cell nuclei, the parameters were set in relation to the nucleus radii $r_i, r_j$ of two interacting objects $i, j$ to $R_N=r_i+r_j$ for the nucleus radius parameter and $R_A=R_M=2\cdot R_N$ for the membrane and adhesive influence radii. The weights of the displacement components were set to the default values mentioned in \cite{Macklin12}: $w_{\text{adh}}=0.52$, $w_{\text{rep}}=1.0$. Moreover, the weight $w_{\text{bdr}}=3.0$ was manually adjusted, such that the interacting objects remained within the spherical boundaries. Note that these parameters were empirically determined and that the presented model does not necessarily represent an accurate physical simulation of the interacting objects. However, the determined parameters produced movement behaviors that were similar to the epiboly movements observed during early zebrafish development due to increased object densities and boundary constraints that caused a directed movement on a sphere surface.
%\begin{gather}
%	F^{\text{adh}}(\mathbf{d}, r_m) = \begin{cases} \left( 1 - \frac{\Vert\mathbf{d}\Vert}{r_m} %\right)^{2}\frac{\mathbf{d}}{\Vert\mathbf{d}\Vert}, & 0 \leq \Vert\mathbf{d}\Vert \leq r_m \\ %\mathbf{0}, & \text{else}. \end{cases} 
%\end{gather}
%\begin{gather}
%	F^{\text{rep}}(\mathbf{d}, r_n, r_m) = \begin{cases} -\left[\left(\left(1-\frac{r_n}{r_m} %\right)^{2}-1\right)\frac{|\mathbf{d}|}{r_n}+1\right]\frac{\mathbf{d}}{\Vert\mathbf{d}\Vert}, & 0 %\leq \Vert\mathbf{d}\Vert \leq r_n \\ 
%	-\left(1-\frac{\Vert\mathbf{d}|}{r_m}\right)^{2}\frac{\mathbf{d}}{\Vert\mathbf{d}\Vert}, & r_n %\leq \Vert\mathbf{d}\Vert \leq r_m
%	\\ \mathbf{0}, & \text{else}. \end{cases} 
%\end{gather}
The simulation discussed so far was only performed in the feature space and did not produce any image output. The actual benchmark images and the corresponding label images were initialized as entirely black images. By iterating over all simulated time points and all simulated objects, both the benchmark images and the ground truth images were successively filled with the simulated objects. The respective intensity and label values of the assigned object video frames were simply copied to the result image and the ground truth image, respectively. Note that every object had a unique ID in the label image, to be able to unambiguously identify objects and their frame-to-frame association. A schematic overview of the benchmark generation processing steps is depicted in \cref{fig:chap4:Benchmark:Illustration}B.

The generation of the benchmark images described so far produced perfectly clear images that were not flawed by any acquisition deficiencies observed in real microscopy images. To approximately model the acquisition process of a real fluorescence microscope, the simulated images were distorted in several steps to obtain a realistic benchmark dataset (\cref{fig:chap4:Benchmark:Illustration}B). First, the intensities of the simulated images were attenuated with respect to the virtual optical axis. The intensities of each slice were multiplied by a linearly decreasing factor, which was set to $1$ at the slice closest to the virtual detection objective and to $0$ at the slice farthest from the detection objective. The next step was to convolve the entire image with a point spread function (PSF). Here, the PSFs published in \cite{Preibisch14} were used, which were measured by imaging fluorescent beads in a light-sheet microscope. Another option frequently found in the literature is to convolve the obtained benchmark images with a three dimensional Gaussian kernel to simulate the point spread function \cite{Mosaliganti12}. To generate multiview images with a single rotation of $180^\circ$, the multiplier used for signal attenuation was inverted and a point spread function that was analogously rotated by $180^\circ$ was used to convolve the images.

To simulate the dark current signal of the detector, a positive offset was added to all intensity values. The offset was empirically determined from light-sheet microscopy data. Furthermore, photon shot noise was simulated by an independent Poisson process at each voxel with the respective image intensities being its average \cite{Preibisch14}. Finally, the readout noise caused by signal amplification was simulated by zero-mean additive Gaussian noise with a standard deviation of $\sigma_{\text{agn}}$. Due to the relatively small size of the simulated embryo, light scattering caused by thick tissue was not considered in this benchmark. However, this might be a reasonable extension for upcoming implementations if simulations should get even more realistic.

The individual steps for modeling the acquisition deficiencies of an input image ${\mathbf{I}^{\text{raw}}}$, which was the raw benchmark image in this case, can be formulated as:
\begin{gather}
	\mathbf{I}^{\text{final}} = P_{\lambda}(\mathbf{I}^{\text{raw}} \ast \mathbf{I}^\text{psf} + \mathbf{I}^{\text{dark}}) + \mathcal{N}(0, \sigma_{\text{agn}}),
\end{gather}
where $\mathbf{I}^\text{psf}$ is the point spread function, $\mathbf{I}^{\text{dark}}$ is the dark current image of the detector, $P_\lambda$ applies a Poisson process to simulate photon shot noise and finally $\mathcal{N}(0, \sigma_{\text{agn}})$ is a normally distributed random variable with zero mean and a specified standard deviation $\sigma_{\text{agn}}$ \cite{Svoboda09}.

Based on manually identified parameters for the noise levels and image content, four benchmark datasets were generated for the validation experiments discussed in the following chapters (\texttt{SBDE1}-\texttt{SBDE4}) and the properties of the datasets are listed in \app\cref{tab:Appendix:BenchmarkDatasetsEmbryo}. Exemplary benchmark images of different developmental stages and using different signal-to-noise ratios are shown in \cref{fig:chap4:Benchmark:TimeSeries}B and \cref{fig:chap4:Benchmark:TimeSeries}C, respectively.
 
%%%%%%%%%%%%%%%%%%%%%%%%%%%%%%%%%%%%%%%%%%%%%%%%%%%%%%%%%%%%%%%%%%%%%%%%%%
%% Performance Assessment
%%%%%%%%%%%%%%%%%%%%%%%%%%%%%%%%%%%%%%%%%%%%%%%%%%%%%%%%%%%%%%%%%%%%%%%%%%
\subsection{Performance Assessment}
\label{sec:chap4:Benchmark:PerformanceAssessment}
To assess the performance of the individual image analysis operators, it was necessary to define specific optimality measures. Due to the available ground truth for the generated benchmark datasets it was possible to compare each of the processing steps to the underlying ground truth information. The following measures were used to assess the quality improvements of the seed detection, segmentation, multiview fusion and the tracking. A more detailed description of the employed measures and the testing platform is given in \cref{sec:Appendix:PerformanceAssessment}.

\subsubsection{Seed Detection} 
The seed detection quality was evaluated using the benchmark datasets \texttt{SBDE1} and \texttt{SBDE2} (\app\cref{tab:Appendix:BenchmarkDatasetsEmbryo}). The intersections of the detected seeds with the labeled ground truth image were calculated. True positives (TP) were counted as ground truth objects that contained at least one seed point. Seed points that were detected in background regions or redundant detections of ground truth objects were considered as false positives (FP). Ground truth objects that did not contain a seed point were counted as false negatives (FN). Using TP, FP and FN, recall, precision and the F-Score (harmonic mean of precision and recall) were calculated (\cref{sec:Appendix:SeedDetectionPerformance}). For all true positives, the average distance to the centroids of the respective ground truth objects was additionally calculated.

\subsubsection{Segmentation}
The segmentation quality was assessed using the benchmark datasets \texttt{SBDE1} and \texttt{SBDE2} (\app\cref{tab:Appendix:BenchmarkDatasetsEmbryo}). As the provided ground truth of the benchmark contained the complete label images of each frame, a detailed quantitative assessment of the automatic segmentation quality could be performed. The set of segmentation validation measures proposed by Coelho \etal was used, namely the Rand index (RI), the Jaccard index (JI), the normalized sum of distances (NSD) and the Hausdorff metric (HM). A detailed description of the measures can be found in \cite{Coelho09} and \cref{sec:Appendix:SegmentationEvaluation}. Topological errors produced by the automatic segmentation were separated into added, missing, split or merged objects. Besides the error counts, this topological information was used to define the number of false positives as the sum of split and added cells and analogously the false negatives as the sum of merged and missing cells. These values were then used to calculate recall, precision and F-Score (\cref{sec:Appendix:SeedDetectionPerformance}).

\subsubsection{Multiview Information Fusion}
The quality assessment of information extracted from multiview experiments was based on the \texttt{SBDE3} benchmark dataset (\app\cref{tab:Appendix:BenchmarkDatasetsEmbryo}). The benchmark dataset contained sequential single view images (SV, one image per time point), sequential multiview images with alternating rotation by $180^\circ$ (SeMV, one image per time point) and simultaneous multiview images (SiMV, two images per time point). As the simulated objects were moving from frame-to-frame, each of the complementary rotation images of the SeMV dataset captured a slightly different scene. The evaluation measures were exactly the same as for the segmentation assessment. However, instead of solely comparing a single view to it's ground truth dataset, the fused segmentation images of two complementary views were compared to the respective ground truth images. In the case of a sequential acquisition, \ie, an acquisition scheme where the specimen was imaged from one side at a time and was rotated to acquire the opposite views, the results of independently obtained segmentations were fused to a single view. This single fused image was compared to both ground truth frames from different time points individually. In contrast to this, the simultaneous acquisition produced two rotation images at each time point, \ie, here the segmentation results of two views of a single time point were fused and compared to the single ground truth image of this particular time point.

\subsubsection{Tracking}
To assess the tracking quality, the \texttt{SBDE4} dataset was used (\app\cref{tab:Appendix:BenchmarkDatasetsEmbryo}). The comparison of the investigated algorithms was performed using the \texttt{TRA} measure as described by Ma{\v{s}}ka \etal \cite{Maska14} (\cref{sec:Appendix:Tracking}). This measure was calculated by considering the tracking result as an acyclic oriented graph and by comparing this graph to the respective ground truth graph. The inverted, weighted and normalized number of required changes to transform the automatically generated graph to the ground truth graph yielded the normalized \texttt{TRA} measure (higher values are better with 1 being ideal). As the centroids of all objects, the complete temporal association and the object ancestry was known for the simulated \texttt{SBDE4} dataset, the required ground truth graph could directly be generated using this data. To obtain a more detailed view on the errors made by the respective tracking algorithms, the number of false positive detections, false negative detections, incorrect edges, missing edges, redundant edges and merged objects were counted. Detailed descriptions of the validation measures are provided in \cref{sec:Appendix:Tracking} and \cite{Maska14}.

%%%%%%%%%%%%%%%%%%%%%%% SEED DETECTION %%%%%%%%%%%%%%%%%%%%%%%%
%\newpage
\section{Seed Point Detection}
\label{sec:chap4:SeedDetection}
In \cref{sec:chap3:SeedDetection:LoGSSMP} a blob detection method that performed an extraction of local extrema in the Laplacian-of-Gaussian maximum intensity projection was introduced. Although, the proposed method worked well in many scenarios, it frequently missed objects that did not exhibit a clear local maximum due to an intensity plateau (\eg, elongated or overexposed objects). To overcome this issue, this section shows how the algorithm was tweaked to be more sensitive to the problematic cases, how redundant seed points could be efficiently fused and how the uncertainty framework proposed in \cref{sec:chap2:UncertaintyFramework} could be incorporated for efficient data filtering. The introduced steps further improved the automatic detection quality of this algorithm and only had a small impact on the processing times ($6\%-35\%$). Images with varying cell counts (\texttt{SBDE1}) and different noise levels (\texttt{SBDE2}) were used to assess the quality and time performance of the algorithms in different realistic scenarios that were similar to the problems observed in the domain of fluorescence microscopy of living embryos.

\subsection{Improved Detection and Fusion of Redundant 3D Seed Points}
\label{sec:chap4:SeedDetection:SeedFusion}
The plain LoG-based seed detection algorithm had difficulties with slightly elongated objects and missed objects that did not have a single strict maximum value, \eg, caused by overexposure of the images. To increase the detection rate even for these problematic cases, the $\leq$-operator was used instead of the $<$-operator to identify local extrema, \ie, the detected maxima did not have to be strictly larger than their neighborhood. This approach, however, yielded many false positive detections in background regions and along elongated objects that had to be further filtered and combined in a subsequent processing step.

As an initial step, an intensity threshold ($t_{\text{wmi}}$) was applied to the detected seed points using the window mean intensity feature as described in \cref{sec:chap3:SeedDetection:LoGSSMP}. The remaining seed points were mostly located properly on the detected objects and remaining false positive detections largely originated from objects that were detected multiple times. To combine redundant objects to a single one, a fusion approach based on hierarchical clustering was used. The hierarchical cluster tree was computed using Ward's minimum variance method to compute distances between clusters, \ie, the within-cluster variance was minimized \cite{WardJr63}. This method tended to produce clusters that were equally sized and thus rendered it well suited to fuse close maxima that were likely to belong to a single object. The outcome of the hierarchical clustering was a complete tree with an unknown amount of real clusters. Using a distance-based cutoff ($t_{\text{dbc}}$), the final clustering could be obtained from the complete tree. An appropriate cutoff distance could again be derived from prior knowledge about the object size and a good rule of thumb was to set it to the smallest expected object radius $r_{\text{min}}$. Using larger radii can cause a fusion of centroids of neighboring cells and may result in a fused detection in the middle of the centroids of the objects of interest. The feature vectors of all detected seeds within a cluster were then averaged to obtain a single detection per object.

\subsection{Extending Seed Detection Algorithms by Uncertainty Handling}
\label{sec:chap4:SeedDetection:UncertaintyIntegration}
To improve the performance of down-stream pipeline components such as a multiview fusion operator, the uncertainty of each extracted object at each pipeline stage was additionally estimated as described in \cref{sec:chap2:PriorKnowledge}. The seed detection stage usually represents one of the first analysis steps and therefore, it was assumed that there was no preceding uncertainty information available for the extracted objects. To estimate the uncertainty of the detected seed points, the window mean intensity, the maximum seed intensity and the z-position features of the objects were used. Besides discarding obvious false positive detections in the background regions, the uncertainties were adjusted such that seed detections in low contrast regions (farther away from the detection objective) had lower membership degrees to the class of correct objects than objects in the high contrast regions (closer to the detection objective).
\begin{figure}[htb]
\centerline{\includegraphics[width=\columnwidth]{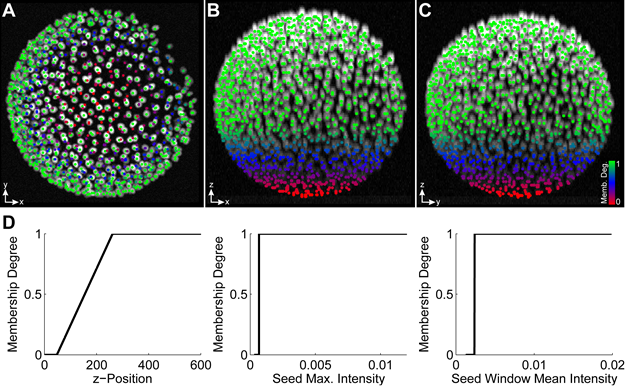}}
\caption[Visualization of the estimated uncertainty for the seed detection stage]{Visualization of the estimated fuzzy set membership functions for the seed detection stage using the z-position, the maximum seed intensity and the window mean intensity features of each seed point. For the intensity-based features, a fixed threshold was used due to the clear separation of background vs.\ foreground objects, \ie, the formulation as a fuzzy set was not explicitly required in this case. Seed points are color-coded according to their membership degree to the class of a correct detection and are superimposed on the respective maximum intensity projections along the Z, Y and X axis (A, B, C). The membership functions used for the individual features are depicted in (D) and the obtained membership degrees were multiplied for the final membership degree. The uncertainty gradient along the z-axis was introduced due to the signal attenuation at locations farther away from the detection objective and was used in later steps to resolve ambiguities during multiview fusion (adapted from \cite{Stegmaier16Arxiv}).}
\label{fig:chap4:SeedDetection:UncertaintyVisualizationAndFuzzySets}
\end{figure}

\noindent The following steps were performed to obtain the parameterization used for the uncertainty estimation depicted in \cref{fig:chap4:SeedDetection:UncertaintyVisualizationAndFuzzySets}:
\begin{itemize}
	\item Select the maximum intensity feature and adjust the first two parameters of the trapezoidal membership function to a threshold value that filters all obvious false positives without filtering any true positives.
	\item Select the seed mean intensity window feature and adjust the first two parameters of the trapezoidal membership function to a threshold value that filters all obvious false positives without filtering any true positives.
	\item Select the z-position feature and set the first parameter of the trapezoidal membership function to the z-position where the darkest objects that still were detected are present. The second parameter should be set somewhere in the bright-to-dark transition, such that ambiguous detections get lower membership values assigned than obviously correct detections.
\end{itemize}

The final fuzzy set membership degree of an object to the fuzzy set of correct objects was determined by the multiplication of all membership degrees of the individual features and was appended as a new feature to the output matrix of the seed detection algorithm (\cref{eq:chap2:AugmentedOutputMatrix}). In this particular example, the seed point maximum intensity and the seed window mean intensity feature are in principal redundant, and could be replaced by only using one or the other. As the window mean intensity should generally be more robust to noise, this feature should be favored. Additionally, no upper limits for the intensity based features were used, which originated from the assumption that the likelihood of a detection being an object of interest does not decrease with increasing intensity. Similarly, the optimal z-position values were not upper-bound as objects closer to the detection objective (higher z-position values in this case) should generally be easier detectable, due to low light attenuation. Of course, these assumptions might have to be adapted if different imaging conditions and specimens are investigated.

The parameterization of the respective fuzzy sets could be easily obtained using the semi-automated analysis approach as described in \cref{sec:chap5:FuzzyParameterAdjustment}. As the GUI provided an instant visual response to parameter changes, it was straightforward to determine the respective transition regions, even for inexperienced users.

\subsection{Validation}
To validate the proposed improvements of the LoG-based seed detection algorithm, the \texttt{SBDE1} and \texttt{SBDE2} benchmark datasets were used (\app\cref{tab:Appendix:BenchmarkDatasetsEmbryo}). The parameterization and description of the respective versions of the investigated algorithms is listed in \cref{tab:chap4:SeedDetection:ParameterizationTable}. Parameters for the scale range to use for the LoG detection were determined empirically, by measuring the radii of the smallest and the largest object that should lead to a valid detection. Here, the minimum and maximum object radii were measured to be $r_{\text{min}}=5.5$ and $r_{\text{max}}=11$, respectively. Using the relationship of $\sigma = r / \sqrt{2}$, yielded the standard deviations $\sigma_{\text{min}}=4$ and $\sigma_{\text{max}}=8$ for the LoG filtering. The intensity threshold parameters ($t_\text{wmi}, \pmb{\theta}_\text{wmi}, \pmb{\theta}_\text{smi}$) as well as the valid z-position range ($\pmb{\theta}_\text{zpos}$) were identified using the semi-automatic tool described in \cref{sec:chap5:FuzzyParameterAdjustment}.
\bgroup
\def\arraystretch{1.5}
\begin{table}[htbp]
\begin{center}
%\rowcolors{2}{white}{gray!25}
\resizebox{\textwidth}{!}{
\begin{tabular}{m{0.16\columnwidth}m{0.3\columnwidth}m{0.45\columnwidth}}
\toprule
\textbf{Method}& \textbf{Parameters}& \textbf{Description} \\
\midrule
LoGSM	& $\sigma_{\text{min}}=4$, $\sigma_{\text{max}}=8$, $\sigma_{\text{step}}=1$, $t_{\text{wmi}}=0.0025$ & Seed detection in the LoG scale space maximum projection with a manually adjusted window mean intensity threshold ($t_{\text{wmi}}$) and a strict maximum detection. \\
\midrule
LoGNSM & $\sigma_{\text{min}}=4$, $\sigma_{\text{max}}=8$, $\sigma_{\text{step}}=1$, $t_{\text{wmi}}=0.0025$ & Seed detection in the LoG scale space maximum projection with a manually adjusted window mean intensity threshold ($t_{\text{wmi}}$) and a non-strict maximum detection (\cref{sec:chap4:SeedDetection:SeedFusion}). \\
\midrule
LoGNSM+F & $\sigma_{\text{min}}=4$, $\sigma_{\text{max}}=8$, $\sigma_{\text{step}}=1$, $t_{\text{wmi}}=0.0025$, $t_{\text{dbc}}=5$ & Same detection as LoGNSM but with additional fusion (F) of redundant detections using a hierarchical clustering approach with a distance-based cutoff value ($t_{\text{dbc}}$). \\
\midrule
LoGNSM+F+U & $\sigma_{\text{min}}=4$, $\sigma_{\text{max}}=8$, $\sigma_{\text{step}}=1$, $t_{\text{dbc}}=5$, $\pmb{\theta}_{\text{wmi}} = (0.0025, 0.0025, \infty, \infty)$, $\pmb{\theta}_{\text{smi}} = (0.0007, 0.0007, \infty, \infty)$, $\pmb{\theta}_{\text{zpos}} = (50, 250, \infty, \infty)$, $\alpha_{11} = 0.0001$, $\beta_{11}=\alpha_{11}$ & Same detection as LoGNSM but with uncertainty-based (U) threshold and the fusion of LoGNSM+F. The forward threshold $\alpha_{11}$ is set slightly above zero, such that obvious false positives are rejected (\cref{sec:chap2:ObjectRejection}). As no further processing was needed $\beta_{11}$ was set to $\alpha_{11}$ (\cref{sec:chap2:InformationPropagation}). \\
\bottomrule
\end{tabular}}
\caption[Abbreviations, parameterizations and descriptions of the investigated seed detection algorithms]{Abbreviations, parameterizations and descriptions of the investigated seed detection algorithms \cite{Stegmaier16Arxiv}.}
\label{tab:chap4:SeedDetection:ParameterizationTable}
\end{center}
\end{table}
\egroup
The extracted seed points of all investigated algorithms were compared to the available ground truth and the quality was assessed using the measures described in \cref{sec:chap4:Benchmark} and \cref{sec:Appendix:PerformanceAssessment}. The obtained values are summarized in \cref{tab:chap4:SeedDetection:DetectionPerformance}, whereas each entry of the table corresponds to the arithmetic mean value of the independently obtained results on the ten benchmark images of \texttt{SBDE1}.
\begin{table}[htb]
\begin{center}
%\rowcolors{2}{white}{gray!25}
\resizebox{\textwidth}{!}{
\begin{tabular}{lccccccccc} %{p{3.5cm}p{6cm}p{5cm}}
\toprule
\textbf{Method}& \textbf{TP} & \textbf{FP} & \textbf{FN} & \textbf{Rec.} & \textbf{Prec.} & \textbf{F-Sc.} & \textbf{Dist.} & \textbf{Time (s)} & \textbf{KVox./s} \\
\midrule
LoGSM & 681.1 &	\textbf{3.3} &	202.8 &	0.77 & \textbf{1.00} & 0.87 & 1.60 & \textbf{7.23} & \textbf{7259.82} \\
LoGNSM & \textbf{813.7} & 160.0 & \textbf{70.2} & \textbf{0.91} & 0.84 & 0.87 & 1.64 & 7.66 & 6858.39 \\
LoGNSM+F & 811.7 & 4.5 & 72.2 & \textbf{0.91} & 0.99 & \textbf{0.95} & \textbf{1.59}	& 7.70 & 6819.58 \\
LoGNSM+F+U & 812.5 & 4.3 & 71.4 & \textbf{0.91} & 0.99 & \textbf{0.95} & \textbf{1.59} & 9.77 & 6170.05 \\
\bottomrule
\end{tabular}}
\caption[Quantitative assessment of the seed detection performance]{Quantitative performance assessment of the LoG-based seed detection methods. The criteria are true positives (TP), false positives (FP), false negatives (FN), recall, precision, F-Score, the distance to the reference (Dist., smaller values are better) as well as the achieved time performance measures in seconds (smaller values are better) and voxels per second (larger values are better). Values represent the arithmetic mean of the results obtained for the individually processed benchmark images (adapted from \cite{Stegmaier16Arxiv}).}
\label{tab:chap4:SeedDetection:DetectionPerformance}
\end{center}
\end{table}

The quantitative analysis confirmed that the proposed extensions of LoGSM could improve the algorithmic performance by up to $9.2\%$ with respect to the F-Score improvement. LoGSM had few false positive detections but on the other hand missed many objects due to the strict maximum detection (recall of $0.77$ and precision of $1.0$). The recall could be improved by $18.2\%$ to a value of $0.91$ by additionally allowing non-strict maxima (LoGNSM). However, this adaption concurrently raised the number of false positives and thus lowered the precision by $16.0\%$ to $0.84$, as objects with maximum plateaus were detected multiple times. These multi-detection errors could be successfully removed using the proposed fusion technique, which was reflected in an F-Score value of $0.95$ for LoGNSM+F(+U), \ie, compared to the LoGSM method, the F-Score was increased by $9.2\%$. Regarding the processing times, the additional effort for a redundant detection was almost negligible, as the non-strict maximum detection simply detected more seed points during the same iteration over the image. The seed point fusion was performed directly in the feature space and was therefore also insignificant compared to the preceding processing steps. For the feature set described here, using the uncertainty-based object rejection (LoGNSM+F+U) only slightly improved the results compared to directly fusing and filtering the data using the hard intensity threshold but increased the processing time by $35\%$. LoGNSM+F yielded almost identical results and required only $6\%$ more processing time compared to LoGSM. Nevertheless, all objects were equipped with an uncertainty value that was propagated through the pipeline and proved to be beneficial to filter, fuse and correct the extracted data in subsequent steps, as described in the respective chapters on image segmentation and multiview fusion. In addition, it should be noted that the processing time required for the image analysis easily exceeds the fuzzy set calculations as soon as the images get larger.

\begin{figure}[tb]
\centerline{\includegraphics[width=\columnwidth]{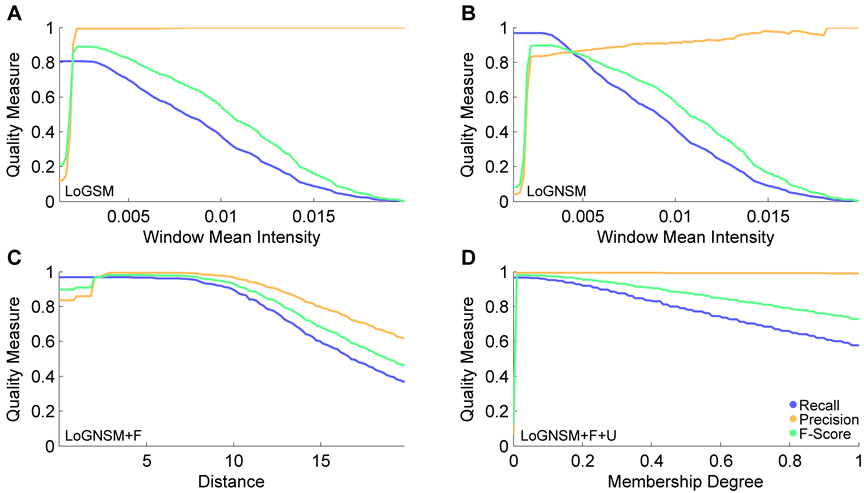}}
\caption[Parameter sensitivity analysis for the different versions of a LoG-based seed detection]{Parameter sensitivity analysis for the different versions of the LoG-based seed detection methods. Depicted curves are the achieved recall, precision and F-Score values for different parameter settings. The panels correspond to the plain seed detection with strict maximum detection (A), non-strict maximum detection (B) and a non-strict maximum detection with additional hierarchical clustering for seed point fusion (C, D). In (C) the threshold for the window mean intensity parameter was fixed at 0.0025 and the cutoff distance of the hierarchical clustering was varied. In (D) the threshold was applied on the membership degree with a fixed cutoff distance of $5$.}
\label{fig:chap4:SeedDetection:ParameterSensitivityLoG}
\end{figure}

To evaluate the influence of the parameter settings on the quality achieved by the different seed detection algorithms, a sensitivity analysis of the most critical parameters for all investigated algorithms was performed on a representative image stack of the benchmark dataset (\texttt{SBDE1}). For the LoG-based seed detection, the intensity-based threshold level was the most critical parameter. Furthermore, the fusion approach additionally required the distance-based cutoff value to be set properly. In \cref{fig:chap4:SeedDetection:ParameterSensitivityLoG}, the parameter sensitivity analysis of the three investigated LoG-based seed detections is depicted. The plain LoG with a strict maximum detection yielded high precision values but only a maximal recall of about $0.8$ in this case, as objects with intensity plateaus were not detected by the algorithm in contrast to the LoGNSM methods. Contrary to this, the non-strict maximum detection approach almost perfectly detected all objects (recall value of 0.99) but had a lower precision due to multiple detections of a single object that could not be filtered by the simple intensity threshold. Using the hierarchical clustering approach for seed point fusion as described earlier, multiple seed points could efficiently be combined to a single seed point based on a distance parameter. As indicated by the broad parameter range with excellent values of both precision and recall in \cref{fig:chap4:SeedDetection:ParameterSensitivityLoG}C, the parameter tuning for the seed point fusion is straightforward and the optimal distance can directly be inferred from the minimum expected radius of objects in the investigated specimen ($\approx$5 pixel in this example). If the distance for the hierarchical clustering was set too small, no fusions were performed and the results were identical to the LoGNSM method (see decreased quality for distances $\leq 2$ in \cref{fig:chap4:SeedDetection:ParameterSensitivityLoG}C). The detection quality achieved by varying the uncertainty-based threshold is depicted in \cref{fig:chap4:SeedDetection:ParameterSensitivityLoG}D and yielded the highest quality for membership degree thresholds slightly above zero. The uncertainty was basically only used to reject highly unlikely objects with an FSMD value of zero ($\alpha_{11}=0.0001$) and the remaining seeds were propagated unaltered to the next step ($\beta_{11}=\alpha_{11}$). Furthermore, all remaining objects preserved and propagated their FSMD value, which enabled improved decisions at down-stream processing steps.

\begin{figure}[htb]
\centerline{\includegraphics[width=\columnwidth]{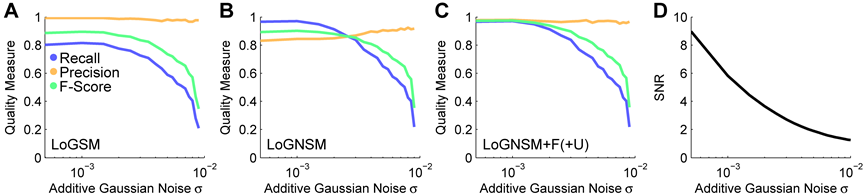}}
\caption[Assessment of the seed detection performance for different noise levels]{Assessment of the seed detection performance for the different noise levels of the \texttt{SBDE2} dataset. The performance measures recall, precision and F-Score are plotted versus the additive Gaussian noise level parameter $\sigma_{\text{agn}}$ for LoGSM (A), LoGNSM (B) and LoGNSM+F(+U) (C). As LoGNSM+F and LoGNSM+F+U produced identical results with respect to recall, precision and F-Score, the plots are combined to a single panel (\cref{tab:chap4:SeedDetection:DetectionPerformance}). The influence of the noise level on the signal-to-noise ratio of the images is plotted in (D) (adapted from \cite{Stegmaier16Arxiv}).}
\label{fig:chap4:SeedDetection:NoiseLevelInfluence}
\end{figure}

The performance of the seed detection under different image noise conditions was tested using the \texttt{SBDE2} dataset, which contained images with different settings for the additive Gaussian noise standard deviation ($\sigma_{\text{agn}} \in [0.0005, 0.01]$). Seed points from these images were extracted using the LoGSM, LoGNSM, LoGNSM+F and LoGNSM+F+U algorithms and the intensity thresholds were determined for each of the noise levels individually using the semi-automatic graphical user interface described in \cref{sec:chap5:FuzzyParameterAdjustment}. An analysis of increasingly blurred images as performed in \cref{sec:chap3:EfficientSegmentation} was skipped here, as the seed detection was highly robust up to smoothing levels of $\sigma^2_\text{smooth} = 30$ that are unlikely to be observed in practice (see \cref{fig:chap3:SeedDetection:NoiseLevelInfluence}). For higher noise levels, the number of detections in background regions heavily increased and it became ambiguous to determine true positive detections in low contrast regions. The manual threshold was therefore adjusted such that the false positive detections were minimized and only unambiguous seeds were considered. This continuous threshold adaption is also the reason for constant (\cref{fig:chap4:SeedDetection:NoiseLevelInfluence}A, C) or increasing precision (\cref{fig:chap4:SeedDetection:NoiseLevelInfluence}B), as it was easier to identify false positives rather than false negative detections in noisy image regions. Objects were robustly detected down to a signal-to-noise ratio of $5$ (\cref{fig:chap4:SeedDetection:NoiseLevelInfluence}), which was close to the visual limit of detection \cite{Murphy12} and emphasized the uncertainty-based improvements. The identified seed points and the associated FSMD values were subsequently propagated to the segmentation step to extract a label image of the actual objects as described in the next section.

%%%%%%%%%%%%%%%%%%%%%%% SEGMENTATION %%%%%%%%%%%%%%%%%%%%%%%%
%\newpage
\section{Segmentation}
\label{sec:chap4:Segmentation}
In \cref{sec:chap3:EfficientSegmentation}, three new algorithms for the automatic segmentation of different geometrical shapes have been presented. The segmentation task required to analyze the simulated benchmark introduced in \cref{sec:chap4:Benchmark} was to efficiently extract individual spherical objects from simulated 3D image stacks, \eg, using the algorithms presented in \cref{sec:chap3:TWANG}. Following the improved seed detection algorithm described in the previous section (LoGNSM+F+U), this section demonstrates how to further improve an algorithm based on adaptive thresholding (OTSUWW). Using propagated information from the seed detection stage and estimated segment FSMDs, both the algorithmic efficiency and the segmentation quality could be improved (OTSUWW+U). Furthermore, the segmentation quality obtained by the TWANG method (\cref{sec:chap3:TWANG}) was quantitatively compared to a TWANG version (TWANG+U) that relied on the improved seed detection introduced in the previous section (LoGNSM+F+U).

%%%%%%%%%%%%%%%%%% FUZZY FEATURE FUSION %%%%%%%%%%%%%%%%%%%
\subsection{Extending Segmentation Algorithms with the Uncertainty Framework}
\label{sec:chap4:Segmentation:UncertaintyEstimation}
As an initial step, fuzzy sets for the segmentation evaluation were parameterized in order to estimate the FSMDs of the individual segments to belong to the class of correct objects. Due to the availability of a ground truth, the statistical quantities of the segments could be directly extracted from the reference images. In the absence of ground truth data, the most critical step to manually obtain the fuzzy set parameterization is to estimate the transition regions by a detailed analysis of objects that deviate from the prior-knowledge-based expectation. In the presented case, the focus was put on the volume and size information of the objects in the benchmark images. Thus, to manually determine the described quantities, it is sufficient to investigate a small number of objects at the lower and the upper size and volume level, \eg, using software tools such as Fiji \cite{Schindelin12}. Of course, the more objects are analyzed, the better the statistics and the resulting fuzzy set membership functions will get. In practice, averaging over about $10$ objects at the lower and the upper end, respectively, should be sufficient.

\begin{table}[htb]
\begin{center}
%\rowcolors{2}{white}{gray!25}
\resizebox{\textwidth}{!}{
\begin{tabular}{lccccccc} %{p{3.5cm}p{6cm}p{5cm}}
\toprule
\textbf{Feature}& \textbf{Min} & \textbf{Max} & \textbf{Mean} & \textbf{Std} & \textbf{Median} & \textbf{5\% quantile} & \textbf{95\% quantile} \\
\midrule
Volume & 449 & 2016 & 993.6 & 247.2 & 990 & 617 & 1405 \\
Width & 13 & 31 & 19.9 & 2.6 & 20 & 15 & 24 \\
Height & 13 & 34 & 19.9 & 2.5 & 20 & 15 & 24 \\
Depth & 3 & 11 & 6.2 & 1.0 & 6 & 5 & 8  \\
\bottomrule
\end{tabular}}
\caption[Statistical quantities of the benchmark dataset]{Statistical quantities of the volume and the extents of all objects contained in the benchmark dataset. Minimum, maximum and quantile values were used to formulate fuzzy sets for each of the features. Individual fuzzy sets were combined using the minimum operator, to obtain a single membership degree value to the fuzzy set of being a valid object (adapted from \cite{Stegmaier16Arxiv}).}
\label{tab:chap4:Segmentation:GroundTruthStatistics}
\end{center}
\end{table}
Based on the extracted statistical quantities of the benchmark images that are summarized in \cref{tab:chap4:Segmentation:GroundTruthStatistics}, the parameter vector $\pmb{\theta}=(a,b,c,d)^\top$ of the trapezoidal fuzzy set membership function for each considered feature was derived using the minimum and maximum values as $a,d$ parameters, respectively. The remaining parameters $b,c$ were set to the 5\%-quantile and the 95\%-quantile. This parameterization ensured that all values smaller or larger than the maximum values got a membership degree of zero and that 90\% of the data range has a membership value of one assigned. The selection of the transition regions depends on the particular purpose the uncertainty should be used for, \ie, it can be used to tweak algorithms to a certain outcome.

\begin{figure}[htb]
\centerline{\includegraphics[width=\columnwidth]{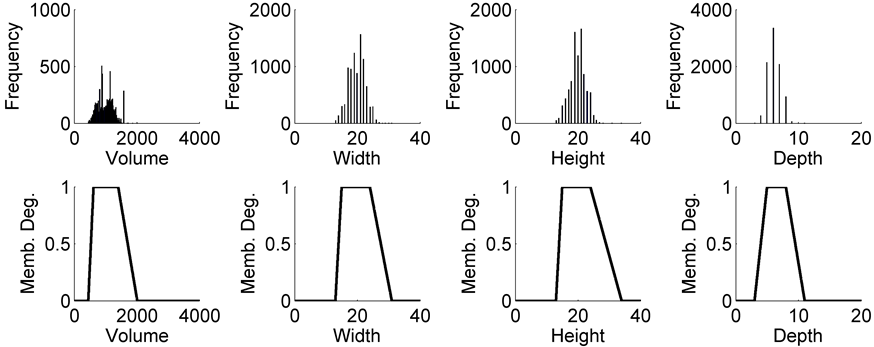}}
\caption[Feature histograms and derived fuzzy set membership functions for segmentation classification]{Feature histograms (top) and the derived fuzzy set membership functions (bottom) for volume ($\pmb{\theta}_\text{vol}=(449,617,1405,2016)^\top$), width ($\pmb{\theta}_\text{w}=(13, 15, 24, 31)^\top$), height ($\pmb{\theta}_\text{h}=(13, 15, 24, 34)^\top$) and depth ($\pmb{\theta}_\text{w}=(3, 5, 8, 11)^\top$). The fuzzy set membership functions were parameterized based on the minimum, maximum values (parameters $a, d$) as well as the 5\% and 95 \%-quantile values (parameters $b, c$) of the objects contained in the benchmark images (adapted from \cite{Stegmaier16Arxiv}).}
\label{fig:chap4:Segmentation:GroundTruthStatistics}
\end{figure}
As depicted in \cref{fig:chap4:Segmentation:GroundTruthStatistics}, the shapes of the fuzzy set membership functions derived from the statistical quantities directly resemble the respective distribution observed in the feature histograms. Another option to derive the membership function parameterization is to infer the transition regions via prior knowledge of the investigated specimen. The uncertainty of the output of the automatic segmentation can directly be assessed by evaluating the membership functions for each feature and by subsequently combining the obtained membership degrees to a single membership value based on the desired combination function (\cref{sec:chap2:MembershipFunctionCombination}). Here, the $\min$-operator was used to combine the individual fuzzy set membership degrees to a single value. The combined value directly corresponded to the membership degree of the feature that deviated the most from the specified expected range. Of course, the size criteria discussed here should only be considered as an exemplary illustration. There are various other features that can potentially be used to assess and improve segmentation results, \eg, integrated intensity, edge information, local entropy, local signal-to-noise ratios, principal components, weighted centroids and many more. Furthermore, if colocalized channels are investigated, complementary information can be used to formulate more complex decision rules. In the case of a fluorescently labeled nuclei and membranes that are imaged in different channels (\cref{sec:chap3:RACE}), rules like \textit{"each cell has exactly one nucleus"} can be formalized in the same way using the fuzzy set membership functions for a quantification of the available prior knowledge. In the next section the estimated uncertainties will be used to improve imperfect segmentation results and to speed up processing operators.

\subsection{Uncertainty Guided Segmentation Performance Improvement}
\label{sec:chap4:Segmentation:PerformanceImprovement}
The naive application of Otsu's method on the benchmark dataset produced an immense amount of merged nuclei. Based on the size criteria used to formulate the uncertainty of extracted objects, it was straightforward to determine, which of the objects needed to be further split to match the expected object size. Existing methods usually directly operated in the image domain to improve the segmentation results. For instance, Khan \etal \cite{Khan14} and Fernandez \etal \cite{Fernandez10} formulated a minimum volume criterion that was used in a seeded watershed segmentation \cite{Beare06}. Seed points were successively removed from the segmentation, until all segments fulfilled the specified criterion. However, as the methods directly operated in the image domain and applied the costly watershed transformation multiple times, their application to larger 3D images was rather limited \cite{Stegmaier16}. However, if high-quality seed points would be provided to the watershed algorithm, multiple executions could be avoided. This was realized in the investigated OTSUWW+U implementation using the seed points of the LoGNSM+F+U method described earlier. Additionally, this method could be further improved by applying the watershed-based splitting technique only onto objects that were known to be larger than expected. The estimated FSMD based on the size and volume criteria of manually investigated nuclei could be perfectly exploited to accomplish this task. Using a parallelization strategy similar to the one discussed in \cref{sec:chap3:TWANG}, all segments with combined FSMD values below $\beta_{21}$ (backward threshold for operator $2$ and linguistic term $1$) that corresponded to objects larger than expected (case 3, \cref{sec:chap2:FuzzySetMembershipFunctions}) were distributed among the available threads and a seeded watershed approach was used for object splitting in each of the cropped regions of the image. This approach was much faster than directly applying the watershed algorithm on the entire image (OTSUWW), due to the uncertainty guided, locally applied processing of erroneous objects. For simplicity, only a straightforward adaptive thresholding was used here. However, the methods can be analogously applied to erroneous segmentation results produced by more complex algorithms such as level-sets or graph-cuts \cite{Li10b, Al-Kofahi10, Stegmaier12a}. After splitting merged objects, the connected components of the image needed to be identified once more and the uncertainty values were re-evaluated as well, to provide the updated information to the subsequent processing operators.

The most critical segmentation error was the presence of many merged regions as described above. However, to further improve the results with respect to false positive detections observed for higher noise levels, segments with combined FSMD values below $\alpha_{21}=0.1$ (forward threshold for operator $2$ and linguistic term $1$), \ie, objects that were smaller than the expected object size (case 2 and case 4, \cref{sec:chap2:FuzzySetMembershipFunctions}) were removed from the label images. To facilitate the implementation of the uncertainty guided segmentation, it was realized using just a single fuzzy set for the correct class of objects (case 1, \cref{sec:chap2:FuzzySetMembershipFunctions}) to identify objects that needed further consideration. To determine if objects with low FSMD values were smaller or larger than the expectation, a comparison to the boundaries of the expected valid range was performed (trapezoidal fuzzy set parameters $b, c$). Threshold values were again identified using the interactive graphical user interface presented in \cref{sec:chap5:FuzzyParameterAdjustment}.

%%%%%%%%%%% VALIDATION ON THE BENCHMARK DATA %%%%%%%%%%%%%%
\subsection{Validation}
\begin{figure}[htb]
\centerline{\includegraphics[width=\columnwidth]{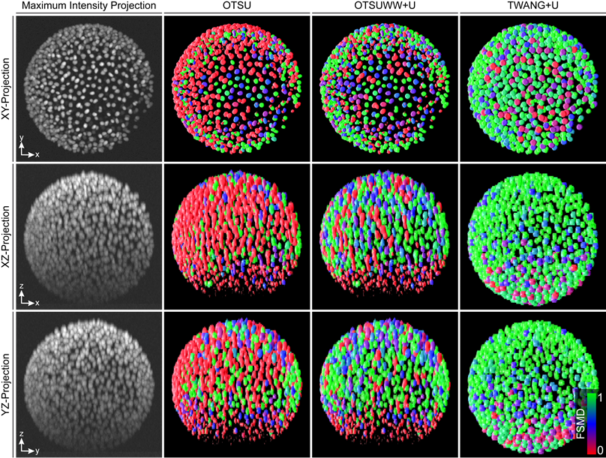}}
\caption[Uncertainty maps for the automatic segmentation]{Maximum intensity projections of the raw image and exemplary volume renderings of the automatic segmentation results produced by OTSU, OTSUWW+U and TWANG+U from different viewpoints (XY, XZ and YZ). The FSMD of individual detected objects was estimated based on morphological criteria such as volume and size (indicated by the color code from red over blue to green for low, medium and high FSMD values)  \cite{Stegmaier16Arxiv}.}
\label{fig:chap4:Segmentation:Comparison}
\end{figure}

To validate the segmentation performance of the described algorithms (OTSU, OTSUWW, OTSUWW+U, TWANG and TWANG+U), all methods were applied to the \texttt{SBDE1} and the \texttt{SBDE2} datasets (\app\cref{tab:Appendix:BenchmarkDatasetsEmbryo}).
In \cref{tab:chap4:Segmentation:ParameterizationTable}, the respective parameterization and a description of each algorithm is provided. The quantitative segmentation quality results obtained on \texttt{SBDE1} using the validation measures described in \cref{sec:chap4:Benchmark} are summarized in \cref{tab:chap4:Segmentation:DetectionPerformance}. The RI value was almost identical for all algorithms and OTSU yielded the highest value. The enhanced adaptive threshold-based techniques yielded an 11.0\% better JI value than TWANG+U and the best NSD value was obtained by OTSUWW+U. Considering the results for RI, JI and NSD, the global threshold-based techniques (OTSU*) produced slightly more accurate results (0.3\%, 5.3\% and 12.0\%, respectively) for the objects they were still able to resolve compared to the best results of the TWANG-based methods. Both TWANG-based methods on the other hand produced the minimal amount of topological errors with respect to split and merged objects compared to all OTSU-based methods. The number of added objects was minimal for OTSUWW+U. TWANG+U produced slightly more added objects than TWANG but efficiently detected far more objects. Thus, the F-Score values achieved by TWANG+U were further increased by $8.4$\% compared to TWANG, \ie, TWANG+U produced the best results with the fewest topological errors (F-Score $0.9$). The low amount of split and merged nuclei for TWANG originate from the single-cell extraction strategy, rather than using a global threshold as performed in the OTSU-based methods. The relatively large amount of added objects detected by the TWANG segmentation were mostly no real false positive detections, but segments where most of the extracted region intersected with the image background instead of the actual object and are thus considered as false positives (see low contrast regions in XZ-Projection and YZ-Projection images in \ref{fig:chap4:Segmentation:Comparison}).
\bgroup
\def\arraystretch{1.5}
\begin{table}[p]
\begin{center}
%\rowcolors{2}{white}{gray!25}
\begin{tabular}{m{0.15\textwidth}m{0.14\textwidth}m{0.63\textwidth}}
\toprule
\textbf{Method}& \textbf{Parameters}& \textbf{Description} \\
\midrule
OTSU	& $\sigma_{\text{smooth}}=1$ & Adaptive thresholding using Otsu's method applied on a regularized raw input image \cite{Otsu79}. \\
\midrule
OTSUWW & $\sigma_{\text{smooth}}=1$ & Same segmentation as OTSU. Merged regions in the entire image were split based on cleaned seeds of the LoGSM method, using a seeded watershed approach. \\
\midrule
OTSUWW+U & $\sigma_{\text{smooth}}=1$, $\alpha_{21}=0.1$, $\beta_{21}=0.1$ & Same segmentation as OTSU. Small noise objects with FSMD values below $\alpha_{21}=0.1$ were rejected from further processing. Merged regions with FSMD below $\beta_{21}=0.1$ that were larger than expected were split locally and in parallel using the cleaned and fused seeds of the LoGNSM+F+U method and a seeded watershed approach in small cropped regions. \\
\midrule
TWANG & $\sigma_{\text{grad}}=3$, $\sigma_{\text{kernel}}=3$, $\omega_{\text{kpm}}=1.41$ & TWANG segmentation as described in \cref{sec:chap3:TWANG} using the cleaned seeds provided by the LoGSM method. The manually optimized parameter $\omega_{\text{kpm}}=1.41$ yielded better results than the default value of $1.0$ for this dataset.  \\
\midrule
TWANG+U & $\sigma_{\text{grad}}=3$, $\sigma_{\text{kernel}}=3$, $\omega_{\text{kpm}}=1.41$, $\alpha_{21}=0$, $\beta_{21}=0$ & TWANG segmentation as described in \cref{sec:chap3:TWANG} using the cleaned and fused seeds provided by the LoGNSM+F+U method. $\alpha_{21}=0$, $\beta_{21}=0$ are adjusted such that all segments are propagated unchanged. \\
\bottomrule
\end{tabular}
\caption[Abbreviations, parameters and descriptions of the segmentation algorithms]{Abbreviations, parameters and descriptions of the segmentation algorithms \cite{Stegmaier16Arxiv}.}
\label{tab:chap4:Segmentation:ParameterizationTable}
\end{center}
\end{table}
\egroup
Regarding the processing times OTSU was the fastest approach, but due to the poor quality without the uncertainty-based extension it was not really an option for a reliable analysis of the image data. TWANG and TWANG+U were $1.8$ times slower than the plain OTSU method but $2.4$ and $3.8$ times faster than OTSUWW and OTSUWW+U, respectively. At the same time, TWANG+U was the most precise approach (F-Score $0.9$). Furthermore, OTSUWW+U was $1.6$ times faster than OTSUWW due to the focused object splitting and additionally produced better results due to the improved seed detection and noise reduction.
\begin{sidewaystable}[p]
\begin{center}
%\rowcolors{2}{white}{gray!25}
\resizebox{\textwidth}{!}{
\begin{tabular}{lccccccccccccc} %{p{3.5cm}p{6cm}p{5cm}}
\toprule
\textbf{Method}& \textbf{RI} & \textbf{JI} & \textbf{NSD ($\times$10)} & \textbf{HM} & \textbf{Split} & \textbf{Merged} & \textbf{Added} & \textbf{Missing} & \textbf{Rec.} & \textbf{Prec.} & \textbf{F-Score} & \textbf{Time (s)} & \textbf{KVoxel/s} \\
\midrule
OTSU & \textbf{97.92} & 27.82 & 3.77 & 6.67 & 25.00 & 370.50 & 42.40 & 264.00 & 0.29 & 0.78 & 0.42 & \textbf{6.11} & \textbf{8589.46} \\
OTSUWW & 97.91 &	\textbf{27.99} &	3.29 &	6.48 &	88.60 &	96.30 &	57.90 &	276.20 &	0.58 &	0.78 &	0.67 &	42.41	&	1237.72 \\
OTSUWW+U & 97.91 & 27.98 & \textbf{3.15} & 5.61 & 50.60 & 57.80 & \textbf{13.30}	& 268.40 & 0.64 & \textbf{0.90} & 0.75 & 26.40 & 2035.66 \\
TWANG & 97.67 &	26.59 &	3.58 &	\textbf{4.62} &	\textbf{0.00} &	\textbf{6.70} &	75.80 &	193.50 &	0.77 &	\textbf{0.90}	& 0.83 &	11.02	& 4774.83 \\
TWANG+U & 97.81 & 25.30 & 3.64 & 4.75 & \textbf{0.00} & 12.40 & 103.70 & \textbf{59.50} & \textbf{0.91} & 0.89 & \textbf{0.90} & 11.15 & 4728.84 \\
\bottomrule
\end{tabular}}
\captionsetup{width=0.96\textheight}
\caption[Quantitative assessment of the segmentation performance]{Quantitative performance assessment of the segmentation methods OTSU, OTSUWW, OTSUWW+U, TWANG and TWANG+U. The criteria used to compare the algorithms are the Rand index (RI), the Jaccard index (JI), the normalized sum of distances (NSD) and the Hausdorff metric (HM) as described in \cref{sec:chap4:Benchmark} and \cref{sec:Appendix:PerformanceAssessment}. Additionally, the topological errors were assessed by counting split, merged, added and missing objects. Precision, recall and F-Score are based on the topological errors by considering split and added nuclei as false positives and merged and missing objects as false negatives, respectively. The achieved time performance was measured in seconds (smaller values are better) and voxels per second (larger values are better). All values represent the arithmetic mean of the individually processed benchmark images (adapted from \cite{Stegmaier16Arxiv}).}
\label{tab:chap4:Segmentation:DetectionPerformance}
\end{center}
\end{sidewaystable}
These results confirmed that the uncertainty information could be efficiently exploited to guide and constrain computationally demanding processing operators to specific locations and thus, to speed up processing operations while the result quality was preserved or even improved.
%\clearpage

Exemplary FSMDs of the final segmentation results of different algorithms are depicted in \cref{fig:chap4:Segmentation:Comparison} and provide a convenient visualization to instantly assess the segmentation quality and to identify potential problems of the methods even by non-experts. All algorithms suffered from the light attenuation in the axial direction. Especially, the techniques that relied on a single global intensity threshold had problems to identify the objects located in these low contrast regions. Besides missing many objects, OTSU additionally merged many of the high intensity objects to a single large blob. Especially, in the z-direction many mergers occurred due to the lower sampling in this direction. However, these merged regions could be successfully split to a large extent using the proposed seed-based splitting techniques (OTSUWW, OTSUWW+U). As TWANG directly operated on the provided seeds, it was still able to extract most of the objects in these regions and yielded even higher recall values using the LoGNSM+F+U seed points. However, due to the low contrast, the segmentation quality of the extracted segments in these regions was reduced. In \cref{tab:chap4:SeedDetection:DetectionPerformance}, this is reflected by the increased number of added objects for TWANG and TWANG+U, which were mostly no real false positives as described above.
\begin{figure}[hbt]
\centerline{\includegraphics[width=\columnwidth]{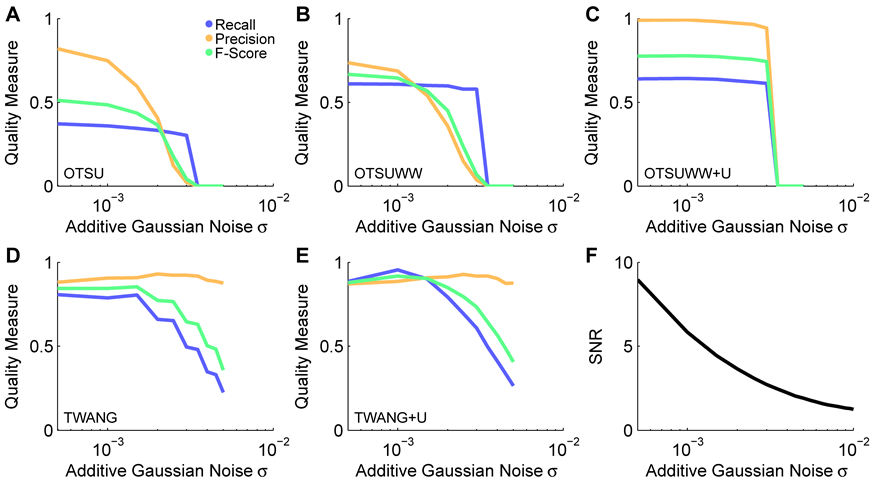}}
\caption[Assessment of the segmentation performance for different noise levels]{Performance evaluation of the segmentation methods OTSU (A), OTSUWW (B), OTSUWW+U (C), TWANG (D) and TWANG+U (E) on images of the \texttt{SBDE2} dataset with different signal-to-noise ratios. The methods based on adaptive thresholding (OTSU, OTSUWW) suffered from high noise levels and produced a successively increased amount of false positive detections, which could be efficiently suppressed using the uncertainty framework-based extension (OTSUWW+U). The result quality of both TWANG versions directly correlated with the quality of the provided seed points, \ie, TWANG+U benefited from the improved detection rate of LoGNSM+F+U (adapted from \cite{Stegmaier16Arxiv}).}
\label{fig:chap4:Segmentation:NoiseLevelInfluence}
\end{figure}

To investigate the impact of the signal-to-noise ratio of the images to the segmentation quality, the benchmark dataset \texttt{SBDE2} was processed using all five algorithms. The segmentation quality of all adaptive thresholding-based methods was heavily affected by the noise level of the images yielding poor precision and recall values even for the lowest noise levels (\cref{fig:chap4:Segmentation:NoiseLevelInfluence}A, B). This was caused by the global threshold that on the one hand merged a lot of objects and on the other hand detected a high amount of false positive segments. The uncertainty-based method OTSUWW+U successfully preserved the increased recall of OTSUWW and at the same time substantially increased the precision to an almost perfect level for noise parameters of $\sigma_\text{agn} < 0.003$ (\cref{fig:chap4:Segmentation:NoiseLevelInfluence}C). The increasing number of small segments observed for OTSU and OTSUWW could be efficiently filtered using the uncertainty-based object rejection. As TWANG heavily depended on the quality of the provided seeds, the observed curves in \cref{fig:chap4:Segmentation:NoiseLevelInfluence}C, D show a high correlation to the seed detection performance (\cref{fig:chap4:SeedDetection:NoiseLevelInfluence}A, C) and render it as a suitable method even for higher noise levels. The improved seed detection of LoGNSM+F+U also directly affected the quality of the TWANG+U segmentation. Note that the seed points were interactively adjusted for each of the noise levels, \ie, subtle variations in the precision and recall (\cref{fig:chap4:Segmentation:NoiseLevelInfluence}D, E) values are caused by a subjective manual threshold adaption. As in the previous section, increasing blur levels were not considered here, as the seed detection and thus the segmentation worked robustly up to unrealistically high blur levels (see \cref{fig:chap3:TWANG:NoiseLevelInfluence}). In the following section, the fusion of redundant segmentation information obtained from multiview experiments will be presented. This represents an approach to partially compensate light attenuation and scattering effects that occur when imaging large specimens.

%\clearpage

%%%%%%%%%%%%%%%%%%%%%%% REGISTRATION %%%%%%%%%%%%%%%%%%%%%%%%
\section{Extended Multiview Information Fusion}
\label{sec:chap4:MultiviewFusion}
The segmentation results of the approaches presented in the previous section were highly influenced by the decreased image quality along the axial direction. An approach to partly overcome these quality deficiencies caused by attenuation and scattering of light is the acquisition of multiple views from different perspectives. Such complementary image stacks of a specimen can be obtained either by using multiple oppositely arranged detection paths \cite{Tomer12, Chhetri15} or by a rotation of the probe using a single camera \cite{Kobitski15}. Both strategies for the acquisition of multiview images require a subsequent fusion step of the information present in the individual views into a single consistent representation. As a first step, the images have to be aligned properly, \ie, the optimal transformation of the images onto one another has to be identified (\cref{sec:chap1:RegistrationAndFusion}). Most methods proceed with a fusion of the images to form a single high quality image that contains only the best parts of each image and perform all further analysis steps on the fused image \cite{Rubio-Guivernau12, Tomer12}. As shown in \cref{sec:chap6:TWANG}, this fusion of the images may introduce artifacts due to object movements that happen during the sequential acquisition of multiple views. Hence, the approach presented here represents a different way to combine the data obtained from different views, namely the seed point detection and segmentation are applied on the individual rotation images and the multiview fusion of the extracted information is performed after the separately performed image analysis. Naturally, only image regions with sufficient quality should contribute to the fused image and the previously estimated uncertainty information is thus exploited in order to obtain a high quality fusion image.

\subsection{Uncertainty-based Fusion of Extracted Objects}
The fusion of extracted objects can either be purely performed on the feature level or incorporate the segmentation results to identify correspondences between complementary image stacks. The fusion in the feature space can be performed analogously to the fusion of redundant seeds (\cref{sec:chap4:SeedDetection:SeedFusion}) and will thus not be analyzed further in this section. One straightforward improvement of the method presented in \cref{sec:chap4:SeedDetection:SeedFusion} with respect to multiview acquisition is to fuse neighboring seeds using a convex combination of the detected seeds based on the respective uncertainty values.

Alternatively, a segmentation-based fusion could be performed to obtain a fused segmentation label image. For convenience, it was assumed that the image transformation of $180^{\circ}$ was already known in advance and the registration process was therefore skipped. The inputs of the fusion algorithm were two complementary labeled segmentation images of the images that needed to be fused. It was crucial to have unique labels in each of the images to avoid ambiguities of the segment association step. Corresponding segments were identified using a histogram-based approach similar to the one presented in \cref{sec:chap3:RACE:IntersectionCalculation}. However, instead of performing a slice-by-slice intersection calculation, the intersection calculations were directly performed in 3D, \ie, the label histogram was filled by iterating over the voxels of both images and by successively increasing the 2D histogram bins indicated by the respective image label pairs. Segments that were only present in one or the other image could be found by searching for empty columns and rows of the histogram, respectively, and by copying these segments without any further consideration into the new result image. In the next step, the assignments of the remaining segments in both images were identified by searching for the respective label pair with the maximum overlap. In the case of segmentation fusion it was not desired to perform a weighted average of the segments but rather to use the better segment for the new image. This was accomplished by simply selecting the segment with a higher FSMD value to the desired class of objects. Here, the size-based fuzzy sets specified in \cref{sec:chap4:Segmentation:UncertaintyEstimation} were used in combination with the information about the axial localization of the extracted objects as specified in \cref{sec:chap4:SeedDetection:UncertaintyIntegration}. The uncertainty information was directly provided for each of the extracted segments from the uncertainty quantification performed in the preceding processing steps. Besides the fusion, no further processing of the data was performed at this step, \ie, the uncertainty framework parameters were set to $\alpha_{31}=0$ (forward threshold of operator $3$ and linguistic term $1$) and $\beta_{31}=0$ (backwards threshold of operator $3$ and linguistic term $1$), to propagate all segments unchanged from the fusion.

\subsection{Validation}
To validate the proposed uncertainty-based multiview segmentation fusion (SF) approach, the two best algorithms identified from the previous section (OTSUWW+U and TWANG+U) were applied to the \texttt{SBDE3} dataset (\app\cref{tab:Appendix:BenchmarkDatasetsEmbryo}). The segmentation quality was assessed for three different acquisition strategies, namely a sequential single view acquisition (SV), a sequential acquisition with a rotation by $180^{\circ}$ after each time point (SeMV) and a simultaneous acquisition scheme with oppositely orientated detection paths (SiMV). In the case of SV and SeMV one image was analyzed per time point and two images were analyzed per time point for SiMV. In order to compare the proposed uncertainty-based segmentation fusion approach to a classical intensity-based fusion on the raw image level (IF), an additional set of multiview images that were combined to a single fused input image using a linear blending of the intensity values with respect to the rotation of both views was segmented with the OTSUWW+U and the TWANG+U algorithms. In \cref{tab:chap4:MultiViewFusion:DetectionPerformance}, the results of the quantitative analysis are summarized.
\begin{sidewaystable}[p]

\begin{center}
\resizebox{\textwidth}{!}{
%%\rowcolors{2}{white}{gray!25}
\begin{tabular}{lccccccccccccc} %{p{3.5cm}p{6cm}p{5cm}}
\toprule
\textbf{Method}& \textbf{RI} & \textbf{JI} & \textbf{NSD*} & \textbf{HM} & \textbf{Split} & \textbf{Merg.} & \textbf{Add.} & \textbf{Miss.} & \textbf{Rec.} & \textbf{Prec.} & \textbf{F-Sc.} & \textbf{Time (s)} & \textbf{KVoxel/s} \\
\midrule
OTSUWW+U (SV) 			& 97.91 & \textbf{27.98} & 3.15 & 5.61 & 50.60 & 57.80 & 13.30 & 268.40 & 0.64 & 0.90 & 0.75 & 26.40 & 2035.66 \\
OTSUWW+U+SF (SeMV)      & 97.93 & 26.03 & 3.21 & 4.91 & 5.90 & 43.30 &  68.80 & 79.70 & 0.85 & 0.91 & 0.88 & 66.40/121.40 & 442.68/809.36 \\
OTSUWW+IF+U (SeMV)      & 98.05 & 25.68 & 3.28 & 4.62 & 1.80 & 64.00 & 44.00 & 17.60 & 0.90 & 0.94 & 0.92 & 21.92 & 2451.28 \\
OTSUWW+U+SF (SiMV)      & 98.26 & 27.19 & 2.97 & 4.10 & 8.90 & 47.40 & 4.60 & 59.10 & 0.87 & \textbf{0.98} & 0.92 & 132.80/242.80 & 221.34/404.68 \\
OTSUWW+IF+U (SiMV)      & \textbf{98.40} & 26.97 & 3.01 & 3.79 & 2.50 & 65.40 & \textbf{3.70} & \textbf{13.70} & 0.90 & 0.99 & \textbf{0.94} & 41.40 & 1298.10 \\

TWANG+U (SV)   			& 97.81 & 25.30 & 3.64 & 4.75 & \textbf{0.00} & 12.40 & 103.70 & 59.50 & 0.91 & 0.89 & 0.90 & \textbf{11.15} & \textbf{4728.84} \\
TWANG+U+SF (SeMV)       & 97.89 & 26.01 & 3.36 & 4.64 & 0.60 & \textbf{5.20} & 91.00 & 80.00 & 0.89 & 0.89 & 0.89 & 51.15/106.15 & 496.72/1030.82 \\
TWANG+IF+U (SeMV)       & 97.82 & 25.47 & 3.76 & 4.86 & 0.40 & 6.80 & 105.40 & 67.20 & 0.91 & 0.88 & 0.89 & 25.68 & 2058.21 \\
TWANG+U+SF (SiMV)       & 98.19 & 27.06 & \textbf{2.94} & \textbf{3.71} & 0.30 & 6.70 & 21.90 & 54.00 & \textbf{0.92} & 0.97 & \textbf{0.94} & 102.30/212.30 & 248.36/515.41 \\
TWANG+IF+U (SiMV)       & 98.14 & 26.69 & 3.13 & 3.83 & \textbf{0.00} & 7.70 & 28.10 & 53.10 & \textbf{0.92} & 0.97 & \textbf{0.94} & 26.15 & 2016.31 \\
\bottomrule
\end{tabular}}
\captionsetup{width=0.96\textheight}
\caption[Quantitative assessment of the multiview fusion performance]{Quantitative performance assessment of the segmentation methods OTSUWW+U and TWANG+U applied on different imaging scenarios. The comparison was performed using sequential single view acquisition (SV), sequential multiview acquisition (SeMV) that rotated the specimen by $180^{\circ}$ after each frame and simultaneous multiview acquisition (SiMV) that performed a simulated imaging at each time point with two oppositely arranged detection paths. Images were either fused by the uncertainty-based segment fusion (SF) or using an intensity-based fusion prior to segmentation (IF). The criteria used to compare the algorithms are the Rand index (RI), the Jaccard index (JI), the normalized sum of distances (NSD $\times 10$) and the Hausdorff metric (HM) as described in \cref{sec:Appendix:SegmentationEvaluation}. Additionally, the topological errors were assessed by counting split, merged, added and missing objects. Precision, recall and F-Score are based on the topological errors by considering split and added nuclei as false positives and merged and missing objects as false negatives, respectively. Performing the fusion in the image space took about $15~s$ per image pair and about $80-190~s$ depending on the number of cells when performed on the segmentation results. The time measurements presented here are the total processing time required to process one frame including segmentation and fusion. Both fusion algorithms were not optimized for speed and implemented in MATLAB. It should be noted that both versions can be easily parallelized and would presumably benefit from a C++-based implementation.}
\label{tab:chap4:MultiViewFusion:DetectionPerformance}
\end{center}
\end{sidewaystable}
The SV results were identical to those obtained in the previous section, where especially the global thresholding methods had difficulties to properly extract the segments in low contrast image regions yielding a low recall value of $0.64$ for OTSUWW+U. Compared to the SV version of OTSUWW+U, the SeMV version's recall was improved by $32.8\%$ to a value of $0.85$ (SF) and by $40.6\%$ to a value of $0.90$ (IF), respectively, due to a significantly lower amount of split and missing objects. Although there were slightly more added objects, the split counts also slightly decreased, yielding a precision improvement of $1.1\%$ (SF) and $4.4\%$ (IF), respectively. The result improvements were even more obvious on the SiMV datasets with a greatly improved precision and recall, yielding an increase of the F-Score value by $22.7\%$ (SF) and $25.3\%$ (IF) to $0.92$ (SF) and $0.94$ (IF) compared to an F-Score of $0.75$ obtained by the SV version of the algorithm. Except for the JI value, all other shape dependent measures (RI, NSD, HM) also benefited from the fusion approach. Comparing the results of the SF and the IF fusion approaches, OTSUWW+U worked slightly better on the IF version of the images and obtained the best values for the SiMV datasets. As the segment-based fusion was $3.2-5.9$-fold slower with the MATLAB-based implementation presented here, the image-based fusion should be preferred if the time resolution is high enough.

The TWANG+U algorithm generally works better in high contrast regions that allow an unambiguous foreground identification. Thus, all shape dependent measures obtained by the TWANG+U segmentation improved in both fusion scenarios, with the best results obtained on the SiMV datasets. Interestingly, the topological segmentation quality of the TWANG+U method was slightly better on the SV dataset as compared to the SeMV dataset (\eg, $1.1\%$ better F-Score on the SV dataset). This was caused by the fact that the seed detection and consequently the segmentation worked already pretty well on the SV images. The comparison of the two different ground truth images to a single fused image therefore performed slightly worse than comparing the two SV segmentations to the two ground truth images with a decreased recall by $2.2\%$ (SF) and a decreased precision by $1.1\%$ (IF). However, this behavior completely changes as soon as the seed detection fails to sufficiently extract the object locations from a single view, where it becomes inevitable to improve the detection quality using multiple views (\cref{sec:chap6:TWANG}). Applying TWANG+U on the SiMV datasets improved most validation measures in comparison to the SV datasets except of a negligible increased amount of split objects. Especially, the number of added objects could be greatly reduced leading to the best overall F-Score of $0.94$. TWANG generally had a lower amount of split and merged cells compared to OTSUWW. As soon as a valid seed point was found, the algorithm successfully extracted the object, due to the explicit prior information about the object size. In contrast to this, OTSUWW initially segmented many merged objects that needed to be split. Some ambiguous split cases that were erroneously split by the watershed algorithm caused the increased number of merged and split objects ($20.2\%$ lower F-Score value compared to the best TWANG results). On the other hand, TWANG entirely relies on the seed detection stage and can not directly influence the detection rate and precision. Thus, TWANG tends to have a higher number of added and missing objects compared to OTSUWW (up to $5.3\%$ lower F-Score value in the worst case scenario). The TWANG results obtained using the SF and IF fusion techniques were almost identical, \ie, the uncertainty-based segmentation fusion represents a reasonable alternative to the intensity-based fusion with respect to the segmentation quality. However, with the current MATLAB-based implementation, the time performance of the segment fusion approach was not yet convincing with $3.9-8.1$-times higher processing times compared to the image-based fusion approach. As the segment fusion approach needed to run two independent segmentations on each of the images, the processing time was already increased 2-fold and required a fusion of the segmented objects afterwards. For both algorithms, approximately $15s$ were needed for the image-based fusion of two stacks and the segment-based fusion took $80-190s$ depending on the number of objects. It should be noted, though, that the same implementation in C++ or even using a GPU implementation will certainly make the segment-based fusion competitive and suitable for large-scale analyses.

These results lead to the general conclusion, that a successful fusion of the segmentation results using the proposed uncertainty framework performs reasonably well in all investigated scenarios. As expected, the simultaneous acquisition technique performed better than the sequential acquisition due to the perfect complementarity of the views. Of course, a sequential multiview acquisition could be improved by a higher sampling rate of the acquired rotation images, such that object movements are compensated sufficiently. Both multiview acquisition approaches significantly improved the segmentation quality obtained by the OTSUWW+U algorithm and finally render this straightforward method with the proposed extensions suitable for the considered image analysis task. Nevertheless, the best choice with respect to a combination of speed and quality was the TWANG+U algorithm that was specifically designed to work on this kind of image data and benefited from both the improved seed points and the additional fusion of complementary frames. The image-based multiview fusion method should be preferred if the temporal resolution is high enough, as the segmentation only needs to be run once in this case. However, if more complex fusion techniques such as wavelet multiview fusion \cite{Rubio-Guivernau12} would be required, the segment-based fusion might actually be faster.

A potential extension of the benchmark that is up to future work would be to generate rotation images with different temporal sampling rates, to identify the maximally allowed time shift between the acquisition of sequential multiview data that still yields valid segmentation results. The next section illustrates how the fused segmentation results can be used to temporally track dynamic objects.

%%%%%%%%%%%%%%%%%%%%%%% TRACKING %%%%%%%%%%%%%%%%%%%%%%%%
\section{Tracking}
\label{sec:chap4:Tracking}
The final step of the investigated image analysis pipeline was the tracking of all detected objects. As described in \cref{sec:chap1:Tracking}. the basic intention of the tracking procedure is to identify the correct correspondences of detected objects in subsequent time points of the acquired images. For illustration purposes, a straightforward nearest-neighbor tracking approach was used, \ie, for each object present in a frame, the spatially closest object in the subsequent frame was identified and stored as a matching partner. This procedure was applied on every frame of the dataset in order to get a complete linkage of all objects. The tracking algorithm obtained the detected, segmented and fused objects directly from the previous step in a matrix representation similar to \cref{eq:chap2:AugmentedOutputMatrix}. The most important features are the centroid of each object as well as the propagated membership degrees if required for the processing. In addition to the comparison of the segmentation methods introduced in \cref{sec:chap4:Segmentation}, an alternative approach that combines a flawed segmentation with provided seeds is presented in this section (OTSU+NN+U).

\subsection{Resolving Tracking Errors using Propagated Information}
\label{sec:chap4:SeededOtsuTracking}
As indicated in the last sections, it was possible to improve segmentation results produced by adaptive thresholding techniques such as OTSU using a watershed-based object splitting (OTSUWW, OTSUWW+U). These improved segmentation results were then propagated to down-stream operators such as multiview fusion or tracking for further processing. Another possibility, however, is to leave the flawed segmentation results produced by the respective algorithms unchanged and additionally provide detected seed points to the tracking algorithm instead of actually using the seed points for splitting in the image domain \cite{Stegmaier12a, Stegmaier16Arxiv}. For illustration purposes, an Otsu-based threshold was applied to both rotation images independently and the resulting binary images were then fused by simply using the maximum pixel value of the two images. As shown in \cref{tab:chap4:Segmentation:DetectionPerformance}, the segmentation quality achieved by OTSU is in principle not reasonably usable without the watershed-based object splitting. Nevertheless, the tracking algorithm could be extended to decide which of the information was reliable and suitable for tracking and could optionally fall back to the provided seed points if the segmentation quality was insufficient. Here, this was simply achieved by the same fuzzy set classifications as used in chapter \cref{sec:chap4:Segmentation:UncertaintyEstimation}. Using an empirically determined threshold of $\beta_{41} = 0.9$ (backward threshold for operator $4$ and linguistic term $1$), all objects with an aggregated FSMD value lower than the threshold were not tracked with the actual segment, but with the seed points the respective segment contained. The forward threshold parameter was set to $\alpha_{41} = 0.0$ in order to report all tracking results. Of course, if a further filtering of the obtained tracking results is desired, the algorithms could be adapted appropriately. The next section summarizes the quantitative results obtained by the different pipelines.

\subsection{Validation}
\begin{figure}[p]
\centerline{\includegraphics[width=\columnwidth]{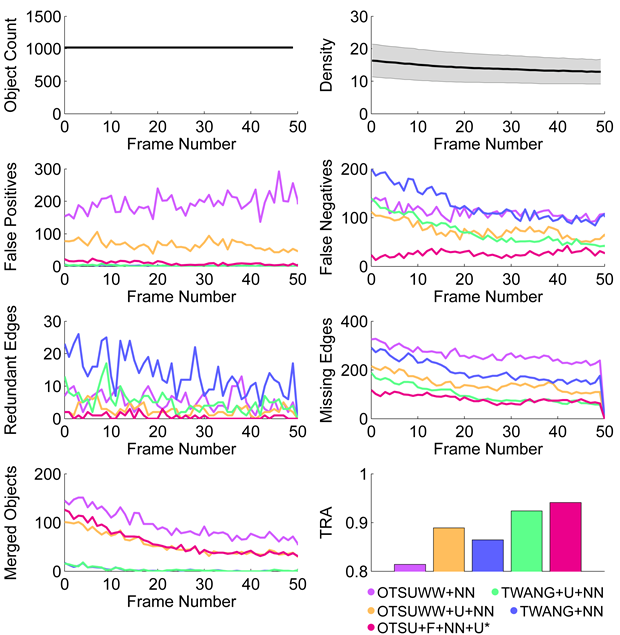}}
\caption[Quantitative assessment of the tracking performance]{Quantitative performance assessment of a nearest neighbor tracking algorithm (NN) applied on different segmentation results obtained on the \texttt{SBDE4} dataset. Two algorithms without uncertainty-based improvements as described in \cref{sec:chap3:TWANG} (OTSUWW, TWANG) were compared to enhanced pipelines that explicitly incorporate prior knowledge-based uncertainty treatment (OUGW+U, TWANG+U) as described in \cref{sec:chap4:Segmentation}. As two rotation images existed per time point (SiMV), the obtained results were fused using the method described in \cref{sec:chap4:MultiviewFusion}. (*) indicates that the respective algorithm did not produce a usable segmentation image, as the correction was solely performed at the tracking step (adapted from \cite{Stegmaier16Arxiv}).}
\label{fig:chap4:Tracking:Results}
\end{figure}

The tracking validation was performed on the \texttt{SBDE4} dataset, which consisted of $50$ frames with two simultaneously acquired rotation images for each frame, yielding a total number of $100$ frames that needed to be processed. Segmentation was performed using OTSUWW, OTSUWW+U, TWANG and TWANG+U and all obtained results were fused using the SF approach introduced in \cref{sec:chap4:MultiviewFusion}. The centroids of all detected objects were then used to perform the nearest neighbor tracking (NN). In addition to tracking centroids obtained from the segmentation step, the method described in \cref{sec:chap4:SeededOtsuTracking} (OTSU+F+NN+U) was applied to the test dataset and the respective tracking results are summarized in \cref{tab:chap4:Tracking:PerformanceTable} and \cref{fig:chap4:Tracking:Results}. 
\begin{sidewaystable}[p]
\begin{center}
%\rowcolors{2}{white}{gray!25}
\resizebox{\textwidth}{!}{
\begin{tabular}{lccccccccccccc} %{p{3.5cm}p{6cm}p{5cm}}
\toprule
\textbf{Method} & \textbf{TP} & \textbf{FP} & \textbf{FN} & \textbf{Red.} & \textbf{Miss.} & \textbf{Merg.} & \textbf{Rec.} & \textbf{Prec.} & \textbf{F-Sc.} & \textbf{TRA} & \textbf{Time (s)} & \textbf{KVoxel/s} \\
\midrule
OTSUWW+NN & 905.33 & 194.10 & 110.67 & 5.51 & 256.88 & 94.33 & 0.89 & 0.82 & 0.86 & 0.81 & 51.36 & 1020.81 \\
OTSUWW+U+NN & 941.94 & 65.65 & 74.06 & 2.63 & 141.39 & 55.04 & 0.93 & 0.93 & 0.93 & 0.89 & 33.68 & 1556.68 \\
TWANG+NN & 889.73 & \textbf{1.73} & 126.27 & 13.75 & 191.16 & 3.29 & 0.88 & \textbf{1.00} & 0.93 & 0.86 & \textbf{14.27} & \textbf{3674.06} \\
TWANG+U+NN & 943.90 & 1.76 & 72.10 & 5.16 & 96.55 & \textbf{3.27} & 0.93 & \textbf{1.00} & 0.96 & 0.92 & 21.55 & 2432.89 \\
OTSU+F+NN+U$^\ast$ & \textbf{989.63} & 9.84 & \textbf{26.37} & \textbf{0.67} & \textbf{79.67} & 59.33 & \textbf{0.97} & 0.99 & \textbf{0.98} & \textbf{0.94} & 18.66 & 2809.72 \\
\bottomrule
\end{tabular}}
\captionsetup{width=0.96\textheight}
\caption[Quantitative assessment of the tracking performance]{Quantitative performance assessment of a nearest neighbor tracking algorithm (NN) applied on different segmentation results. Two algorithms without uncertainty-based improvements as described in \cref{sec:chap3:TWANG} (OTSUWW, TWANG) were compared to enhanced pipelines that explicitly incorporated prior knowledge-based uncertainty treatment (OTSUWW+U, TWANG+U) as described in \cref{sec:chap4:SeedDetection}, \cref{sec:chap4:Segmentation}, \cref{sec:chap4:MultiviewFusion}. Furthermore, an OTSU-based segmentation with additional seed points from LoGNSM+F+U (OTSU+F+NN+U) was used with an adapted tracking algorithm (\cref{sec:chap4:SeededOtsuTracking}). Note that the segmentation produced by OTSU+F+NN+U was not usable for other purposes than tracking due to many merged regions (indicated by ($^\ast$)). The validation measures correspond to true positives (TP), false positives (FP), false negatives (FN), redundant edges (Red.), missing edges (Miss.) and merged objects (Merg.). Furthermore, recall, precision, F-Score and the TRA measure were calculated as described in \cref{sec:Appendix:PerformanceAssessment}. Processing times are average values for applying segmentation and tracking on a single image and were measured in seconds (lower values are better) and voxels per second (higher values are better). For better comparison, the constant processing time offset required for image fusion was skipped here (adapted from \cite{Stegmaier16Arxiv}).}
\label{tab:chap4:Tracking:PerformanceTable}
\end{center}
\end{sidewaystable}
Both pipelines without uncertainty treatment reached the lowest tracking accuracy with respect to the TRA measure due to an increased number of missing objects (recall of $0.89$ for OTSUWW+NN, and $0.88$ for TWANG+NN) and a high number of false positive detections (precision of $0.82$ for OTSUWW+NN). Of course, these missing objects directly correlated with the number of missing edges and explain the $222.4\%$ (OTSUWW+NN) and $139.9\%$ (TWANG+NN) higher amount of missing edges compared to the best scoring algorithm in this category (OTSU+F+NN+U). Furthermore, OTSUWW+NN suffered from many merged regions that also contributed to the $8.2\%$ lower recall value compared to OTSU+F+NN+U. In contrast to this, all uncertainty-enhanced methods provided comparable results, with the best results achieved by OTSU+F+NN+U and TWANG+U+NN. Although the amount of false negatives of TWANG+U+NN was almost halved in comparison to the TWANG+NN method, the detection rate was still the largest problem of this pipeline resulting in a $21.2\%$ higher amount of missing edges compared to OTSU+F+NN+U. On the other hand, OTSU+F+NN+U still suffered from the same under-segmentation tendency as observed for OTSUWW+NN and OTSUWW+U+NN and consequently missing edges were its main problem. With respect to false positive detections both TWANG-based methods provided the best results, with precision values of $1.0$ due to a high quality seed detection that benefitted from the simultaneous multiview acquisition. This reflects an improvement of the precision obtained by the TWANG-based methods of $22.0\%$ compared to OTSUWW+NN, which did not have an uncertainty-based object exclusion and thus an increased amount of false positive detections in background regions. The reason for the slightly higher number of redundant edges observed for the two TWANG-based methods is not yet fully clear. Most likely, the seed detection already provided a redundant seed to the segmentation method, which produced two nearby segments that were in turn counted as a redundant segment by the tracking evaluation. However, even for the worst algorithm in this category (TWANG+NN) redundant edges were only observed for $1.4\%$ of the tracked objects and thus play a minor role compared to the other tracking errors. As shown in \cref{fig:chap4:Tracking:Results}, the number of false negatives and merged objects directly correlated with the density of the objects, \ie, the closer the objects are to each other, the more under-segmentation errors occurred. However, this was not the case for TWANG-based methods due to the explicit prior knowledge about the object size that is incorporated to the algorithm. The OTSU+F+NN+U method produced the best tracking result and was the second fastest method (TRA value of $0.94$ and average processing time of $18.7s$) closely followed by TWANG+U+NN (TRA value of $0.92$ and average processing time of $21.6s$). Compared to OTSUWW+NN, TWANG+U+NN and OTSU+F+NN+U provide superior quality in all categories and in particular an increase of the TRA measure by $ 13.6\%$ and $16.0\%$, and a decrease of processing times by $58.0\%$ and $63.7\%$, respectively. Thus, the two latter methods  represent the best quality vs.\ speed trade-off and are suitable for large-scale analyses. Although OTSU+F+NN+U provided excellent results in this comparison, it should be noted that the extracted segmentation masks were largely merged and an object splitting approach as performed for OTSUWW would be required if object properties need to be known. An additional object splitting approach, however, would eradicate the performance benefit of the method and, \eg, the TWANG+U+NN pipeline should be favored in this case.

To fall back on seed point information has no benefit for segmentation methods like TWANG, where the algorithmic design already only extracts a single segment per seed point and literally no merged objects exist. However, an interesting extension to consider in upcoming work might be a combination of the LoGNSM+F+U and the OTSU-based segmentation for seed detection and to feed these seeds to the TWANG algorithm, to reach both a further reduced amount of missed objects and a reduced amount of merged objects. Moreover, the temporal coherence was not yet considered in the investigated framework, \ie, additionally allowing a nearest neighbor matching over multiple frames could potentially also help to reduce the number of missing and redundant detections.

%\input{Content/4_AlgorithmEnhancement/CompletePipeline.tex}

%%%%%%% FINAL DISCUSSION %%%%%%%%%
\section{Discussion}
In this chapter, an image analysis pipeline comprised of seed point detection, segmentation, multiview fusion and tracking was systematically equipped with prior know\-ledge-based enhancements according to the uncertainty framework proposed in \cref{sec:chap2:UncertaintyFramework}. The performance of all improvements was quantitatively assessed on a new and comprehensive validation benchmark inspired by light-sheet microscopy recordings of live specimen. In particular, this framework was successfully employed for the following tasks:
\begin{itemize}
	\item \textbf{Filtering}: The seed detection as well as the segmentation algorithms benefitted from the uncertainty-based feature rejection. Objects that were classified as clearly deviating from the expected object properties were removed and thus the number of false positive detections could effectively be reduced.
	\item \textbf{Splitting}: The segmentation methods based on adaptive thresholding produced many merged segments. Using morphology-based fuzzy sets to describe the properties of a correct detection, objects that were significantly larger could be reliably identified. The improved seed detection methods combined with a watershed-based object splitting successfully separated merged objects.
	\item \textbf{Fusion}: The FSMD values of all objects that were calculated at the seed point detection and the segmentation stage could be used to fuse the information present in complementary views to a single consistent representation of high quality. The feature-based fusion provided almost identical results compared to an intensity-based fusion in the image domain.
	\item \textbf{Tracking}: Similar to the splitting approach, the propagated information was used to resolve tracking conflicts that originated by merged segmentations. By skipping the costly object splitting in the image domain, this approach yielded both improved tracking quality and improved time performance.
\end{itemize}
For illustration purposes, mostly straightforward processing methods have been used to illustrate the general concepts. However, extending more complex segmentation or tracking algorithms with the presented concepts of filtering, splitting and fusion should work analogously if the uncertainty-based corrections are considered as a post-processing strategy of each processing operator. Especially the tracking step offers a lot of potential to be further improved. The uncertainty framework could be exploited to classify movement events, \eg, to detect object divisions or to reconstruct missing objects using the temporal coherence of the objects. However, this was beyond the scope of this thesis and will be addressed in upcoming work. In addition to the actual algorithmic improvements, it has been shown how the respective fuzzy sets can be parameterized based on available prior knowledge, such as feature histograms or knowledge about acquisition deficiencies. The respective shape of the fuzzy sets has to be determined based on the desired outcome. For instance, the false positive suppression at the seed point detection stage could be performed with a fixed threshold instead of an explicit usage of fuzzy sets for the intensity based features. However, to model the increasing uncertainty (decreasing FSMD value) in regions farther away from the detection objective using a trapezoidal shape was more appropriate. Further work has to be put on the automatic determination of the involved fuzzy sets, \eg, using a semi-automatic approach for a manual classification of a representative subset of data. The remaining sections present the implementations of the methods described so far as well as the biological experiments the methods were applied to.
 \cleardoublepage

%% Implementations

%%%%%%%%%%%%%%%%%%%%%%% IMPLEMENTATION AND NUMERICAL TOOLS %%%%%%%%%%%%%%%%%%%%%%%%
\chapter{New Implementations and Numerical Tools}
\label{sec:chap5:Implementations}
After having introduced the theoretical details of the image analysis pipelines considered throughout this thesis, the custom-made software tools that were developed for the practical application of the new methods primarily to large-scale fluorescence microscopy image data are covered in this chapter. The implementations can be separated into XPIWIT (\textbf{X}ML \textbf{Pi}peline \textbf{W}izard for the \textbf{I}nsight \textbf{T}oolkit), a C++-based software tool for efficient large-scale image analysis \cite{Bartschat16} and extensions to the MATLAB toolbox Gait-CAD, which contains numerous tools for sophisticated data analysis \cite{Mikut08Biosig, Stegmaier12}. 

\section{XPIWIT - XML Pipeline Wizard for the Insight Toolkit} % \footnote{This chapter is partly based on a publication by Stegmaier \etal \cite{Stegmaier15a}.}
\label{sec:chap5:XPIWIT}
The Insight Toolkit (ITK) offers plenty of features for multidimensional image analysis and has an active community in the biomedical field that constantly improves and extents the functionality of ITK \cite{Ibanez05}. Current implementations, however, suffer either from flexibility due to specifically compiled C++ pipelines for a certain task or by slow execution times caused by multiple read/write operations for separate filter execution. To overcome these limitations an XML-based wrapper application for the Insight Toolkit has been developed that enables a graphical setup and rapid prototyping of image analysis pipelines while preserving the performance of a pure C++ implementation. Created XML pipelines can directly be used to interface XPIWIT in console mode on large clusters. The current version of XPIWIT already incorporates 79 wrappers to existing ITK filters as well as 27 new filters that were developed in the scope of this thesis. The portfolio of filters comprises IO filters, preprocessing filters, edge detectors, morphological operators, segmentation filters and many more. To ensure XPIWIT can constantly be extended with new functionality, template files that facilitate the implementation of new filters for XPIWIT have been created. Filters that are already implemented as image-to-image filters in ITK can be included into XPIWIT in a matter of minutes and are instantly accessible via the XML interface. The software tool was successfully applied for the automated analysis of terabyte-scale, time-resolved 3D image data of developing zebrafish embryos.

\subsection{XML Pipeline Creation}
\label{sec:chap5:XMLPipeline}
As described in the previous chapters, a general concept of automated image processing, which is also inherent to ITK, is the arrangement of processing operators in a feed-forward pipeline structure. Hence, an XML-based pipeline format has been specifically developed, which allows to create flexible image analysis pipelines using a variety of different processing filters.

In \cref{lst:chap5:XMLPipeline}, a simple example image processing pipeline consisting of an image reader and a median filter is depicted. Besides basic tags about the XML version, encoding and the XPIWIT specifier (Line 1-2), the \texttt{<pipeline>} tag encapsulates the actual pipeline. A pipeline in turn is split into separate processing items with the tag \texttt{<item>} that can be further customized. Each of the specified filters has a set of inputs that have to point to the item ID of the respective output of a preceding filter. Outputs of the filter are specified by the filter definition in XPIWIT and do not need to be explicitly defined by the user. In addition to image inputs and outputs, XPIWIT also has an internal metadata handling system that can be used to share metainformation such as extracted object locations or image statistics between different processing operators. The optimal execution order of the individual processing operators is internally determined by XPIWIT and ensures that the requested image and meta inputs are available as soon as a processing operator is executed. Moreover, the XML structure allows to adjust all available parameters of each filter using key-value pairs. The XML pipelines are decoupled from the IO parameters, which facilitates the sharing of customized pipelines with other users and additionally allows to use a single pipeline to process multiple images with the same protocol.

\lstset{language=XML, numbers=left, breaklines=true, tabsize=2, backgroundcolor=\color{lightgray}}
\begin{lstlisting}[caption={Exemplary XML Pipeline.},label=lst:chap5:XMLPipeline]
<?xml version="1.0" encoding="UTF-8"?>
<xpiwit>
<pipeline>
	<item item_id="item_0001">				
		<name>ImageReader</name>					
		<input>
			<image item_id_ref="cmd" number_of_output="0">
		</input>
		<arguments>
			<parameter key="SpacingX" value="1.0">
			<parameter key="SpacingY" value="1.0">
			<parameter key="SpacingZ" value="5.0">
			<parameter key="MaxThreads" value="12">
			<parameter key="BufferOriginalImage" value="0">
			<parameter key="WriteResult" value="0">
		</arguments>
	</item>
	<item item_id="item_0002">
		<name>MedianImageFilter</name>
		<input>
			<image item_id_ref="item_0001" number_of_output="1">
		</input>
		<arguments>
			<parameter key="WriteResult" value="0">
			<parameter key="Radius" value="2">
			<parameter key="FilterMask3D" value="0">
			<parameter key="MaxThreads" value="12">
		</arguments>
	</item>
</pipeline>
</xpiwit>
\end{lstlisting}

\subsection{Data Generation}
To apply a predefined XML processing pipeline to a desired set of images, XPIWIT has to be executed from the command prompt with command line input arguments. Alternatively, a configuration text file can be piped to the executable. This configuration file needs to contain the output path, one or more input paths, the path to the XML file to be processed, and further optional parameters. This concept allows to heavily parallelize the data processing, \eg, using distributed computing environments such as Apache Hadoop, where one XPIWIT configuration file is generated per image and all jobs are distributed among available processing nodes (\cref{sec:Appendix:HadoopCluster}).

All filters can optionally write their processed output to disk. However, this is mainly used for pipeline debugging and in most cases only the last filter needs to save its output to disk. Metadata that is generated by one of XPIWIT's processing operators is saved in the CSV format (Comma Separated Value). This simple text format can be used to store large tables of values, where each column is separated by a specific character (\eg, "," or ";") and each row is separated by a simple line break. Many powerful data analysis tools (\eg, Excel, MATLAB, Gait-CAD, SPSS, Knime, RapidMiner) provide functionality to import such CSV files and therefore allow to explore the generated data in more detail. Optionally, XPIWIT is able to store the column specifiers in the first row, which can directly be used, for instance, as identifiers for single features and time series in Gait-CAD.

\subsection{Special Filters}
Besides the various wrappers for ITK-internal filters, the XPIWIT implementation also comprises the following special filters and pipelines:
\begin{itemize}
	\item Seed detection pipelines (\cref{sec:chap3:SeedDetection})
	\item TWANG pipeline for efficient segmentation of spherical objects in 3D images (\cref{sec:chap3:TWANG})
	\item RACE pipeline to efficiently segment locally plane-like objects in 3D images (\cref{sec:chap3:RACE})
	\item Interface to various noise and point spread function simulations for artificial image disruption (\cref{sec:chap3:SeedDetection}, \cref{sec:chap3:TWANG}, \cref{sec:chap3:RACE} and \cref{sec:chap4:Benchmark})
	\item Utility filters to convert segmentation label images to feature maps (\cref{fig:chap4:Segmentation:Comparison})
	\item Filter for watershed-based object splitting (\cref{sec:chap4:Segmentation:PerformanceImprovement})
\end{itemize}

\subsection{Gait-CAD Compatibility}
The main intention of XPIWIT was to speed up time costly processing steps that directly operate on large-scale image data. However, to exploit the powerful data analysis capabilities of the open-source MATLAB toolbox Gait-CAD \cite{Mikut08Biosig, Stegmaier12}, it is desirable to have a convenient interface between Gait-CAD and XPIWIT. There are in principal two ways to accomplish the synergy of both software tools. The first option is to extract data from large-scale image data solely using XPIWIT and to perform the analysis of extracted information by importing the generated \texttt{*.csv} metadata files into Gait-CAD data structures. As a second option, all filters that are compiled into the respective XPIWIT executable can be used within the usual Gait-CAD plugin-sequence dialog. Gait-CAD automatically takes care of generating temporary input files and XML pipelines for XPIWIT and automatically imports the results generated by XPIWIT back to Gait-CAD.

However, the general concept of pipelines in Gait-CAD assumes strictly linear pipelines with a single predecessor and a single successor. To achieve more flexibility and to be able to exploit the entire functionality of XPIWIT, a dedicated GUI has been developed, which is described in more detail in the following section \cite{Huebner14}.

\subsection{Graphical User Interface for Rapid Prototyping}
\begin{figure}[htb]
\centerline{\includegraphics[width=\columnwidth]{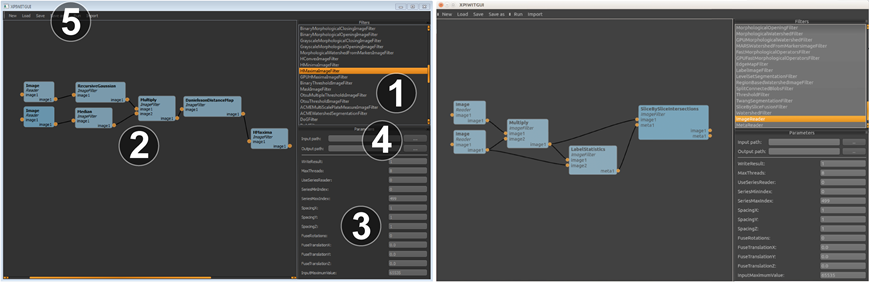}}
\caption[Exemplary screenshot of the XPIWIT GUI on Windows and Ubuntu]{Exemplary screenshot of the XPIWIT GUI on Windows (left) and on Ubuntu (right). The main control elements of the GUI are the filter list (1), the pipeline drawing area (2), the parameter adjustment panel (3), the input/output definition panel used for pipeline test runs (4) and the usual controls for saving, loading and executing pipelines (5). Generated pipelines are independent from the GUI, \eg, to perform large-scale analyses on computing clusters.}
\label{fig:chap5:XPIWITGUIScreenshot}
\end{figure}

An easy-to-use graphical user interface (GUI) has been developed that allows to create and modify XML pipelines based on the filters compiled into the XPIWIT executable. In \cref{fig:chap5:XPIWITGUIScreenshot}, the basic layout of the GUI is depicted. Filters can be placed via drag and drop from the filter list (1) to the pipeline drawing area (2). All parameters can be customized by selecting the respective filter in the drawing area and by adjusting the parameters in the parameter customization panel (3). The GUI is implemented using the open-source version of the Qt framework, which allows us to compile the GUI for Windows, Linux and Max OS X. Once the input and output parameters (4) are defined, the created pipeline can immediately be run on the specified data and can of course be saved and re-opened using the menu bar (5).

\subsection{Implementation Details} XPIWIT is implemented in C++ using the Insight Toolkit (\url{http://www.itk.org/}) and the Qt SDK (\url{http://qt-project.org/downloads}). Platform independent project files have been realized using the CMake build tool (\url{http://www.cmake.org/}). Both, XPIWIT and the associated GUI have been successfully compiled and tested under Windows, Linux and Mac OS X. Software and documentation are licensed under the GNU-GPL and will be made publicly available for download under \url{https://bitbucket.org/jstegmaier/xpiwit} as soon as the corresponding manuscript is published.

\subsection{Comparison to Existing Software Solutions}
To compare XPIWIT to other existing software tools, \cref{tab:chap5:SoftwareComparison} shows an extended version of the software comparison presented by Kankaanp\"a\"a \etal \cite{Kankaanpaa12}. Three new categories that are crucial for the analysis of large-scale imaging experiments were added. Namely, algorithms for large-scale analyses need to be computationally efficient to guarantee reasonable processing times (\textit{Fast}) and they have to be implemented properly to make the applications usable with possible memory constraints (\textit{Memory Efficiency}). Additionally, the usage of distributed processing on computing clusters often requires the applications to be runnable in headless mode and parametrizable via command line (\textit{Parallelizable}). Besides the new evaluation criteria, two new applications (ICY \cite{Chaumont12} and XPIWIT \cite{Bartschat16}) were included in the table. Commercial software tools (Imaris, FluoView, Volocity and ZEN Lite) and applications that are solely suitable for visualization or other special purposes (BioImage Suite, VisBio) were excluded from the overview table. The performance estimation with respect to speed and memory efficiency was performed by applying both a median and a Gaussian filter to a large image dataset.
As indicated in \cref{tab:chap5:SoftwareComparison}, the strengths of XPIWIT clearly lie in the domain of efficiently processing large-scale data, potentially on computing clusters. Of course, these criteria can also be achieved with special purpose implementations such as \cite{Amat14}, but XPIWIT additionally provides the flexibility of using different pipelines with a single static executable. Further work for XPIWIT has to be put into adding more wrappers to ITK internal filters as well as further improvements of the GUI with respect to result previewing, parameter adjustments and the possibility to use VTK-based visualizations. With respect to the new measurements, Vaa3D was the most powerful alternative to XPIWIT. As both XPIWIT and {Vaa3D} build on the foundation of ITK, the processing times were mostly comparable. However, {Vaa3D} encapsulates a lot of functionality into black box operators that lack parameter adjustment possibilities, \ie, the algorithmic behavior is not always transparent to the user.
\begin{table}[htb]
\begin{center}
\resizebox{\textwidth}{!}{
%\rowcolors{2}{white}{gray!25}
\small\begin{tabular}{lccccccccc}%{m{0.13\linewidth}m{0.0.13\linewidth}m{0.68\linewidth}}
\toprule
\textbf{Name} & \textbf{Open} & \textbf{Ext.} & \textbf{Usable} & \textbf{Adj.} & \textbf{Appl.} & \textbf{Exp.} & \textbf{\emph{Fast}} & \textbf{\emph{Mem.}} & \textbf{\emph{Parall.}} \\
\midrule
BioImageXD & $\bullet$ $\bullet$ $\bullet$ $\bullet$ & $\bullet$ $\bullet$ $\bullet$ $\bullet$ & $\bullet$ $\bullet$ $\bullet$ $\bullet$ & $\bullet$ $\bullet$ $\bullet$ $\bullet$ & $\bullet$ $\bullet$ $\bullet$ $\bullet$ & $\bullet$ $\bullet$ $\bullet$ $\bullet$ & $\bullet$ $\bullet$ & $\bullet$ & $\bullet$ \\
Fiji & $\bullet$ $\bullet$ $\bullet$ $\bullet$ & $\bullet$ $\bullet$ & $\bullet$ $\bullet$ & $\bullet$ $\bullet$ $\bullet$ & $\bullet$ $\bullet$ & $\bullet$ $\bullet$ $\bullet$ $\bullet$ & $\bullet$ $\bullet$ & $\bullet$ $\bullet$ $\bullet$ & $\bullet$ $\bullet$ $\bullet$ $\bullet$ \\
Vaa3D & $\bullet$ $\bullet$ $\bullet$ & $\bullet$ $\bullet$ & $\bullet$ $\bullet$ $\bullet$ & $\bullet$ $\bullet$ $\bullet$ & $\bullet$ & $\bullet$ $\bullet$ $\bullet$ $\bullet$ & $\bullet$ $\bullet$ $\bullet$ $\bullet$ & $\bullet$ $\bullet$ $\bullet$ $\bullet$ & $\bullet$ $\bullet$ $\bullet$ $\bullet$ \\
Matlab & $\bullet$ $\bullet$ & $\bullet$ $\bullet$ & $\bullet$ $\bullet$ $\bullet$ & $\bullet$ $\bullet$ $\bullet$ & $\bullet$ & $\bullet$ $\bullet$ & $\bullet$ & $\bullet$ & $\bullet$ $\bullet$ $\bullet$ \\
\emph{ICY} & $\bullet$ $\bullet$ $\bullet$ $\bullet$ & $\bullet$ $\bullet$ $\bullet$ $\bullet$ & $\bullet$ $\bullet$ $\bullet$ & $\bullet$ $\bullet$ $\bullet$ $\bullet$ & $\bullet$ $\bullet$ $\bullet$ & $\bullet$ $\bullet$ $\bullet$ $\bullet$ & $\bullet$ & $\bullet$ & $\bullet$ \\
\emph{XPIWIT} & $\bullet$ $\bullet$ $\bullet$ $\bullet$ & $\bullet$ $\bullet$ & $\bullet$ $\bullet$ $\bullet$ & $\bullet$ $\bullet$ $\bullet$ & $\bullet$ $\bullet$ $\bullet$ $\bullet$ & $\bullet$ $\bullet$ $\bullet$ $\bullet$ & $\bullet$ $\bullet$ $\bullet$ $\bullet$ & $\bullet$ $\bullet$ $\bullet$ $\bullet$ & $\bullet$ $\bullet$ $\bullet$ $\bullet$ \\
\bottomrule
\end{tabular}}
\caption[Comparison of image analysis software tools]{Comparison of the software tools BioImageXD \cite{Kankaanpaa12}, Fiji \cite{Schindelin12}, Vaa3D \cite{Peng14}, Matlab (The MathWorks, Inc.), ICY \cite{Chaumont12} and XPIWIT \cite{Bartschat16}. As described in \cite{Kankaanpaa12}, the evaluation criteria are open-source (\emph{Open}), extensibility (\emph{Ext.}), usability (\emph{Usable}), adjustability (\emph{Adj.}), applicability (\emph{Appl.}), expandability (\emph{Exp.}), processing speed (\emph{Fast}), memory efficiency (\emph{Mem.}) and the possibility to run the software on distributed computing environments (\emph{Parall.}). The properties are rated from one to maximally four points. For the criteria \emph{Fast} and \emph{Mem.}\,processing times and memory consumption were measured for a median and a Gaussian filter applied on a large image dataset. It should be noted that all assessments are of objective nature and each of the applications has its own benefits and drawbacks in specific fields of application (extended from \cite{Kankaanpaa12}, additional table entries are highlighted in italic font).}
\label{tab:chap5:SoftwareComparison}
\end{center}
\end{table}

\section{Extensions of the Open-Source MATLAB Toolbox Gait-CAD} %\footnote{This chapter is partly based on a publication by Stegmaier \etal \cite{Stegmaier12}.}
Gait-CAD is a convenient open-source MATLAB toolbox that is being developed at the Institute for Applied Computer Science at KIT since 2006. It is especially suited for interactive data analysis of both single features and time series \cite{Mikut08Biosig}. To extend the functionality of Gait-CAD with respect to the automated analysis data extracted from large-scale images, several additional toolboxes have been developed to facilitate image analysis, tracking and benchmarking as well as the visualization and export of related experimental results. The following sections provide a brief overview of the individual toolboxes and a more detailed description can be found in \cite{Stegmaier12, Stegmaier14a}. The most recent releases of Gait-CAD including the ImVID and the Tracking extension packages can be obtained from \url{http://sourceforge.net/projects/gait-cad/}. Furthermore, the SpinalCord extension package is hosted at \url{http://sourceforge.net/projects/zebrafishimage/files/}. The remaining extension packages are still under active development and the current versions are available upon demand.

\subsection{The ImVID Extension Package}
\label{sec:chap5:ImVID}
The ImVID Extension Package extends the powerful data mining capabilities of Gait-CAD by comprehensive functionality for automatic image processing and image analysis. To facilitate the input and output of the toolbox and to make it accessible to a broad range of scientific images the BioFormats library is used \cite{Linkert10}. Imported image data is immediately linked with Gait-CAD's internal data structures, such that analyses can always benefit from both data and visualizations as well as raw  and processed images. The toolbox provides convenient access to use and parameterize numerous filters of the MATLAB Image Processing Toolbox as well as custom-made filters such as the methods described in \cref{sec:chap3:SeedDetection} and \cref{sec:chap3:TWANG} as well as all Gait-CAD compatible XPIWIT filters that map a single input image to a single output image. All implemented filters can be used and combined to form complex image analysis pipelines using Gait-CAD's plugin sequence functionality. Once a pipeline is created it can be easily applied to all imported images automatically. The extension package can also be used in combination with Gait-CAD's batch processing system to easily scale projects up, \eg, to analyze thousands of images of a high-throughput screen in an automatic fashion.

\subsection{The Tracking Extension Package}
\label{sec:chap5:Tracking}
The Tracking Extension Package can be used to identify corresponding objects in multidimensional time series of object locations. Based on available prior knowledge about the objects' movement behavior, several constraints such as bounding box size and maximum velocity can be specified. Objects missing in a single frame that reappear in the next frame can be reconstructed by a linear interpolation scheme. Furthermore, a heuristic was implemented to repair incomplete tracks that originate from objects missing in multiple frames or by events such as object divisions. Tracked data can either be viewed using Gait-CAD's internal visualization functionality or exported in the \texttt{*.vtk} format to generate sophisticated videos of the object tracking, \eg, in Kitware's scientific visualization tool ParaView \cite{Henderson07}. The toolbox was successfully applied to the automated analysis of embryonic development as described in \cite{Weger14, Takamiya16, Kobitski15} and in \cref{sec:chap6:TWANG}.

\subsection{The SpinalCord Extension Package}
\label{sec:chap5:SpinalCord}
The SpinalCord Extension Package basically condenses all functionality to perform a quantitative analysis of laterally oriented zebrafish embryos in fluorescence microscopy images using the methodology described in \cref{sec:chap3:SpinalCord}. It is comprised of components for 
\begin{itemize}
	\item splitting and merging RGB channels
	\item rotation correction of the images
	\item automatic focus detection and extended focus image generation
	\item detection, extraction and transformation of elongated objects
	\item extraction of intensity profiles
	\item extraction of object distribution histograms.
\end{itemize}
Although the implementation was specifically designed for the particular purpose of automatically analyzing the spinal cord region of laterally oriented zebrafish embryos, the modular design in combination with all available ImVID filters allows a straightforward adaption to similar problems, where the main task is to detect elongated objects (possibly with colocalized information in multiple color channels) and to transform the extracted regions to a uniform and comparable representation. The method was successfully applied to the automated analysis of the impact of small molecules and the effect of environmental Acetylcholinesterase (AChE) inhibitors to particular neuron populations in specifically designed transgenic zebrafish embryos \cite{Stegmaier14a, Shahid16}.

\subsection{The Embryo3DT Extension Package}
\label{sec:chap5:Embryo3DT}
The Embryo3DT Extension Package comprises numerous functions and tools that facilitate the automated analysis of data extracted from large-scale microscopy images using XPIWIT. The following functionality has been implemented:
\begin{itemize} 
	\item a registration algorithm for the fusion of extracted spatial information of dynamic objects in different rotation images (\cref{sec:chap4:MultiviewFusion})
	\item functions to reject false positive detections based on fuzzy set membership values (\cref{sec:chap4:SeedDetection})
	\item fusion heuristics to get rid of redundant detections (\cref{sec:chap4:SeedDetection:SeedFusion})
	\item multiple specialized visualization options to optimally view the extracted data. 
\end{itemize}

\begin{figure}[htb]
\centerline{\includegraphics[width=\columnwidth]{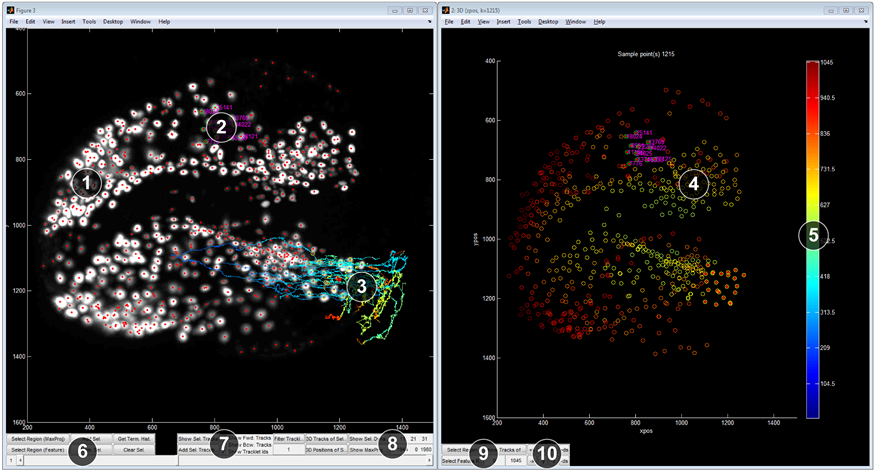}}
\caption[Exemplary screenshot of an Embryo3DT visualization]{Exemplary screenshot of an Embryo3DT visualization that shows extracted objects superimposed on the respective maximum intensity projection (1) with associated tracklet IDs (2) and a group of tracked objects (3). The right panel depicts a color-coded feature visualization (4, 5) that can be used for selection and tracking visualization in the feature space. The control elements are used to select data and to navigate in the temporal domain (6), to visualize tracks of selected objects (7), to generate object videos (8), to perform feature based selections and visualizations (9) and to manually refine the registration of complementary rotation views (10).}
\label{fig:chap5:Embryo3DTScreenshot}
\end{figure}

In \cref{fig:chap5:Embryo3DTScreenshot}, an exemplary visualization for the interactive analysis of embryonic development is shown. Besides the possibility to instantly visualize the dynamics of extracted objects, the Embryo3DT Extension Toolbox can export object locations and associated feature values to a \texttt{*.vtk}-based format to exploit the interactive visualization capabilities of ParaView.

\subsection{The Benchmark Extension Package}
\label{sec:chap5:Benchmark}
The Benchmark Extension Toolbox includes all the functionality for generating a simulated benchmark of moving and dividing objects as described in \cref{sec:chap4:Benchmark}. All parameters can be easily adjusted using a dedicated Gait-CAD dialog. It can be used to generate simulated 2D and 3D images of time series of dynamic objects or independent frames with a fixed number of objects. Next to the simulated image, segmentation label images as well as object metadata are stored as ground truth, to enable subsequent evaluation and comparison of automatically obtained processing quality of different image analysis operators. The toolbox comes with a database of simulated fluorescent nuclei that were extracted from the data generated by Svoboda \etal \cite{Svoboda12}. The artificial images can be distorted using different noise models, such as Poisson noise, Gaussian noise or salt and pepper noise and can be convolved with an arbitrary point spread function \cite{Gonzalez03}.
As MATLAB still tends to be slow when working with large 3D images, the acquisition simulation was additionally implemented in XPIWIT, to speed up the process for the creation of large-scale benchmarks as discussed in \cref{sec:chap4:Benchmark}.
In addition to the generation of simulated objects, functions to quantitatively compare the ground truth information to the respective results of automatic analysis operators are provided (\cref{sec:chap4:Benchmark:PerformanceAssessment}).

\subsection{Semi-Automatic Uncertainty-based Image Analysis}
\label{sec:chap5:FuzzyParameterAdjustment}
\begin{figure}[htb]
\centerline{\includegraphics[width=\columnwidth]{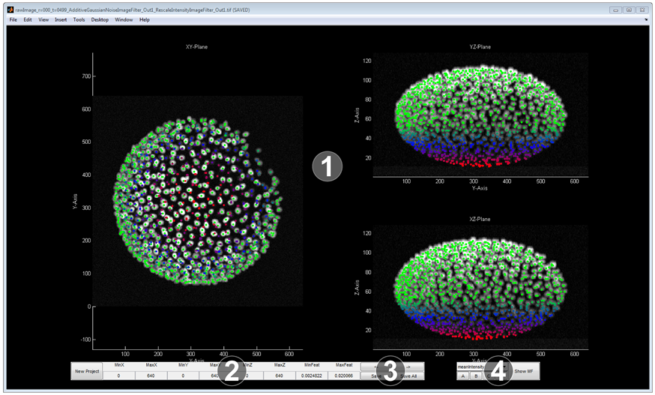}}
\caption[Exemplary screenshot of the semi-automatic data correction GUI]{Exemplary screenshot of the semi-automatic data correction GUI. The three main panels depict the maximum intensity projection views of the XY, XZ and YZ plane of an image containing simulated cell nuclei (white objects) with superimposed centroids of detected objects (1). Undesired objects can be filtered based on the feature values, \ie, spatial constraints or intensity constraints (2). Parameters can be propagated to all other images of the project and filtered results can be saved as Gait-CAD compatible \texttt{*.csv} files (3). Parameters used in the uncertainty framework, \ie, the fuzzy set membership parameterization of any of the available features can be performed in this GUI. The color-code from red (low FSMD) over blue (medium FSMD) to green (high FSMD) provides an instant visual feedback of the membership degree obtained for the individual objects (4).}
\label{fig:chap5:ManualSegmentationScreenshot}
\end{figure}

Although a high degree of automation is indispensable for large-scale image analysis projects, in some cases it makes sense to proofread and curate extracted information manually. For this purpose, a semi-automatic procedure has been developed that starts with a very sensitive automatic detection using XPIWIT and proceeds with a manual post-processing step to refine the results produced by the automatic processing. This allows to process large datasets in one run with potentially sub-optimal parameters and to correct the results afterwards instead of performing all calculations over and over again until optimal parameters are found. The developed GUI is perfectly suited to adjust spatial constraints and intensity thresholds to largely suppress false positive detections. Moreover, it can be used to estimate prior knowledge-based fuzzy set membership functions for all available features. Using an intuitive mouse wheel-based adjustment of a selected feature threshold, an instant visual feedback of filtered results is provided to the user and the respective threshold values can be used to set the parameters of a trapezoidal membership function (see (4) in \cref{fig:chap5:ManualSegmentationScreenshot}) as well as to identify the uncertainty framework filter parameters $\alpha_{il}$ and $\beta_{il}$ (\cref{sec:chap2:UncertaintyPropagation}). The individual objects are color-coded in red, blue and green to visualize low, medium and high degree of membership to the current fuzzy set.
The GUI allows navigating through all files of a project, allows to propagate parameter settings to all other time points or to adjust all parameters individually. Filtered data can finally be exported to a Gait-CAD compatible \texttt{*.csv} file. To enable browsing through large datasets, internally only file references are stored and images are loaded on the fly. \cref{fig:chap5:ManualSegmentationScreenshot} shows an exemplary screenshot of the GUI where a 3D benchmark image as discussed in \cref{sec:chap4:Benchmark} is analyzed. The semi-automatic tool was used for counting motoneurons in zebrafish embryos at different developmental stages and to refine segmentation results obtained from large-scale analyses of embryonic development that were extracted using XPIWIT.
 
\cleardoublepage

%% Applications

\chapter[Automated Quantitative Analysis of Embryonic Development]{Automated Quantitative Analysis of Embryonic Development in Large-scale Fluorescence Microscopy Images}
\label{sec:chap6:Applications}
After the new methodology and the associated implementations have been presented in the previous parts of the thesis, this chapter introduces three application examples of the developed methods that all stem from the field of developmental biology. New imaging techniques such as light-sheet microscopy \cite{Huisken09, Keller11, Tomer12} as well as new methods to automate image-based experiments \cite{Peravali11} offer unprecedented possibilities for detailed new studies in this scientific field. However, the immensely large amount of image data that is generated routinely by these novel techniques has to be stored, processed and analyzed with a high degree of automation as well to keep pace with the fast acquisition and to overcome the limitations of manual investigations. The challenges observed in the presented examples are a large amount of multi-channel 2D images that needed to be quantitatively compared to each other (\cref{sec:chap6:SpinalCord}) and terabyte-scale time series of 3D images produced by state-of-the-art light-sheet microscopes (\cref{sec:chap6:TWANG}, \cref{sec:chap6:RACE}).

%Initially, the segmentation for elongated structures (\cref{sec:chap3:SpinalCord}) is applied to analyze fluorescently labeled neuron populations in the spinal cord of zebrafish to quantify the impact of both known and unknown chemicals on the embryonic development. The second project aims at detecting, segmenting and tracking fluorescently labeled nuclei in large-scale 3D+t images using the newly introduced TWANG segmentation method (\cref{sec:chap3:TWANG}) with the aim of identifying developmental differences in early embryos of zebrafish at the single cell level. Finally, the RACE algorithm for fast plane-like structure segmentation (\cref{sec:chap3:RACE}) is employed to rapidly analyze fluorescently labeled cellular membranes in large-scale time series of 3D microscopy images of fruit fly, zebrafish and mouse embryos, to obtain a time-resolved characterization of cell morphology within a living system.

%%%%%%%%%%%%%%%%%%%%%%%%%%%%%%%%%%%%%%%%%%%%%%%%%%%%%%%%%%%%%%%%%%%%%%%%%%%%%%%%%
%%%%%%%%%%%%%%%%%%%%%%%%%%%%% Spinal Cord Project %%%%%%%%%%%%%%%%%%%%%%%%%%%%%%%
\section[Quantification of Neuronal Patterns in the Spinal Cord of Zebrafish]{Automated Quantification of Neuronal Patterns in the Spinal Cord of Zebrafish} % \footnote{This chapter is partly based on a publication by Stegmaier \etal \cite{Stegmaier14a}.}
\label{sec:chap6:SpinalCord}
The aim of this project was to automatically quantify the impact of unknown chemicals to different types of neurons in the spinal cord of laterally oriented zebrafish embryos \cite{Stegmaier14a}. The analysis was performed on a newly established transgenic line of zebrafish that exhibited fluorescence upon gene expression of \textit{dbx1b} (dI6, V0 and V1 interneurons tagged with eGFP) and \textit{ngn1} (Rohon-Beard cells, motoneurons, tagged with RFP) at two distinct excitation wavelengths \cite{Stegmaier14a}. The advantage of using this particular double-transgenic zebrafish line in the experiment was to have as many as possible types of central nervous system neurons (sensory-, inter- and motoneurons) for quantitative assessment in a single biological system, to possibly identify and thus narrow down the interesting candidates which may have effects on spinal cord neurons upon treatment. Treatment effects could for example be an increase or decrease in the number of a specific neuron type or an observable positional shift of the investigated cells. To enable high-throughput screening experiments where thousands of embryos might be screened, the developed algorithms should provide a strongly automated readout and extract both quantitative and qualitative information reliably and fast. As there was no existing solution to the specific problems observed for this screen, a new image analysis pipeline had to be developed (\cref{sec:chap3:SpinalCord}). Here, a pilot study to confirm the applicability of the proposed analysis framework is presented. Therefore, a systematic validation was performed using embryos that were treated with different concentrations ($0.5-10~\mu\text{M}$) of the Notch inhibitor {LY-411575} with a known effect on the development of moto- and interneurons \cite{Fauq07, Shin07}.

\subsection{Dataset Description}
\label{sec:chap6:SpinalCord:DatasetDescription}
\begin{figure}[htb]
\centerline{\includegraphics[width=\columnwidth]{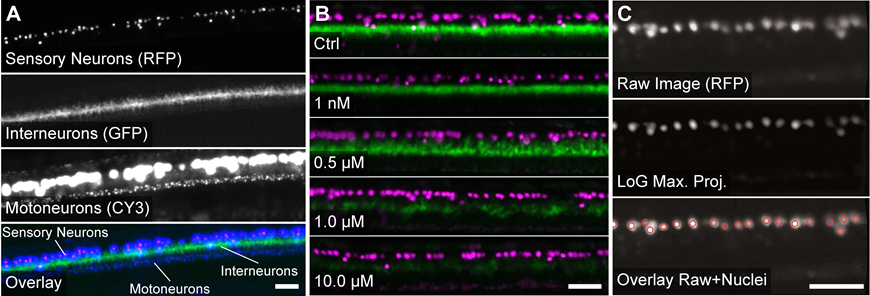}}
\caption[Overlay of RFP and GFP images of control embryos and {LY-411575} treated embryos]{(A) Raw images of the sensory neurons (RFP channel), the interneurons (GFP channel), the motoneurons (CY3 channel) and an overlay of all three channels. (B) Exemplary straightened crop regions obtained by the ROI extraction algorithm described in \cref{sec:chap3:SpinalCord:ROIExtraction}. The overlay images of the RFP and the GFP channel are shown for a control embryo and embryos treated with different concentrations of {LY-411575}. (C) Exemplary results of the LoG-based seed detection applied on the straightened ROI images (algorithm from \cref{sec:chap3:SeedDetection:LoGSSMP}). Scale bar: $50~\mu\text{m}$ (adapted from \cite{Stegmaier14a}).}
\label{fig:chap6:SpinalCord:ClassifiedOverviewReduced}
\end{figure}

Images were acquired using compound fluorescence microscopy (Leica DM5000 at 10$\times$ magnification). The embryos were excited with two different wavelengths (GFP: 484~nm, RFP: 585~nm) and the emitted fluorescence was captured using RFP, GFP and CY3 filters. The CY3 filter was used to facilitate the acquisition procedure, as the RFP signal was also visible in this channel due to fluorescent crosstalk and the additional channel allowed to use a different exposure times for the CY3 channel to make the dimmer motoneuron signal sufficiently visible (RFP: 40~ms, GFP: 200~ms, CY3: 410~ms). The considered images were acquired using a dynamic range of 8 bit at a resolution of $1392\times 1040$ pixels (px) with a physical pixel size of $1\text{px} \mathrel{\widehat{=}}0.64~\mu\text{m}$. An overview of the acquired raw image data is shown in \cref{fig:chap6:SpinalCord:ClassifiedOverviewReduced}A. 
In-focus images were selected manually to have three high-quality images per concentration and repeat, which yielded images of 108 embryos (3 repeats, 12 different concentrations and 3 embryos per concentration). Further details on the sample preparation, microscopy parameters and the experimental procedure can be found in \cite{Stegmaier14a}.

\subsection{Automated Image Analysis Framework}
\label{sec:chap6:SpinalCord:AnalysisFramework}
The automated analysis framework was developed using as much prior knowledge as possible. In particular, the interneuron signal (GFP channel) was used to estimate the location of the spinal cord. The other neuron populations were localized relative to this identified reference line, \ie, sensory neurons and motoneurons were detected above and below the spinal cord center line, respectively. A more detailed explanation of the specific properties of the individual acquired channels is summarized \cref{tab:chap6:SpinalCord:PriorKnowledge}.
The first step of the analysis procedure was the extraction of a comparable region of interest of the embryos. The interneuron signal appeared as a dense elongated object along the spinal cord of the embryo and was extracted using the method described in \cref{sec:chap3:SpinalCord}. Based on the determined spinal cord in the GFP channel, the same crop region along the regression curve with radius $r=70$ was extracted from all channels, to obtain a colocalized straightened version of the region of interest (\cref{sec:chap3:SpinalCord:ROIExtraction}). To identify sensory neurons and motoneurons in the RFP and the CY3 channel, respectively, the LoG-based seed detection method was used with $\sigma_{\text{min}}=\{2,3\}$, $\sigma_{\text{max}}=\{3,4\}$ and $\sigma_{\text{step}}=\{1,1\}$, where the parameters were determined on the basis of minimum and maximum radii of the respective neurons (\cref{tab:chap6:SpinalCord:PriorKnowledge}) as described in \cref{sec:chap3:SeedDetection:LoGSSMP}. For each of the channels, all single features described in \cref{sec:chap3:SpinalCord} were extracted and used for further analysis and comparison. Exemplary overlay images of straightened crop regions for different concentrations as well as detected sensory neurons in an RFP channel image are shown in \cref{fig:chap6:SpinalCord:ClassifiedOverviewReduced}B, C. The performance of the region of interest extraction and the LoG-based neuron detection was validated as described in \cite{Stegmaier14a}. The software required less than 6 seconds per image on a single processor of a common desktop computer, \ie, even larger screens comprised of several thousands of images can be processed within a few hours.
\bgroup
\def\arraystretch{1.5}
\begin{table}[hbt]
\begin{center}
%\rowcolors{2}{white}{gray!25}
\resizebox{\textwidth}{!}{
\begin{tabular}{m{0.13\linewidth}m{0.13\linewidth}m{0.66\linewidth}}
\toprule
\textbf{Type} & \textbf{Channel} & \textbf{Description} \\
\midrule
Inter\-neurons & $\texttt{GFP}$ &  Thick band of cells between 		  								  sensory- and motoneurons, individual cells not detectable, cubic polynomial to model the spinal cord \\
\midrule
Sensory Neurons & $\texttt{RFP}$, $\texttt{CY3}$ & Mostly well separated nuclei on dorsal side of the spinal cord, mainly overexposed in $\texttt{CY3}$, good signal in $\texttt{RFP}$ channel, diameter: $7-10~\mu \text{m}$ \\
\midrule
Moto\-neurons & $\texttt{RFP}$, $\texttt{CY3}$ & Densely packed nuclei on the ventral side of the spinal cord, weak signal in $\texttt{RFP}$ channel, good signal in $\texttt{CY3}$ channel, diameter: $5-7~\mu \text{m}$ \\
\bottomrule
\end{tabular}}
\caption[Prior knowledge used for the algorithmic design]{Prior knowledge used for the algorithmic design. Individual channels were perfectly colocalized and exhibited the fluorescent signal of stained nuclei from three different neuron types with channel-specific exposure times and parameters as described in \cref{sec:chap6:SpinalCord:DatasetDescription} (adapted from \cite{Stegmaier14a}).}
\label{tab:chap6:SpinalCord:PriorKnowledge}
\end{center}
\end{table}
\egroup

\subsection{Results}
\label{sec:chap6:SpinalCord:Results}
\begin{figure}[htb]
\centerline{\includegraphics[width=\columnwidth]{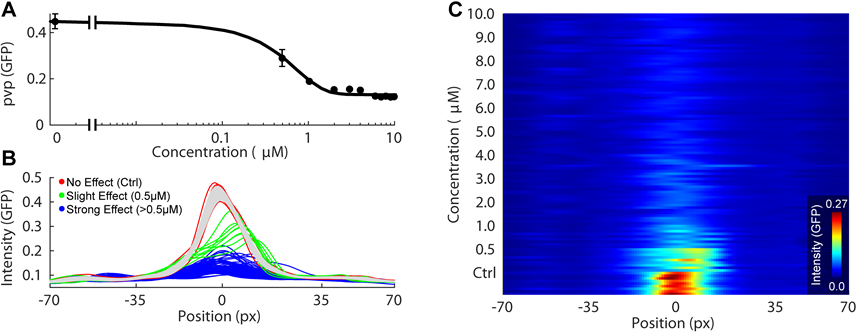}}
\caption[Dose-response graphs for {LY-411575} treated embryos]{ (A) The dose-response curve of the GFP peak intensity ($\texttt{pvp}_\texttt{GFP}$) shows an exponential decrease for increased concentration of {LY-411575}. (B) GFP channel intensity profiles of the pilot screen classified into \textit{no effect} (red), \textit{slight effect} (green) and \textit{strong effect} (blue). The interneuron intensity was slightly reduced for a concentration of $0.5~\mu\text{M}$ and was significantly reduced by higher concentrations of {LY-411575}. The gray surface represents the norm corridor of one standard deviation around the average control embryo profile (\cref{sec:chap3:SpinalCord:SpatialDistributionAnalysis}). (C) Color-coded intensity profiles of the GFP channel. Each row of pixels corresponds to a single intensity profile and the curves are sorted by increasing {LY-411575} concentration (bottom to top). In (B) and (C), a shift of the interneuron peak location towards the ventral side of the embryo can be observed for a treatment of $0.5~\mu\text{M}$. This effect can also be visually observed \cref{fig:chap6:SpinalCord:ClassifiedOverviewReduced}B. (adapted from \cite{Stegmaier14a}).}
\label{fig:chap6:SpinalCord:IntensityProfilesOverview}
\end{figure}

On the basis of visual inspection and the results reported in \cite{Fauq07, Shin07}, a constant number of sensory neurons, a decreased GFP intensity of the interneurons and an increase in the number and intensity of motoneurons upon treatment with various concentrations of {LY-411575} was anticipated. This was confirmed by the performed dose-response experiment and is exemplarily summarized in \cref{fig:chap6:SpinalCord:IntensityProfilesOverview} for the GFP signal of the interneuron population.
As suspected, the GFP intensity of interneurons was significantly decreased upon {LY-411575} treatment compared to the control embryos. The effect was reflected in all GFP related features (\eg $\texttt{ip}_{\texttt{GFP}}$, $\texttt{pvp}_\texttt{GFP}$ or $\texttt{nd}_{\texttt{GFP}}$). The dose-response curve depicted in \cref{fig:chap6:SpinalCord:IntensityProfilesOverview}A shows an exponentially decreasing GFP peak intensity value for increased treatment concentrations. Additionally, a tendency to an increased number of motoneurons could be observed in the CY3 channel for LY-411575 concentrations larger than $4.0~\operatorname{\mu\text{M}}$, \eg, using the features $\texttt{pvp}_{\texttt{CY3}}$ or $\texttt{nd}_\texttt{CY3}$ as described in \cite{Stegmaier14a}. One possibility to visualize the extracted information of the entire experiment in a single plot is the superposition of all intensity profiles as depicted in \cref{fig:chap6:SpinalCord:IntensityProfilesOverview}B. The shaded area in \cref{fig:chap6:SpinalCord:IntensityProfilesOverview}B indicates the norm corridor of one standard deviation. Profiles that clearly deviate from this area may indicate a potential treatment effect and should be investigated in more detail. In the case of the investigated LY-411575 concentrations, for instance, the reduction of interneuron GFP intensity could roughly be separated into the classes \textit{no effect}, \textit{slight effect} and \textit{strong effect} for control embryos, a concentration of $0.5~\operatorname{\mu\text{M}}$ and larger concentrations, respectively (\cref{fig:chap6:SpinalCord:IntensityProfilesOverview}B). The decreased GFP intensity was caused by a decreased number of interneurons. As the counting of interneurons was not possible using the considered image material, this was confirmed using 3D image stacks acquired with confocal microscopy as described in \cite{Stegmaier14a}. Another powerful visualization of the extracted information is the side-by-side alignment of the intensity profiles of the GFP channel images as shown in \cref{fig:chap6:SpinalCord:IntensityProfilesOverview}C. Using this visualization a comprehensive comparison of all intensity profiles can be obtained within just one overview image. Effects like a ventrally oriented shift of the interneurons for a concentration of $0.5~\operatorname{\mu\text{M}}$ could be easily identified with this approach. Further results obtained on the remaining image channels and the associated features that were used to quantify the effects can be found in \cite{Stegmaier14a}.

\subsection{Discussion}
Various methods for the automated analysis of zebrafish related screens were presented in the past \cite{Liu06, Liu08, Chen11a, Mikut13}. However, the specific demands and challenges of the presented experiment could not be handled by any existing methods and therefore needed to be newly developed. The generally introduced method for the extraction of elongated objects and their environment presented in \cref{sec:chap3:SpinalCord} was inspired by the demands of this screen and worked well in practice as confirmed by the results presented in \cref{sec:chap6:SpinalCord:Results}. Besides the validation performed on the simulated benchmark in \cref{sec:chap3:SpinalCord}, manual investigation confirmed that the algorithm properly extracted $98\%$ of the $108$ images in the pilot screen using the regression-based ROI extraction method and thus represents a robust solution to this specific problem. Additionally, $90.0~\pm~3.3\%$ of manually annotated sensory neurons were successfully detected by the LoG-based approach confirming its validity and sufficient quality to detect trends in changed sensory neuron counts. It should be noted that even manual counting was ambiguous in some cases and intra- and inter-expert variations were observed. The resolution used in the pilot screen was too low to unambiguously resolve the individual motoneurons. Thus, to quantify the effects on the motoneuron development it was more reasonable to use the intensity based features such as intensity profiles, spatial positioning of intensity peaks with respect to the spinal cord center or the spatial distribution analysis approaches as described in \cref{sec:chap3:SpinalCord:SpatialDistributionAnalysis} and \cite{Stegmaier14a}. Alternatively, a higher magnification level of the microscope or even a different microscopy technique could potentially improve the motoneuron detection as detailed in \cite{Stegmaier14a}.

Overall, the presented approach represents a well-suited method to quantitatively analyze different types of neurons in the spinal cord of zebrafish. The versatility of the extracted features could be used to reliably characterize treatment effects in a reproducible manner and enabled the extraction of image properties on the pixel or intensity level that were beyond visual inspection capabilities of human experts. Besides the application on the described pilot screen, the proposed image analysis framework was already successfully applied to analyze a small molecules screen of a large chemical library consisting of about 1600 chemicals with unknown toxicological effects. Furthermore, with minor parameter adjustments, the presented method was applied to quantitatively assess the effect of environmental Acetylcholinesterase (AChE) inhibitors in a specifically designed transgenic zebrafish line that served as a fluorescent biosensor \cite{Shahid16}.

%%%%%%%%%%%%%%%%%%%%%%% BIOLOGICAL RESULTS %%%%%%%%%%%%%%%%%%%%%%%%
\newpage
\section[Automated Analysis of Embryonic Development in 3D+T Microscopy Images]{Automated Quantitative Analysis of Embryonic Development in 3D+T Microscopy Images} % \footnote{This chapter is partly based on a publication by Stegmaier \etal \cite{Stegmaier14}.}
\label{sec:chap6:TWANG}

As sketched in the introduction, light-sheet fluorescence microscopy offers a great potential for in-depth analyses of living specimens such as the fruit fly, zebrafish or mouse embryos at the single cell level \cite{Keller13, Stegmaier14}. The ultimate goal would be to reconstruct an accurate lineage tree of cellular ancestry starting at the fertilized egg and going down to the various differentiated cell types of a living organism \cite{Amat14, Takamiya16}. The resolution and image quality achieved by light-sheet microscopy allows to detect and quantify phenotypic alterations in mutant embryos or upon drug exposure at the single cell level \cite{Otte13, Kobitski15}. However, the large amount of acquired image data cannot be manually analyzed and requires sophisticated and fast algorithms to exhaustively explore the recorded data. Although many methods for the automatic analysis of such datasets have already been presented, most of them were developed for 2D applications and do not scale properly. Thus, the applicability of most existing methods to large-scale datasets routinely acquired in developmental biology imaging experiments is rather limited (\cref{sec:chap1:ImageAnalysis}).

In this section, image datasets acquired using state-of-the-art light-sheet microscopy and the developed pipeline for their automatic analysis are introduced. While the new imaging technique allows to capture detailed 3D+t images of the early embryonic development of zebrafish embryos, there were multiple challenges that had to be solved to automatically extract biologically relevant data accurately and fast. The theoretically introduced methods derived in the last chapters were applied on these large-scale datasets to demonstrate the successful operation for seed detection, segmentation, multiview fusion and tracking on an actual biological problem and to emphasize the potential of quantitative approaches in developmental biology.

\subsection{Dataset Description}
\begin{figure}[!p]
\centerline{\includegraphics[width=\columnwidth]{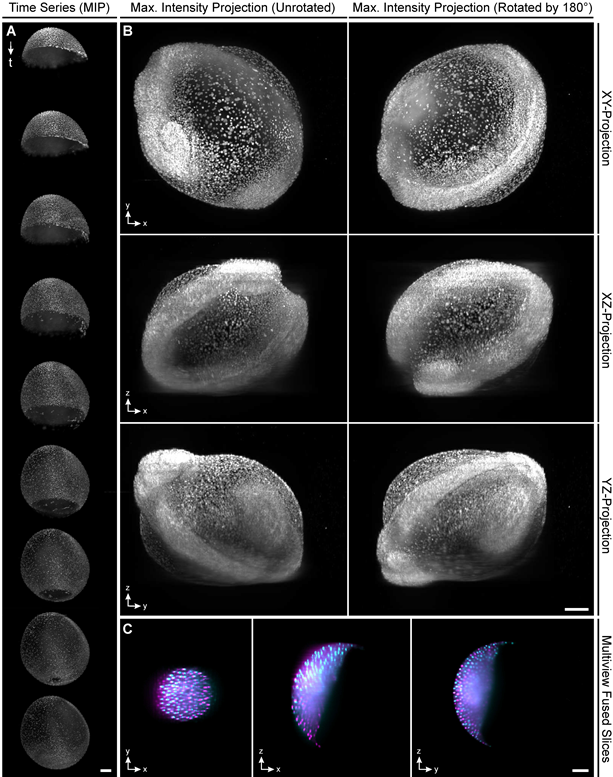}}
\caption[Exemplary light-sheet microscopy images of early zebrafish development]{(A) Snapshots of early embryonic development in zebrafish (4--10~hpf) using an axial maximum intensity projection (MIP). (B) MIPs of two complementary rotation images projected along the z, y and x direction, respectively (12~hpf). (C) Slices of two complementary views that have been registered using a bead-based registration approach. Displacement of cyan vs.\ magenta objects indicates cell movement happening between two frames. Scale bar: $50~\mu\text{m}$.}
\label{fig:chap6:TWANG:RealToxOverview}
\end{figure}

The image data used for the automated analysis of early embryonic development in zebrafish as described in this chapter were acquired using a custom-made  digital scanned laser light-sheet fluorescence microscope (DSLM) \cite{Otte13, Kobitski15} (\cref{sec:chap1:ImageAcquisition}). The microscope was equipped with two-sided illumination and a rotatable sample chamber that allowed a rotation by $180^\circ$ around the vertical axis. Dechorionated transgenic zebrafish embryos with fluorescently labeled nuclei were embedded in low melting agarose, which was enriched with fluorescent polystyrene beads with a diameter of $0.1~\mu\text{m}$ that served as landmarks for the rigid registration of the different view angles. In a typical experiment for the analysis of an early zebrafish embryo, the image acquisition was started between 1-2 hours post fertilization (hpf, 8-64 cell stage) and continuous imaging was performed for 10-16 hours in one minute time intervals (about 30 seconds for a single view). A single image stack acquired using this procedure had a resolution of 2560$\times$2160$\times$500 voxels, a dynamic range of 12 bit (stored as 16 bit unsigned char), a file size of about $5.16$~GB and covered a physical volume of roughly 1~$\text{mm}^3$. Thus, the amount of image data that had to be analyzed was on the order of 6-10~TB for a single embryo experiment. Several exemplary snapshots and maximum intensity projections of different views of a developing zebrafish are shown in \cref{fig:chap6:TWANG:RealToxOverview}A and \cref{fig:chap6:TWANG:RealToxOverview}B, respectively. Further information on the microscopy setup, the experimental procedure and the animal handling are provided in \cite{Otte13, Kobitski15}. Information on the infrastructure for storing and processing the large-scale datasets is summarized in \cref{sec:Appendix:Infrastructure}.

\subsection{Automated Image Analysis Framework}
In addition to the terabyte-scale dataset size, the main problems observed during the automated analysis were the reduced axial resolution (\cref{fig:chap6:TWANG:RealToxOverview}B) as well as light scattering and intensity reduction in regions farther away from the detection objective (\cref{fig:chap6:TWANG:RealToxOverview}B). The employed rotation of the specimen at every second frame helped to obtain a reasonable image quality of the entire embryo (analogous to the SeMV images used in \cref{sec:chap4:MultiviewFusion}). However, the use of this multiview acquisition required an additional fusion step to combine the extracted information of two complementary views into a single high-quality representation. Besides the problem of determining the exact transformation used for the registration of the two views, the dynamics of the investigated specimen had to be considered, \eg, if investigated objects were moving during the acquisition of the complementary views (\cref{fig:chap6:TWANG:RealToxOverview}C).

\subsubsection{Multiview Information Fusion}
\begin{figure}[htb]
\centerline{\includegraphics[width=\columnwidth]{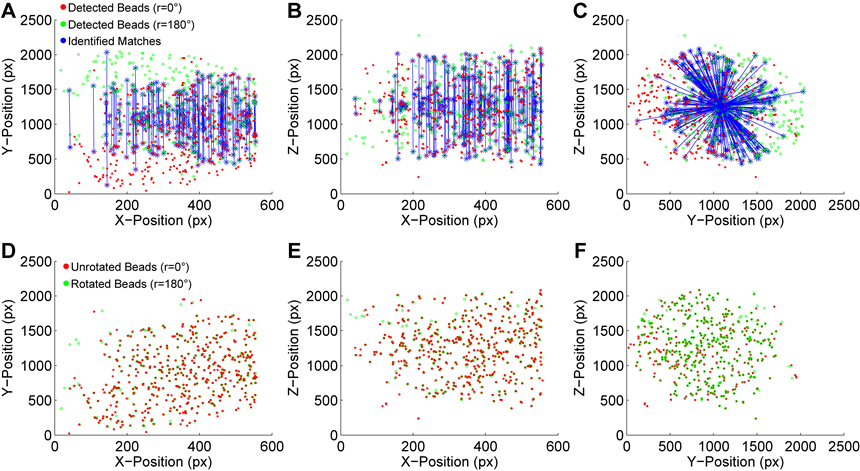}}
\caption[Registration of beads detected in the Laplacian-of-Gaussian maximum projection]{Registration of beads detected in the Laplacian-of-Gaussian maximum projection using the parameters $\sigma_{\text{min}}=4, \sigma_{\text{max}}=5, \sigma_{\text{step}}=1$. As the beads are smaller than the nuclei, the LoG parameters were adjusted to match the manually measured radii of the beads (\cref{sec:chap3:SeedDetection}). The first row (A-C), shows the unregistered beads and the identified correspondences of the seeds detected in two complementary rotation images. The second row shows the registered beads (D-F). The average distance of registered beads was $3.07 \pm 1.43 \text{px}$ in image coordinates, which corresponds to $1.22 \pm 0.58 \mu\text{m}$ in physical coordinates.}
\label{fig:chap6:TWANG:RegisteredBeads}
\end{figure}

The identification of the rigid transformation between the complementary rotation views was accomplished using a bead-based registration algorithm similar to the one proposed by Preibisch \etal \cite{Preibisch10}. The fluorescent beads were detected using the LoG-based seed detection method as described in \cref{sec:chap3:SeedDetection}. Correspondences between detected beads were identified by a matching of descriptors that characterized the local relations of each bead to its direct neighbors \cite{Preibisch10}. The iterative RANSAC optimization algorithm used in \cite{Preibisch10} was replaced by an analytic solution to the least-squares estimation problem based on the SVD-based (Singular Value Decomposition) method proposed by Umeyama \etal \cite{Umeyama91}, to speed up the transformation calculation. In \cref{fig:chap6:TWANG:RegisteredBeads}, the detected beads, the identified correspondences and the final registration result are shown.

If the complementary images were acquired simultaneously or if the investigated specimen was fixed, the identified transformation can be used to directly fuse the views to a single high-quality image. For the image material considered here, however, a direct fusion of the acquired rotation images to a single image stack did not lead to satisfactory results. This was caused primarily by object movements between rotationally complement frames and reduced image quality in the central regions of the fusion images (\cref{fig:chap6:TWANG:RealToxOverview}C). Thus, pre-processing, seed detection and segmentation were performed independently for each of the rotation images and then fused to a single representation of the entire embryo using the identified rigid transformation. To eliminate redundant information, \ie, objects that were detected in both of the views, the information fusion scheme described in \cref{sec:chap4:MultiviewFusion} was used.

\subsubsection{Fast Detection and Segmentation of Fluorescently Labeled Nuclei}
The seed detection was performed using the LoG-based approach as described in \cref{sec:chap3:SeedDetection} and \cite{Stegmaier14}. The images were down-scaled by a factor of 0.5 in the xy-direction, which still provided sufficiently high resolution to set individual nuclei apart from each other and helped to reduce the required processing time. The LoG parameters ($\sigma_{\text{min}}=6$, $\sigma_{\text{max}}=9$ and $\sigma_{\text{step}}=3$) were determined using manually measured minimum and maximum radii of the imaged nuclei as described in \cref{sec:chap3:SeedDetection}. Based on the extracted seed points, the fluorescently labeled nuclei were segmented using the TWANG method with the default parameters $\sigma_{\text{grad}}=3.0$, $\sigma_{\text{kernel}}=3.0$ and $\omega_{\text{kpm}}=1.5$ as described in \cref{sec:chap3:TWANG} and \cite{Stegmaier14}. To save both time and storage space, the actual segmentation images were not written to disk and only the extracted properties of detected nuclei were stored in a table format and used for further processing.

\begin{figure}[htb]
\centerline{\includegraphics[width=\columnwidth]{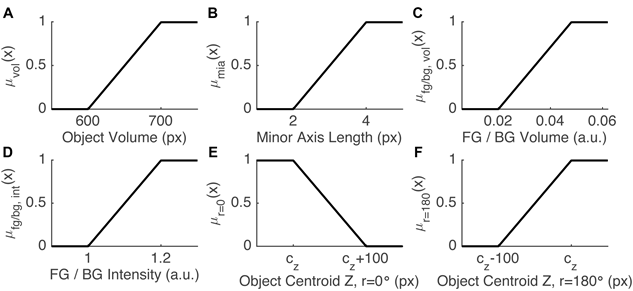}}
\caption[Membership functions for false positive reduction]{Fuzzy set membership functions that were used to discard false positive detections during the multiview fusion. Panels (A-D) show the membership functions for the object volume ($\pmb{\theta}_{\text{vol}}=(600, 700, \infty, \infty)^\top$), the minor axis ($\pmb{\theta}_{\text{mia}}=(2, 4, \infty, \infty)^\top$), the ratio of foreground vs.\ background volume ($\pmb{\theta}_{\text{fg/bg, vol}}=(0.02, 0.048, \infty, \infty)^\top$) and the ratio of foreground vs.\ background intensity ($\pmb{\theta}_{\text{fg/bg, int}}=(1.0, 1.2, \infty, \infty)^\top$). Parameter values were identified using minimum, maximum, 5\%-quantile and 95\%-quantile values of the feature histograms as described in \cref{sec:chap4:Segmentation:UncertaintyEstimation}. The fuzzy sets to discard detections in poor quality image regions (E-F) were parameterized with $\pmb{\theta}_{\text{r=0}}=(-\infty, -\infty, c_\text{z}, c_\text{z}+100)^\top$ and $\pmb{\theta}_{\text{r=180}}=(c_\text{z}-100, c_\text{z}, \infty, \infty)^\top$), respectively, with $c_z$ being the $z$-component of the embryo centroid. Thus, objects could be excluded based on their distance to the respective detection objective with a transition region in the center of the embryo. Note that fuzzy set parameters $-\infty$ or $\infty$ indicate plateau regions of the fuzzy sets towards the lower or the upper value range, respectively.}
\label{fig:chap6:TWANG:FusionFSMD}
\end{figure}

The extracted information of complementary views was registered using the identified transformation of the bead-based registration and then fused to a single representation (see \cref{sec:chap4:MultiviewFusion} and \cite{Kobitski15}). To eliminate redundant objects, each detected nucleus was associated with a fuzzy set membership degree of being a \textit{desired object/nucleus} (high value) or a \textit{false positive/noise} (low value). The most discriminative features for the extracted segments were identified by manual inspection of the feature histograms at different developmental time points. As the fluorescent signal was not yet strongly expressed at the early developmental stages, the algorithm detected mostly false positives, which could be used to characterize the expected feature values of false positive detections. Contrary to this, later time points contained almost no false positives and were thus used to determine feature values of true positive detections. The object volume, the minor axis, the ratio of foreground vs. background volume and the foreground vs. background intensity ratio of each nucleus were identified to be the most discriminative features. A visualization of the corresponding fuzzy set membership functions $\mu_\text{vol}(x), \mu_\text{mia}(x), \mu_\text{fg/bg, vol}(x), \mu_\text{fg/bg, int}(x)$ as well as the respective parameter vectors are provided in \cref{fig:chap6:TWANG:FusionFSMD}A-D. Additionally, depending on the rotation of the image, the membership values were reduced if a detected object was found in an imaged region of poor quality. This was accomplished by using the axial centroid location $c_z$ of the embryo ($z$-component of the mean centroid location of all detected nuclei), to define an additional fuzzy set for each of the rotations. Regions towards the respective detection objective obtained higher membership values, whereas low membership values were assigned for regions that were farther away (\cref{fig:chap6:TWANG:FusionFSMD}E, F). The individual membership functions of the features of every detected segment were combined using a fuzzy conjunction and potential false positive objects with a combined membership value of less than 0.5 were discarded. To remove redundant detections, objects that were detected in both views were combined using a weighted sum of the feature vectors if their distance was smaller than their minor axis \cite{Kobitski15}. The weighted average for the object combination was based on the fuzzy set membership degree (\cref{sec:chap2:ResolvingAmbiguities}). Based on the artificial benchmark validation presented in \cref{sec:chap4:MultiviewFusion} this approach proofed to work well in practice (\cref{tab:chap4:MultiViewFusion:DetectionPerformance}, TWANG+U+SF (SeMV), F-Score of $0.89$). For the real microscopy data, however, the quality could only be assessed using visual inspection of an overlay of the fusion results and the maximum intensity projection due to the immense number of objects (up to $\approx 25000$ cells for the investigated zebrafish embryos). 

\subsubsection{Tracking}
Object correspondences in neighboring frames were identified using a nearest
neighbor tracking approach as described in \cref{sec:chap4:Tracking} and \cite{Takamiya16, Kobitski15, Weger14}. Depending on movement speed of the investigated nuclei, the maximum allowed distance for the nearest neighbor matching was set to 15-30 pixels, which corresponded to 0.1-0.2~$\mu\text{m}/s$ for the considered image material \cite{Takamiya16, Kobitski15}. To correct tracks where an object was missing in a single frame but re-appeared in the next frame, the missing object location was interpolated. However, to obtain tracks that ranged over a larger time span all available tracklets (short partial tracks of a single nucleus) were extracted and a subsequent nearest neighbor matching was employed to connect tracklet terminals both in the spatial and the
temporal domain. The maximally allowed temporal displacement was set to 300~$s$ and a maximum displacement speed of 0.17~$\mu\text{m}/s$ was used \cite{Kobitski15}. These parameters were identified via manual investigation of average cell movements in consecutive images and represent the maximum distance a cell can travel without causing ambiguous correspondences. Following the assumption that closely located nuclei migrate to the same destination, a branching of tracks was permitted if a maximum of two reasonable matching partners had been identified at the terminal of a tracklet \cite{Takamiya16}. It should be noted, though, that these associations do not necessarily correspond to real mitotic events of the nuclei but helped to identify tracks that range over the entire time window and thus provided a convenient way to highlight regional migration patterns. A customized graphical user interface was used to qualitatively assess the correct operation of the automatic tracking by superimposing selected traces of individual nuclei on the respective maximum intensity projection images (\cref{fig:chap5:Embryo3DTScreenshot}).

\subsection{Results}
\begin{figure}[htb]
\begin{center}
\includegraphics[width=0.7\textwidth]{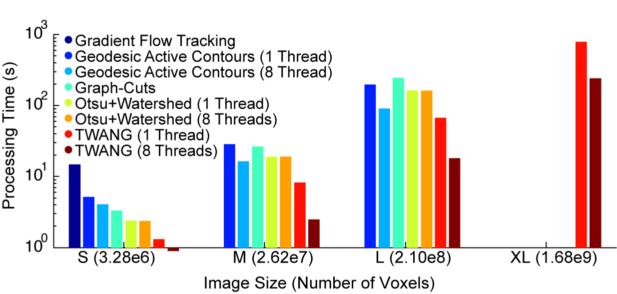}
\end{center}
\caption[Nucleus segmentation performance on differently sized 3D images]{Processing times for the compared segmentation algorithms in seconds (lower values are better) applied on 3D microscopy images of an early zebrafish embryo with fluorescently labeled cell nuclei. Image sizes correspond to 256x256x50 (S), 512x512x100 (M), 1024x1024x200 (L) and 2048x2048x400 (XL) voxels. Missing bars indicate that the respective algorithms failed to process the given image size (sizes M, L and XL for Gradient Flow Tracking and only size XL for Geodesic Active Contours (*), Graph-Cuts and Otsu+Watershed (*)). The opposed direction of TWANG (8 Threads) for image size S originated from a processing time below one second. In all tested categories, TWANG segmentation delivered peak performance and was the only method that could process the XL images with the given memory constraint of 32GB (adapted from \cite{Stegmaier14}).}
\label{fig:chap6:TWANG:SegmentationPerformance}
\end{figure}

\begin{figure}[htb]
\begin{center}
\includegraphics[width=\columnwidth]{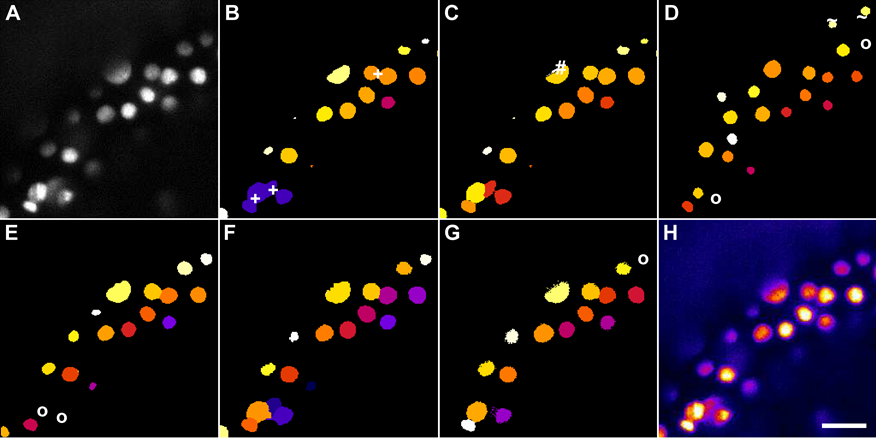}
\end{center}
\caption[Segmentation quality on a 3D image of labeled nuclei of a zebrafish embryo]{Segmentation quality achieved on a 3D image of labeled nuclei of a zebrafish embryo. The panels show the maximum intensity projections of 3 neighboring z-slices. (A) Original image, (B) adaptive thresholding using Otsu's method \cite{Otsu79}, (C) Otsu's method combined with watershed-based blob splitting \cite{Otsu79, Beare06}, (D) geodesic active contours \cite{Caselles97}, (E) gradient vector flow tracking \cite{Li07b}, (F) graph-cuts segmentation \cite{Al-Kofahi10}, (G) TWANG segmentation and (H) a false-colored original image. Segmentation errors indicated by the symbols are split (\#), merged (+), missing (o) or spurious ($\sim$) \cite{Stegmaier14}. Scale bar: $10~\mu\text{m}$.}
\label{fig:chap6:TWANG:SegmentationComparison3DPerformance}
\end{figure}

A comparison of the performance required by a bunch of nucleus segmentation methods on the described light-sheet microscopy images of zebrafish embryos was performed by analyzing differently sized crop regions of a full-size 3D image stack. The time needed to process 3D images with resolutions of 256x256x50, 512x512x100, 1024x1024x200, 2048x2048x400 voxels for sizes S, M, L and XL, respectively, was measured in seconds. The plain OTSU method was excluded from this test, as it did not produce satisfactory results in any of the benchmarks presented in \cref{sec:chap3:TWANG} and \cite{Stegmaier14}. \cref{fig:chap6:TWANG:SegmentationPerformance} summarizes the performance obtained by OTSUWW, GAC, GFT, GC and TWANG to segment a single stack of the differently sized 3D image stacks.
TWANG clearly dominated the field and was up to one order of magnitude faster than the compared algorithms. Even using a non-threaded implementation, the performance benefit of TWANG held true. Despite the good results of OTSUWW achieved on the 2D and 3D benchmark images \cite{Stegmaier14}, OTSUWW was 3-fold slower than the single threaded TWANG and up to 10-fold slower compared to our threaded implementation of TWANG applied on the L dataset due to the missing parallelization that becomes more significant for large images. Additionally, TWANG has a relatively low memory footprint due to the local extraction of objects. This property allowed even to process the largest image category (XL), which could not be handled by any of the compared algorithm within the memory limitation of 32GB. Analyzing a single experiment of about 10~TB of image data required approximately 12 hours on a small computing cluster consisting of 30 nodes (\cref{sec:Appendix:Infrastructure}). The same task would have taken multiple weeks up to months with previously existing approaches. As shown in \cref{fig:chap6:TWANG:SegmentationComparison3DPerformance}, the obtained segmentation quality was comparable to the results of the more complex segmentation algorithms. Due to the lack of a ground truth for the 3D+t data, only a qualitative comparison could be performed.

\begin{figure}[htb]
\centerline{\includegraphics[width=\columnwidth]{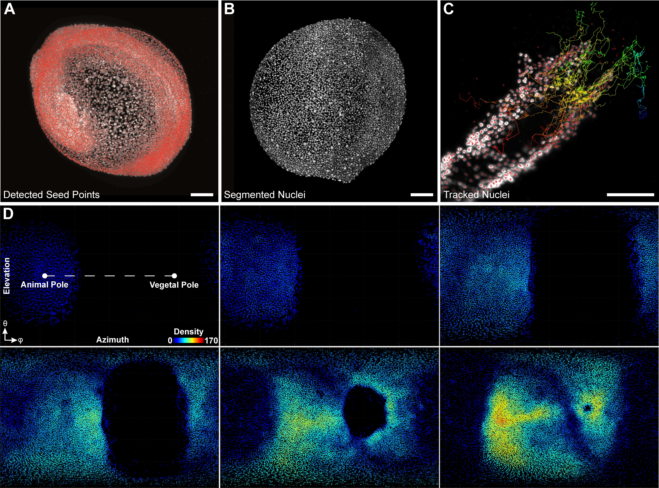}}
\caption[Automatic analysis results for a developing zebrafish embryo]{(A) Detected seed points using the LoG-based method described in \cref{sec:chap3:SeedDetection} at 12~hpf. (B) Segmented nuclei using the TWANG segmentation algorithm as described in \cref{sec:chap3:TWANG} at 11~hpf. (C) Traces of selected nuclei that were identified using a nearest neighbor tracking algorithm with tracklet fusion as described in \cref{sec:chap4:Tracking}. Centroids for the tracking were identified using the TWANG algorithm and the color code indicates time (increasing from red to blue). (D) Cell density distribution in a developing zebrafish embryo with a time difference of 200 minutes between each of the frames. Extracted coordinates of the labeled nuclei were transformed into spherical coordinates \cite{Weger14} and the panels show a selection of snapshots using the azimuth and elevation angles of each nucleus for plotting (time increases from left to right and top to bottom). Color encodes the number of neighbors of each nucleus that lie within a bounding sphere of 100px (40$\mu$m). Scale bar: $100\mu\text{m}$.}
\label{fig:chap6:TWANG:TWANGResultsOverview}
\end{figure}

The obtained segmentation quality of different investigated algorithms on an exemplary slice of a zebrafish dataset is shown in \cref{fig:chap6:TWANG:SegmentationComparison3DPerformance}.
Moreover, the results of the seed detection and the segmentation applied on the whole embryo level are provided in \cref{fig:chap6:TWANG:TWANGResultsOverview}A, B. Both visualizations emphasize the suitability of the proposed TWANG segmentation for a reliable extraction of fluorescently labeled nuclei and the associated properties from large-scale 3D+t image data. An exemplary visualization of a selection of tracked nuclei is shown in \cref{fig:chap6:TWANG:TWANGResultsOverview}C. In \cref{fig:chap6:TWANG:TWANGResultsOverview}D, several snapshots of an early zebrafish embryo are visualized by plotting the azimuth and elevation angles derived from a spherical projection of all nuclei in a 2D plot \cite{Weger14}. This digitally performed unwrapping of the embryo allowed to map the 3D organism to a 2D representation. The color code used in \cref{fig:chap6:TWANG:TWANGResultsOverview}D reflects the density of nuclei in their local environment, \ie, blueish values indicate a low density, whereas reddish values indicate higher density. If combined with a tracking approach, this color code can also be used to visualize developmental quantities such as migration speed, migration direction and division events of cells or even the cellular ancestry as demonstrated in \cite{Amat14, Stegmaier16}.

\subsection{Discussion}
In this section, the entire pipeline for the automated analysis of large-scale light-sheet microscopy image data was presented. The newly developed TWANG method implemented in XPIWIT represents a reasonable approach to obtain a reliable and fast segmentation of labeled nuclei in distributed computing environments. Furthermore, a new information fusion approach was presented that is capable of combining information from different complementary views of a specimen. Extracted cell centroids were temporally connected to obtain detailed cell trajectories that allow characterizing cellular dynamics on the whole-embryo level. Due to the limited space, however, only proof-of-principle results were presented in this section. The described pipeline was successfully applied to multiple projects in the field of developmental biology. In particular, the developed methods were used to:
\begin{itemize}
	\item count fluorescently labeled nuclei in the spinal cord of laterally oriented zebrafish embryos using confocal microscopy data \cite{Stegmaier14a}
	\item track fluorescently labeled neural crest precursor cells to their final destination in the eye of zebrafish using light-sheet microscopy data \cite{Takamiya16}
	\item assess the impact of Bisphenol A (BPA) onto early embryonic development of zebrafish using light-sheet microscopy data \cite{Otte13}
	\item quantify number, density and movement behavior of cells in the \textit{mondoa} morphant using light-sheet microscopy data \cite{Weger14}
	\item to derive an ensemble averaged reference embryo that can be used to precisely unveil developmental differences of mutant and wild-type embryos using light-sheet microscopy data \cite{Kobitski15}.
\end{itemize}
In all scenarios, the new approach allowed to qualitatively and quantitatively investigate developmental processes at the single cell level, which are for instance, cellular movements, density changes, cell proliferation rates, cell counts and epiboly stage to name but a few. The detailed biological description and results can be found in the respective publications \cite{Otte13, Stegmaier14a, Weger14, Kobitski15, Takamiya16}.

\clearpage

%%%%%%%%%%%%%%%%%%%%%%% BIOLOGICAL RESULTS %%%%%%%%%%%%%%%%%%%%%%%%
\section{Automated Segmentation of Fluorescently Labeled Membrane Images}
\label{sec:chap6:RACE}
The third application example deals with the automatic segmentation of fluorescently labeled cellular membranes. Extracted cell shapes can significantly contribute to better understanding of cell-cell interactions, morphological changes during cell migration as well as phenotypic alterations on the whole embryo level \cite{Lecuit07, Oates09}.

As indicated in \cref{sec:chap3:RACE}, several methods for automatic membrane segmentation were presented in the past \cite{Khan14, Mosaliganti12, Fernandez10, Michelin13}. However, all of the described methods rely on isotropic image data, \ie, for anisotropic image data as mostly observed in fluorescence microscopy-based experiments, an upsampling of the data is performed to obtain isotropic resolution. Additionally, the use of computationally demanding processing operators renders most of the existing methods impractical for the analysis of large-scale experiments due to enormous memory consumption and processing times. In particular, terabyte-scale time series of light-sheet microscopy data that are routinely acquired in the field of developmental biology cannot be analyzed to their full extent and only excerpts of the data could be considered so far \cite{Khan14}.

In this section, the newly developed RACE method for the segmentation of locally plane-like structures as described in \cref{sec:chap3:RACE} is applied to obtain cell shape information from large-scale images of fluorescently labeled membranes. The new algorithm is able to directly process anisotropic datasets, involves only efficient processing operators and offers a high degree of parallelization to perform new segmentations reliably and fast. At the same time, the segmentation quality is comparable or even better than the quality obtained by existing methods \cite{Stegmaier16}.

\subsection{Dataset Description}
\begin{figure}[!bp]
\centerline{\includegraphics[width=\columnwidth]{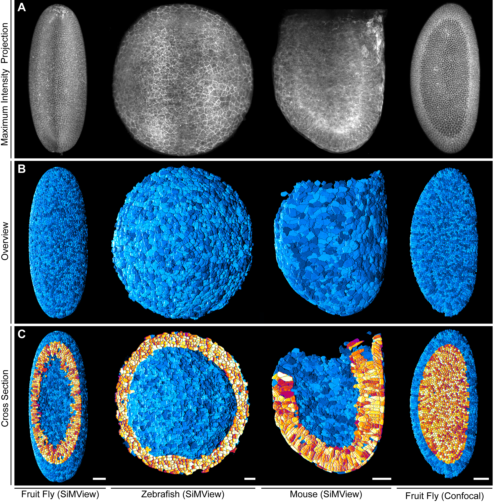}}
\caption[Whole embryo membrane segmentation results]{Exemplary segmentation results obtained by applying the RACE algorithm to 3D light-sheet microscopy images fo fruit fly, zebrafish and mouse embryos and a confocal image of a fruit fly embryo. (A) shows maximum intensity projections of each specimen. The remaining rows show volume renderings of the segmentation results from the outside of the embryo (B) and for a cross section that makes the interior of the embryos visible (C). The segmentation images show that the individual cells are nicely resolved, which allows to characterize cellular morphology for subsequent quantitative analyses. Scale bar: $50~\mu\text{m}$ (adapted from \cite{Stegmaier16}).}
\label{fig:chap6:RACE:WholeEmbryoSegmentation}
\end{figure}

The RACE algorithm was tested on different model organisms (fruit fly, zebrafish and mouse) and using both light-sheet (SiMView \cite{Tomer12}) and confocal microscopy techniques (Zeiss 710 laser-scanning microscope). The respective image sizes of one channel of a single frame ranged from 640$\times$1300$\times$23 (37~MB, 16 bit) for the confocal fruit fly dataset to 2048$\times$2048$\times$120 (960~MB, 16 bit) for a light-sheet image of a zebrafish embryo. All of the presented images contained both fluorescently labeled nuclei and membranes and were acquired at different developmental stages to test the algorithm under various different circumstances. Exemplary maximum intensity projections of the acquired image data are shown in \cref{fig:chap6:RACE:WholeEmbryoSegmentation}A. Further information on the microscopy setup, the experimental procedure and the animal handling are provided in \cite{Stegmaier16}.

\subsection{Automated Analysis Framework}
The automatic membrane segmentation was performed using the new RACE algorithm as described in \cref{sec:chap3:RACE}. Seeds were obtained from the membrane channel (MS) or from an additional nucleus channel (NS) and used for the fusion of the 2D segments that were identified by the slice-by-slice watershed algorithm applied on a membrane-enhanced input image. Essentially three intensity dependent parameters needed to be adapted and optimized for the different specimens and an overview of the manually identified optimal values are provided in \app\cref{tab:Appendix:RaceParameters} and \cite{Stegmaier16}.

In addition to the performance validation on the simulated benchmark data presented in \cref{sec:chap3:RACE:Validation}, a comparison to manually labeled ground truth images was performed on a set of representative image regions of different model organisms and using different microscopy techniques. The quantitative comparison confirmed the suitability of the proposed RACE approach with improved quality and speed compared to state-of-the-art methods. Further information about the ground truth generation, the comparison procedure as well as the details on the additional validation examples can be found in \cite{Stegmaier16}. Due to the laboriousness of the manual ground truth generation, however, the algorithmic validation had to be performed on small representative image regions rather than on the whole embryo level.

\subsection{Results}
\begin{figure}[htb]
\centerline{\includegraphics[width=\columnwidth]{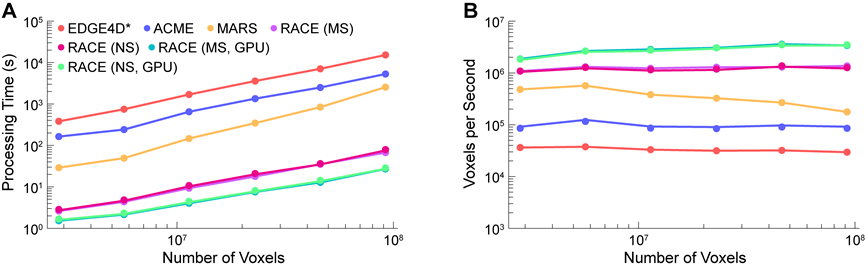}}
\caption[Membrane segmentation performance comparison]{Performance comparison of the proposed RACE algorithm to other state-of-the-art membrane segmentation methods (ACME \cite{Mosaliganti12}, MARS \cite{Fernandez10} and EDGE4D \cite{Khan14}). Differently sized crop regions of an early \textit{Drosophila} embryo were processed and the obtained processing times in seconds are plotted against the respective number of voxels (A). In (B), the speed was measured by means of voxels per second. Note that ACME, MARS and EDGE4D work on isotropic image data, \ie, the isotropic amount of voxels was used for the speed estimation. All versions of the RACE segmentation algorithm clearly outperformed the compared methods by at least an order of magnitude. The dedicated GPGPU implementation additionally pushes the performance by effectively eradicating existing bottlenecks of the algorithm. Note that the respective processing speed for each CPU-based and each GPU-based implementation of RACE was almost identical. Thus, the lines for RACE (NS) and RACE (MS) as well as RACE (NS,GPU) and RACE (MS, GPU) overlap in the performance plot and appear as a single line. Due to available implementation for Mac OS X, the performance of EDGE4D was assessed on an Apple MacBook Pro (Intel Core i7-3615QM CPU @ 2.3GHz, 16GB, Mac OS X 10.9.5) and is indicated by (*). For all other algorithms a workstation computer (2x Intel Xeon E5 CPUs @ 3.1 GHz, 196 GB RAM, NVidia Tesla K20 GPU, Windows 7 Professional 64-bit) was used to assess the performance (adapted from \cite{Stegmaier16}).}
\label{fig:chap6:RACE:PerformanceAnalysis}
\end{figure}

Volume renderings of the segmentation results produced by the RACE algorithm are shown in \cref{fig:chap6:RACE:WholeEmbryoSegmentation}B, C, for different model organisms (fruit fly, zebrafish and mouse) and different microscopy techniques (light-sheet and confocal microscopy). In all investigated scenarios, the RACE method nicely extracted cell shapes in the entire embryo that were suitable for further quantitative analyses. As demonstrated in \cref{fig:chap6:RACE:WholeEmbryoSegmentation}C, even the interior of the hollow embryos only contained a negligible amount of false positive detections due to autofluorescence. All segmentation results could be processed within a few minutes on a single workstation computer instead of hours as required by existing methods. To quantitatively assess the time performance of the algorithms, differently sized crop regions of a fruit fly embryo were processed using the segmentation algorithms MARS \cite{Fernandez10}, ACME \cite{Mosaliganti12}, EDGE4D \cite{Khan14} and the different versions of the RACE algorithm \cite{Stegmaier16}. The results of the performance comparison are depicted in \cref{fig:chap6:RACE:PerformanceAnalysis}. The RACE algorithm was by a large margin the fastest method, being one to two orders of magnitude faster than the compared algorithms. For a terabyte-scale dataset this essentially means that processing times are reduced from years to a few days with high segmentation accuracy. 
 
\begin{figure}[!htb]
\centerline{\includegraphics[width=\columnwidth]{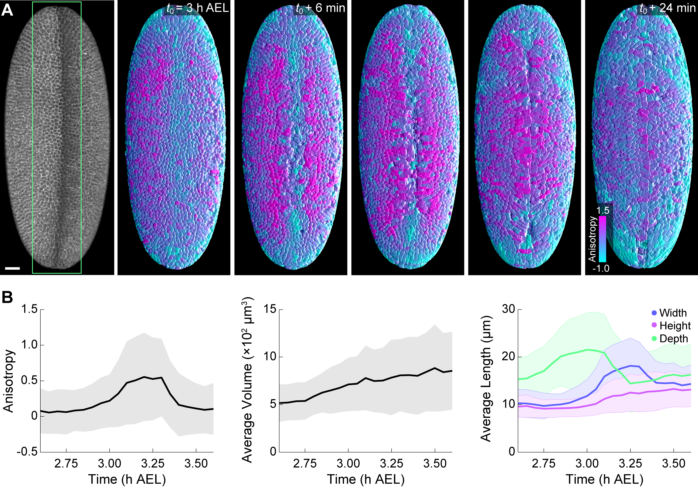}}
\caption[Anisotropy maps of the ventral fold formation in \textit{Drosophila} embryos]{(A) Visualization of the tissue anisotropy observed during ventral fold formation of a \textit{Drosophila} embryo starting at roughly $2.5$ hours after egg laying (AEL). The color code reflects the anisotropy level of the cells, \ie, cyan cells are elongated along the anteroposterior axis (height larger than width), purple cells are almost circular (similar height and width) and bright magenta cells are elongated along the mediolateral axis (width larger than height). In (B), average time courses of the anisotropy measure, the volume and the cell dimensions are shown for a selected region of interest (green rectangle in (A)). The furrow formation commences with a dorsoventral elongation of the cells (increased depth measure), followed by a mediolateral elongation of the cells (increased width measure). Upon closure of the ventral fold, both values decrease again. The same effect can be observed for the anisotropy measure, which serves as an indicator for mediolateral elongation of the cells. Shaded area indicates one standard deviation. Scale bar: $25~\mu\text{m}$ (adapted from \cite{Stegmaier16}).}
\label{fig:chap6:RACE:AnisotropyMaps}
\end{figure}

In \cref{fig:chap6:RACE:AnisotropyMaps}, an exemplary analysis of the obtained cell shape segmentation is shown. The visualization shows the color coded tissue anisotropy during \textit{Drosophila} gastrulation, which can be nicely used to emphasize cell shape changes like the medio-lateral elongation of cells near the ventral furrow (\cref{fig:chap6:RACE:AnisotropyMaps}A). The anisotropy measure was calculated by
\begin{gather}
	\text{Anisotropy} = \begin{cases} (w/h)-1 ,& w/h > 1 \\ 1-(h/w) ,& otherwise \end{cases},
\end{gather}
where $w$ and $h$ correspond to the medio-lateral and the anterior-posterior extent of the cells, respectively \cite{Stegmaier16}. To obtain comparable extents at all locations of the embryo, each extracted cell was projected to an ellipsoid that was fit to the outer shell of the embryo. Extents were then estimated by projecting the segment to the medio-lateral tangent (width), the anterior-posterior tangent (height) and the surface normal (depth) at the respective centroid location. In addition to visualization of various features, properties of the extracted segments provided quantitative information of cell shape changes, such as anisotropy levels, volume or cell extents as shown in \cref{fig:chap6:RACE:AnisotropyMaps}B. As demonstrated in \cite{Stegmaier16}, the approach could also successfully be linked to a nucleus tracking approach \cite{Amat14}, which provides a combination of cellular shape information with the ancestry of each cell. In upcoming projects this could for instance be used to perform quantitative analysis of the temporal cell shape changes on the whole embryo level. 

\subsection{Discussion}
In this section, exemplary results of the RACE segmentation method were presented. The method successfully segmented images of different specimens and different microscope types. While the results offered similar or even better segmentation quality, the new method represents a high-performance approach to cell shape segmentation with processing speed that exceeds the acquisition duration of common 3D microscopy images. In most cases where the fluorescently labeled nuclei were nicely distinguishable, the use of this additional channel for seed detection could directly improve the precision of the algorithm, by discarding many false positive detections in background regions \cite{Stegmaier16}. However, the two channels had to be highly colocalized to guarantee that detected seeds were located within the correct compartment. For sequential acquisition of a membrane and a nucleus channel, for instance, it has to be ensured that the acquired membrane image still encompasses the correct nucleus during the slightly later imaged second channel and vice versa.

In image regions with poor signal, the presented algorithm might potentially fail to close gaps in the object boundaries and produce 2D under-segmentation errors. Thus, more complex signal reconstruction approaches such as tensor voting might be inevitable if large gaps are present \cite{Mosaliganti12, Michelin13}. An approach to potentially speed up the tensor voting operation would be to perform it also in a 2D slice-based manner as for the slice-by-slice watershed segmentation presented earlier. Moreover, the presented approach involves parameters related to the global image intensity. If the intensity of the input image heavily varies between slices it might be necessary to perform an additional preprocessing step that normalizes the intensity levels across the considered slices. An additional option might be to automatically determine separate intensity dependent threshold parameters for each of the slices.

In conclusion, the presented method is the first cell shape segmentation algorithm providing real-time processing capabilities to quantitatively analyze and visualize the vast amount of image data produced by modern 3D+t microscopy experiments in developmental biology.
 \cleardoublepage
 
\cleardoublepage

%% Conclusions

%\part{Conclusions (5)}
\chapter{Conclusions}
\label{sec:chap7:Conclusions}
As multidimensional image acquisition techniques are rapidly advancing, automated image analysis has become an indispensable tool to be able to handle the vast amount of generated image data. Especially in cases where datasets of a single experiment are comprised of multiple terabytes of image data, most existing algorithms tend to be far too computationally demanding or only offer limited result quality. Thus, the analyses had to be constrained to small subsets of the data or required an enormous amount of computing power to provide quantitative results reasonably fast. A central intention of the work summarized in this thesis was to develop new concepts and methods that permitted a reliable analysis of large-scale microscopy images significantly faster than existing methods without lowering the quality of automatically obtained results. The main idea to accomplish this goal was to rely on efficient processing operators where possible and to use as much available prior knowledge as possible to compensate potential weaknesses of the algorithms. 

In \cref{sec:chap2:UncertaintyFramework}, a general concept for the mathematical formulation of prior knowledge was proposed and it was shown, how image analysis pipelines can be equipped with the formalized prior knowledge to make more elaborate decisions. Furthermore, the framework includes the propagation of estimated result uncertainties, to be able to inform and improve downstream pipeline operators about the validity of their input data. Besides these general concepts, three new segmentation algorithms were developed that are able to reliably detect geometrical shapes like lines, spheres or plane-like structures in terabyte-scale multidimensional images (\cref{sec:chap3:EfficientSegmentation}). To further improve the result quality, \cref{sec:chap4:AlgorithmEnhancement} demonstrated how existing pipelines can be systematically extended by uncertainty considerations, in order to filter, repair and fuse produced data. A quantitative assessment of the results produced by the extensions proofed their superior performance compared to the plain pipelines and thus render it a feasible approach to efficiently improve the result quality of image analysis pipelines. All methods were implemented in open-source software tools (\cref{sec:chap5:Implementations}) and were successfully applied to analyze various challenging image-based experiments in the field of developmental biology (\cref{sec:chap6:Applications}). In a nutshell, the major contributions presented in this thesis are:
\begin{enumerate}
	\item A new concept on how to formalize, estimate and combine the uncertainty observed in a variety of image analysis problems based on fuzzy set membership degree (\cref{sec:chap2:UncertaintyFramework}).
		
	\item A general framework for the propagation of estimated uncertainty values in image analysis operators. This can be used to efficiently filter extracted data or to allow more complex information sharing compared to linear image analysis pipelines without uncertainty propagation (\cref{sec:chap2:UncertaintyPropagation}).
		
	\item A highly efficient new algorithm for the detection and segmentation of (hyper-) spherical objects such as fluorescently labeled cell nuclei in large 3D microscopy images. The newly developed algorithm reaches segmentation quality of state-of-the-art algorithms while outperforming the existing approaches by a factor of up to ten with respect to processing time (\cref{sec:chap3:SeedDetection}, \cref{sec:chap3:TWANG}).
	
	\item A new segmentation method that can extract comparable regions of interest along arbitrarily oriented line-like objects from multi-channel 2D images. A customized set of image features was developed in order to qualitatively and quantitatively compare intensity and object distributions within and between multiple color channels (\cref{sec:chap3:SeedDetection}, \cref{sec:chap3:SpinalCord}).
	
	\item A new algorithm for the segmentation of locally plane-like structures such as fluorescently labeled cell membranes in 3D microscopy images. The algorithm was specifically developed for the application on large 3D datasets, \ie, it was optimized to be memory efficient and fast compared to existing methods (\cref{sec:chap3:RACE}).
	
	\item The thorough validation of the functionality of the proposed methods on specialized validation benchmarks that were newly developed or adapted if suitable datasets were already available. The robustness of all methods was assessed using variable levels of image noise and image blur (\cref{sec:chap3:SeedDetection}-\cref{sec:chap3:RACE}).
		
	\item An extension of existing and newly developed image analysis operators to estimate and handle fuzzy uncertainty information. The quality of efficient operators could be significantly improved using the proposed fuzzy uncertainty propagation framework (\cref{sec:chap4:AlgorithmEnhancement}).
			
	\item A comprehensive new 3D cell simulation benchmark to model each step of an exemplary image analysis pipeline with the associated ground truth. This benchmark allows to draw conclusions on the performance of the entire processing pipeline instead of only considering individual operators independently (\cref{sec:chap4:Benchmark}).
	
	\item The implementation of all developed methods for the efficient analysis of 3D images in a new open-source software tool named XPIWIT that can be used for large-scale analyses of multidimensional images. The software runs on multiple platforms including Windows, Linux and Mac OS X, and can be executed on cluster systems like Hadoop. An easy-to-use XML-based pipeline definition format is used to execute, share and archive created image analysis protocols (\cref{sec:chap5:XPIWIT}).
	
	\item Extensions of the open-source MATLAB toolbox Gait-CAD to be applicable for processing of multidimensional images (\cref{sec:chap5:ImVID}), object tracking (\cref{sec:chap5:Tracking}), quantification of neuron populations in the spinal cord of zebrafish (\cref{sec:chap5:SpinalCord}), trajectory visualization (\cref{sec:chap5:Embryo3DT}), benchmark generation (\cref{sec:chap5:Benchmark}) and semi-automatic parameter optimization (\cref{sec:chap5:FuzzyParameterAdjustment}).
	
	\item The successful application of the newly developed and implemented spi\-nal cord extension for Gait-CAD to automatically analyze a high-through\-put screen. In particular, the impact of more than $2000$ small molecules on the neuronal development in the spinal cord of zebrafish embryos was quantitatively analyzed using the new software (\cref{sec:chap6:SpinalCord}).
	
	\item An entire processing framework for 3D+t microscopy image data, which comprises registration, segmentation, tracking, visualization and statistical analysis. In close collaboration with the Institute of Toxicology and Genetics at KIT and Janelia Research Campus (Virginia, USA), hundreds of terabytes of 3D+t image data of fruit fly, zebrafish and mouse embryos were successfully analyzed and interpreted (\cref{sec:chap6:TWANG}, \cref{sec:chap6:RACE}). In particular, features like cell counts, cell movements, cell shape changes as well as density variations were analyzed in time-resolved 3D microscopy images. Thus, it was confirmed that the presented methods are capable of extracting quantitative information both reliably and fast even for large terabyte-scale datasets.
\end{enumerate}

%% Extended validation benchmarks
Although the image material considered throughout in this thesis came from different sources and was flawed with different acquisition deficiencies such as noise and blur, the disruptions of images within one dataset were comparable. In real experiments, however, the imaging conditions can vary significantly between different experiments or even change during a single experiment. Causes for varying image content can be, for instance, temporally inhomogeneous expression of fluorescent reporters, varying experimental conditions or inherent variations between different probes or model organisms. Nevertheless, successful image analysis algorithms should be as robust as possible to be able to cope with these varying circumstances. An exemplary initiative aiming to provide standard methods for algorithmic validation and comparison is the Cell Tracking Challenge that was performed during the International Symposium on Biomedical Imaging 2013-2015 \cite{Maska14}. The provided datasets comprise simulated as well as manually annotated datasets of various specimens and microscope types to perform quantitative comparisons of cell segmentation and tracking algorithms. However, the covered simulated datasets tended to be too easy with a small amount of cells and real large-scale microscopy datasets, \eg, of a fruit fly embryo could only be partially annotated due to the enormous manual effort required for the ground truth generation. Thus, more realistic simulated benchmark datasets are required in order to assess and compare the quality of different processing strategies in this challenging scenario. The artificial embryo simulation benchmark presented in this thesis and a recent extension of it presented in \cite{Stegmaier16a} already provide a first step in this direction but need to be further advanced to mimic even more realistic acquisition flaws at different levels of detail including tissue-dependent light scattering, realistic light attenuation as well as object distances and object sizes learned from real datasets.

%% interactive parameter adjustment
To be able to cope with varying imaging conditions and different datasets, algorithmic parameters need to be optimized for the respective application scenario to obtain reasonable results. For the seed point detection step, the concept of interactive post-processing corrections was introduced in \cref{sec:chap5:FuzzyParameterAdjustment}. Algorithmic parameters were initially adjusted such that false negative rates were minimized while the false positive rates potentially increased. In a post-correction step, the user could adapt various filter parameters to interactively clean the data. As the filtering was solely performed on the feature level, an instant visual feedback of the obtained results could be provided to the user without the necessity to reprocess any data on the image level. Thus, laborious parameter adjustments could be performed after the actual processing. This concept can potentially also be used to improve the result quality of other image analysis operators such as segmentation, multiview fusion and tracking, especially with respect to large-scale datasets of varying image content and quality. However, more sophisticated data structures are required, \eg, to temporally store potential segmentation or tracking hypotheses that can be used for subsequent visualization and feature extraction. Moreover, a combination of filters can potentially be used to improve the discrimination capabilities and to be able to also resolve subtle object differences. The uncertainty framework presented \cref{sec:chap2:UncertaintyFramework} could be adapted to work with this kind of post-processing approach. In addition, an interesting extension would be to investigate if and how parameters of the fuzzy uncertainty framework can be automatically adapted or interpolated to match the specific dynamically changing requirements of the acquired image material. For instance, an interactive graphical user interface similar to the one presented in \cref{sec:chap5:FuzzyParameterAdjustment} could be used for sparse manual annotations that in turn help to automatically determine the uncertainty framework parameters before they are applied to the datasets.

%% Tracking quality still not sufficient
With respect to the investigated tracking algorithms, there is still a need for even more accurate methods. Even the best published tracking algorithms have a finite error rate and are thus not yet capable of generating error-free cell lineages automatically without laborious manual proofreading and correction of identified correspondences \cite{Amat14, Faure16}. Hence, a substantial effort is required in the future to further improve the quality of object detection, object segmentation and the detection of events like cell divisions. The manual effort should be reduced to a minimum and supported with interactive and smart correction frameworks that guide the user to specific sites of potential errors and allow for easy corrections.

%% real time processing
As microscopy systems are continuously advancing, it is important to regularly update developed implementations, \eg, to make use of new hardware architectures, parallel programming paradigms or distributed computing solutions, to keep pace with latest developments. Especially, the recently presented time-resolved functional imaging techniques on the whole-embryo level will soon raise new challenges for fast and accurate automatic analysis. The membrane segmentation method presented in this thesis, for instance, provides real-time capabilities when executed on a high-end workstation. In the present state, however, the acquisition and the analysis were still considered as separate steps and were performed individually. An instant qualitative and quantitative feedback as well as live visualizations of biologically relevant features during the experiment would offer unprecedented possibilities to the experimentalists. For instance, to further improve the analysis of single-cell morphology changes, large-scale tissue reorganizations, temporal changes of gene expression patterns or signal molecule transduction at the whole-embryo level. This would allow to study effects of chemical treatments or other perturbations applied to the model systems in detail and at the time of the experiment. Being able to automatically extract error-free lineages and segmentations in real-time with a minimal amount or optimally no user intervention will ultimately enable truly interactive biological experiments on a daily basis.
 
\cleardoublepage

%% Appendix

%%%%%%%%%%%%%%%%%%%%%%% APPENDIX %%%%%%%%%%%%%%%%%%%%%%%%
\appendix

%%%%%%%%%%%%%%%%%%%%%%% APPENDIX %%%%%%%%%%%%%%%%%%%%%%%%
%%%%%%%%%%%%%%%%%%%%%%% SYMBOLS %%%%%%%%%%%%%%%%%%%%%%%%
\chapter{Nomenclature and Symbols}

In this section the notation used throughout the thesis is summarized. The initial enumeration demonstrates the basic concepts used for nomenclature and the subsequent list of symbols explains all used symbols within their context. The nomenclature is partly based on the formulation given in \cite{Mikut08Habil}.

\begin{itemize}
	\item Lower case letters are used for scalars, parameters, indexing variables and functions (\eg, $a,b,c,d$ are common names for parameters, $f(\cdot),g(\cdot), h(\cdot)$ are common names for functions or functionals, $i,j,k,l$ are common names for indices and finally $x,y,z$ are common names for scalars or variables)
	\item Upper case letters are used for number of entities (\eg, $N,N_{\text{f}}$ for the number of data tuples and the number of features, respectively) 
	\item Upper case bold face letters represent matrices (\eg, $\mathbf{M,N}$)
	\item Lower case bold face letters represent vectors (\eg, $\mathbf{x,y}$ for column vectors and $\mathbf{x^\top,y^\top}$ for row vectors)
	\item Upper case script and Greek letters represent sets (\eg, $\Omega, \Theta, \mathcal{X}, \mathcal{L}$)
	\item Accessing elements of data structures is denoted by square brackets (\eg, $\mathbf{x}[1]$ accesses the first element of vector $\mathbf{x}$, $\mathbf{M}[1,2]$ accesses the matrix element in the first row and second column)
	\item Elements topped with a hat refer to estimates of a certain quantity (\eg, $\hat{x}$ represents an estimate of the scalar $x$, $\mathbf{\hat{x}}$ represents an estimate of vector $\mathbf{x}$)
	\item Elements topped with a tilde refer to a changed quantity (\eg, $\tilde{\mathcal{X}}$ represents a modified subset of $\mathcal{X}$)
	\item Optimal solutions are indicated by an asterisk (\eg, $x^{\ast}$ would denote the optimal solution to $\min(f(x))$)
\end{itemize}

%\newpage

%\bgroup
%\def\arraystretch{1.5}

\begin{longtable}{m{.25\columnwidth}m{.7\columnwidth}}%[!htb]
%\rowcolors{2}{white}{gray!25}
%\begin{tabular}{m{.25\columnwidth}m{.7\columnwidth}}
\toprule
\textbf{Symbol}&\textbf{Description} \\
\midrule
$|x|$												& Absolute value of $x$ \\
$\langle \mathbf{x} , \mathbf{y} \rangle$			& Dot product of two vectors $\mathbf{x}$ and $\mathbf{y}$, equivalent to $\mathbf{x}^\top \cdot \mathbf{y}$ \\
$\ast$												& The convolution operator. Discrete convolution is used for images. \\
$\parallel \mathbf{x} \parallel_2$					& Euclidean norm of a vector $\mathbf{x}$ \\
$\bigtriangledown$									& The gradient operator. Finite differences are used to calculate image gradients. \\
$\circ$												& Element-wise multiplication of two vectors or matrices (Hadamard product) \\
$\wedge$											& Logical AND-operator \\
$\lor$												& Logical OR-operator \\
$ \frac{\partial f}{\partial x}$					& Partial derivative of function $f$ with respect to $x$ \\
$\mathbf{0}$                  						& Matrix or vector containing only zeros \\
$\mathbf{1}$                  						& Matrix or vector containing only ones \\
2D                            						& Two dimensional \\
2D+t												& Two spatial dimensions and temporal dimension, e.g. 2D videos \\
3D                            						& Three dimensional \\
3D+t                         						& Three spatial dimensions and temporal dimension, e.g. 3D videos \\
AEL													& \textbf{A}fter \textbf{E}gg \textbf{L}aying (used to specify the age of \textit{Drosophila} embryos) \\
$c$													& Index variable for image channels \\
$c_\text{z}$										& z component of the centroid of an investigated specimen \\
CCD													& \textbf{C}harge-\textbf{c}oupled \textbf{D}evices \\
CMOS												& \textbf{C}omplementary \textbf{M}etal \textbf{O}xide \textbf{S}emiconductor-based Detectors \\
CNR												& \textbf{C}ontrast-to-\textbf{N}oise \textbf{R}atio \\
$\mathbf{d}_s(\mathbf{x})$							& Vector pointing from seed location $\mathbf{x}_s$ to $\mathbf{x}$ \\
$D(i)$												& Minimum distance of pixel $i$ to the closest edge of a reference shape (used to calculate HM and NSD)\\
DNA													& \textbf{D}eoxyribo\textbf{n}ucleic \textbf{A}cid \\
DoG													& \textbf{D}ifference-\textbf{o}f-\textbf{G}aussian \\
DOM													& \textbf{D}egree \textbf{o}f \textbf{m}embership in the context of fuzzy sets \\
%\bottomrule
%\end{tabular}
%\end{table}

%\begin{table}[htb]
%\rowcolors{2}{white}{gray!25}
%\begin{tabular}{m{.25\columnwidth}m{.70\columnwidth}}
%\toprule
%\textbf{Symbol}&\textbf{Description} \\
%\midrule
DoM													& \textbf{D}ifference-\textbf{o}f-\textbf{M}ean \\
DSLM												& \textbf{D}igital \textbf{S}canned Laser \textbf{L}ight Sheet Fluorescence \textbf{M}icroscopy \\
EDM													& \textbf{E}uclidean \textbf{D}istance \textbf{M}ap \\
EDT													& \textbf{E}uclidean \textbf{D}istance \textbf{T}ransform \\
$\text{exp}(\cdot), e^x$							& Exponential function \\
$f$													& Index variable for features \\
FSMD												& \textbf{F}uzzy \textbf{S}et \textbf{M}embership \textbf{D}egree \\
$\mathbf{g}_s(\mathbf{x}, \sigma_{\text{grad}})$	& Image gradient for seed point $s$ at location $\mathbf{x}$ with regularization scale $\sigma_{\text{grad}}$ \\
$\mathbf{G}(\mathbf{x}, \sigma)$					& Gaussian kernel with standard deviation $\sigma$ \\
$\texttt{gi}^{*}_{c}$								& Global intensity feature like min, max, mean, median or variance of channel $c$ of an image \\
GFP													& \textbf{G}reen \textbf{F}luorescent \textbf{P}rotein \\
GNU-GPL												& GNU \textbf{G}eneral \textbf{P}ublic \textbf{L}icense \\
$h$													& The height of maxima/minima that are suppressed by the $h$-maximum and $h$-minimum filters \\
$\mathbf{H}$										& 2D histogram matrix containing label correspondences of a reference image and an automatic segmentation \\
HM													& \textbf{H}ausdorff \textbf{M}etric \\
hpf													& \textbf{H}ours \textbf{P}ost \textbf{F}ertilization (used to specify the age of zebrafish embryos) \\
$i,j,k,l,m,n$										& Index variables defined by the context \\
$\texttt{ip}_c$										& Area under the curve of an intensity profile $\mathbf{p}_c$ of channel $c$ \\
$\texttt{ip}^{*}_c$									& Features such as area under the curve, min, max, mean, median or variance of an intensity profile $\mathbf{p}_c$ of channel $c$ \\
$i_\text{x}, i_\text{y}, i_\text{z}$				 		 				& Index variables for x,y,z direction \\
$\mathbf{I}$										& General image variable \\
$\mathbf{I}^{\text{cl}}_r$							& Iteratively closed image using a maximum structuring element radius $r$ \\
$\mathbf{I}^{\text{cr}}$						    & Cropped image \\
$\mathbf{I}^{\text{cr}}_c$						    & Channel $c$ of a cropped image \\
$\mathbf{I}^{\text{cr}}_s$						    & Cropped image for seed point $s$ \\
%\bottomrule
%\end{tabular}

%\begin{table}[htb]
%\rowcolors{2}{white}{gray!25}
%\begin{tabular}{m{.25\columnwidth}m{.70\columnwidth}}
%\toprule
%\textbf{Symbol}&\textbf{Description} \\
%\midrule

$\mathbf{I}^{\text{dark}}$						    & Simulated dark current image of the detector \\
$\mathbf{I}^{\text{edm}}$						    & Euclidean distance map image \\
$\mathbf{I}^{\text{final}}$						    & Final simulated microscopy image containing all acquisition simulation steps \\
$\mathbf{I}^{\text{hee}}$						    & Image of the Hessian-based objectness filter response \\
$\mathbf{I}^{\text{iedm}}$						    & Euclidean distance map image of an inverted input image \\
$\mathbf{I}^{\text{LoG}}$						    & Laplacian-of-Gaussian filtered image \\
$\mathbf{I}^\text{LoG}(\mathbf{x}, \sigma)$			& Laplacian-of-Gaussian with standard deviation $\sigma$  at location $\mathbf{x}$ \\
$\mathbf{I}^{\text{LoGMP}}$						    & Maximum intensity projection of multiple Laplacian-of-Gaussian filtered images \\
$\mathbf{I}^\text{LoGMP}(\mathbf{x}, \sigma_{\text{min}}, \sigma_{\text{max}})$	& Laplacian-of-Gaussian maximum intensity projection of scales $\sigma_{\text{min}}$ to $\sigma_{\text{max}}$ at location $\mathbf{x}$ \\
$\mathbf{I}^{\text{med}}$						    & Median filtered image defined by the context \\
$\mathbf{I}^{\text{MS}}$						    & Image containing the scale that yielded respective maximum value for $\mathbf{I}^{\text{LoGMP}}$ \\
$\mathbf{I}^\text{MS}(\mathbf{x}, \sigma_{\text{min}}, \sigma_{\text{max}})$	& The scale yielding the maximum value of $\text{LoGMP}(\mathbf{x}, \sigma_{\text{min}}, \sigma_{\text{max}})$ \\
$\mathbf{I}^{\text{psf}}$						    & Image of a point spread function (PSF) \\
$\mathbf{I}^{\text{raw}}$						    & Unprocessed raw image \\
$\mathbf{I}^{\text{ref}}$						    & Reference image, \ie, a labeled ground truth image \\
$\mathbf{I}^{\text{seg}}$						    & Segmentation image obtained with an automatic segmentation algorithm \\
$\mathbf{I}^{\text{ws}}$						    & Slice-based 2D watershed segmentation image \\
ITK													& \textbf{I}nsight \textbf{T}ool\textbf{k}it \\
JI	 												& \textbf{J}accard \textbf{I}ndex \\
$k$												   	& Index variable for time-discrete time series \\
$l$												   	& Index variable for linguistic terms \\
LoG												   	& \textbf{L}aplacian-\textbf{o}f-\textbf{G}aussian \\
LSFM												& \textbf{L}ight-\textbf{S}heet \textbf{F}luorescence \textbf{M}icroscopy \\
$M$													& General number of data tuples, objects or dimensions specified by the context \\
MBF													& Abbreviation for \textbf{m}em\textbf{b}ership \textbf{f}unction \\
%\bottomrule
%\end{tabular}
%\end{table}

%\begin{table}[htb]
%\rowcolors{2}{white}{gray!25}
%\begin{tabular}{m{.25\columnwidth}m{.70\columnwidth}}
%\toprule
%\textbf{Symbol}&\textbf{Description} \\
%\midrule
$\min(\cdot)$, $\max(\cdot)$						& Minimum and maximum functions \\
$n$													& Index variable for data tuples \\
$\mathbf{n}_s(\mathbf{x})$							& Normal vector of seed point $s$ at image location $\mathbf{x}$ \\
$N$													& General number of data tuples, objects or dimensions specified by the context \\
$N_{\text{c}}$												& Number of channels of a multi-channel image \\
$N_{\text{cl}}$							  				& General number of combined linguistic terms \\
$N_{\text{cl}, i}$							  				& Number of combined linguistic terms for operator $i$  \\
$N_\text{d}$												& Number of dimensions \\
$N_{\text{f}}$												& General number of features \\
$N_{\text{f},i}$											& Number of extracted features produced by operator $i$ (i.e. columns in the result matrix) \\
$N_{i}$							  			 		& Number of data tuples produced by operator $i$ (i.e. rows in the result matrix) \\
$N_{\text{k}}$												& Number of sampling points of time series or videos \\
$N_{\text{l}}$							  			 		& General number of linguistic terms \\
$N_{\text{l},i}$							  				& Number of linguistic terms used by operator $i$ \\
$N_{\text{ts}}$											& Number of time series \\
$N_{\text{ts}, i}$											& Number of time series produced by operator $i$ \\
$N_{\text{op}}$											& Number of operators within an analysis pipeline \\
$N_{\text{p}}$												& Dimensionality of the a parameter vector, \ie, the number of parameters \\
$N_{\text{s}}$												& Number of seeds detected in an image \\
$N_{\text{x}},N_{\text{y}},N_{\text{z}}$	    							& Number of pixels in x,y,z direction \\
$\mathcal{N(\mu, \sigma)}$							& Normally distributed random variable with mean $\mu$ and standard deviation $\sigma$ \\
$\mathbf{nd}_c$										& Vector of variance normalized mean deviation of an intensity profile from a reference profile \\
NSD													& \textbf{N}ormalized \textbf{S}um of \textbf{D}istances \\
$O(\pmb{\lambda})$									& Response of the objectness filter for the eigenvalue vector $\pmb{\lambda}$ \\
$\mathbf{p}_c$ 										& Vector containing the intensity profile of image channel $c$ \\
%\bottomrule
%\end{tabular}
%\end{table}

%\begin{table}[htb]
%\rowcolors{2}{white}{gray!25}
%\begin{tabular}{m{.25\columnwidth}m{.70\columnwidth}}
%\toprule
%\textbf{Symbol}&\textbf{Description} \\
%\midrule
$\bar{\mathbf{p}}_{c}^{\text{ctrl}}$ 				& Vector containing the averaged intensity profiles of a control group of image channel $c$ \\
PCL													& \textbf{P}oint \textbf{C}loud \textbf{L}ibrary \\
PDE													& \textbf{P}artial \textbf{D}ifferential \textbf{E}quation \\
$\texttt{plp}_c$ 									& Peak location of intensity profile $\mathbf{p}_c$ of channel $c$ \\
PSF													& \textbf{P}oint \textbf{S}pread \textbf{F}unction \\
$\texttt{pvp}_c$									& Peak value of intensity profile $\mathbf{p}_c$ of channel $c$ \\
$\texttt{pwp}_c$									& Peak width of intensity profile $\mathbf{p}_c$ of channel $c$ \\
$\mathcal{P}_{\text{bg}}$							& The set of background pixels in a binary image \\
$\mathcal{P}_{\text{fg}}$							& The set of foreground pixels in a binary image \\
$P_\lambda(.)$										& Poisson process to simulate photon shot noise using the image intensities as mean \\
$q_{\text{l}}, q_{\text{u}}$ 						& Parameters for lower and upper quantile saturation \\
$r$													& General variable for radius \\
$\mathbb{R}$										& The set of real numbers \\
$\mathbb{R}^{N_d}$									& The $N_d$-dimensional vector space of real numbers \\
$r_\text{min}, r_\text{max}$						& Minimum and maximum radius of objects specified in the context \\
$r_s$												& Approximate radius of seed point $s$ detected in the LoGMP image \\
RACE												& \textbf{R}eal-time \textbf{A}utomated \textbf{C}ell shape \textbf{E}xtractor \\ 
$R_A, R_N, R_M$										& Parameters used for simulated object interaction terms \\
RFP													& \textbf{R}ed \textbf{F}luorescent \textbf{P}rotein \\
RGB													& \textbf{R}ed \textbf{G}reen \textbf{B}lue (color specification convention) \\
RI													& \textbf{R}and \textbf{I}ndex \\
ROI													& \textbf{R}egion \textbf{o}f \textbf{I}nterest \\
$\mathbf{r}_\text{x}, \mathbf{r}_\text{y}$			& Vectors containing the abscissa and the ordinate of a regression curve \\
$s$													& Index variable for seed points \\
$\mathbf{s}$										& Vector containing the physical spacing of an image along each dimension \\
%\bottomrule
%\end{tabular}
%\end{table}

%\begin{table}[htb]
%\rowcolors{2}{white}{gray!25}
%\begin{tabular}{m{.25\columnwidth}m{.70\columnwidth}}
%\toprule
%\textbf{Symbol}&\textbf{Description} \\
%\midrule
$S$													& Frobenius norm of the vector of eigenvalues \\
$\mathbf{S}_r$										& Euclidean sphere structuring element with radius $r$ \\
SDK													& \textbf{S}oftware \textbf{D}evelopment \textbf{K}it \\
SNR													& \textbf{S}ignal-to-\textbf{N}oise \textbf{R}atio \\
$\text{sgn}(\cdot)$									& Signum function that returns the sign of a real number \\ 
SPIM												& \textbf{S}elective/\textbf{S}ingle \textbf{P}lane \textbf{I}llumination \textbf{M}icroscopy \\
$t_\text{dbc}$										& Distance-based cutoff value used for feature fusion \\			
$t_\text{wmi}$										& Threshold for the window mean intensity feature \\
TWANG												& \textbf{T}hreshold of \textbf{W}eighted Intensity \textbf{A}nd Seed-\textbf{N}ormal \textbf{G}radient Dot Product Image \\
VTK													& \textbf{V}isualization \textbf{T}ool\textbf{k}it \\
$w_s(\mathbf{x})$									& TWANG weighting kernel at image location $\mathbf{x}$ for seed point $s$\\
$w_{\text{adh}}, w_{\text{rep}}, w_{\text{bdr}}$ & Weights for the adhesive displacement, the repulsive displacement and the boundary displacement \\
$x$													& General variable defined by the context \\
$\mathbf{x}$										& General vector variable defined by the context \\
$\mathbf{x}_s$										& Position of seed point $s$ \\
$x_i[n,l]$											& $l$-th feature in the $n$-th data tuple produced by the $i$-th operator \\
$\mathbf{x}_{i}[n]$		 							& $n$-th data tuple produced by the $i$-th operator containing $N_{f,i}$ feature entries \\
$\mathbf{X}_{i}$					  				& Result matrix of the $i$-th operator containing $N_i$ data tuples with $N_{f,i}$ features \\
$\mathcal{X}_i$										& Variable for an output set of operator $i$\\
XML													& E\textbf{x}tensible \textbf{M}arkup \textbf{L}anguage \\
XPIWIT												& \textbf{X}ML \textbf{Pi}peline \textbf{W}rapper for the \textbf{I}nsight \textbf{T}oolkit \\
$\alpha_{il}$										& Forward threshold of operator $i$ and linguistic term $l$ \\
%$\alpha_{i_x}$										& Angle at pixel $i_x$ used for the rotation correction of elongated objects \\
$\beta_{il}$										& Backwards threshold of operator $i$ and linguistic term $l$ \\
$\beta_{\text{hee}}, \gamma_{\text{hee}}$			& Parameters to control the response of the Hessian-based objectness filter \\
%\bottomrule
%\end{tabular}
%\end{table}

%\begin{table}[t]
%\rowcolors{2}{white}{gray!25}
%\begin{tabular}{m{.25\columnwidth}m{.70\columnwidth}}
%\toprule
%\textbf{Symbol}&\textbf{Description} \\
%\midrule
$\Delta\textbf{x}^{\text{adh}}$							& Adhesive displacement vector of two neighboring objects \\
$\Delta\textbf{x}^{\text{bdr}}$							& Boundary constraint for dynamic object simulation \\
$\Delta\textbf{x}^{\text{rep}}$							& Repulsive displacement vector of two neighboring objects \\
$\Delta\textbf{x}^{\text{tot}}$							& Sum of all displacement vectors of a single object \\
$\theta$											& General parameter specified in the context \\
$\pmb{\theta}$										& General parameter vector specified in the context \\
$\lambda_1, \lambda_2, ..., \lambda_{N_d}$ 			& Eigenvalues sorted in ascending order of a $N_d \times N_d$ matrix defined in the context \\
$\pmb{\lambda}$										& $N_d$-dimensional vector of eigenvalues sorted in ascending order \\
$\pmb{\lambda}(\mathbf{x}, \sigma_{\text{hee}})$ 	& Vector of eigenvalues of the Hessian calculated at location $\mathbf{x}$ of a $\sigma_{\text{hee}}$-regularized input image \\
$\mu$												& Arithmetic mean value \\
$\mu_{\mathcal{A}}(\cdot)$						  			& Membership function for the fuzzy set $\mathcal{A}$  \\
$\pmb{\mu}_{\mathcal{A}_l}(\cdot)$					& Vector of membership functions for all features of the $l$-th linguistic term  \\
$\pmb{\mu}_{{\mathcal{A}_{il}}}(\cdot)$				& Vector of membership functions for all features of the $l$-th linguistic term of processing operator $i$  \\
$\mu_{A_{ifl}}(\cdot)$								& Membership function of feature $f$ and linguistic term $l$ of the $i$-th processing operator \\
$\mu_{c}^{\text{nd}}$ 								& Average deviation of an intensity profile from an averaged reference profile \\
$\mu_{\text{fg}}$ 									& Mean intensity of the foreground signal (used for SNR calculation) \\
$\pi$												& Mathematical constant (approximately $3.14159$) \\
$\sigma$											& Standard deviation \\
$\sigma^2$											& Variance \\
$\sigma_{\text{agn}}$								& Zero-mean additive Gaussian noise standard deviation \\
$\sigma_{\text{bg}}$ 								& Standard deviation of the background signal of an image (used for SNR calculation) \\
$\pmb{\sigma}_c^{\text{ctrl}}$ 						& Standard deviation of the control profiles of channel $c$ \\
$\sigma_{\text{grad}}$								& TWANG segmentation regularization scale used to smooth the vector field \\
%\bottomrule
%\end{tabular}
%\end{table}

%\begin{table}[htb]
%\rowcolors{2}{white}{gray!25}
%\begin{tabular}{m{.25\columnwidth}m{.70\columnwidth}}
%\toprule
%\textbf{Symbol}&\textbf{Description} \\
%\midrule
$\sigma_{\text{hee}}$								& Regularization scale of the Hessian-based objectness filter \\
$\sigma_{\text{kernel}}$							& TWANG segmentation weighting kernel standard deviation \\
$\sigma_{\text{min}}$								& Minimum scale used for the LoG-based seed detection \\
$\sigma_{\text{max}}$								& Maximum scale used for the LoG-based seed detection \\
$\sigma_{\text{seed}}$								& The scale of the LoG-filtered image where a seed point was detected at \\
$\sigma_{\text{smooth}}$							& The standard deviation of a Gaussian smoothing kernel \\
$\sigma^2_{\text{smooth}}$							& The variance of a Gaussian smoothing kernel \\
$\sigma_{\text{step}}$								& Step size between adjacent scales used for the LoG-based seed detection \\
$\pmb{\Sigma}$										& Covariance matrix \\
$\pmb{\Sigma}_{\text{row}}$, $\pmb{\Sigma}_{\text{col}}$ & Vectors containing the row sums of all columns and the column sums of all rows of the 2D histogram $\mathbf{H}$, respectively \\
$\phi_s(\mathbf{x})$								& Normalized dot product of image gradient and seed point normal at image location $\mathbf{x}$ for seed point $s$\\
$\Phi$												& Variable for a set \\
$\chi_{\mathcal{C}}(x)$								& Characteristic function of the set $\mathcal{C}$ \\
$\omega_{\text{kpm}}$								& TWANG segmentation weighting kernel plateau multiplier \\
$\Omega_i$											& Variable for set that combines subsets of the outputs of operator $i$ and operator $i-1$ \\
$\tilde{\Omega}_i$	  								& Variable for a modified input set of operator $i+2$ \\
\bottomrule
%\end{tabular}
%\vspace*{18cm}
%\end{table}
\end{longtable}

 \cleardoublepage

%%%%%%%%%%%%%%%%%%%%%%% APPENDIX %%%%%%%%%%%%%%%%%%%%%%%%
%%%%%%%%%%%%%%%%%%%%%%% METHODS %%%%%%%%%%%%%%%%%%%%%%%%%
\chapter{Infrastructure}
\label{sec:Appendix:Infrastructure}
Although the algorithms and analysis methods presented in this thesis were developed with the focus on time and memory efficiency, the tremendous amount of multidimensional image data produced by modern microscopes demands for a specialized infrastructure to store, process, analyze and archive the acquired image data. This chapter introduces the platforms and services hosted by the Steinbuch Center for Computing (SCC) at the Karlsruhe Institute of Technology (KIT) that were used to establish the automated processing pipeline described in \cref{sec:chap6:Applications}. Additionally, it is demonstrated how XPIWIT was used for parallel processing within this framework and which workarounds were necessary to cope with observed hardware limitations.

\section{Large Scale Data Facility for Data Storage and Processing}
\label{sec:Appendix:LSDF}
The Large Scale Data Facility (LSDF) hosted at KIT has been providing storage, data management services as well as a computing facility that could be used for data analysis since 2009 \cite{Garcia11IPDP}. It has been established in order to support large-scale scientific experiments, such as high-throughput microscopy as discussed in \cref{sec:chap6:Applications}. Essentially, the LSDF framework provided storage capacity of multiple petabytes that was made accessible through high-speed network connections. Acquired microscopy data was transferred to this large storage system via optical fibers and was archived there for further processing and analysis of the images \cite{Kobitski15}. Data stored on the LSDF could be mounted locally via SSHFS (Secure SHell File System) or a Windows network drive to provide easy access to the raw data and the generated results.

To accelerate the result generation and automated analysis of the acquired image data, the processing was parallelized using an Apache Hadoop-based cluster for distributed computing hosted at SCC. The cluster consisted of two head nodes for user interaction equipped with two quad-core Intel Xeon E5520 CPUs @ 2.27~GHz and $96$~GB of main memory. The actual work was distributed over $30$ worker nodes each containing the same processors as the head-nodes, $36$~GB of memory, 2~TB hard disk drives and a Gigabit network connection. Scientific Linux $5.5$ was used as an operating system on all nodes of the cluster. To be able to access raw image data and to store processed data, a gigabit network connection permitted fast data transfer between the LSDF and the respective processing nodes \cite{Garcia11ICDIM}.

\section{Hadoop Streaming to Parallelize XPIWIT}
\label{sec:Appendix:HadoopCluster}
To speed up the image analysis pipeline implemented in XPIWIT, it was executed on a Hadoop-based cluster. Hadoop represents an open-source implementation of Google's map reduce framework and allows to easily parallelize applications, in case the computations are divisible into map tasks that independently compute intermediate results and reduce tasks that combine intermediate results to a final outcome \cite{Dean08}. Usually, Hadoop is used by specialized Java implementations of the map and reduce functions on the basis of the Hadoop libraries for Java. As XPIWIT was implemented in C++, Hadoop's streaming functionality was used instead of a native implementation in Java. Hadoop streaming essentially allows to specify arbitrary executables to be used as mappers or reducers. For the segmentation of thousands of images, processing each of the individual images was considered as an independent map task and could thus be processed in parallel on different nodes. A reducer was not used in this case as the segmentation did not need to combine information of individual images. However, the reducer might be used to combine obtained results, \eg, to perform object tracking after all separately executed mappers have successfully processed the required object centroids.

\begin{figure}[htbp]
\centerline{\includegraphics[width=\columnwidth]{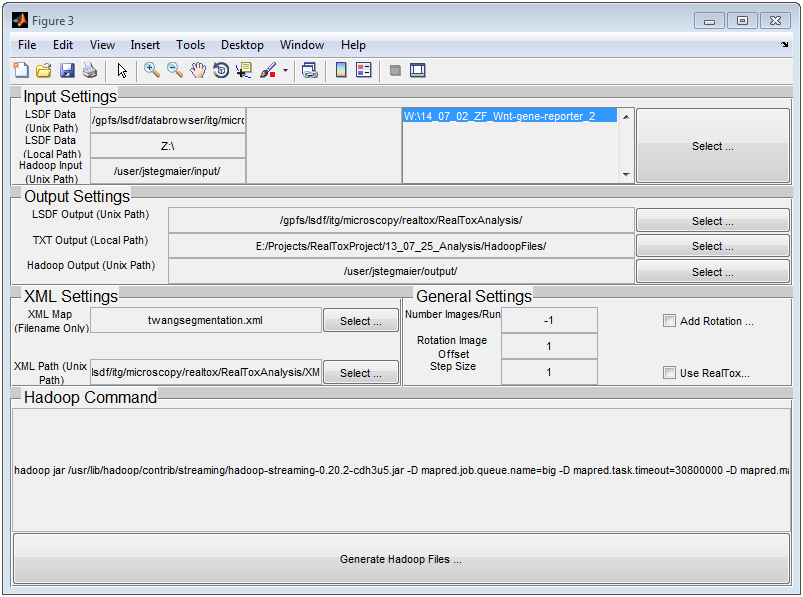}}
\caption[Screenshot of the Hadoop input file generator]{Screenshot of the Hadoop input file generator implemented in MATLAB. Based on network drive access to the LSDF, the desired folders for input images, result output and the XML pipeline used for processing can be specified. Additionally, the Hadoop command and folder related settings can be adjusted if necessary.}
\label{fig:Appendix:HadoopFileGenerator}
\end{figure}

To inform the Hadoop job about the data to be processed, an XPIWIT input file had to generated for each of the images. A special purpose MATLAB GUI was created to facilitate this task and is shown in \app\cref{fig:Appendix:HadoopFileGenerator}.
A selected input folder on the LSDF was parsed for valid image files and for each of the images a unique XPIWIT input file was generated. An exemplary input file containing the required input, output and XML paths on the LSDF is listed in \app\cref{lst:Appendix:XPIWITInputFile}.
\bgroup
\lstset{language=TXT, numbers=left, breaklines=true, tabsize=2, numbersep=25pt, backgroundcolor=\color{lightgray}}
\begin{lstlisting}[caption={XPIWIT Input Text Files Distributed by Hadoop.},label=lst:Appendix:XPIWITInputFile]
--output /gpfs/lsdf/.../14_07_09_ZF_nc/Processed/, result 
--input 0, /gpfs/lsdf/.../14_07_09_ZF_nc/myimage_t=0012.btf, 3, float 
--xml /gpfs/lsdf/.../XMLFiles/mysegmentation.xml
--seed 792
--lockfile on 
--subfolder filterid, filtername 
--outputformat imagename, filtername 
--end
\end{lstlisting}
\egroup

Once all input files had been generated and copied to the Hadoop file system (HDFS), the actual streaming job was started. In \app\cref{lst:Appendix:StreamingCommand}, the command used to start the Hadoop streaming batch process is shown. Besides the essential call of the Hadoop streaming application, several parameters had to be provided to successfully start XPIWIT-based jobs on the cluster. Most importantly, the input directory, which was located on the HDFS, had to be specified. It contained the previously generated XPIWIT input files for each of the images about to be processed and a map task was generated for each of these files. The job scheduler took care of equally distributing the map tasks among all available processing nodes. The corresponding input parameters for the mapper, reducer and their possible dependencies to shared libraries or other files are specified in Lines 9-14 in \app\cref{lst:Appendix:StreamingCommand}.

\lstset{language=TXT, numbers=left, breaklines=true, tabsize=2, numbersep=25pt,	backgroundcolor=\color{lightgray}}
\begin{lstlisting}[caption={Hadoop Streaming Command used for Processing.},label=lst:Appendix:StreamingCommand]
hadoop jar /usr/lib/.../hadoop/contrib/streaming/hadoop-streaming-0.20.2-cdh3u5.jar 
-D mapred.map.max.attempts=50 
-D mapred.job.queue.name=big 
-D mapred.task.timeout=30800000 
-D mapred.map.tasks=1 
-D mapred.reduce.tasks=0 
-input "/user/.../input/14_08_05_ZF_nc/" 
-output "/user/.../output/14_08_05_ZF_nc/" 
-mapper "XPIWIT" 
-reducer 0 
-file "/gpfs/lsdf/software/.../XPIWIT/XPIWIT" 
-file "/gpfs/lsdf/software/.../XPIWIT/libtiff.so.5" 
-file "/gpfs/lsdf/software/.../XPIWIT/libmng.so.1" 
-file "/gpfs/lsdf/software/.../XPIWIT/liblcms.so.1"
\end{lstlisting}

Several problems were observed with the Hadoop streaming framework with respect to processing huge datasets. Most importantly, the job scheduler of the cluster was configured to start multiple map tasks at once on each of the cluster nodes. For applications with low memory demands, this might be a reasonable behavior, as the processing times generally should decrease by starting multiple jobs simultaneously. However, processing one of the large 3D images already required to make use of the entire main memory that was available on the respective cluster node. To prevent multiple tasks from being started concurrently on a single node, a so called lockfile-feature was added to XPIWIT, \ie, a node specific text file was generated as soon as a node was occupied by an instance of XPIWIT (\app\cref{lst:Appendix:XPIWITInputFile}, Line 5). Each idle process frequently checked if the lockfile for the node it was running on still was present on disk and remained in the idle state as long as the file existed. As soon as the lockfile got deleted, the next instance of XPIWIT started processing its data. However, one problem still occurred, if two or more processes wrote their lockfile exactly at the same time to the disk. In this case, both map-tasks were allowed to process and the node memory started swapping to disk. To prevent this rare event, a randomized sleep timer was used, which made the processes sleep for a short random period of time, before they started to search for the lockfile. However, if processes were generated at the same time on the same node with exactly the same starting time used as seed for the random number generator, the problem persisted. Therefore, the random numbers were not seeded with the current time but with a special seed that was provided from the XPIWIT input file to each individual map-task (\app\cref{lst:Appendix:XPIWITInputFile}, Line 4). This solved the issue of simultaneous processes on a single node and finally allowed to instantly start several thousands of tasks with a single call of Hadoop Streaming. Segmenting an entire embryo dataset with the TWANG algorithm described in \cref{sec:chap3:TWANG} could be processed approximately within 12 hours using the described cluster and parallelization strategy. Performing the same task on a single high-end workstation computer would have required approximately two weeks, \ie, the use of efficient parallelization is crucial in order to obtain experimental results in a contemporary way.

To further automate the image analysis and to facilitate the processing for non-experts, it is planned to directly integrate access to processing software via the data ingest tool developed at the Institute for Data Processing and Electronics (IPE) at KIT \cite{Jejkal12}.

%%%%%%%%%%%%%%%%% MATHEMATICS (Reconsider if really necessary...) %%%%%%%%%%%%%%%%%%%
%\chapter{Mathematics}
%\begin{itemize}
%	\item discrete convolution
%	\item image gradient operation
%	\item dot product ($\langle.,.\rangle$-operator)
%	\item image regularization using a gaussian
%	\item Hessian matrix of an image ([Ix,Iy,Iz] = gradient(I); [Ixx,Iyx,Iyz] = gradient(Ix); [Iyx,Iyy,Iyz] = gradient(Iy); [Izx,Izx,Izz] = gradient(Iz))%
%	\item Eigenvalues of the hesse matrix?
%\end{itemize}

\chapter{Performance Assessment}
\label{sec:Appendix:PerformanceAssessment}
An important component of algorithmic development is the quality assessment of the results produced by new methods and to compare these methods to existing approaches. The quantification and identification of errors allows to identify problems of algorithms and the possibility to improve their outcome by focusing on the particular weaknesses. This chapter summarizes the validation measures that were used to quantitatively compare seed detection, segmentation and tracking approaches as described in \cref{sec:chap3:EfficientSegmentation} and \cref{sec:chap4:AlgorithmEnhancement}.

%%%%%%%%%%%%%%%%%%%%%%%%%%%%%%%%%%%%%%%%%%%%%%%%%%%%%%%%%%%%%%%%%%%%%%%%%%
%% Performance Evaluation of the Seed Point Detection
%%%%%%%%%%%%%%%%%%%%%%%%%%%%%%%%%%%%%%%%%%%%%%%%%%%%%%%%%%%%%%%%%%%%%%%%%%
\section{Seed Point Detection}
\label{sec:Appendix:SeedDetectionPerformance}
To quantify the quality of a seed detection algorithm, the extracted spatial positions were compared to the original centroids of the ground truth objects. Due to the availability of segmentation ground truth, it was possible to simply check if a detected seed was localized within one of the object segments. If this was the case the seed was considered a \textit{true positive (TP)} detection. If a detected seed was located in the background area, it was considered a \textit{false positive (FP)}. The case of a \textit{false negative (FN)} detection corresponds to an object present in the image data that was not associated to any of the detected seeds. The last remaining possibility, \ie, a \textit{true negative (TN)} is not considered, as in principal each background pixel that is not classified as a seed point corresponds to this case. 

On the basis of the \textit{TP, FP} and \textit{FN}, the statistical measures recall, precision and F-Score were calculated as:
\begin{gather}
 	\text{recall} = \frac{TP}{TP+FN} \label{eq:Appendix:Recall}
\end{gather}
\begin{gather}
 	\text{precision} = \frac{TP}{TP+FP} \label{eq:Appendix:Precision}
\end{gather}
\begin{gather}
 	\text{F-Score} = 2 \cdot \frac{\text{recall} \cdot \text{precision}}{\text{recall}+\text{precision}} \label{eq:Appendix:FScore}
\end{gather}
The values of recall, precision and F-Score all lie within the interval $[0,1]$, with larger values being better. Recall reflects the relative amount of objects that were retrieved, \ie, a value of 1 indicates that all objects were successfully identified. Contrary, precision measures the relative amount of correct objects out of all detected objects, \ie, a value of 1 indicates that only correct objects and no false positives were detected. F-Score combines both measures to a single value and represents the harmonic mean of recall and precision. 

For all correct detections, the Euclidean distance of the detected seed point to the centroid of the respective segment was additionally calculated to estimate how close the automatic detections are to the true object locations.

%%%%%%%%%%%%%%%%%%%%%%%%%%%%%%%%%%%%%%%%%%%%%%%%%%%%%%%%%%%%%%%%%%%%%%%%%%
%% Segmentation
%%%%%%%%%%%%%%%%%%%%%%%%%%%%%%%%%%%%%%%%%%%%%%%%%%%%%%%%%%%%%%%%%%%%%%%%%%
\section{Segmentation}
\label{sec:Appendix:SegmentationEvaluation}
Contrary to the seed detection evaluation, the terms \textit{false positive, false negative, true positive} and \textit{true negative} are insufficient to solely quantify the quality obtained by a segmentation algorithm, \ie, besides the topological errors objects might be correctly detected but possess different shapes or may be fragmented. To assess the correspondence between automatically derived segmentation results and the ground truth, the pixel-based Rand index (RI), the Jaccard index (JI), the Hausdorff metric (HM) and the normalized sum of distances (NSD) were used as defined in \cite{Coelho09}. Additionally, topological errors are assessed as \textit{split, merged, spurious} or \textit{missing} objects. The extension of the 2D implementation by Coelho \etal \cite{Coelho09} to 3D images is straightforward. The only necessary change is using a 3D implementation of the Euclidean distance map for distance calculations of segmented to reference pixels for HM and NSD. 

\paragraph{Rand-Index (RI):} Analogous to the definition in \cite{Coelho09}, $\mathbf{I}^\text{ref}$ and $\mathbf{I}^\text{seg}$ correspond to a labeled reference image and a labeled segmentation image, respectively, with $\max(\mathbf{I}^\text{ref})$ and $\max(\mathbf{I}^\text{seg})$ unique labels for each of the detected objects. Moreover, let $i,j$ be indexing all possible pixel pairs with $i \neq j$, then $a,b,c,d$ are defined as the amount of pixel pairs that meet the criteria $(\mathbf{I}^\text{ref}{[}i{]} = \mathbf{I}^\text{ref}{[}j{]} \wedge \mathbf{I}^\text{seg}{[}i{]} = \mathbf{I}^\text{seg}{[}j{]})$, $(\mathbf{I}^\text{ref}{[}i{]} = \mathbf{I}^\text{ref}{[}j{]} \wedge \mathbf{I}^\text{seg}{[}i{]} \neq \mathbf{I}^\text{seg}{[}j{]})$, $(\mathbf{I}^\text{ref}{[}i{]} \neq \mathbf{I}^\text{ref}{[}j{]} \wedge \mathbf{I}^\text{seg}{[}i{]} = \mathbf{I}^\text{seg}{[}j{]})$ or $(\mathbf{I}^\text{ref}{[}i{]} \neq \mathbf{I}^\text{ref}{[}j{]} \wedge \mathbf{I}^\text{seg}{[}i{]} \neq \mathbf{I}^\text{seg}{[}j{]})$. Following the implementation of Coelho \etal, the values for $a,b,c,d$ can be calculated using a 2D image histogram. 
\begin{figure}[htb]
\centerline{\includegraphics[width=0.5\columnwidth]{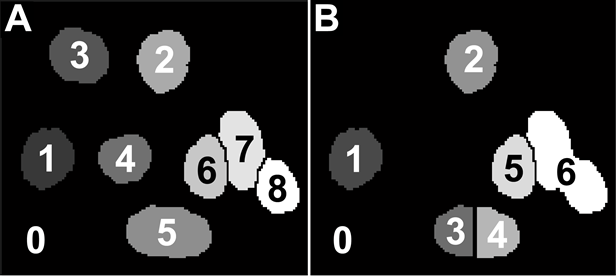}}
\caption[Labeled reference image and an artificially distorted segmentation image]{The image shows a labeled reference image (A) and an artificially distorted segmentation label image (B). In the segmentation image two nuclei were removed, two nuclei were merged and one nucleus was split. Numbers indicate the object IDs used as column and row indicators in the intersection histogram (\app\cref{tab:Appendix:SegmentationComparisonExample}).}
\label{fig:Appendix:SegmentationComparisonExample}
\end{figure}
Considering for example the reference and segmentation images depicted in \app\cref{fig:Appendix:SegmentationComparisonExample}, the label histogram listed in \app\cref{tab:Appendix:SegmentationComparisonExample} is obtained.
\begin{table}[htb]
\begin{center}
%\resizebox{\textwidth}{!}{
%\rowcolors{2}{white}{gray!25}
\begin{tabular}{lccccccccc}
\toprule
\textbf{$R/S$} & \textbf{0} & \textbf{1} & \textbf{2} & \textbf{3} & \textbf{4} & \textbf{5} & \textbf{6} & $\pmb{\Sigma}_{\text{row}}$ \\
\midrule
\textbf{0} & 13922 & 0   & 0    & 0   & 0   & 0   & 24 & 13946\\
\textbf{1} & 0     & 272 & 0    & 0   & 0   & 0   & 0  & 272 \\
\textbf{2} & 0     & 0   & 552  & 0   & 0   & 0   & 0  & 552 \\
\textbf{3} & 578   & 0   & 0    & 0   & 0   & 0   & 0  & 578 \\
\textbf{4} & 449   & 0   & 0    & 0   & 0   & 0   & 0  & 449 \\
\textbf{5} & 46    & 0   & 0    & 353 & 394 & 0   & 0  & 793\\
%\textbf{6} & 0     & 0   & 0    & 0   & 560 & 0   & 0   & 0  & 560 \\
\textbf{6} & 0     & 0   & 0    & 0   & 0   & 468 & 0  & 468 \\
\textbf{7} & 0     & 0   & 0    & 0   & 0   & 0   & 567 & 567 \\
\textbf{8} & 0     & 0   & 0    & 0   & 0   & 0   & 391 & 391 \\
\midrule
$\pmb{\Sigma}_{\text{col}}$ & 14995 & 272   & 552    & 353  & 394   & 468   & 982 \\
\bottomrule
\end{tabular}
\caption[Intersection histogram for segmentation quality assessment]{Histogram of the reference and segmentation images shown in \app\cref{fig:Appendix:SegmentationComparisonExample}. Reference labels are determining the number of rows, whereas the segmentation image labels determine the number of columns. The label $\pmb{0}$ is used as the image background label. Using the calculation scheme described in the main text yields $a=97847186, b=15489759, c=507572$ and $d=48434603$, which corresponds to a Rand Index of $0.90$ (\cref{eq:Appendix:RandIndex}) and a Jaccard Index of $2.27$ (\cref{eq:Appendix:JaccardIndex}).}
\label{tab:Appendix:SegmentationComparisonExample}
\end{center}
\end{table}
Histogram entries of the 2D histogram $\mathbf{H}$ are generated by iterating over all pixel locations of the reference and the segmentation image and by using the respective object labels as column and row index. The histogram bins are increased by one for each index pair that points to them. For lowest possible memory usage it is suggested to use sequentially increasing IDs for segments instead of uniformly distributed IDs over the whole value range, unless there really are that much objects to be evaluated. In the case of a valid ID range of 16 bit, this could for example lead to a histogram with $65535$ x $65535$ bins, which would require 16 GB of memory to be processed. To calculate the values $a,b,c,d$ from the histogram, the binomial coefficient can be used. Thus, the first case can be calculated as 
\begin{gather}
a = \sum_{i=0}^{\max(\mathbf{I}^\text{seg})} \sum_{j=0}^{\max(\mathbf{I}^\text{ref})}{\binom{\mathbf{H}{[}i,j{]}}{2}}
\end{gather}
 for histogram entries $\mathbf{H}{[}i,j{]} > 1$. This reflects the sum of all possible index combinations that have equal labels in reference and segmentation, respectively. Similarly, the values for $b$ and $c$ can be calculated using the vectors for row and the column sums $\pmb{\Sigma}_{\text{row}}$, $\pmb{\Sigma}_{\text{col}}$:
\begin{gather}
b = \left( \sum_{i=0}^{\max(\mathbf{I}^\text{ref})} {\binom{\pmb{\Sigma}_\text{row}{[}i{]}}{2}} \right) - a\\
c = \left( \sum_{j=0}^{\max(\mathbf{I}^\text{seg})} {\binom{\pmb{\Sigma}_\text{col}{[}j{]}}{2}} \right) - a
\end{gather}
The value $b$ reflects the number of index pairs where $(\mathbf{I}^\text{ref}{[}i{]} = \mathbf{I}^\text{ref}{[}j{]} \wedge \mathbf{I}^\text{seg}{[}i{]} \neq \mathbf{I}^\text{seg}{[}j{]})$, whereas $c$ vice versa represents the number of index pairs with $(\mathbf{I}^\text{ref}{[}i{]} \neq \mathbf{I}^\text{ref}{[}j{]} \wedge \mathbf{I}^\text{seg}{[}i{]} = \mathbf{I}^\text{seg}{[}j{]})$. The value for $d$ can be directly calculated from the remaining cases as:
\begin{gather}
d = {\binom{N}{2}}-a-b-c,
\end{gather}
with $N$ being the number of pixels in the images. Finally, the Rand index \cite{Rand71} is defined by:
\begin{gather}
\operatorname{RI}(\mathbf{I}^\text{ref}, \mathbf{I}^\text{seg}) = \frac{a+d}{a+b+c+d},
\label{eq:Appendix:RandIndex}
\end{gather}
which is the fraction of index pairs that have the same labeling in reference and segmentation, respectively, versus all possible pixel pairs. This measure is normalized, \ie, having a range of $[0,1]$, where $1$ indicates perfect agreement of all index pairs in the reference and the segmentation image \cite{Coelho09}. In the result tables of this thesis, the RI values are scaled by 100, to obtain percentage values.

\paragraph{Jaccard-Index (JI):} Using the same calculation of $a,b,c,d$ as for the Rand index, the Jaccard index is defined as:
\begin{gather}
\operatorname{JI}(\mathbf{I}^\text{ref}, \mathbf{I}^\text{seg}) = \frac{a+d}{b+c+d}.
\label{eq:Appendix:JaccardIndex}
\end{gather}
The higher the value attained by the Jaccard index, the better is the alignment of the reference image $\mathbf{I}^\text{ref}$ to the segmentation image $\mathbf{I}^\text{seg}$ \cite{Coelho09}.

\paragraph{Normalized Sum of Distances (NSD):} The NSD is a normalized quantity that reflects the average distance of labeled pixels that do not agree in reference and segmentation:
\begin{gather}
\operatorname{NSD}(\mathbf{I}^\text{ref}, \mathbf{I}^\text{seg}) = \frac{\sum_{i=1}^N \left((\mathbf{I}^\text{ref}{[}i{]} \neq \mathbf{I}^\text{seg}{[}i{]}) \cdot D(i) \right)}{\sum_{i=1}^N D(i)}.
\end{gather}
Here $D(i)$ is the minimum distance of pixel $i$ to the edge of the reference shape being compared. The index $i$ is running over all non-zero entries in the union of images $\mathbf{I}^\text{ref}$ and $\mathbf{I}^\text{seg}$. For the comparison $(\mathbf{I}^\text{ref}{[}i{]} \neq \mathbf{I}^\text{seg}{[}i{]})$ binary images are assumed. To avoid ambiguities of reference and segmentation, the calculation of the NSD is performed on individual labeled regions that are binarized and subsequently used for the NSD calculation. Values of the obtained NSD values of all considered objects are averaged for each image.

\paragraph{Hausdorff Metric (HM):} To assess the deviation of reference and segmentation object, the Hausdorff metric is defined as the maximum of the set of minimal distances of two compared shapes \cite{Bamford03}.
\begin{gather}
\operatorname{HM}(\mathbf{I}^\text{ref}, \mathbf{I}^\text{seg}) = \max_{i=1,...,N} \left( D(i) \cdot (\mathbf{I}^\text{ref}{[}i{]} \neq \mathbf{I}^\text{seg}{[}i{]})  \right).
\end{gather}
The measure is iteratively calculated for all overlapping shapes of reference and segmentation image and averaged by the number of reference objects afterwards. For both the NSD and the Hausdorff metric, the distance to the closest reference point at each pixel location $i$ can be efficiently computed using the Euclidean distance map of the binarized reference image ($\mathbf{I}^\text{edm}$) and the inverted binarized reference image ($\mathbf{I}^\text{iedm}$) \cite{Cuisenaire99}:
\begin{gather}
 D(i) = \max \begin{cases} \mathbf{I}^\text{edm}{[}i{]}   &, \mathbf{I}^\text{ref}{[}i{]} \neq 0 \\
 						   \mathbf{I}^\text{iedm}{[}i{]}  &, otherwise \end{cases}
\end{gather}
Two cases are required due to the fact that a misaligned segment can be either smaller or larger than the reference object. If the segmented object is smaller than the reference object, the maximum distance reflects the distance of the segmented object to the reference object from its inside ($\mathbf{I}^\text{edm}{[}i{]}$). In the other case, the distance has to be calculated from a point outside of the reference shape to its border ($\mathbf{I}^\text{iedm}{[}i{]}$). 

The formulas for the evaluation measures are summarized in \app\cref{tab:Appendix:SegmentationPerformanceMeasures}.
\begin{table}[htb]
%\begin{center}
%\rowcolors{2}{white}{gray!25}
%\small
\resizebox{\textwidth}{!}{
\begin{tabular}{m{0.18\columnwidth}ccm{0.36\columnwidth}}
\toprule
\textbf{Measure} 					& \textbf{Formula} 		& \textbf{Range} & \textbf{Description} \\
\midrule
Rand Index (RI) & $\frac{a+d}{a+b+c+d}$ & $[0,1]$ & Fraction of pixel combinations that have equal labels to all possible pixel combinations. \\
\midrule
Jaccard Index (JI) & $\frac{a+d}{b+c+d}$ & $[0,\infty)$ & Fraction of pixel combinations that have equal labels to all non-equally labeled pixel combinations. \\
\midrule
Hausdorff Metric (HM) & $\max \left( D(i) \cdot (\mathbf{I}^\text{ref}{[}i{]} \neq \mathbf{I}^\text{seg}{[}i{]}) \right)$ & $[0,\infty)$ & Describes the maximum of the set of shortest distances between two segments \cite{Bamford03}. \\
\midrule
Normalized Sum of Distances (NSD) & $\frac{\sum_{i=1}^N \left((\mathbf{I}^\text{ref}{[}i{]} \neq \mathbf{I}^\text{seg}{[}i{]}) \cdot D(i) \right)}{\sum_{i=1}^N D(i)}$ & $[0,1]$ & The normalized distance of erroneously segmented pixels to the reference border \cite{Coelho09}. \\
\bottomrule
\end{tabular}}
\caption[Performance evaluation criteria for segmentation algorithms]{Criteria for the quantitative comparison of a labeled ground truth segmentation with an automatically obtained segmentation image as defined in \cite{Rand71, Coelho09, Bamford03}.}
\label{tab:Appendix:SegmentationPerformanceMeasures}
%\end{center}
\end{table}

\paragraph{Split Objects:} Objects are considered as split if multiple fragments of the segmentation image are mapped to a single segment in the reference image. This can be again computed using the 2D histogram by searching for rows with multiple non-zero values (See \eg, row 5 in \ref{tab:Appendix:SegmentationComparisonExample}).

\paragraph{Merged Objects:} Merged objects are defined as multiple objects in the reference image and as a single object in the segmentation image. By searching for columns with multiple non-zero values, these merged objects can be identified in the histogram representation (See \eg, column 6 in \ref{tab:Appendix:SegmentationComparisonExample}).

\paragraph{Spurious Objects:} Objects that were erroneously added by the segmentation algorithm can be identified by searching for columns that have their maximum value at the background row of the reference.

\paragraph{Missing Objects:} Missing objects can be identified in the 2D histogram by searching for rows, where the maximum values lie in the column of the segmentation background. In \ref{tab:Appendix:SegmentationComparisonExample}, the third and fourth row correspond to missing objects of \ref{fig:Appendix:SegmentationComparisonExample}.

\paragraph{Recall, Precision and F-Score:} The values for recall, precision and F-Score as defined in \cref{eq:Appendix:Recall}, \cref{eq:Appendix:Precision} and \cref{eq:Appendix:FScore} were also calculated for the segmentation evaluation by considering correctly identified objects as true positives ($TP$), missing and merged objects as false negative detections ($FN$), and split and added objects as false positive detections ($FP$).

%%%%%%%%%%%%%%%%%%%%%%%%%%%%%%%%%%%%%%%%%%%%%%%%%%%%%%%%%%%%%%%%%%%%%%%%%%
%% Tracking Performance Evaluation
%%%%%%%%%%%%%%%%%%%%%%%%%%%%%%%%%%%%%%%%%%%%%%%%%%%%%%%%%%%%%%%%%%%%%%%%%%
\section{Tracking}
\label{sec:Appendix:Tracking}
To evaluate and compare the quality of the investigated tracking algorithms, the \texttt{TRA} measure described by Ma{\v{s}}ka \etal was used \cite{Maska14}. This measure is calculated by considering the tracking result as an acyclic oriented graph. Each of the nodes within the graph corresponds to a detected object in a single frame and the edges between objects correspond to their temporal association. The idea of the tracking evaluation measure is to compare the graph of a manually annotated ground truth dataset to the automatically generated tracking graph. A high-quality automatic tracking should contain as few as possible differences to the ground truth graph, \ie, counting and weighting the number of changes necessary to transform the automatically obtained graph to the ground truth graph can be used to assess the tracking quality. Due to the fact that some correction operations are more difficult than others, the required correction steps are weighted by the effort needed for manual correction. The weights defined in \cite{Maska14} are as follows with the required operation in brackets:
\begin{itemize}
	\item delete a false positive node or edge: $w_\text{FP}=1$ or $w_\text{ED} = 1$ (single deletion of a node (FP) or edge (ED))
	\item change an erroneous edge: $w_\text{EC}=1$ (single reassignment of the edge terminal (EC))
	\item add a missing temporal association: $w_\text{EA}=1.5$ (linkage of two objects necessary (EA))
	\item split an under-segmented node: $w_\text{NS}=5$ (manual splitting the merged object (NS))
	\item add a missing object: $w_\text{FN}=10$ (manual segmentation of the missing object (FN))
\end{itemize}
The weighted sum of required change operations can then be formulated as
\begin{gather}
	\texttt{TRA}_P = 
	w_\text{NS}\cdot\text{NS} + 
	w_\text{FN}\cdot\text{FN} + 
	w_\text{FP}\cdot\text{FP} + 
	w_\text{ED}\cdot\text{ED} + 
	w_\text{EA}\cdot\text{EA} + 
	w_\text{EC}\cdot\text{EC}.
\end{gather}

The obtained $\texttt{TRA}_P$ value is compared to the number of operations that would be required to manually generate the ground truth graph
\begin{gather}
\texttt{TRA}_E = 
	w_\text{FN}\cdot |N| + 
	w_\text{EA}\cdot |E|,
\end{gather}
with $|N|$ and $|E|$ being the number of nodes and edges, respectively. Finally, a normalization of the measure is performed using the following formulation:
\begin{gather}
	\texttt{TRA} = 1 - \frac{\min{(\texttt{TRA}_P, \texttt{TRA}_E)}}{\texttt{TRA}_E}.
\end{gather}
The value range of the \texttt{TRA} is thus scaled to a range of $[0,1]$ to be able to compare results obtained on different datasets with higher values being better and with 1 being ideal. In addition to this single value quality measure, a more detailed view on the errors is obtained by looking at the number of false positive detections, false negative detections, incorrect edges, missing edges, redundant edges and merged objects. This information is also provided by the respective log file produced for the \texttt{TRA} measure calculation. A more detailed description of the measure, implementation details and exemplary calculations are provided in the original publication \cite{Maska14}.

\section{Evaluation Platform}
\label{sec:Appendix:EvaluationPlatform}
Unless otherwise stated, all measurements with respect to processing times were performed on a desktop PC equipped with an Intel Core i7-2600 CPU @ 3.4GHz and 32GB of memory installed using the Windows 7 (x64) operating system.
 \cleardoublepage

\chapter{Benchmark Datasets and Parameters}
\bgroup
\def\arraystretch{1.5}

\begin{table}[htb]
\begin{center}
%\rowcolors{2}{white}{gray!25}
% \begin{tabular}{m{0.05\columnwidth}m{0.06\columnwidth}m{0.08\columnwidth}m{0.1\columnwidth}m{0.08\columnwidth}m{0.05\columnwidth}m{0.33\columnwidth}}
\resizebox{\textwidth}{!}{
\small\begin{tabular}{lcccccm{0.36\columnwidth}}
\toprule
\textbf{Name} & \textbf{Im.} & \textbf{N./B.} & \textbf{Resolution} & $\sigma_{\text{agn}}$ & $\sigma^2_{\text{smooth}}$ & \textbf{Description} \\
\midrule
\texttt{SBDS1} & 30 & 1 & 314$\times$220$\times$50 & - & - & 3D images of simulated fluorescent nuclei by Svoboda \etal \cite{Svoboda12}. Used in \cref{sec:chap3:SeedDetection} and \cref{sec:chap3:TWANG}. \\
\midrule
\texttt{SBDS2} & 50 & 10 & 314$\times$220$\times$50 & $[0.0,0.128]$ & - & Selected images of \texttt{SBDS1} with 10 different noise levels. Used in \cref{sec:chap3:SeedDetection} and \cref{sec:chap3:TWANG}. \\
\midrule
\texttt{SBDS3} & 50 & 10 & 314$\times$220$\times$50 & - & $[0,100]$ & Selected images of \texttt{SBDS1} with 10 different blur levels. Used in \cref{sec:chap3:SeedDetection} and \cref{sec:chap3:TWANG}. \\
\midrule
\texttt{SBDL} & 100 & 5 & 1024$\times$768$\times$2 & $[10^{-5},10^{-1}]$ & - & Simulated multi-channel 2D images containing elongated objects with spherical objects in the close vicinity. Used in \cref{sec:chap3:SpinalCord}. \\
\midrule
\texttt{SBDP1} & 4 & 1 & 128$\times$128$\times$51 & - & - & 3D images of simulated fluorescent membrane structures by Mosaliganti \etal \cite{Mosaliganti12}. Used in \cref{sec:chap3:RACE}. \\
\midrule
\texttt{SBDP2} & 40 & 10 & 128$\times$128$\times$51 & $[0.0,0.008]$ & - & Images of \texttt{SBDP1} with 10 different noise levels. Used in \cref{sec:chap3:RACE}. \\
\midrule
\texttt{SBDP3} & 40 & 10 & 128$\times$128$\times$51 & - & $[0,100]$ & Images of \texttt{SBDP1} with 10 different blur levels. Used in \cref{sec:chap3:RACE}. \\
\bottomrule
\end{tabular}}
\caption[Benchmark datasets used for the validation experiments in \cref{sec:chap3:EfficientSegmentation}]{The benchmark datasets used for the validation experiments in \cref{sec:chap3:EfficientSegmentation}. Columns list the dataset identifier (Name), the number of images (Im.), the number of noise/blur levels (N./B.), the image resolution, the noise/blur parameter ranges for $\sigma_{\text{agn}}$ and $\sigma^2_{\text{smooth}}$ and the dataset description.}
\label{tab:Appendix:BenchmarkDatasets}
\end{center}
\end{table}

\begin{table}[htb]
\begin{center}
%\rowcolors{2}{white}{gray!25}
%\begin{tabular}{m{0.05\columnwidth}m{0.06\columnwidth}m{0.06\columnwidth}m{0.08\columnwidth}m{0.1\columnwidth}m{0.1\columnwidth}m{0.4\columnwidth}}
\resizebox{\textwidth}{!}{
\small\begin{tabular}{lcccccm{0.28\columnwidth}}
\toprule
\textbf{Name} & \textbf{Im.} & \textbf{N.} & \textbf{Objects} & \textbf{Resolution} & $\sigma_{\text{agn}}$ & \textbf{Description} \\
\midrule
\texttt{SBDE1} & 10 & 1 & 300--1000 & 640$\times$640$\times$128 & 0.0007 & Five pairs of subsequent frames with different numbers of objects. Used for seed detection and segmentation validation (\cref{sec:chap4:SeedDetection} and \cref{sec:chap4:Segmentation}). \\
\midrule
\texttt{SBDE2} & 20 & 20 & 1000 & 640$\times$640$\times$128 & $[0.0005,0.01]$ &  Single frame of \texttt{SBDE1} with 20 different noise levels. Used for seed detection and segmentation validation (\cref{sec:chap4:SeedDetection} and \cref{sec:chap4:Segmentation}). \\
\midrule
\texttt{SBDE3} & 10/20 & 1 & 300--1000 & 640$\times$640$\times$128 & 0.0007 &  Images of \texttt{SBDS1} but with additional rotation simulation of sequential (SeMV) and simultaneous acquisition (SiMV). Images are rotated by $180^\circ$ at every second frame for SeMV (10 images) and every image is rotated for SiMV (20 images). Used for the validation of the multiview information fusion (\cref{sec:chap4:MultiviewFusion}). \\
\midrule
\texttt{SBDE4} & 100 & 1 & 1000 & 640$\times$640$\times$128 & 0.0007 &  Sequential time points containing $1000$ objects each. Images are rotated by $180^\circ$ at every frame for SiMV ($2 \times 50$ images). Used for the validation of the tracking (\cref{sec:chap4:Tracking}). \\
\bottomrule
\end{tabular}}
\caption[Benchmark datasets used for the validation experiments in \cref{sec:chap4:AlgorithmEnhancement}]{The benchmark datasets used for the validation experiments in \cref{sec:chap4:AlgorithmEnhancement}. Columns list the dataset identifier (Name), the number of images (Im.), the number of noise levels (N.), the number of objects (Objects), the image resolution, the noise parameter ranges for $\sigma_{\text{agn}}$ and the dataset description.}
\label{tab:Appendix:BenchmarkDatasetsEmbryo}
\end{center}
\end{table}

\egroup

% Table generated by Excel2LaTeX from sheet 'Tabelle1'
\begin{sidewaystable}[htbp]
	%\begin{center}
	%\rowcolors{2}{white}{gray!25}
\resizebox{\textwidth}{!}{
    \begin{tabular}{lcccccc}
    \toprule
    \textbf{Dataset} & \multicolumn{2}{c}{\textbf{Drosophila (SIMView)}} & \multicolumn{2}{c}{\textbf{Mouse (SIMView)}} & \multicolumn{2}{c}{\textbf{Drosophila (Confocal)}} \\
    \midrule
    {Parameter/Algorithm} & $\text{RACE}_{\text{MS}}$  & $\text{RACE}_{\text{NS}}$  & $\text{RACE}_{\text{MS}}$  & $\text{RACE}_{\text{NS}}$  & $\text{RACE}_{\text{MS}}$  & $\text{RACE}_{\text{NS}}$ \\
    {XY vs. Z Ratio} & 5 & 5 & 5 & 5 & 5 & 5 \\
    {2D Median Radius} & 2 & 2 & 2 & 2 & 2 & 2 \\
    {Binary Threshold (MS)} & \multicolumn{1}{c}{0.0008} & \multicolumn{1}{c}{-} & \multicolumn{1}{c}{0.00004} & \multicolumn{1}{c}{-} & \multicolumn{1}{c}{0.002} & \multicolumn{1}{c}{-} \\
    {Laplacian-of-Gaussian Sigma (NS)} & \multicolumn{1}{c}{-} & \multicolumn{1}{c}{4} & \multicolumn{1}{c}{-} & \multicolumn{1}{c}{4} & \multicolumn{1}{c}{-} & \multicolumn{1}{c}{4} \\
    {Binary Threshold (NS)} & \multicolumn{1}{c}{-} & \multicolumn{1}{c}{OTSU} & \multicolumn{1}{c}{-} & \multicolumn{1}{c}{0.00002} & \multicolumn{1}{c}{-} & \multicolumn{1}{c}{OTSU} \\
    {HMaximaFilter Height} & 20 & 20 & 40 & 45 & 20 & 20 \\
    {HMaxima Binarization Threshold} & 0.00001 & 0.00001 & 0.00001 & 0.00001 & 0.00001 & 0.00001 \\
    {HessianToObjectnessFilter Sigma} & 2 & 2 & 2 & 2 & 2 & 2 \\
    {HessianToObjectnessFilter Alpha} & 1 & 1 & 1 & 1 & 1 & 1 \\
    {HessianToObjectnessFilter Beta}  & 1 & 1 & 1 & 1 & 1 & 1 \\
    {HessianToObjectnessFilter Gamma} & 0.1 & 0.1 & 0.1 & 0.1 & 0.1 & 0.1 \\
    {Morphological Closing MinRadius} & 4 & 4 & 1 & 1 & 4 & 4 \\
    {Morphological Closing MaxRadius} & 4 & 4 & 4 & 4 & 4 & 4 \\
    {Morphological Closing FilterMask3D} & 1 & 1 & 1 & 1 & 1 & 1 \\
    {Slice-By-Slice Watershed Level} & 39 & 39 & 2 & 2 & 50 & 50 \\
    {Slice-By-Slice RegionProps MinSeedArea} & 3 & 3 & 3 & 3 & 0 & 0 \\
    {Slice-By-Slice FusionFilter MaxSegmentArea} & 1000 & 1000 & 2000 & 2000 & 1000 & 1000 \\
    {Slice-By-Slice FusionFilter MinVolume} & 1500 & 1500 & 2300 & 2300 & 1500 & 1500 \\
    {Slice-By-Slice FusionFilter MaxVolume} & 4000 & 4000 & 13000 & 13000 & 4000 & 4000 \\
    \bottomrule
    \end{tabular}}%
		\captionsetup{width=0.96\textheight}
  \caption[Parameterization of the RACE algorithm]{Parameterization of the RACE algorithm applied on fluorescence microscopy images of different specimens and microscopy techniques. The three most important intensity dependent parameters that need to be adjusted for each image are \textit{Binary Threshold (MS, NS)}, \textit{HMaximaFilter Height} and \textit{Slice-By-Slice Watershed Level}. The remaining parameters can be inferred from the experimental setup or may remain constant (adapted from \cite{Stegmaier16}).}
  \label{tab:Appendix:RaceParameters}%
%\end{center}
\end{sidewaystable}%
 \cleardoublepage
 
\cleardoublepage

\backmatter

\listoffigures 
\listoftables

\lstlistoflistings
\addcontentsline{toc}{chapter}{List of Listings}

%\nocite{*}
\bibliographystyle{mydiss}
\bibliography{E:/Literature/biosignal}

\printindex

\end{document}